\documentclass{article}
\usepackage{iclr2026_conference,times}

\usepackage{cmap}               
\usepackage[utf8]{inputenc}
\usepackage[T1]{fontenc}
\usepackage{accsupp}
\usepackage{hyperref}       \usepackage{url}            \usepackage{booktabs}       \usepackage{amsfonts}       \usepackage{nicefrac}       \usepackage{microtype}      \usepackage[x11names]{xcolor}
\usepackage{placeins}

\usepackage{textcomp}
\usepackage{tcolorbox}
\usepackage{listings}

\definecolor{cellbg}{RGB}{247,247,247}
\definecolor{cellborder}{RGB}{207,207,207}
\definecolor{kwcolor}{RGB}{0,119,170}
\definecolor{strcolor}{RGB}{170,68,0}
\definecolor{commentcolor}{RGB}{100,100,100}
\definecolor{builtincolor}{RGB}{155,0,155}
\definecolor{numcolor}{RGB}{0,150,0}

\lstdefinestyle{python}{
  language=Python,
  basicstyle=\fontfamily{lmtt}\selectfont\tiny,
  keywordstyle=\color{kwcolor}\bfseries,
  stringstyle=\color{strcolor},
  commentstyle=\color{commentcolor}\itshape,
  numberstyle=\color{numcolor},
  emph={True,False,None,self,cls},
  emphstyle=\color{builtincolor},
  morekeywords={print,range,len,int,str,float,list,dict,set,tuple},
  showstringspaces=false,
  columns=fullflexible,
  keepspaces=true,
  breaklines=true,
  literate={␣}{{\BeginAccSupp{method=hex,unicode,ActualText=2423}\textcolor{cellbg}{\textvisiblespace}\EndAccSupp{}}}1,
}

\tcbuselibrary{listings, breakable}

\newcounter{cellnum}
\newtcblisting{ipycell}{
  listing only, breakable,
  colback=cellbg, colframe=cellborder,
  arc=2pt, left=4pt, right=4pt, top=2pt, bottom=2pt,
  before={\stepcounter{cellnum}\noindent\texttt{In [\thecellnum]:}\par\vspace{2pt}},
  listing options={style=python}
}

\usepackage{amsmath}
\usepackage{amssymb}
\usepackage{mathtools}
\usepackage{amsthm}

\usepackage{lipsum}

\usepackage{listings}
\usepackage{textcomp}
\usepackage{fancyvrb}

\usepackage{amsmath,amsfonts,bm}

\def\eqref#1{equation~\ref{#1}}

\def\1{\bm{1}}

\DeclareMathAlphabet{\mathsfit}{\encodingdefault}{\sfdefault}{m}{sl}
\SetMathAlphabet{\mathsfit}{bold}{\encodingdefault}{\sfdefault}{bx}{n}

\usepackage{hyperref}
\usepackage{url}

\usepackage{tabularx}
\usepackage{booktabs}

\usepackage{enumitem}
\usepackage{mdframed}

\iclrfinalcopy
\title{Unified Neural Scaling Laws}

\author{Ethan Caballero
\\ Mila, University of Montreal
\\ \texttt{ethan.victor.caballero@gmail.com}
\\ \texttt{ethan.caballero@mila.quebec}
\And
Priyank Jaini
\\ Google DeepMind
\And
David Krueger
\\ Mila, University of Montreal
\And
\hspace{140pt}
Irina Rish
\\ \hspace{140pt} Mila, University of Montreal
}

\input{glyphtounicode}
\pdfglyphtounicode{asteriskmath}{002A}
\pdfglyphtounicode{minus}{002D}

\begin{document}

\maketitle

\begin{abstract}
We present a functional form (that we refer to as a \textit{Unified Neural Scaling Law (UNSL)}) that accurately models and extrapolates the scaling behaviors of deep neural networks as multiple dimensions all vary simultaneously (i.e. how the evaluation metric of interest varies as one simultaneously varies the number of model parameters, training dataset size, number of training steps, number of inference steps, and various hyperparameters) for various architectures and for each of various tasks within a varied set of upstream and downstream tasks. When compared to other functional forms for neural scaling, this functional form yields \textbf{extrapolations} of scaling behavior that are considerably more accurate on this set. 
\end{abstract}

\vspace{-1mm}

\section{Introduction}
\label{intro}

\vspace{-1mm}

Training today's state-of-the-art neural networks requires  significant  amounts of   computational resources and training data. Given a wide range of available methods and architectures to choose from, accurate forecasting of  their performance  is essential for selecting those that are likely to perform best at scale, especially since the top-performing methods at smaller scales often fail to maintain their performance at larger scales \citep{sutton2019bitter,tolstikhin2021mlp}. Moreover, accurate forecasting of neural network behaviors at scale is critical not only for identifying the top-performing  approaches but also for ensuring AI safety, as predicting the emergence of novel capabilities at scale is essential for responsible development and deployment of advanced AI systems.
This realization motivated the study of {\em neural scaling laws} \citep{cortes1994learning,2017arXiv171200409H,DBLP:journals/corr/abs-1909-12673,icm2020arXiv200108361K,DBLP:journals/corr/abs-2106-04560,abnar2021exploring,brown2020language, bahri2021explaining, Alabdulmohsi2022revisiting, caballero2023broken} which aim to predict the behavior of large-scale models as the amount of compute, data, and model parameters increases.

Clearly, the  accuracy,  as well as the confidence of predictions made by neural scaling laws can only increase (or remain the same) as a larger number of relevant predictors are included, due to the standard conditional entropy inequality, $H(Y|{\bf X}) \leq H(Y)$, where ${\bf X}$ is the vector of predictive variables and $Y$ is the performance evaluation metric. Namely, as  the number of predictive variables $X_i, i = 1,...,m$  increases, the conditional entropy $H(Y|(X_1, ... , X_m))$ can only decrease (or remain the same). Ultimately, to obtain the maximal  achievable reduction in the entropy of $Y$, one would need to identify the set of all possible $X_i$ that are causally related to $Y$, and develop a complete model $P(Y|{\bf X})$ that can serve as a ``unified functional form'' of neural network behavior(s) at scale.

To address this need for a (more) unified functional form, we present \textit{Unified Neural Scaling Laws (UNSL)},
a functional form that accurately models and extrapolates the scaling behaviors of deep neural networks as multiple dimensions all vary simultaneously. When compared to other functional forms for neural scaling, this functional form yields \textbf{extrapolations} of scaling behavior that are considerably more accurate on this set. Additionally, this functional form accurately models and extrapolates multivariate scaling behavior that other functional forms are incapable of expressing such as the nonmonotonic transitions present in the scaling behavior of overfitting and hyperparameters (such as learning rate and standard deviation of weights at initialization) that have a nonmonotonic relationship with the performance evaluation metric.

\section{The Functional Form of Unified Neural Scaling Laws}
\label{section:unsl}

\begin{figure*}[t!]    \centering

\raisebox{0mm}{\includegraphics[width=0.33\textwidth]{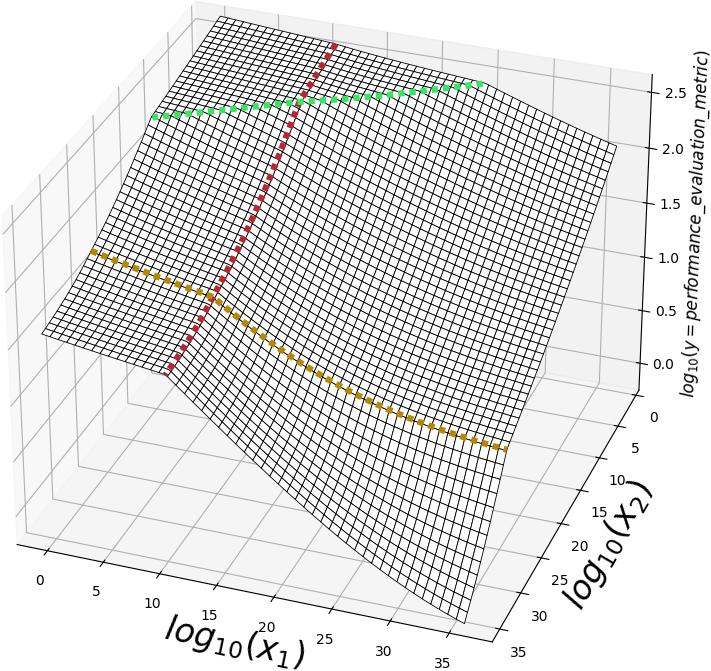}}
\hspace*{.3mm}\raisebox{9.51mm}{\includegraphics[width=0.661\textwidth]{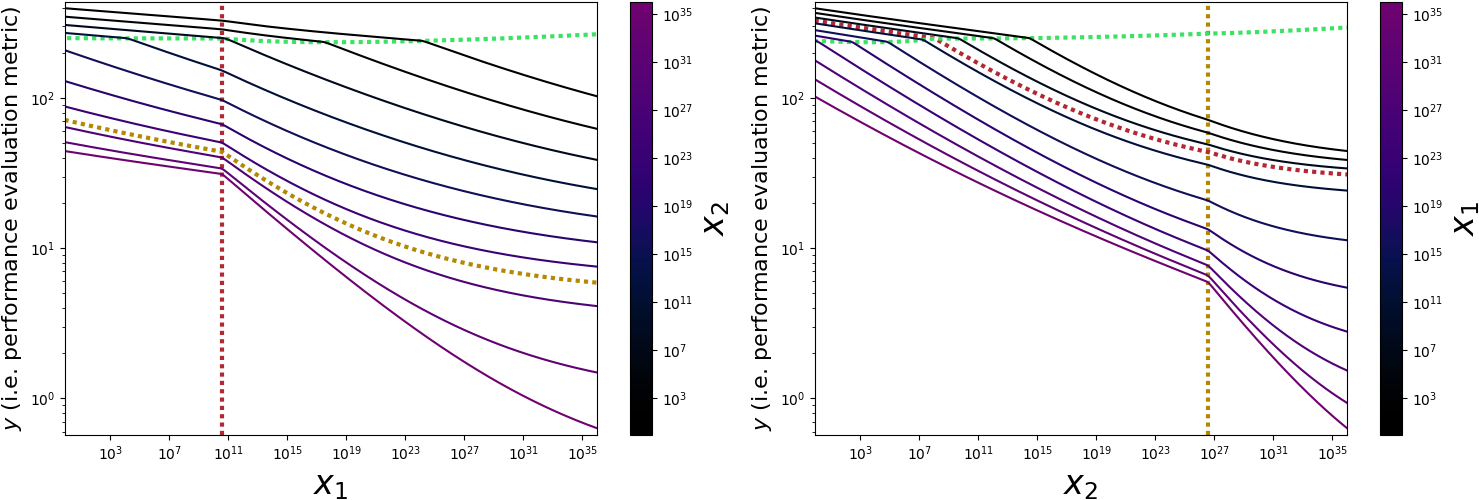}}
    \caption{An illustration of a Unified Neural Scaling Law (UNSL)  (dark solid lines) with two input dimensions, $x_1$ and $x_2$; the central and the right plots show the projections on each of the input dimensions, respectively. In this particular example, an UNSL contains 3 hyperbreaks highlighted by brighter dotted lines - orange, yellow, and green.  The green hyperbreak is created by a non-bottleneck component. The orange hyperbreak is created by an $x_1$ bottleneck component. The yellow hyperbreak is created by an $x_2$ bottleneck component. See Section \ref{section:unsl} for detailed explanation of hyperbreaks.}
    \label{fig:figure_1}
\end{figure*}

\FloatBarrier
Let  $y$ denote a performance evaluation metric of interest, e.g. prediction error or cross-entropy, "upstream" (i.e., measured on the validation dataset from the pretraining data distribution) or "downstream" (i.e., measured on new data and/or tasks that the model does not encounter during pretraining), and let $(x_i)_{i=1}^m \in \overline{\mathbb{R}}_{>0}^{\,m}$ denote a tuple of $m$ quantities that can be viewed as predictors of $y$, e.g. number of model parameters, training dataset size, number of training steps, number of inference steps, and values of various hyperparameters.

We present the following general functional form of a unified neural scaling law (UNSL):

\begin{equation}
y = a_0 + \left(  \bigg(Q(3) + \underbrace{\left(Q(S+4) + {a_1}^{-1}\right)^{-1}}_{\text{oppositional force of overfitting}} \bigg)^{-1} + {a_2}^{-1} \right)^{-1},
\label{eq:unsl1}
\end{equation}

where $Q$ is defined as follows:
\begin{equation}
\begin{split}
Q(&q) = \left(\left(R(q)\right)^{-1} + {a_q}^{-1}\right)^{-1}+ \underbrace{\sum_{s=1}^{S} \left(R(q+s) + {a_{q+s}}^{-1}\right)^{-1}}_{\text{oppositional force of hyperparameters}},
\end{split}
\label{eq:unsl2}
\end{equation}

where $R$ is defined as follows:
\begin{equation}
\begin{split}
R(&r) = \underbrace{K( \, U_r,\,\, n_{r_0},\,\, r \! \cdot \! (m \! + \! 1))}_{\text{non-bottleneck component}} \, + \, \underbrace{\sum_{\mathclap{ \kern 0.9mm t \in T_r}} K(\{t\},\,\, n_{r_t},\,\, r \! \cdot \! (m \! + \! 1) \! + \! t)}_{\text{bottleneck components}}, \\  & \hspace{-3.8mm} \text{where } U_r,T_r\subseteq\{1,\ldots,m\},
\end{split}
\label{eq:unsl3}
\end{equation}

and where $K$ is a {\em Multivariate Broken Neural Scaling Law (MBNSL)}, defined as follows:

\begin{equation}
K(M, n, k) = b_{k} \cdot \left(\prod_{i \in M} x_i^{-c_{i_{0_{k}}}}\right)
\prod_{j=1}^n \left( 1 + \left( \frac{\prod_{\substack{i \in M}} x_i^{c_{i_{j_{k}}}}}{d_{j_{k}}} \right)^{\left|\frac{1}{f_{j_{k}}}\right|} \right)^{-f_{j_{k}}}.
\label{eq:mbnsl}
\end{equation}

The parameters whose values are unknown constants that must be estimated by fitting the above functional form to the ($x_1 ... x_m, y$) data points are all those whose base is one of these: $a, b, c, d, f$. 

\clearpage  
\vspace*{-7mm}

The purpose of the variables $i$, $j$, $k$, $q$, $r$, $s$, $t$ is indexation. $n$ is a bound of a product operator; as a result, each of $n_{r_0}$ and $n_{r_t}$ implicitly is a bound of a product operator. $S$ is a bound of a summation operator. $M \subseteq \{1,  \ldots,  m\}$. $M$ is a product index set; as a result, $U_r$ implicitly is a product index set. $T_r$ is a summation index set. $K, Q, R$ are functions and the contents of the parentheses in $K(\cdot), Q(\cdot), R(\cdot)$ are arguments of those functions. Whenever an argument of $K$, $Q$, or $R$ is obtained via addition(s) and/or multiplication(s), the sole reason that those additions and multiplications occur is to cause each instantiation of $K$ to have a unique value for $k$.

\textbf{Equations \ref{eq:unsl1}, \ref{eq:unsl2}, \ref{eq:unsl3}, and \ref{eq:mbnsl} are interpreted as follows.}

We use the term \textbf{\textit{multi-log space}} to refer to the (m+1)-dimensional space obtained by applying the logarithmic transformation to each of every dimension ($x_1 ... x_m, y$).

\textbf{Equation \ref{eq:mbnsl}} is an extension of the univariate \textit{broken neural scaling law (BNSL)} of \citet{caballero2023broken} to multivariate settings. When $|M| = 1$, its expressivity is identical to the univariate broken neural scaling law functional form (with the performance limit term subtracted out) from \citet{caballero2023broken}. When $|M| > 1$, Equation \ref{eq:mbnsl} defines a sequence of $n + 1$ smoothly connected hyperplanes in multi-log space. Constant $n$ corresponds to the number of (smooth) ``hyperbreaks'' (i.e. transitions) between $n + 1$ consecutive hyperplanes in multi-log space; the dimensionality of each hyperplane is $|M|$, and the dimensionality of each hyperbreak is $|M| - 1$. When $n=0$, Equation \ref{eq:mbnsl} becomes $b_{k} \prod_{i \in M} x_i^{-c_{i_{0_{k}}}}$.
In multi-log space, the initial exponent for each input dimension $(c_{i_{0_{k}}})_{i \in M}$ corresponds to the gradient of the first hyperplane with respect to the input dimensions $(x_i)_{i \in M}$. In multi-log space, $b_{k}$ corresponds to the offset of the output of Equation \ref{eq:mbnsl}.
The j-th hyperplane smoothly transitions to the (j+1)th hyperplane at the values of $(x_i)_{i \in M}$ for which this equality is true: $d_{j_k} = \prod_{i \in M} x_i^{c_{i_{j_k}}}$. The j-th exponent for each input dimension $(c_{i_{j_{k}}})_{i \in M}$ multiplied by $\operatorname{sign}(f_{j_k})$ corresponds to the change in gradient (with respect to the input dimensions $(x_i)_{i \in M}$) between the j-th hyperplane and the (j+1)th hyperplane in multi-log space. Constant $f_{j_k}$ represents the sharpness of the hyperbreak between the j-th and the (j+1)th hyperplane in multi-log space; smaller values of $|f_{j_k}|$ yield a sharper hyperbreak and regions (before and after the j-th hyperbreak) that have less curvature in
multi-log space; larger values of $|f_{j_k}|$ yield a smoother (wider) hyperbreak and regions (before and after the
j-th hyperbreak) that have more curvature in multi-log space.

\textbf{Equation \ref{eq:unsl3}} consists of 2 kinds of components. The component $K( U_r,\,\, n_{r_0},\,\, r \! \cdot \! (m \! + \! 1))$ is referred to as a ``non-bottleneck'' component and corresponds to the smoothly connected hyperplanes (in multi-log space) as described in the previous paragraph. Each of the components summed together in the summation 
$\sum_{\substack{\\\\ \kern -3.9mm t \in T_r}} \kern -2.2mm K(\{t\},\, n_{r_t},\, r \! \cdot \! (m \! + \! 1) \! + \! t)$ 
is referred to as a ``bottleneck'' component and corresponds to each of the performance limits when bottlenecked by each of the dimensions $(x_t)_{t \in T_r}$.

\textbf{Equation \ref{eq:unsl2}} is as follows. $R(q)$ represents everything that has been discussed thus far in this Section \ref{section:unsl}; $a_q$ represents a misperformance limit (e.g., the cross-entropy or test error rate of random guessing). The remaining contents of Equation \ref{eq:unsl2} represent the ``oppositional force'' of hyperparameters (such as learning rate and standard deviation of weights at initialization) that have an oppositional relationship with the performance evaluation metric; for example, when learning rate and/or standard deviation of weights at initialization are too large, they exert an ``oppositional force'' on the value of $Q(q)$. $S$ represents the number of misperformance limits of the ``oppositional force'' of hyperparameters; $S$ does \textbf{not} represent any other quantities (e.g. \textbf{$S$ does not represent the number of hyperparameters}). \textbf{In practice, $S \le 1$ except in relatively contrived scenarios} (e.g. scenarios in which it is simultaneously true that number of training steps is very small (e.g. smaller than 5 steps) and learning rate is a value greater than 1) such as the scaling behavior shown in Figure \ref{fig:multiple_upper_limits} of Appendix \ref{subsection:empirical_desiderata_5}.

\textbf{Equation \ref{eq:unsl1}} is as follows. $Q(3)$ represents everything that has been discussed thus far in this Section \ref{section:unsl}. The constant $a_0$ corresponds to the limit as to how far the value of $y$ can be reduced (or maximized) even if all of $x_1 ... x_m$ go to the values of $(x_i)_{i=1}^m \in \overline{\mathbb{R}}_{>0}^{\,m}$ that yield the global optimum of $y$. The constant $a_2$ corresponds to a misperformance limit that is caused by the particular performance evaluation metric that is used. For example, when using a performance evaluation metric (such as cross-entropy) that is unbounded above, $a_2 = \infty$ (i.e. $a_2^{-1} = 0$); and when using a performance evaluation metric (such as error rate) that is bounded above, $a_2 < \infty$. The remaining contents (of Equation \ref{eq:unsl1}), i.e. the inner reciprocal $\left(Q(S+4) + {a_1}^{-1}\right)^{-1}$, correspond to the ``oppositional force'' exerted by overfitting. When one trains a model for more than one epoch, this inner reciprocal becomes a non-negligible number that is considerably larger than zero.


\subsection{The Additive Symmetry}
\label{symmetry}


\FloatBarrier

\begin{figure*}[htpb]    \centering

\raisebox{0mm}{\includegraphics[width=0.33\textwidth]{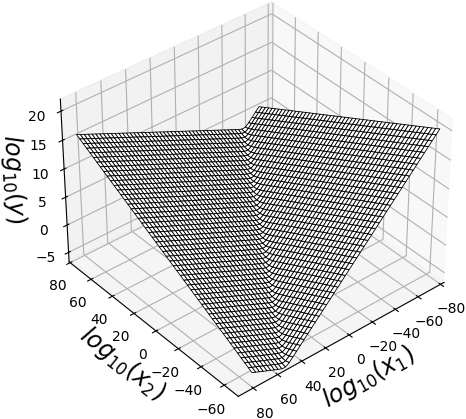}}
\hspace*{.3mm}\raisebox{9.51mm}{\includegraphics[width=0.661\textwidth]{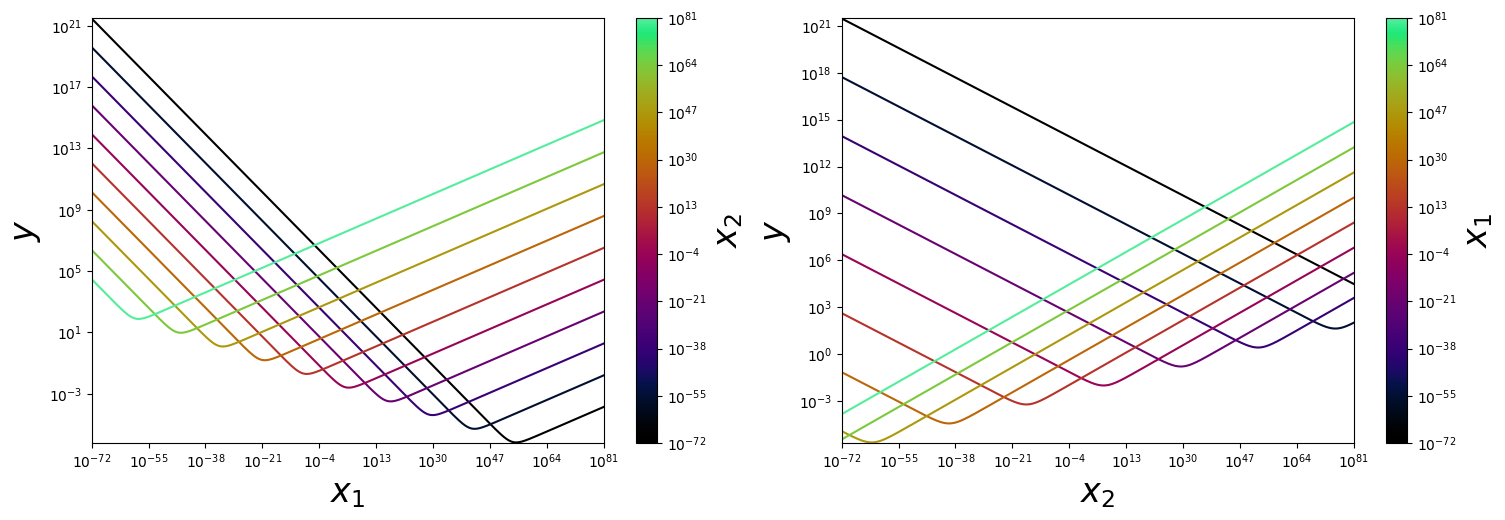}}
    \caption{An illustration of an example configuration of Equation \ref{eq:mbnsl_additive} with two input dimensions, $x_1$ and $x_2$. All 3 plots are of the same scaling behavior. See Section \ref{symmetry} for more details.}
    \label{fig:figure_2}
\end{figure*}

The following expression\footnote{In Equation \ref{eq:mbnsl_additive},  $b$, $c_{i_0}$, $g$, and $h_{i}$ are constants estimated by fitting Equation \ref{eq:mbnsl_additive} to ($x_1 ... x_m, y$) data points.} implicitly shows up in several places (when an addition takes place) in Equations \ref{eq:unsl1}, \ref{eq:unsl2}, and \ref{eq:unsl3}:
\begin{equation}
\hspace{-18.5mm} y = b \cdot \left(\prod_{\substack{i=1}}^{m} x_i^{-c_{i_{0}}}\right) + g \cdot \left(\prod_{\substack{i=1}}^{m} x_i^{h_{i}}\right),
\label{eq:mbnsl_additive}
\end{equation}

and is equivalent to a ($n=1, M=\{1,\ldots,m\}$) version of Equation \ref{eq:mbnsl}:
\begin{equation}
y = b \cdot \left(\prod_{\substack{i=1}}^{m} x_i^{-c_{i_{0}}}\right)
 \left( 1 + \left( \frac{\prod_{\substack{i=1}}^{m} x_i^{c_{i_{1}}}}{d} \right)^{\left|\frac{1}{f}\right|} \right)^{-f},
\label{eq:mbnsl_single_break}
\end{equation}

when all these equalities are simultaneously true:
\begin{align*}
f &= -1\hspace{2mm},      \hspace{1.1cm}
c_{i_{1}} = c_{i_{0}} + h_{i}\hspace{2mm},      \hspace{1.1cm}
d = b/g \hspace{2mm}.
\end{align*}

Equation \ref{eq:mbnsl_additive} is different from Equation \ref{eq:mbnsl_single_break} in that (assuming $b$, $g$, and $d$ are positive numbers):

\begin{enumerate}[leftmargin=2em, itemsep=1.5mm, parsep=0pt]

\item For Equation \ref{eq:mbnsl_additive}, the change in gradient (with respect to the input dimensions $x_{1},...,x_{m}$ as any $x_{i}$ increases) between the 1st hyperplane and the 2nd hyperplane in multi-log space is always nonnegative; meanwhile, for Equation \ref{eq:mbnsl_single_break}, this change in gradient can be any amount.

\item For Equation \ref{eq:mbnsl_additive}, the sharpness of the hyperbreak between the 1st and the 2nd hyperplane in multi-log space is dependent solely on the amount of change in gradient between the 1st hyperplane and the 2nd hyperplane in multi-log space; meanwhile, for Equation \ref{eq:mbnsl_single_break}, this sharpness is dependent on the value of $f$ (and as a result is decoupled from the amount of change in gradient between the 1st hyperplane and the 2nd hyperplane in multi-log space).

\end{enumerate}

Empirically, we observe that nonmonotonic transitions always seem to be characterized by Equation \ref{eq:mbnsl_additive} rather than \ref{eq:mbnsl_single_break}. As a result, (when an addition takes place in the center) in Equations \ref{eq:unsl1} and \ref{eq:unsl2}, we implicitly use Equation \ref{eq:mbnsl_additive} to model phenomena (e.g. overfitting and hyperparameters such as learning rate and standard deviation of weights at initialization) that are capable of exhibiting a nonmonotonic relationship with the performance evaluation metric.

Empirically, we observe that transitions to or from regions in which the gradient (with respect to at least one of the input dimensions $x_{1},...,x_{m}$) is equal to zero always seem to be characterized by a version of Equation \ref{eq:mbnsl_additive} in which each $h_{i}$ (in $h_{1},...,h_{m}$) for which the gradient with respect to $x_{i}$ (in $x_{1},...,x_{m}$) is equal to zero is equal to zero. As a result, we implicitly use that version of Equation \ref{eq:mbnsl_additive} when addition takes place in Equation \ref{eq:unsl3} and when addition takes place with a parameter whose base is $a$ in parts of Equations \ref{eq:unsl1} and \ref{eq:unsl2}.

Note that Equation \ref{eq:mbnsl_additive} sums two $n=0$ versions of MBNSL of Equation \ref{eq:mbnsl}. To extend the relations discussed in this Section \ref{symmetry} thus far to a summation of two MBNSLs that each have an arbitrary number of hyperbreaks $n$, see Appendix \ref{section:additive_symmetry_extension}.

\subsection{Desiderata}
\label{section:desiderata}

The UNSL functional form satisfies all of the following desiderata:

\begin{enumerate}[leftmargin=2em, itemsep=1.5mm, parsep=0pt]
    \item \label{item:1} Each univariate scaling behavior is a univariate \textit{broken neural scaling law (BNSL)} of \citet{caballero2023broken}. This means that (as discussed in Section \ref{symmetry}) for a significant subset of transitions between consecutive hyperplanes (in multi-log space) the sharpness needs to be decoupled from the amount of change in gradient (i.e. via the extra expressivity granted by $f$ in Equation \ref{eq:mbnsl_single_break} (and Equation \ref{eq:mbnsl})).
    \item \label{item:2} The position of break(s) (within univariate scaling behaviors) within hyperbreak(s) created by non-bottleneck components are shifted via multiplication in a way that is dependent on other input dimensions.
    \item \label{item:3} Whenever all but one $x_{i}$ dimension in $x_{1}...x_{m}$ simultaneously go to the values of $(x_i)_{i=1}^m \in \overline{\mathbb{R}}_{>0}^{\,m}$ that yield the global optimum of $y$, that performance limit is dependent on the value of that single $x_{i}$ dimension (that is bottlenecking performance) and no other dimension in $x_{1}...x_{m}$.
    When sufficiently close to the global optimum of $y$, the transition to that performance limit is characterized by the functional form $y = a + \sum_{t \in T_r} b_t \cdot x_t^{-c_{t}}$.
    \item \label{item:4} The performance limit as all $x_{i}$ dimensions in $x_{1}...x_{m}$ simultaneously go to the values of $(x_i)_{i=1}^m \in \overline{\mathbb{R}}_{>0}^{\,m}$ that yield the global optimum of $y$ is dependent on a constant (e.g. the irreducible entropy or Bayes error). The transition to this performance limit is characterized by summing an entire functional form with a constant (e.g. $a_0$).
    \item \label{item:5} The misperformance limit (e.g. upper limits when using metrics such as error or loss for which a lower value of that metric is considered better) when the amount of misperformance is not bottlenecked by any $x_i$ in $x_1 ... x_m$ is dependent on a constant. The transition to this misperformance limit is characterized by raising to the -1 power the sum of a functional form and a constant. Examples of such misperformance limits in some scenarios are the loss or error of a random guessing (maximum entropy) model and in other scenarios are a value much larger than the loss or error of a random guessing (maximum entropy) model.
    \item \label{item:6} Whenever all but one $x_{i}$ dimension in $x_{1}...x_{m}$ simultaneously go to the values of $(x_i)_{i=1}^m \in \overline{\mathbb{R}}_{>0}^{\,m}$ that yield the globally worst value of $y$, that misperformance limit (e.g. upper limits when using metrics such as error or loss for which a lower value of that metric is considered better) is dependent on the value of that single $x_{i}$ dimension (that is bottlenecking misperformance) and no other dimension in $x_{1}...x_{m}$.
    When sufficiently far from the global optimum of $y$, the transition to that misperformance limit is characterized by the functional form $y = \left(a + \sum_{t \in T_r} b_t \cdot x_t^{-c_{t}}\right)^{-1}$. Examples of such misperformance limits are the high loss or error obtained when training dataset size is too small (i.e. such that overfitting occurs).
    \item \label{item:7} Nonmonotonic transitions (e.g. due to overfitting and hyperparameters such as learning rate and standard deviation of weights at initialization) are characterized by the additive symmetry discussed in Section \ref{symmetry}.
    \item \label{item:8} The ``oppositional forces'' of hyperparameters oppose ``good learning'' (i.e. the subset of learning that is not considered to be overfitting) and ``bad learning'' (i.e. the subset of learning that is considered to be overfitting).
\end{enumerate}

\textbf{See Appendix \ref{section:desiderata_satisfaction_explanation} for explanation of how UNSL functional form satisfies all of these desiderata. \\ See Appendix \ref{section:empirical_desiderata} for evidence that all of these desiderata are empirically true.}

\section{Related Work}
\label{section:related_work}
To the best of our knowledge, \citet{DBLP:journals/corr/abs-1909-12673} was the first to describe a functional form with multivariate input; this functional form is $y = a + b_1{x_1}^{-c_1} + b_2{x_2}^{-c_2}$ in which $x_1$ is number of model parameters and $x_2$ is training dataset size. \citet{icm2020arXiv200108361K} (and others such as \citet{hoffmann2022training}) used this same functional form, but had $x_2$ be number of training steps multiplied by training batch size; we refer to this functional form as ``CF''.

\citet{muennighoff2023scaling} introduced this functional form (that we refer to as ``DC'') with trivariate input:

$y = a + b_1 \cdot (U_N + U_N \cdot d_1 \cdot (1 - e^{(-1 \cdot R_N/(d_1))}))^{-c_1} + b_2 \cdot (x_3 + x_3 \cdot d_2 \cdot (1 - e^{(-1 \cdot R_D/(d_2))}))^{-c_2}$  ;

in that functional form:

$R_D = \max(0, (x_2 / x_3) - 1)$ ,

$U_N = \min(x_1, (x_3 \cdot ((c_1 \cdot b_1)/(c_2 \cdot b_2))^{(1/(c_1+c_2))})^{(c_2/c_1)} \cdot ((c_1 \cdot b_1)/(c_2 \cdot b_2))^{(1/(c_1+c_2))})$ ,

$R_N = \max(0, (x_1 / U_N ) - 1)$ ,

$x_1$ is number of model parameters, $x_2$  is number of training steps multiplied by training batch size, and $x_3$ is training dataset size. When training dataset size is so large that one only trains for one epoch, functional form ``DC'' is mathematically identical to functional form ``CF''.

See Appendix \ref{sec:additional_related_work} for additional related work.

\newcommand{\framedtext}[1]{\par\noindent\fcolorbox{Magenta2}{white}{
    \parbox{\dimexpr\linewidth-2.18\fboxsep-2\fboxrule}{#1}}}


\section{Empirical Results: Fits \& Extrapolations of Functional Forms}
\label{section:functional_form_fits}

\begin{mdframed}[linecolor=Magenta2]

We now show the fits \& \textbf{extrapolations} of various functional forms. \textbf{In all plots here, onward, \& in the appendix, triangle-shaped points are points used for fitting a functional form, circle-shaped points are held-out points used for evaluating extrapolation of functional form fit to the triangle-shaped points, \& lines are the functional form that has been fit to triangle-shaped points. The color of each line and (the inside of) each point represents its value along the color bar dimension. Lines of the functional form are intentionally only rendered at the values of the color bar dimension for which there exists at least one (triangle-shaped or circle-shaped) point; this means that the vertical distance of each point from the line (that is the same color as that point) represents the error of the functional form when fitting (or extrapolating to) that point. 100\% of the plots in this paper here, onward, \& in the appendix contain circle-shaped point(s) for evaluating extrapolation.}

\end{mdframed}

See Appendix \ref{section:fitting_unsl} for details on fitting UNSL. 
See Appendix \ref{sec:implementation} for code that implements UNSL.
See Appendix \ref{section:num_point} for an analysis of how the number of observed points used for fitting affects extrapolation accuracy.
See Appendix \ref{section:oom_extrap} for an example of UNSL accurately extrapolating to scales an order of magnitude larger in multiple dimensions simultaneously. See Appendix \ref{section:compute_optimal} for how to obtain the compute-optimal values of the input dimensions from a fitted UNSL.

All the extrapolation evaluations reported in the tables (that have {\small$\downarrow$} symbol in the top row) are reported in terms of root mean squared log error (RMSLE) ± root standard log error. See Appendix \ref{section:definition_of_Root_Mean_Squared_Log_Error} for definition of RMSLE and Appendix \ref{section:definition_of_Root_Standard_Log_Error} for definition of root standard log error.

\subsubsection{Ablation functional forms}
\label{section:ablation_forms}

A1 functional form refers to the baseline ablation functional form in which all the additive symmetries discussed in Section \ref{symmetry} have been removed such that this A1 baseline functional form is Equation \ref{eq:mbnsl}, i.e. :
$$
y = b \cdot \left(\prod_{i=1}^{m} x_i^{-c_{i_{0}}}\right) \prod_{j=1}^n \left(1 + \left(\frac{\prod_{i=1}^m x_i^{c_{i_{j}}}}{d_j}\right)^{\left|\frac{1}{f_j}\right|}\right)^{-f_j} .
$$

A2 functional form refers to the baseline ablation functional form that consists solely of Equation \ref{eq:unsl3} (which consists of Equation \ref{eq:mbnsl}) plus the constant $a_0$ (which corresponds to the limit as to how far the value of $y$ can be reduced (or maximized) even if all of $x_1 ... x_m$ go to the values of $(x_i)_{i=1}^m \in \overline{\mathbb{R}}_{>0}^{\,m}$ that yield the global optimum of $y$):
$$
y = a_0 + K( \, U_0,\,\, n_{0_0},\,\, 0) \, + \, \sum_{\mathclap{ \kern 0.9mm t \in T_0}} K(\{t\},\,\, n_{0_t},\,\, t),  \hspace{3.8mm} \text{where } U_0,T_0\subseteq\{1,\ldots,m\}.
$$

A2 functional form incorporates more of the additive symmetries discussed in Section \ref{symmetry} than A1 functional form does.

A3 functional form refers to the baseline ablation functional form that consists solely of Equation \ref{eq:unsl2} (which consists of Equation \ref{eq:unsl3} (which consists of Equation \ref{eq:mbnsl})) plus the constant $a_0$ (which corresponds to the limit as to how far the value of $y$ can be reduced (or maximized) even if all of $x_1 ... x_m$ go to the values of $(x_i)_{i=1}^m \in \overline{\mathbb{R}}_{>0}^{\,m}$ that yield the global optimum of $y$):
$$
y = a_0 + \left(\left(\left(\left(R(0)\right)^{-1} + {a_1}^{-1}\right)^{-1}+ \sum_{s=1}^{S} \left(R(s) + {a_{s+2}}^{-1}\right)^{-1}\right)^{-1} + a_2^{-1}\right)^{-1} .
$$

A3 functional form incorporates more of the additive symmetries discussed in Section \ref{symmetry} than A2 functional form does. UNSL functional form incorporates all of the additive symmetries discussed in Section \ref{symmetry} (i.e. more than A3 functional form does).


\subsubsection{Summary of Results}
\label{section:results_summary}

\begin{table}[hbt!]
  \centering
  \footnotesize
      \begin{tabular}{ |c|c|c|c|c|c|c| }
\hline
Domain & CF $\uparrow$ & DC $\uparrow$ & A1 $\uparrow$ & A2 $\uparrow$ & A3 $\uparrow$ & UNSL $\uparrow$\\
\hline
Downstream Image Classification & 0.00\% & 0.00\% & 8.70\% & 8.70\% & 21.74\% & \bfseries 60.87\%\\
Language (Downstream \& Upstream) & 0.00\% & 0.00\% & 0.00\% & 11.11\% & 0.00\% & \bfseries 88.89\%\\
\hline
\end{tabular}
  \caption{
  Percentage of tasks by domain where each functional form is the best for \textbf{extrapolation} of scaling behavior.  See Sections \ref{section:vision} and \ref{section:language} for more details.
  }
\label{table:scaling_laws_benchmark_dataset__summary}
\end{table}


A1, A2, A3, and UNSL all have the exact same supremal expressivity. As a result, the fact that UNSL is better for extrapolation than A1, A2, and A3 in Table \ref{table:scaling_laws_benchmark_dataset__summary} is due to the fact that UNSL enforces more of the desiderata (of Section \ref{section:desiderata}) (e.g., via incorporating all of the symmetries discussed in Section \ref{symmetry}) than A1, A2, and A3 do.

\FloatBarrier

\begin{figure*}[h]    \centering
\includegraphics[width=0.486\textwidth]{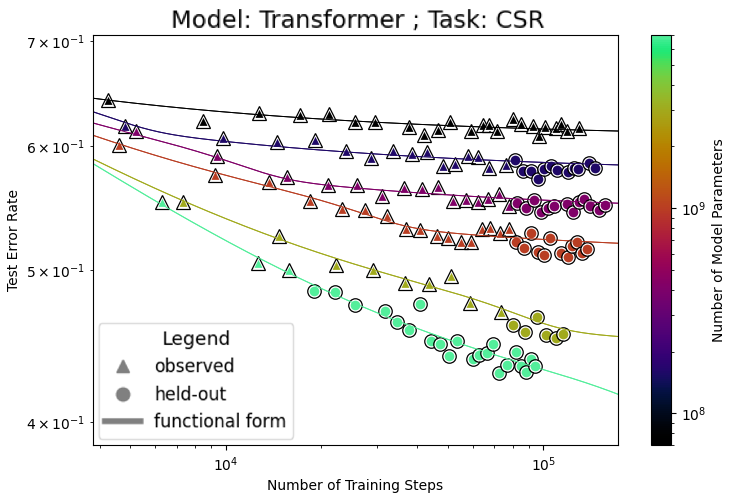}
\includegraphics[width=0.504\textwidth]
{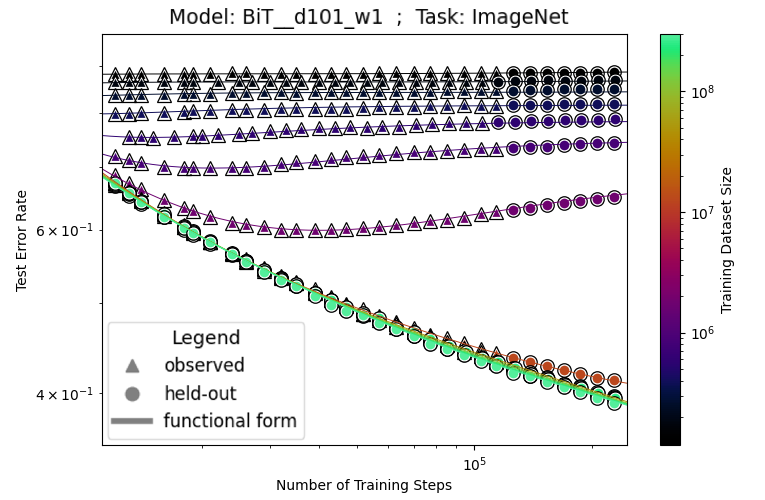}
    \caption{
    UNSL accurately Extrapolating Downstream Performance; there are many additional accurate extrapolation results in Appendix \ref{section:extrapolation_benchmark}. Experimental data of scaling behavior in left plot is downstream performance on CSR (Common Sense Reasoning), i.e. downstream zero-shot mean test error rate across HellaSwag, ARC (easy and challenge), PIQA, WinoGrande, OpenBookQA, SIQA, and BoolQ; see Section \ref{section:language} for more details. Experimental data of scaling behavior in right plot is few-shot downstream performance on ImageNet; see Section \ref{section:vision} for more details.
    }
    \label{fig:extrapolation_examples}
\end{figure*}

\FloatBarrier


\subsection{Vision}
\label{section:vision}


We evaluate how well various functional forms extrapolate performance on downstream vision tasks as multiple dimensions vary simultaneously. The tasks that are evaluated are test error rate on each of various few-shot downstream image classification tasks; the downstream tasks are: Birds 200 \citep{welinder2010caltech}, Cars 196 \citep{krause20133d}, and ImageNet \citep{deng2009imagenet}. The following architectures of various sizes are pre-trained on subsets of JFT-300M \citep{sun2017revisiting}: vision transformers (ViT) \citep{dosovitskiy2020image}, MLP mixers (MiX) \citep{tolstikhin2021mlp}, and big-transfer residual neural networks (BiT) \citep{kolesnikov2020big}. The bivariate subset of this scaling behavior data is obtained via correspondence with authors of \cite{Alabdulmohsi2022revisiting}; the simultaneously varying dimensions of the bivariate scaling behavior are training dataset size and number of training steps. The trivariate subset of this scaling behavior data is obtained from the ViT/16 results of \cite{zhai2022scaling}; the simultaneously varying dimensions of the trivariate scaling behavior are training dataset size, number of training steps, and number of model parameters. As can be seen in Tables \ref{table:scaling_laws_benchmark_dataset__summary},  \ref{table:downstream_vision}, and \ref{table:downstream_vision_trivariate}, UNSL yields extrapolations with the lowest RMSLE (Root Mean Squared Logarithmic Error) for 60.87\% of tasks of any of the functional forms, while the next best functional form performs the best on only 21.74\% of the tasks. To view plots of UNSL, DC, A1, A2, and A3 on each of these bivariate scaling behaviors, in Appendix \ref{subsection:extrapolation_benchmark_vision} respectively see Figures \ref{fig:unsl_downstream_vision}, \ref{fig:dc_downstream_vision}, \ref{fig:a1_downstream_vision}, \ref{fig:a2_downstream_vision}, \ref{fig:a3_downstream_vision}.
To view plots of UNSL, DC, A1, A2, and A3 on each of these trivariate scaling behaviors, in Appendix \ref{subsection:extrapolation_benchmark_vision} respectively see Figures \ref{fig:unsl_trivariate_vision}, \ref{fig:dc_trivariate_vision},  \ref{fig:a1_trivariate_vision},  \ref{fig:a2_trivariate_vision}, \ref{fig:a3_trivariate_vision}.

In Appendix \ref{subsection:extrapolation_reinforcement_learning}, we additionally show that UNSL accurately extrapolates the multivariate scaling behavior of reinforcement learning.

\vspace{-1.6mm}

In Appendix \ref{subsection:extrapolation_width_depth}, we additionally show that UNSL accurately extrapolates multivariate scaling behavior as width and depth vary simultaneously.

\vspace{-1.6mm}

In Appendix \ref{subsection:extrapolation_batch_size}, we additionally show that UNSL accurately extrapolates multivariate scaling behavior when batch size is an input dimension to UNSL.

\vspace{-1.6mm}

In Figure \ref{fig:unsl_additive_symmetry} of Appendix \ref{subsection:empirical_desiderata_7}, we additionally show that UNSL accurately extrapolates the trivariate scaling behavior as learning rate, standard deviation of weights at initialization, and number of training steps all vary simultaneously.

\begin{table}[hbt!]
    \centering
    \scriptsize
        \begin{tabular}{|c|c|c|c|c|c|}
    \hline
    Task & DC $\downarrow$ & A1 $\downarrow$ & A2 $\downarrow$ & A3 $\downarrow$ & UNSL $\downarrow$ \\
    \hline
        Birds   & 2.65e-1 ± 2.49e-2 & 7.38e-2 ± 1.41e-2 & 6.51e-2 ± 1.79e-2 & 4.77e-2 ± 7.51e-3 & \textbf{4.03e-2 ± 5.51e-3} \\
    Imagenet   & 2.54e-1 ± 2.02e-2 & 4.61e-2 ± 1.01e-2 & 3.39e-2 ± 1.06e-2 & 2.20e-2 ± 3.63e-3 & \textbf{1.70e-2 ± 2.76e-3} \\
    \hline
    \end{tabular}
    \caption{Extrapolation Results for trivariate scaling behavior of downstream vision performance. See Section \ref{section:vision} for more details.}
    \label{table:downstream_vision_trivariate}
\end{table}

\begin{table}[hbt!]
\centering
\scriptsize
\begin{tabular}{|c c|c|c|c|c|c|}
\hline
Task & Model & DC $\downarrow$ & A1 $\downarrow$ & A2 $\downarrow$ & A3 $\downarrow$ & UNSL $\downarrow$ \\
\hline
Birds    & BiT/d101/w3 & 3.97e-1 ± 9.84e-3 & 1.80e-2 ± 1.16e-3 & 2.12e-2 ± 1.43e-3 & 1.49e-2 ± 1.24e-3 & \textbf{1.41e-2 ± 1.11e-3} \\
Birds    & BiT/d50/w3  & 4.45e-1 ± 1.16e-2 & 4.29e-2 ± 2.94e-3 & 1.67e-2 ± 1.13e-3 & \textbf{1.46e-2 ± 1.22e-3} & 1.47e-2 ± 1.07e-3 \\
Birds    & MiX/L/16    & 4.92e-1 ± 1.15e-2 & \textbf{1.39e-2 ± 8.46e-4} & 1.72e-2 ± 1.05e-3 & 2.47e-2 ± 1.93e-3 & 2.62e-2 ± 1.56e-3 \\
Birds    & MiX/B/16    & 3.91e-1 ± 9.73e-3 & 2.12e-2 ± 1.88e-3 & 2.05e-2 ± 1.44e-3 & 1.90e-2 ± 1.36e-3 & \textbf{1.89e-2 ± 1.23e-3} \\
Birds    & BiT/d50/w1  & 3.50e-1 ± 9.39e-3 & \textbf{1.16e-2 ± 7.67e-4} & 1.36e-2 ± 1.03e-3 & 2.05e-2 ± 1.76e-3 & 1.59e-2 ± 1.38e-3 \\
Birds    & ViT/B/16    & 3.40e-1 ± 8.03e-3 & 5.86e-2 ± 6.60e-3 & 3.08e-2 ± 1.57e-3 & 2.98e-2 ± 1.72e-3 & \textbf{2.27e-2 ± 1.38e-3} \\
Birds    & BiT/d101/w1 & 3.97e-1 ± 9.84e-3 & 1.80e-2 ± 1.16e-3 & 1.29e-2 ± 9.28e-4 & 1.32e-2 ± 1.04e-3 & \textbf{1.21e-2 ± 9.10e-4} \\
Cars     & MiX/L/16    & 6.23e-1 ± 1.36e-2 & 5.83e-2 ± 5.45e-3 & 4.54e-2 ± 2.49e-3 & 2.15e-2 ± 2.23e-3 & \textbf{2.13e-2 ± 2.13e-3} \\
Cars     & MiX/B/16    & 7.05e-1 ± 1.42e-2 & 3.96e-2 ± 2.42e-3 & 2.46e-2 ± 2.15e-3 & \textbf{2.30e-2 ± 1.83e-3} & 3.31e-2 ± 3.26e-3 \\
Cars     & ViT/B/16    & 1.05e+0 ± 1.64e-2 & 1.36e-1 ± 9.15e-3 & 8.74e-2 ± 4.71e-3 & 4.77e-2 ± 3.14e-3 & \textbf{2.74e-2 ± 1.73e-3} \\
Cars     & BiT/d101/w3 & 3.03e-1 ± 7.80e-3 & 2.24e-2 ± 1.61e-3 & 2.12e-2 ± 1.43e-3 & \textbf{1.75e-2 ± 1.44e-3} & 1.91e-2 ± 1.35e-3 \\
Cars     & BiT/d101/w1 & 5.91e-1 ± 1.02e-2 & 3.89e-2 ± 1.97e-3 & 2.77e-2 ± 1.68e-3 & 2.46e-2 ± 1.66e-3 & \textbf{2.33e-2 ± 1.69e-3} \\
Cars     & BiT/d50/w3  & 3.87e-1 ± 1.29e-2 & 2.66e-2 ± 2.05e-3 & \textbf{2.55e-2 ± 2.00e-3} & 3.13e-2 ± 2.46e-3 & 2.87e-2 ± 2.29e-3 \\
Cars     & BiT/d50/w1  & 6.71e-1 ± 1.32e-2 & 1.99e-2 ± 1.45e-3 & \textbf{1.93e-2 ± 1.28e-3} & 2.51e-2 ± 1.90e-3 & 2.83e-2 ± 2.18e-3 \\
Imagenet & MiX/L/16    & 4.30e-1 ± 9.59e-3 & 7.81e-3 ± 5.69e-4 & 1.13e-2 ± 8.20e-4 & 1.49e-2 ± 1.33e-3 & \textbf{7.23e-3 ± 6.44e-4} \\
Imagenet & BiT/d101/w1 & 2.50e-1 ± 6.00e-3 & 9.52e-3 ± 8.50e-4 & 4.77e-3 ± 3.07e-4 & 5.97e-3 ± 5.66e-4 & \textbf{3.74e-3 ± 2.45e-4} \\
Imagenet & BiT/d50/w1  & 2.17e-1 ± 5.36e-3 & 7.77e-3 ± 4.45e-4 & 4.40e-3 ± 2.47e-4 & 3.73e-3 ± 3.04e-4 & \textbf{2.61e-3 ± 2.58e-4} \\
Imagenet & ViT/B/16    & 3.69e-1 ± 8.98e-3 & 1.41e-2 ± 1.05e-3 & 1.04e-2 ± 9.16e-4 & 1.43e-2 ± 1.02e-3 & \textbf{8.57e-3 ± 7.78e-4} \\
Imagenet & MiX/B/16    & 3.21e-1 ± 8.18e-3 & 1.06e-2 ± 1.21e-3 & 7.86e-3 ± 4.83e-4 & 5.07e-3 ± 4.97e-4 & \textbf{3.35e-3 ± 2.56e-4} \\
Imagenet & BiT/d101/w3 & 3.26e-1 ± 8.10e-3 & 5.44e-3 ± 4.89e-4 & 1.00e-2 ± 8.45e-4 & \textbf{4.17e-3 ± 2.93e-4} & 5.31e-3 ± 4.71e-4 \\
Imagenet & BiT/d50/w3  & 3.09e-1 ± 8.07e-3 & 2.95e-2 ± 1.69e-3 & 6.19e-3 ± 3.91e-4 & \textbf{3.79e-3 ± 2.50e-4} & 4.20e-3 ± 2.40e-4 \\
\hline
\end{tabular}
\caption{Extrapolation Results for bivariate scaling behavior of downstream vision performance. See Section \ref{section:vision} for more details.}
\label{table:downstream_vision}
\end{table}


\subsection{Language}
\label{section:language}


We evaluate how well various functional forms extrapolate performance on downstream (and upstream) language tasks as multiple dimensions vary simultaneously. As can be seen in Tables \ref{table:scaling_laws_benchmark_dataset__summary}, \ref{table:llm_trivariate}, and \ref{table:llm_bivariate}, UNSL yields extrapolations with the lowest RMSLE (Root Mean Squared Logarithmic Error) for 88.89\% of tasks of any of the functional forms, while the next best functional form performs the best on only 11.11\% of the tasks. To view plots of UNSL, DC, A1, A2, and A3 on trivariate scaling behavior, in Appendix \ref{subsubsection:extrapolation_benchmark_language_trivariate} respectively see Figures \ref{fig:unsl_llm_trivariate}, \ref{fig:dc_llm_trivariate}, \ref{fig:a1_llm_trivariate}, \ref{fig:a2_llm_trivariate}, \ref{fig:a3_llm_trivariate}; this trivariate scaling behavior data is from scaling behavior data released by \citet{muennighoff2023scaling}, and the simultaneously varying dimensions of these trivariate scaling behaviors are number of model parameters, number of tokens processed, and number of tokens in training dataset. To view plots of UNSL, CF, A1, and A2 on each of these bivariate scaling behaviors, in Appendix \ref{subsubsection:extrapolation_benchmark_language_bivariate} respectively see Figures \ref{fig:unsl_llm_bivariate}, \ref{fig:cf_llm_bivariate}, \ref{fig:a1_llm_bivariate}, and \ref{fig:a2_llm_bivariate}; the simultaneously varying dimensions of these bivariate scaling behaviors are number of model parameters and number of training steps (or number of tokens processed). There is no A3 in Table \ref{table:llm_bivariate} because UNSL becomes A3 in the scenario of Table \ref{table:llm_bivariate}, i.e. the scenario in which training dataset size is effectively infinite such that one only trains for one epoch. The bivariate scaling behaviors that are referred to as "constant" are obtained from the LLaMA and HGRN2 portions of Figures 1 and 2 of \cite{shen2024scaling}; they are referred to as "constant" because the learning rate is held constant and a learning rate schedule is not used. The bivariate scaling behaviors that are referred to as "chinchilla" are obtained via correspondence with authors of \cite{hoffmann2022training}; they are called "chinchilla" because they use "chinchilla-scaling" (i.e. a learning rate schedule that is chosen to be training compute optimal as in \cite{hoffmann2022training}) and are the scaling behavior data from \cite{hoffmann2022training}. CSR (Common Sense Reasoning) is zero-shot mean test error rate across HellaSwag \citep{zellers2019hellaswag}, ARC (easy and challenge) \citep{clark2018arc}, PIQA \citep{bisk2019piqa}, WinoGrande \citep{sakaguchi2019winogrande}, OpenBookQA \citep{mihaylov2018openbookqa}, SIQA \citep{sap2019siqa}, and BoolQ \citep{clark2019boolq}.

In Appendix \ref{subsection:extrapolation_inference_scaling}, we additionally show that UNSL accurately extrapolates the multivariate scaling behavior of inference (i.e. test-time) scaling.

\begin{table}[hbt!]
    \centering
    \footnotesize
        \begin{tabular}{ |c|c|c|c|c| }
\hline
DC $\downarrow$ & A1 $\downarrow$ & A2 $\downarrow$ & A3 $\downarrow$ & UNSL $\downarrow$\\
 \hline
6.24e-2 ± 6.00e-3 & 2.00e-2 ± 1.90e-3 & 1.96e-2 ± 3.62e-3 & 1.49e-2 ± 3.45e-3 & \bfseries 7.82e-3 ± 1.33e-3\\
 \hline
\end{tabular}
    \caption{
    Extrapolation Results for trivariate scaling behavior of language performance. See Section \ref{section:language} for more details.
    }
    \label{table:llm_trivariate}
\end{table}

\begin{table}[hbt!]
    \centering
    \scriptsize
    \begin{tabular}{ |ccc|c|c|c|c|c| }
    \hline
Task & Model & Scaling & CF $\downarrow$ & A1 $\downarrow$ & A2 $\downarrow$ & UNSL $\downarrow$\\
 \hline

Upstream & Transformer & Chinchilla & 1.72e-2 ± 1.69e-3 & 9.85e-3 ± 1.30e-3 & 4.43e-3 ± 6.15e-4 & \bfseries 3.81e-3 ± 6.52e-4 \\ LAMBADA & Transformer & Chinchilla & 2.08e-2 ± 2.48e-3 & 1.45e-2 ± 1.89e-3 & 1.30e-2 ± 1.80e-3 & \bfseries 1.13e-2 ± 1.60e-3 \\

CSR & Transformer & Constant & 4.50e-2 ± 3.72e-3 & 1.43e-2 ± 1.24e-3 & 1.66e-2 ± 1.08e-3 & \bfseries 1.28e-2 ± 1.05e-3 \\
LAMBADA & Transformer & Constant & 3.06e-2 ± 3.92e-3 & 4.17e-2 ± 3.52e-3 & 3.12e-2 ± 2.48e-3 & \bfseries 2.24e-2 ± 1.71e-3 \\
Upstream & Transformer & Constant & 7.15e-2 ± 5.03e-3 & 3.98e-2 ± 3.27e-3 & 2.89e-2 ± 1.58e-3 & \bfseries 7.95e-3 ± 6.63e-4 \\

CSR & Recurrent & Constant & 5.20e-2 ± 3.32e-3 & 2.65e-1 ± 2.87e-2 & \bfseries 1.15e-2 ± 9.39e-4 & 1.22e-2 ± 9.15e-4 \\
LAMBADA & Recurrent & Constant & 3.02e-2 ± 2.63e-3 & 3.75e-2 ± 2.60e-3 & 4.31e-2 ± 3.62e-3 & \bfseries 1.66e-2 ± 1.38e-3 \\
Upstream & Recurrent & Constant & 3.13e-2 ± 2.36e-3 & 3.07e-2 ± 1.99e-3 & 1.92e-2 ± 1.63e-3 & \bfseries 4.66e-3 ± 3.51e-4 \\

 \hline
\end{tabular}
    \caption{
    Extrapolation Results for bivariate scaling behavior of downstream (and upstream) language performance. See Section \ref{section:language} for more details.
    }
    \label{table:llm_bivariate}
\end{table}

\FloatBarrier

\FloatBarrier

\section{The Limit of the Predictability of Scaling Behavior}
\label{section:limit}

\FloatBarrier

\begin{figure*}[h]    \centering
\includegraphics[width=0.495\textwidth]{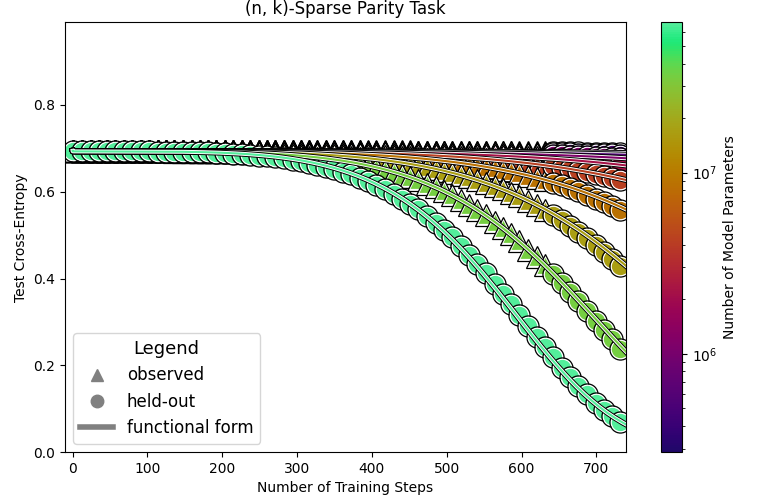}
\includegraphics[width=0.495\textwidth]{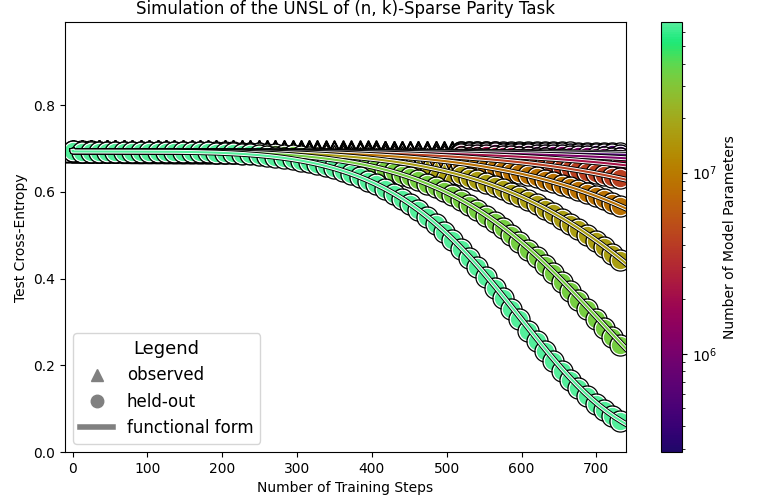}
    \caption{
    Extrapolation of UNSL on scaling behavior of an MLP trained for a single epoch on the (n, k)-sparse parity task (with $n=40$ and $k=4$) of \cite{barak2022hidden}. Each point in the left plot is the mean of greater than 100 seeds. In the left plot, each point is gathered from an MLP trained for a single epoch on the (n, k)-sparse parity task (with $n=40$ and $k=4$) of \cite{barak2022hidden}. In the right plot, each point is gathered from a noiseless simulation of the UNSL of the scaling behavior of that (n, k)-sparse parity task.  See Section \ref{section:limit} and Appendix \ref{section:experimental_details} for more details.
    }
    \label{fig:limit_of_predictability}
\end{figure*}

\FloatBarrier

We use UNSL to glean insights about the limit of the predictability of scaling behavior. In Figure \ref{fig:limit_of_predictability} left, UNSL accurately extrapolates the scaling behavior of the sparse parity task of \cite{barak2022hidden}, despite the fact that this task famously does not exhibit any observable progress in loss (nor error) for the first few hundred training steps. In Figure \ref{fig:limit_of_predictability} right, we use a noiseless simulation of the UNSL of the scaling behavior of the sparse parity task to show what would happen if one had infinitely many training runs / seeds to average out all the noisy deviation between runs such that one could recover (i.e. learn via curve-fitting) the learned constants of the UNSL as well as possible. We observe the following:


\begin{itemize}
    \item To accurately extrapolate past each hyperbreak, the shortest distance to each hyperbreak from (the convex hull of) the points used for fitting must be sufficiently small.
\end{itemize}


\section{Discussion}
\label{section:discussion}


We have presented the unified neural scaling law (UNSL) functional form that accurately models and extrapolates the scaling behaviors of deep neural networks as multiple dimensions all vary simultaneously (i.e. how the evaluation metric of interest varies as one simultaneously varies the number of model parameters, training dataset size, number of training steps, number of inference steps, and various hyperparameters) for various architectures and for each of various tasks within a varied set of upstream and downstream tasks. When compared to other functional forms for neural scaling, this functional form yields \textbf{extrapolations} of scaling behavior that are considerably more accurate on this set.


\subsubsection*{Acknowledgments}


We are thankful for useful feedback and assistance from Ben Adlam, Ibrahim Alabdulmohsin, Sebastian Borgeaud, Kevin Clark and others.

\newpage

\section*{Appendix}

\section{Extension of the Additive Symmetry relations discussed in Section \ref{symmetry} to a summation of two MBNSLs that each have an arbitrary number of hyperbreaks $n$}
\label{section:additive_symmetry_extension}

Note that Equation \ref{eq:mbnsl_additive} sums two $n=0$ versions of MBNSL of Equation \ref{eq:mbnsl}. To extend the relations discussed in Section \ref{symmetry} to a summation of two MBNSLs that each have an arbitrary number of hyperbreaks $n$, for each of those two MBNSLs one needs to obtain the $n=0$ version $(w_b \prod_{\substack{i=1}}^{m} x_i^{w_{c_{i}}})$ of MBNSL of Equation \ref{eq:mbnsl} that is the tangent hyperplane in multi-log space. The values of $w_b$ and $w_{c_{i}}$ that yield the tangent hyperplane in multi-log space are:

$$
w_{c_{i}} = - c_{i_{0}} - \sum_{j=1}^n \operatorname{sign}(f_j) \cdot c_{i_{j}} \cdot \left(1 + \left(\frac{\prod_{l=1}^m x_l^{c_{l_{j}}}}{d_j}\right)^{-\left|\frac{1}{f_j}\right|}\right)^{-1} ,
$$

$$
w_b = b \cdot \left(\prod_{l=1}^{m} x_l^{-c_{l_{0}}}\right) \left(\prod_{j=1}^n \left(1 + \left(\frac{\prod_{l=1}^m x_l^{c_{l_{j}}}}{d_j}\right)^{\left|\frac{1}{f_j}\right|}\right)^{-f_j}\right) \prod_{i=1}^m x_i^{-w_{c_{i}}} .
$$

\section{Definition of Root Mean Squared Log Error}
\label{section:definition_of_Root_Mean_Squared_Log_Error}

\[Root\_Mean\_Squared\_Log\_Error = RMSLE = \sqrt{\frac{1}{N}\sum_{i=1}^{N}(\log(y_{i})-\log(\hat{y}_{i}))^2}\]

\section{Definition of Root Standard Log Error}
\label{section:definition_of_Root_Standard_Log_Error}

\[error_i = (\log(y_{i})-\log(\hat{y}_{i}))^2\]
\[\mu_{error} = \frac{1}{N}\sum_{i=1}^N error_i\]
\[\sigma_{error} = \sqrt{\frac{1}{N-1}\sum_{i=1}^N(error_i-\mu_{error})^2}\]
\[Root\_Standard\_Log\_Error = \sqrt{\mu_{error} + \frac{\sigma_{error}}{\sqrt{len(\hat{y})}}} - \sqrt{\mu_{error}}\]

\section{Experimental Details of Fitting UNSL}
\label{section:fitting_unsl}
We fit the UNSL by implementing it in KFAC-JAX \citep{kfac-jax2022github} and minimizing mean squared log error (MSLE):

\begin{equation}
MSLE = \frac{1}{N}  \sum_{i=1}^N (\log(y_i+\epsilon) - \log(\hat{y_i}+\epsilon))^2 ,
\label{eq:msle}
\end{equation}

with $\epsilon = 10^{-16}$. We also employ L2 regularization on the exponents of the UNSL with a weighting of $\lambda$ relative to the MSLE loss term.

The values of $n$ (from Equation \ref{eq:mbnsl}) (and $S$ from Equation \ref{eq:unsl2}) and $\lambda$ that yield the lowest extrapolation error can be obtained as follows.
Split the set of observed points (i.e. the triangle shaped points) used for fitting into two sets, a validation set and a training set; for each of every point in the validation set, the training set should not contain a point that is simultaneously larger than each of every $x$ dimension ($x_1 ... x_m$) of that validation set point. The values of $n$, $S$, and $\lambda$ with the lowest validation error when fitting on the remaining training points are then used. Note that once the values of $n$, $S$, and $\lambda$ are identified, the validation set is added back to the training set; (and the hold-out points (i.e. the circle shaped points) are still held out to evaluate extrapolation RMSLE). In practice, $S \le 1$ unless the scaling behavior of interest is an extravagant scaling behavior that is similar to the scaling behavior shown in Figure \ref{fig:multiple_upper_limits} of Appendix \ref{subsection:empirical_desiderata_5}.

It takes 20000 training steps and 20 seeds of random initialization for KFAC-JAX to converge when fitting a UNSL. We use the JAX default ``LeCun Normal'' initialization as the distribution from which each random initialization (for each seed) is drawn from for parameters of UNSL. Unlike the values of $n$ (from Equation \ref{eq:mbnsl}) (and $S$ from Equation \ref{eq:unsl2}) and $\lambda$, the optimal seed that is selected is that which yields the lowest training error (not the lowest validation error).

\section{Experimental Details of Sections \ref{section:limit}, \ref{section:num_point}, \ref{section:empirical_desiderata} (besides Figure \ref{fig:upper_limit_overfit}), and \ref{section:oom_extrap}}
\label{section:experimental_details}

For all figures in Sections \ref{section:limit}, \ref{section:num_point}, \ref{section:empirical_desiderata} (besides Figure \ref{fig:upper_limit_overfit}), and \ref{section:oom_extrap}:

\begin{itemize}
    \item The batch size is 80000. No regularization is used because training dataset size is $\sim$infinite such that model is only trained for a single epoch. Adam is used. Adam hyperparameters are ${\beta}_1=0$ and ${\beta}_2=0$ (except for Figures \ref{fig:unsl_additive_symmetry} and \ref{fig:baseline_additive_symmetry} (and Table \ref{table:trivariate_desiderata}) in Section \ref{subsection:empirical_desiderata_7} in which ${\beta}_1=0.9$ and ${\beta}_2=0.999$). Except when learning rate and/or standard deviation of weights at initialization are explicitly varied in the plots of figures, learning rate and standard deviation of weights at initialization are held constant.
\end{itemize}

In Figure \ref{fig:limit_of_predictability}, number of model parameters is varied by varying width when depth is held constant.

\section{Obtaining the Compute-Optimal Values of the Input Dimensions}
\label{section:compute_optimal}
Let $\mathcal{D}$ be the index set that contains the indexes of dimensions of $(x_1,..., x_m)$ that directly contribute to amount of training compute used (e.g. number of model parameters, number of training steps, etc.). Let $\mathcal{H}$ be the index set that contains the indexes of dimensions of $(x_1,..., x_m)$ that do not directly contribute to amount of training compute used (e.g. learning rate, standard deviation of weights at initialization, etc.). $C$ is amount of training compute used. $C_0$ is a constant (e.g. equal to $6$ in \cite{hoffmann2022training}) such that $C_0 = C / (\prod_{i \in \mathcal{D}} x_i).$ $\lambda$ is a Lagrange multiplier.

To obtain the values of $(x_1,..., x_m)$ that yield the lowest value of $y$ for a given value of $C$, one solves following system of equations:  
$$
\frac{\partial y}{\partial x_{\ell}} + \lambda\frac{C}{x_{\ell}} = 0, \quad \ell \in \mathcal{D},
$$
$$
\frac{\partial y}{\partial x_v} = 0, \quad v \in \mathcal{H},
$$
$$
C - C_0 \prod_{\ell \in \mathcal{D}} x_{\ell} = 0.
$$

\section{Effect of varying the number of observed points used for fitting UNSL functional form}
\label{section:num_point}

\FloatBarrier

\begin{figure*}[h]    \centering

\includegraphics[width=0.24\textwidth]{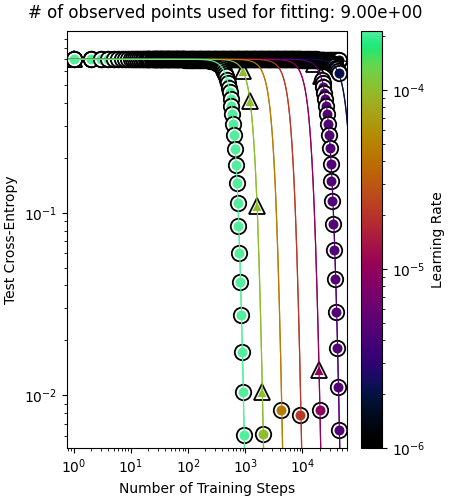}
\includegraphics[width=0.24\textwidth]{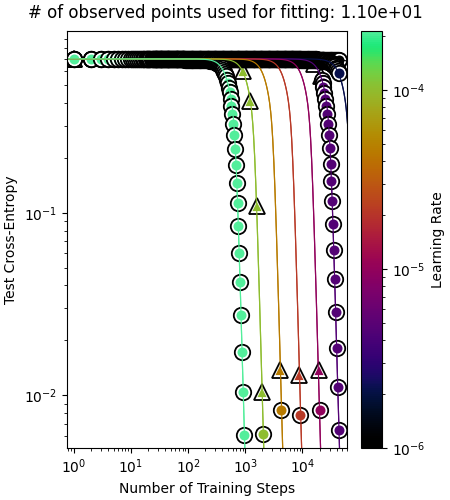}
\includegraphics[width=0.24\textwidth]{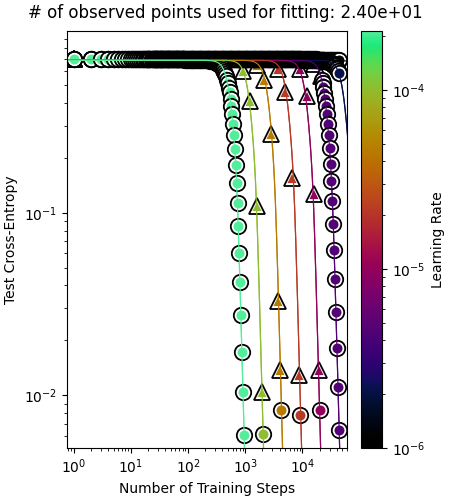}
\includegraphics[width=0.24\textwidth]{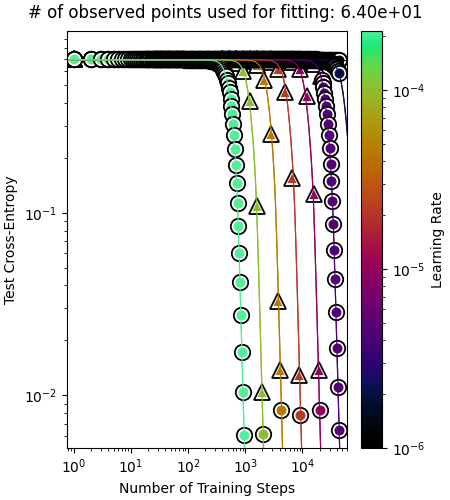}
\includegraphics[width=0.24\textwidth]{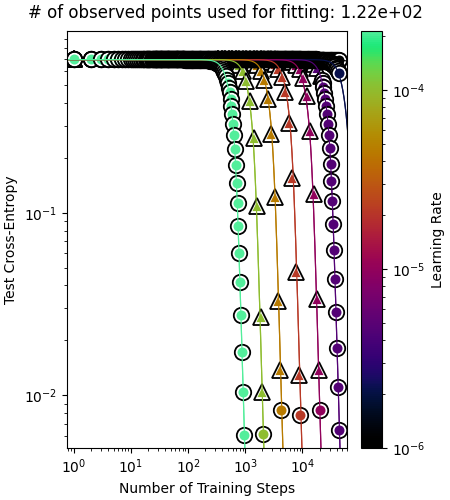}
\includegraphics[width=0.24\textwidth]{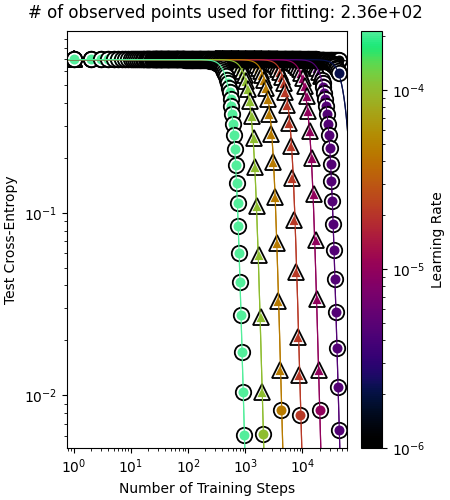}
\includegraphics[width=0.24\textwidth]{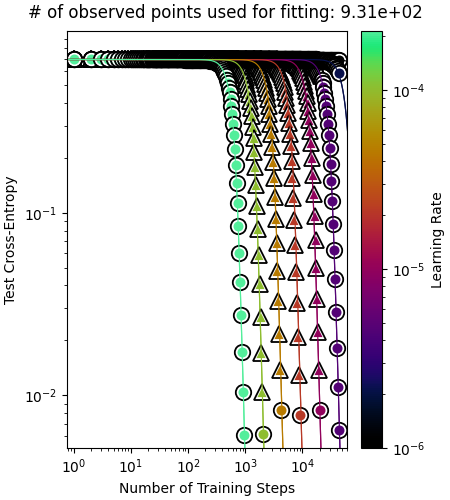}
\includegraphics[width=0.24\textwidth]{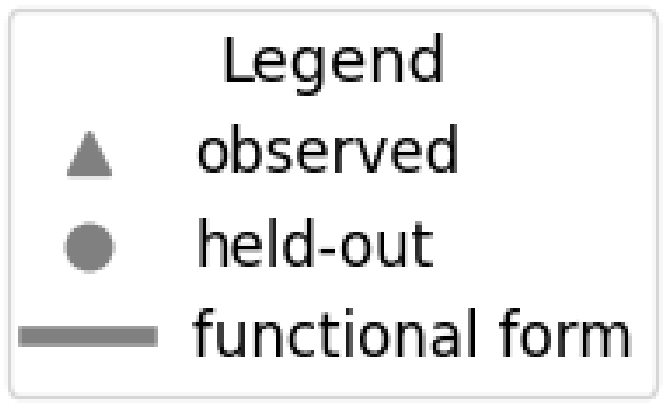}
    \caption{
    Varying the number of observed points used for fitting UNSL functional form from $9\mathrm{e}0$ (in top left plot) to $9\mathrm{e}2$ (in bottom right plot). Scaling behavior is that of an MLP trained for a single epoch on the (n, k)-sparse parity task (with $n=40$ and $k=4$) of \cite{barak2022hidden}. See Appendix \ref{section:num_point} for more details.
    }
    \label{fig:num_point}
\end{figure*}


\FloatBarrier

In Figure \ref{fig:num_point}, we observe that UNSL accurately extrapolates scaling behavior when only a small number of observed points are used for fitting UNSL functional form.

\FloatBarrier

\section{UNSL accurately extrapolating to scales an order of magnitude larger in multiple dimensions simultaneously}
\label{section:oom_extrap}

\begin{figure*}[h]    \centering
\includegraphics[width=0.79\textwidth]{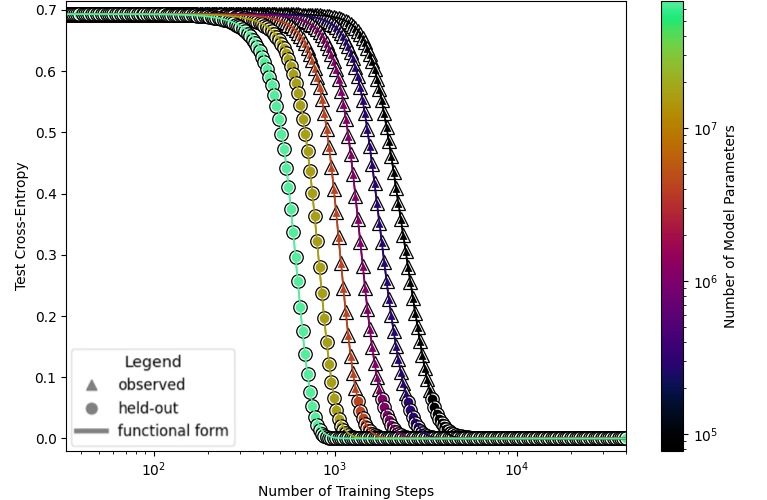}
    \caption{
        Extrapolation of UNSL on scaling behavior of an MLP trained for a single epoch on the (n, k)-sparse parity task (with $n=40$ and $k=4$) of \cite{barak2022hidden}. Each point in the plot is the mean of greater than 100 seeds. See Section \ref{section:oom_extrap} and Appendix \ref{section:experimental_details} for more details.
    }
    \label{fig:oom_extrap}
\end{figure*}

In Figure \ref{fig:oom_extrap}, UNSL accurately extrapolates to scales an order of magnitude larger in multiple dimensions simultaneously.

\FloatBarrier


\section{Supremal Expressivity Equivalence of A1, A2, A3, and UNSL}
\label{section:supremal_expressivity}

In multi-log space, MBNSL (i.e. Equation \ref{eq:mbnsl} and A1) with $|M|=m$ and $n$ hyperbreaks is
$$
\log y \;=\; \log b \;-\; \left(\sum_{i=1}^{m} c_{i_{0}}\,\log x_i\right) \;-\; \sum_{j=1}^{n} f_{j} \cdot \mathrm{softplus}\!\left(\left|\frac{1}{f_{j}}\right|\!\left(\sum_{i=1}^{m} c_{i_{j}}\,\log x_i \;-\; \log d_{j}\right)\right).
$$
which is a single-hidden-layer feedforward network with softplus activation, linear skip connection, and $n$ hidden units. Since the softplus function is continuous and non-polynomial, the universal approximation theorem for non-polynomial activations \citep{leshno1993multilayer, cybenko1989approximation, hornik1991approximation} ensures that $\{\text{A1} : n \in \mathbb{N}\}$ is dense in $C(\Omega, \mathbb{R}_{>0})$ for any compact $\Omega \subseteq \mathbb{R}_{>0}^m$.

A2, A3, and UNSL generate positive continuous functions of $(x_1, \ldots, x_m)$, which can therefore be arbitrarily well approximated by A1. Conversely, A1 is derived from each by specifying the corresponding parameters: $a_i^{-1}=0$ for all $i$, ignoring various $K$ components using their parameters, and (for UNSL) $S=0$. Hence A1, A2, A3, and UNSL have identical supremal expressivity.

\FloatBarrier


\section{Explanation of how UNSL functional form satisfies all of the desiderata of Section \ref{section:desiderata}}
\label{section:desiderata_satisfaction_explanation}

\subsection{Explanation of how UNSL functional form satisfies Desideratum \ref{item:1}}
\label{subsection:satisfaction_explanation_desiderata_1}

Desideratum \ref{item:1} says that for each single input dimension $x_i$, the scaling behavior follows a univariate broken neural scaling law of \cite{caballero2023broken}, i.e.:
\[
y = b \cdot x_i^{-{c_i}_0} \prod_{j=1}^n \left(1 + \left(\frac{x_i^{{c_i}_j}}{d_j}\right)^{\left|\frac{1}{f_j}\right|}\right)^{-f_j},
\]
where $b$, ${c_i}_0$, ${c_i}_1 ... {c_i}_j$, $d_1 ... d_j$, and $f_1 ... f_j$ are learned parameters. (Note that ``performance limit'' term $a$ from \cite{caballero2023broken} is intentionally removed here because it is addressed by other desiderata.)

This is implemented in Equation \ref{eq:mbnsl}, where each univariate scaling behavior is modeled as a Broken Neural Scaling Law (BNSL):
\begin{equation*}
K(M, n, k) = b_{k} \cdot \left(\prod_{i \in M} x_i^{-c_{i_{0_{k}}}}\right)
\prod_{j=1}^n \left( 1 + \left( \frac{\prod_{\substack{i \in M}} x_i^{c_{i_{j_{k}}}}}{d_{j_{k}}} \right)^{\left|\frac{1}{f_{j_{k}}}\right|} \right)^{-f_{j_{k}}}.
\end{equation*}

For the pedagogical purposes of this Section \ref{subsection:satisfaction_explanation_desiderata_1}, by setting $M=\{1, \ldots, m\}$ and removing subscript $k$ one can simplify Equation \ref{eq:mbnsl} to:
$$
y = b \cdot \left(\prod_{i=1}^{m} x_i^{-c_{i_{0}}}\right) \prod_{j=1}^n \left(1 + \left(\frac{\prod_{i=1}^m x_i^{c_{i_{j}}}}{d_j}\right)^{\left|\frac{1}{f_j}\right|}\right)^{-f_j} .
$$

In that equation, when one varies only a single input dimension $x_i$, all $x$ with a subscript in ``$\{1, \ldots, m\} \setminus \{i\}$'' become constants that can be folded into $b$ or $d_j$, hence recovering the univariate broken neural scaling law of \cite{caballero2023broken}, i.e.:
\[
y = b \cdot x_i^{-{c_i}_0} \prod_{j=1}^n \left(1 + \left(\frac{x_i^{{c_i}_j}}{d_j}\right)^{\left|\frac{1}{f_j}\right|}\right)^{-f_j}.
\]

\subsection{Explanation of how UNSL functional form satisfies Desideratum \ref{item:2}}
\label{subsection:satisfaction_explanation_desiderata_2}

In Equation \ref{eq:mbnsl}, the j-th hyperbreak (i.e. smooth transition from the j-th hyperplane to the (j+1)-th hyperplane) occurs at the values of $(x_i)_{i \in M}$ for which this equality is true: 

\[
d_{j_k} = \prod_{i \in M} x_i^{c_{i_{j_k}}}.
\]

As can be seen from this equality, the location at which each hyperbreak occurs is shifted via multiplicative interaction between (exponentiations of) input dimensions $(x_i)_{i \in M}$.

\subsection{Explanation of how UNSL functional form satisfies Desideratum \ref{item:3}}
\label{subsection:satisfaction_explanation_desiderata_3}

For the pedagogical purposes of this Section~\ref{subsection:satisfaction_explanation_desiderata_3}, by removing subscript $k$ one can simplify Equation \ref{eq:mbnsl} to:

$$
y = b \cdot \left(\prod_{i \in M} x_i^{-c_{i_{0}}}\right)
\prod_{j=1}^n \left( 1 + \left( \frac{\prod_{\substack{i \in M}} x_i^{c_{i_{j}}}}{d_{j}} \right)^{\left|\frac{1}{f_{j}}\right|} \right)^{-f_{j}}.
$$

When $c_{i_j}$ and $f_j$ are constrained to force that functional form to always be nonmonotonic
(and assuming $x_i>0,\;y>0,\;b>0,\;d_j>0$), that functional form effectively becomes the following
monomial when maximally close to the global minima with respect to $y$:

$$
y = \left( b \cdot \prod_{\substack{j=1 \\ f_j > 0}}^{n} d_j \right)
    \cdot \prod_{\substack{i \in M}} x_i^{-\left( c_{i_0} +
    \sum_{\substack{j=1 \\ f_j > 0}}^{n} c_{i_j} \right)}.
$$

When Equation~\ref{eq:mbnsl} becomes a monomial,
the expressivity of Equation~\ref{eq:unsl3} becomes equivalent to the expressivity of this functional form:

$$
b \cdot \left(\prod_{i\in U_r} x_i^{-c_{i_{0_0}}} \right)
\;+\; \sum_{t\in T_r} b_t\,x_t^{-c_{t_{0_t}}},
$$

which effectively becomes

$$
\sum_{t\in T_r} b_t\,x_t^{-c_{t_{0_t}}}
$$

when

$$
b\prod_{i\in U_r} x_i^{-c_{i_{0_0}}} \ll \sum_{t\in T_r} b_t\,x_t^{-c_{t_{0_t}}}.
$$

As a result, it is also true that
$$
a \;+\; b \cdot \left(\prod_{i\in U_r} x_i^{-c_{i_{0_0}}} \right)
\;+\; \sum_{t\in T_r} b_t\,x_t^{-c_{t_{0_t}}}
$$

effectively becomes

$$
a \;+\; \sum_{t\in T_r} b_t\,x_t^{-c_{t_{0_t}}}
$$

when

$$
b\prod_{i\in U_r} x_i^{-c_{i_{0_0}}} \ll \sum_{t\in T_r} b_t\,x_t^{-c_{t_{0_t}}}.
$$

Additionally, the following functional form

$$
a \;+\; \sum_{t\in T_r} b_t\,x_t^{-c_{t_{0_t}}}
$$

effectively becomes

$$
a \;+\; b_v\,x_v^{-c_{v_{0_v}}},\,\,\,\, (\text{where } v \in T_r)
$$

when

$$
\Bigg(a + \sum_{t\in T_r \setminus \{v\}} b_t\,x_t^{-c_{t_{0_t}}} \Bigg) \ll \Bigg(a + b_v\,x_v^{-c_{v_{0_v}}} \Bigg).
$$

\subsection{Explanation of how UNSL functional form satisfies Desideratum \ref{item:4}}
\label{subsection:satisfaction_explanation_desiderata_4}

Recall that Equation \ref{eq:unsl1} is:
\[
y = a_0 + \left(\left(Q(3) + \left(Q(S+4) + a_1^{-1}\right)^{-1}\right)^{-1} + a_2^{-1}\right)^{-1}.
\]

Desideratum \ref{item:4} is captured by the addition between $a_0$ and $\left(\left(Q(3) + \left(Q(S+4) + a_1^{-1}\right)^{-1}\right)^{-1} + a_2^{-1}\right)^{-1}$ in this equation, where $a_0$ represents the ultimate limit of performance.

\subsection{Explanation of how UNSL functional form satisfies Desideratum \ref{item:5}}
\label{subsection:satisfaction_explanation_desiderata_5}

This is captured in Equations \ref{eq:unsl1} and \ref{eq:unsl2} when (the reciprocal of) each of every variable in the set $\{ a_i \}_{i > 0}$ is summed with a functional form and each resultant sum is then raised to the $-1$ power. The set $\{ a_i \}_{i > 0}$ contains multiple variables rather than a single variable because misperformance caused by different phenomena often have different misperformance limits. For example, misperformance caused by overfitting often has a misperformance limit that is significantly worse than the performance of random guessing; meanwhile, misperformance caused by nonoptimal hyperparameters often has at least one misperformance limit that is equal to the performance of random guessing. The reason that in Equation \ref{eq:unsl2} a value of $S$ greater than $1$ (rather than equal to $1$) is sometimes used is that there sometimes are multiple misperformance limits $a_{q+s}$ (e.g. as in Figure \ref{fig:multiple_upper_limits} of Appendix \ref{subsection:empirical_desiderata_5}): a misperformance limit that is significantly larger than random guessing (that usually is noticeable when the number of training steps is small) and a misperformance limit that approximately is less than or equal to random guessing (that usually is noticeable when the number of training steps is large).

\subsection{Explanation of how UNSL functional form satisfies Desideratum \ref{item:6}}
\label{subsection:satisfaction_explanation_desiderata_6}

Recall Appendix \ref{subsection:satisfaction_explanation_desiderata_3}. As a result of Appendix \ref{subsection:satisfaction_explanation_desiderata_3}, Desideratum \ref{item:6} is captured by each of every instance in which $R(r)$ of Equation \ref{eq:unsl3} is effectively raised to the $-1$ power; an instance in which $R(r)$ occurs is considered ``effectively raised to the $-1$ power'' if the count of reciprocal operations whose scope contains that instance is odd. Instances in which this occurs are $\left({R(q+s) + a_{q+s}}^{-1}\right)^{-1}$ from Equation \ref{eq:unsl2} and $\left(Q(S+4) + {a_1}^{-1}\right)^{-1}$ from Equation \ref{eq:unsl1} (which contains $\left(\left(R(q)\right)^{-1} + {a_q}^{-1}\right)^{-1}$ from Equation \ref{eq:unsl2}).

\subsection{Explanation of how UNSL functional form satisfies Desideratum \ref{item:7}}
\label{subsection:satisfaction_explanation_desiderata_7}

This desideratum is captured when $\left(\left(R(q)\right)^{-1} + {a_q}^{-1}\right)^{-1}$ is summed with the ``oppositional force of hyperparameters'' in Equation \ref{eq:unsl2}, and when $Q(3)$ is summed with the ``oppositional force of overfitting'' in Equation \ref{eq:unsl1}.

\subsection{Explanation of how UNSL functional form satisfies Desideratum \ref{item:8}}
\label{subsection:satisfaction_explanation_desiderata_8}

UNSL (i.e. Equation \ref{eq:unsl1}) functional form (expanded out for pedagogical purposes) is:

\[
\begin{aligned}
y =\ & a_0 + \Bigg(\bigg(\left(\left(R(3)\right)^{-1} + {a_3}^{-1}\right)^{-1}
+ \underbrace{\sum_{s=1}^{S} \left(R(3+s) + {a_{3+s}}^{-1}\right)^{-1}}_{\text{oppositional force of hyperparameters}} \\
& + \underbrace{\left({a_1}^{-1} + \left(\left(R(S+4)\right)^{-1}
+ {a_{S+4}}^{-1}\right)^{-1}
+ \underbrace{\sum_{s=1}^{S} \left(R(S+4+s) + {a_{S+4+s}}^{-1}\right)^{-1}}_{\text{oppositional force of hyperparameters}}\right)^{-1}}_{\text{oppositional force of overfitting}}\bigg)^{-1} + {a_2}^{-1}\Bigg)^{-1}.
\end{aligned}
\]

As can be seen in that expansion of UNSL, oppositional force(s) of hyperparameters oppose the ``oppositional force of overfitting'' and the subset of the UNSL functional form that is not the ``oppositional force of overfitting''. Note that each of every ``oppositional force'' is nonnegative and that what each of every ``oppositional force'' opposes is nonnegative.

\section{Empirical Evidence of Desiderata of Section \ref{section:desiderata}}
\label{section:empirical_desiderata}

\subsection{Empirical Evidence of Desideratum \ref{item:1}}
\label{subsection:empirical_desiderata_1}

\FloatBarrier

\FloatBarrier

\begin{figure*}[h]    \centering
\includegraphics[width=0.495\textwidth]{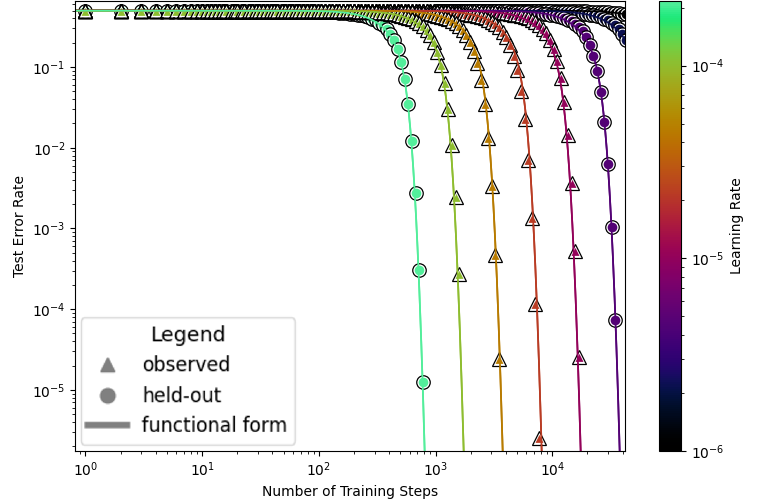}
\includegraphics[width=0.495\textwidth]{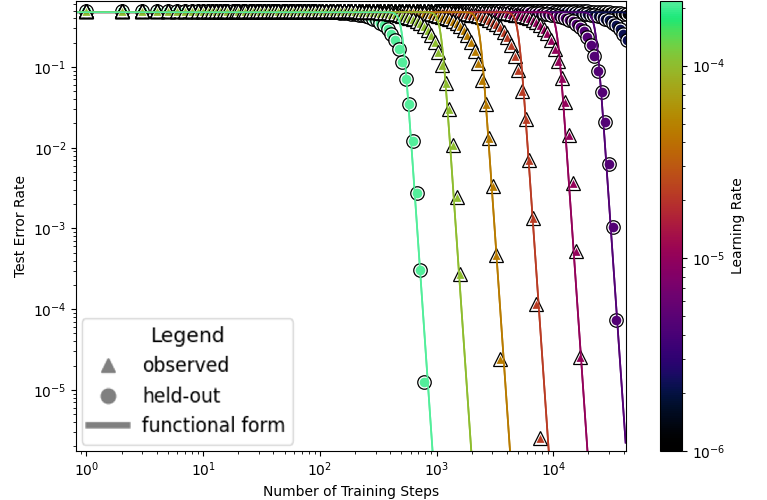}
    \caption{
        Extrapolation Results on scaling behavior of an MLP trained for a single epoch on the (n, k)-sparse parity task (with $n=40$ and $k=4$) of \cite{barak2022hidden}. Left figure fits the functional form $y = \left(\left(\left(b \prod_{\substack{i=1}}^{m} x_i^{-c_{i_{0}}}\right)\left( 1 + \left( \frac{\prod_{\substack{i=1}}^{m} x_i^{c_{i_{1}}}}{d} \right)^{\left|\frac{1}{f}\right|} \right)^{-f}\right)^{-1} + a^{-1} \right)^{-1}$. Right figure fits the functional form of left figure when $f$ is constrained to be $1$ such that the functional form of right figure is \hspace{3.8mm} $y = \left(\left(\left(b \prod_{\substack{i=1}}^{m} x_i^{-c_{i_{0}}}\right)\left( 1 + \left( \frac{\prod_{\substack{i=1}}^{m} x_i^{c_{i_{1}}}}{d} \right)^{\left|\frac{1}{1}\right|} \right)^{-1}\right)^{-1} + a^{-1} \right)^{-1}$. Observe that the fits and extrapolations in the top right quadrant of right figure are unsatisfactory. See Section \ref{subsection:empirical_desiderata_1} for more details.
    }
    \label{fig:sharpness_desiderata}
\end{figure*}

\FloatBarrier

In Figure \ref{fig:sharpness_desiderata}, Desideratum \ref{item:1} is true empirically. As can be seen in Figure \ref{fig:sharpness_desiderata}, the sharpness needs to be decoupled from the amount of change in gradient (i.e. via the extra expressivity granted by $f$ in Equation \ref{eq:mbnsl_single_break} (and Equation \ref{eq:mbnsl})) in order to accurately fit and accurately extrapolate the scaling behavior.

\FloatBarrier

\FloatBarrier

\subsection{Empirical Evidence of Desideratum \ref{item:2}}
\label{subsection:empirical_desiderata_2}

\FloatBarrier

Recall that in Equation \ref{eq:mbnsl} (with subscript $k$ removed for pedagogical purposes) the j-th hyperbreak (i.e. smooth transition from the j-th hyperplane to the (j+1)-th hyperplane) occurs at the values of $(x_i)_{i \in M}$ for which this equality is true: 
\[
d_j = \prod_{i \in M} x_i^{c_{i_{j}}}.
\]

As a result, desideratum \ref{item:2} is true empirically because the functional form $y = \left(\left(\left(b \prod_{\substack{i=1}}^{m} x_i^{-c_{i_{0}}}\right)\left( 1 + \left( \frac{\prod_{\substack{i=1}}^{m} x_i^{c_{i_{1}}}}{d} \right)^{\left|\frac{1}{f}\right|} \right)^{-f}\right)^{-1} + a^{-1} \right)^{-1}$ accurately fits and accurately extrapolates the scaling behavior in Figure \ref{fig:sharpness_desiderata} left.

\FloatBarrier

\subsection{Empirical Evidence of Desideratum \ref{item:3}}
\label{subsection:empirical_desiderata_3}

Note that $(x_i)_{i=1}^m \in \overline{\mathbb{R}}_{>0}^{\,m}$ and that $(x_t)_{t \in T_r} \in \overline{\mathbb{R}}_{>0}^{\,T_r}$.

Desideratum \ref{item:3} is observed empirically in several prior works such as \cite{hoffmann2022training} which empirically show that the scaling behavior follows $y = a + \sum_{t \in T_r} b_t \cdot x_t^{-c_{t}}$ when sufficiently close to the global optimum of $y$.

\subsection{Empirical Evidence of Desideratum \ref{item:4}}
\label{subsection:empirical_desiderata_4}
Desideratum \ref{item:4} is observed empirically in several prior works such as \cite{hoffmann2022training} in which the transition to the performance limit as all $x_{i}$ dimensions in $x_{1}...x_{m}$ simultaneously go to the values of $(x_i)_{i=1}^m \in \overline{\mathbb{R}}_{>0}^{\,m}$ that yield the global optimum of $y$ is characterized by summing an entire functional form with a constant.

\subsection{Empirical Evidence of Desideratum \ref{item:5}}
\label{subsection:empirical_desiderata_5}

\FloatBarrier

\begin{figure*}[h]    \centering
\includegraphics[width=0.85\textwidth]{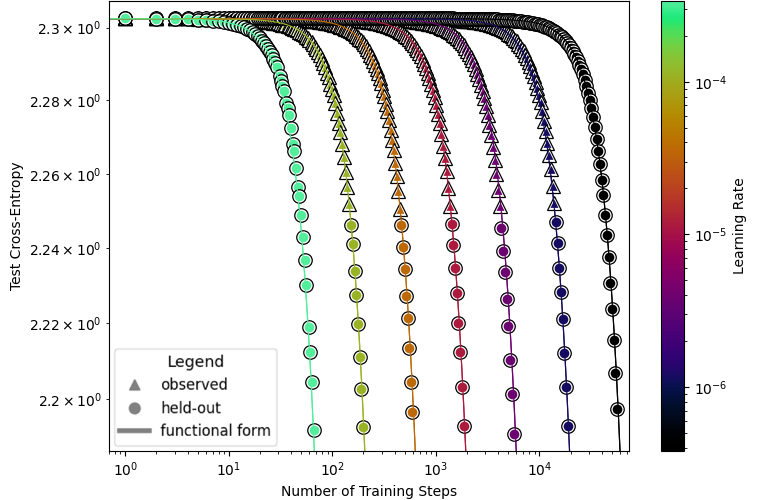}
    \caption{
    Extrapolation Results of functional form $y=\left(\left(b \prod_{\substack{i=1}}^{m} x_i^{-c_{i}}\right)^{-1} + a^{-1} \right)^{-1}$. Scaling behavior is (top left region of) that of an MLP trained for a single epoch on dataset of \cite{greydanus2024scaling}. See Section \ref{subsection:empirical_desiderata_5} for more details.
    }
    \label{fig:first_upper_limit}
\end{figure*}

\FloatBarrier

In Figure \ref{fig:first_upper_limit}, Desideratum \ref{item:5} is true empirically.
As can be seen in Figure \ref{fig:first_upper_limit}, the scaling behavior is characterized by raising to the -1 power the sum of a functional form and a constant.

\FloatBarrier

\begin{figure*}[h]    \centering
\includegraphics[width=0.99\textwidth]{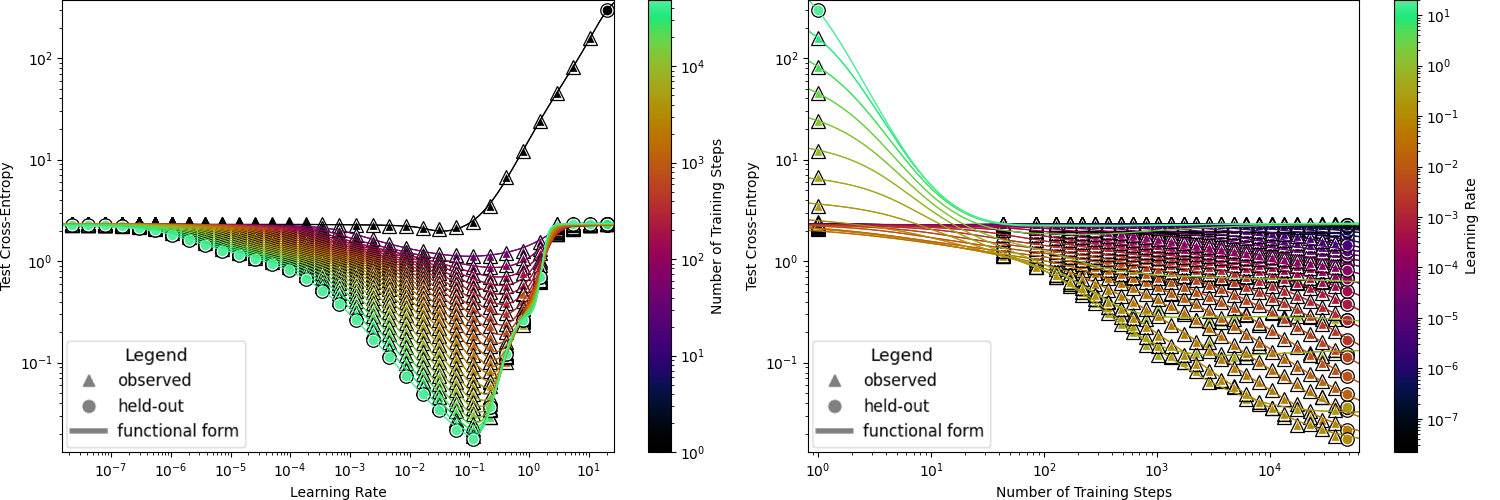}
    \caption{
    Extrapolation Results of UNSL functional form. Scaling behavior is that of an MLP (when standard deviation of weights at initialization is large) trained for a single epoch on dataset of \cite{greydanus2024scaling}. See Section \ref{subsection:empirical_desiderata_5} for more details.
    }
    \label{fig:multiple_upper_limits}
\end{figure*}

\FloatBarrier

In Figure \ref{fig:multiple_upper_limits} with regards to Desideratum \ref{item:5}, there is a misperformance limit \textasciitilde equal to random guessing performance (cross-entropy of 2.3) when it is simultaneously true that learning rate is large (i.e. greater than 3) and number of training steps is large (i.e. greater than 100); and an additional misperformance limit equal to a value significantly larger (i.e. larger than the largest $y$-axis value of Figure \ref{fig:multiple_upper_limits}) than random guessing performance (cross-entropy of 2.3) occurs when it is simultaneously true that learning rate is large (i.e. significantly greater than 20) and number of training steps is small (i.e. less than 2). As a result, $S$ (from Equation \ref{eq:unsl2}) is equal to $2$ in Figure \ref{fig:multiple_upper_limits}.

\FloatBarrier

\subsection{Empirical Evidence of Desideratum \ref{item:6}}
\label{subsection:empirical_desiderata_6}

\FloatBarrier

\begin{figure*}[h]    \centering
\includegraphics[width=0.99\textwidth]{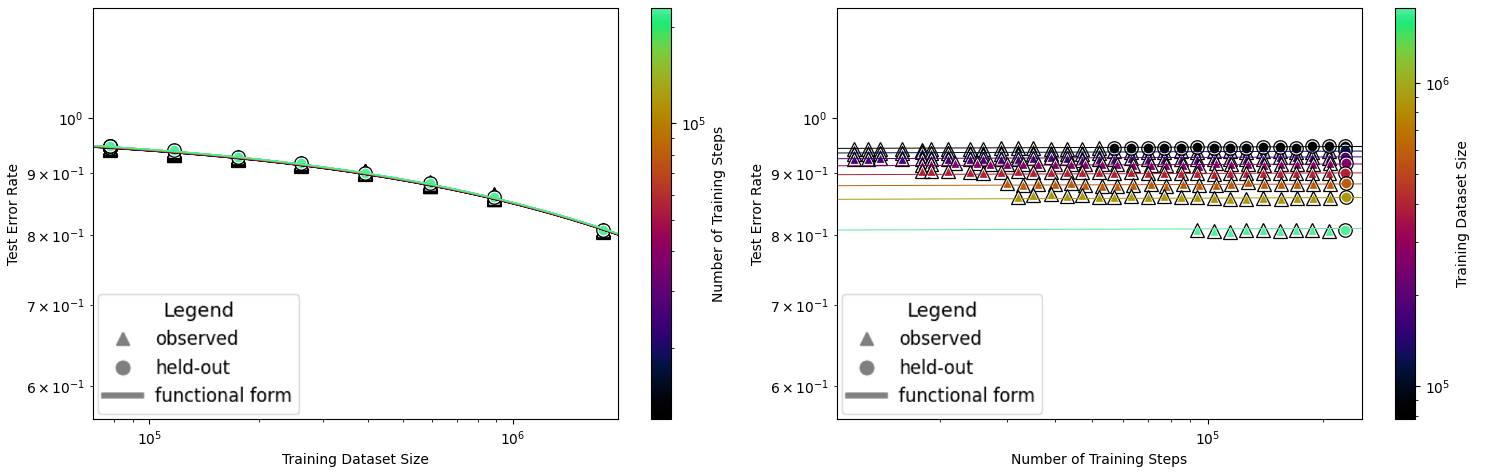}
    \caption{
    Extrapolation Results of functional form $y = \left(a + \sum_{t \in T_r} b_t \cdot x_t^{-c_{t}}\right)^{-1}$. Scaling behavior is (top region of the scaling behavior of) downstream ImageNet test error rate of ViT pre-trained on JFT. See Sections \ref{subsection:empirical_desiderata_6} and \ref{section:vision} for more details.
    }
    \label{fig:upper_limit_overfit}
\end{figure*}

\FloatBarrier

In Figure \ref{fig:upper_limit_overfit}, Desideratum \ref{item:6} is true empirically.
As can be seen in Figure \ref{fig:upper_limit_overfit}, the scaling behavior is characterized by the functional form $y = \left(a + \sum_{t \in T_r} b_t \cdot x_t^{-c_{t}}\right)^{-1}$.

\FloatBarrier

\subsection{Empirical Evidence of Desideratum \ref{item:7}}
\label{subsection:empirical_desiderata_7}

\begin{figure*}[h]
    \centering
\begin{minipage}{0.32\textwidth}
\includegraphics[width=1.0\textwidth]{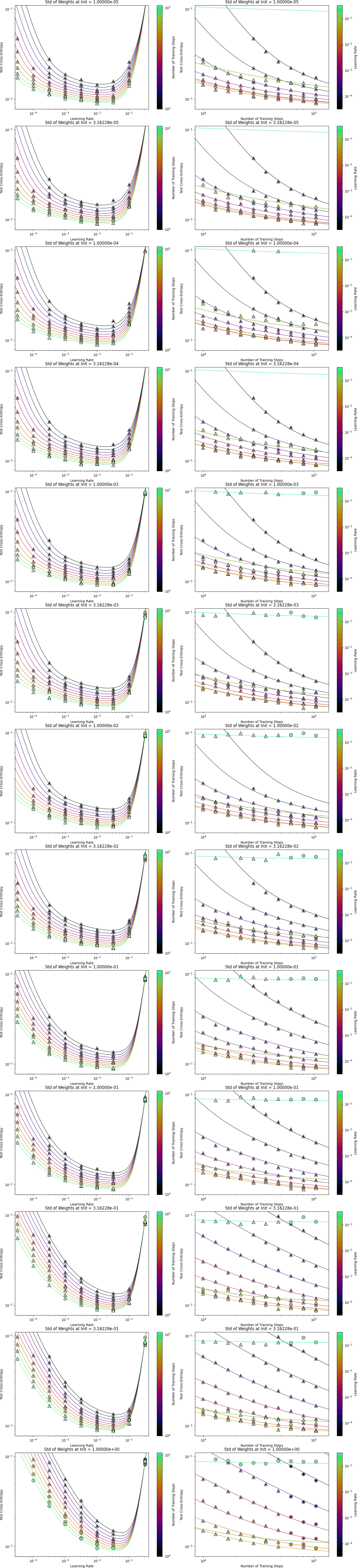}
\end{minipage}
\begin{minipage}{0.32\textwidth}
\includegraphics[width=1.0\textwidth]{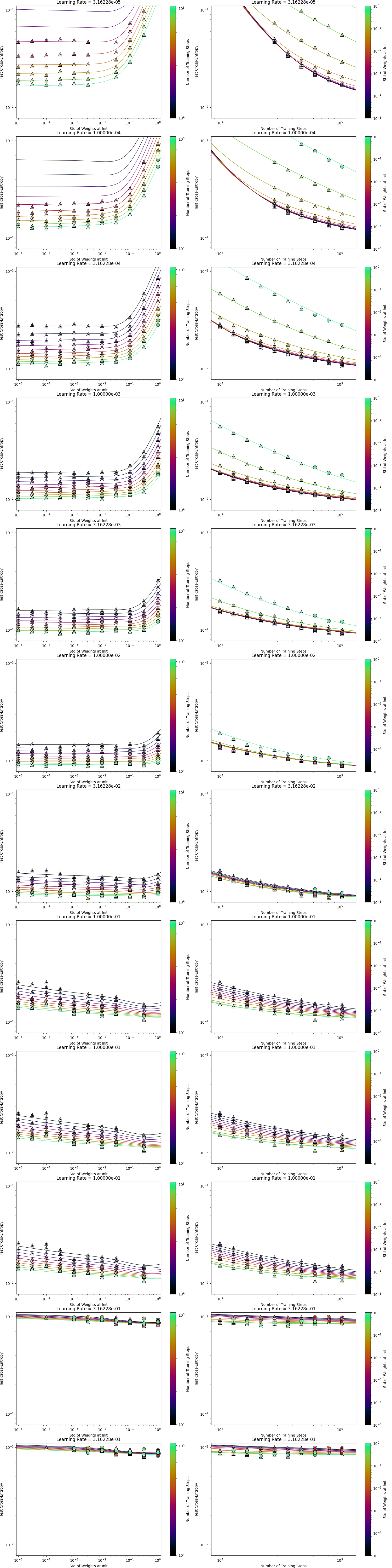}
\end{minipage}
\begin{minipage}{0.32\textwidth}
\includegraphics[width=0.7\textwidth]{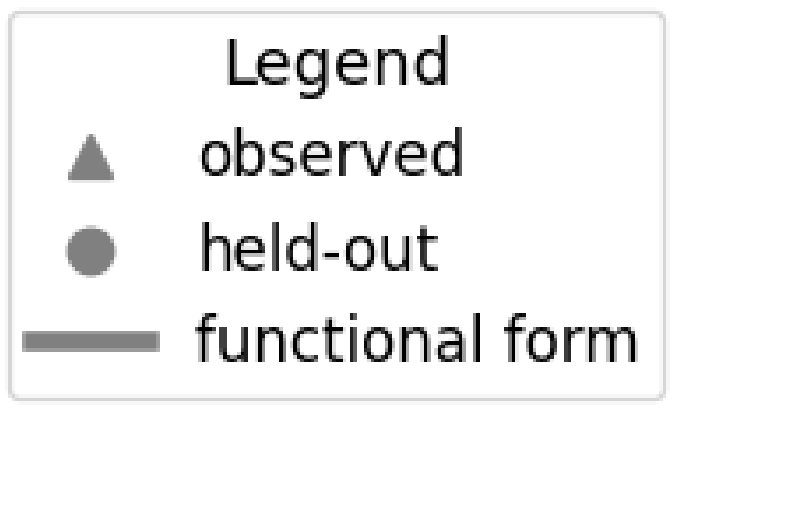}
\includegraphics[width=1.0\textwidth]{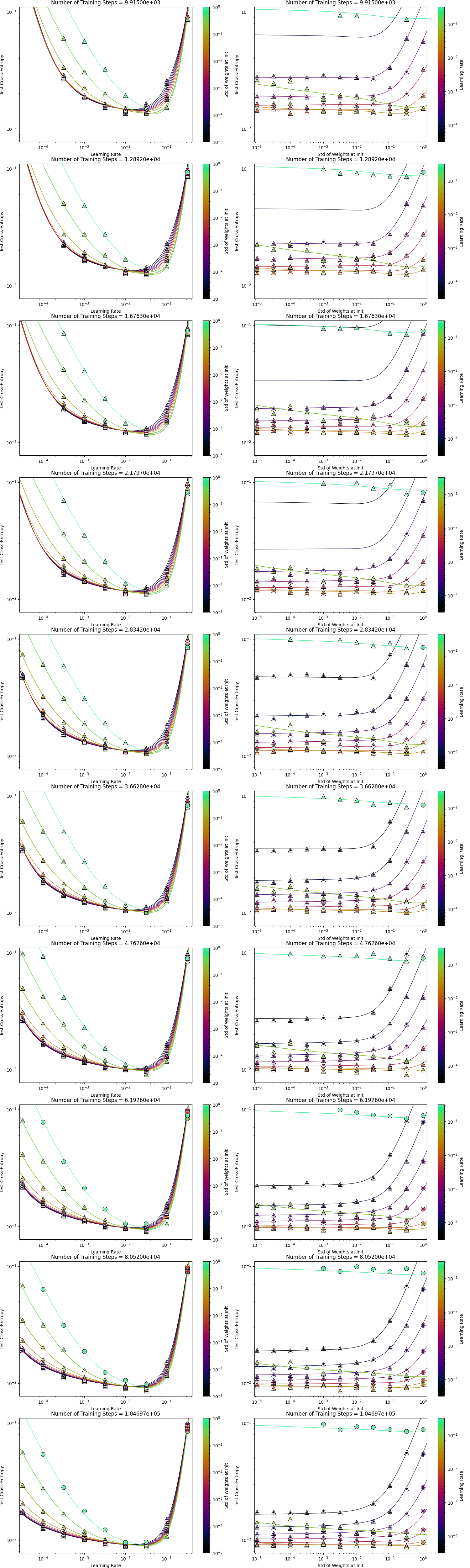}
\end{minipage}
    \caption{
    Extrapolation Results of UNSL. This trivariate scaling behavior is that of an MLP trained for a single epoch on dataset of \cite{greydanus2024scaling}. See Section \ref{subsection:empirical_desiderata_7} for more details.
    }
    \label{fig:unsl_additive_symmetry}
\end{figure*}

\begin{figure*}[h]
    \centering
\begin{minipage}{0.32\textwidth}
\includegraphics[width=1.0\textwidth]{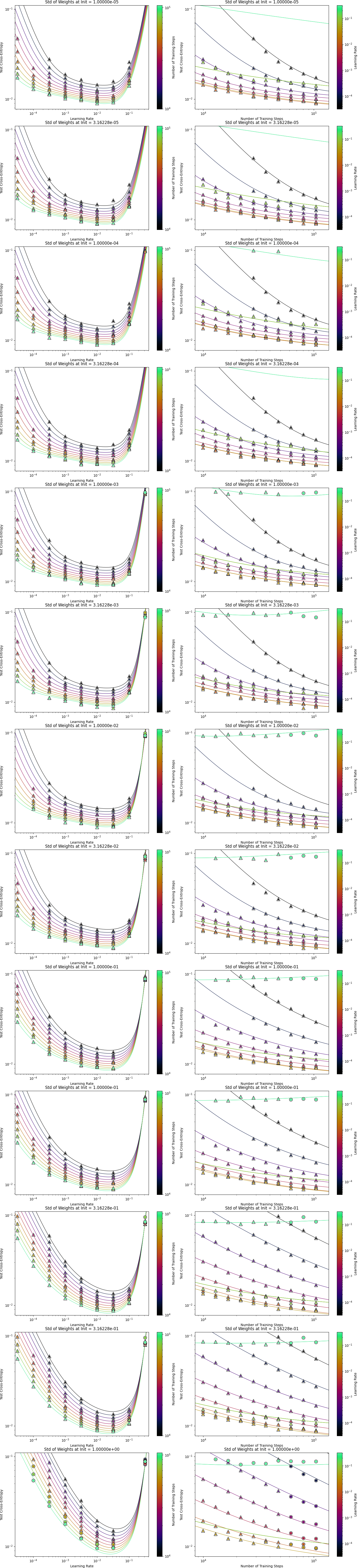}
\end{minipage}
\begin{minipage}{0.32\textwidth}
\includegraphics[width=1.0\textwidth]{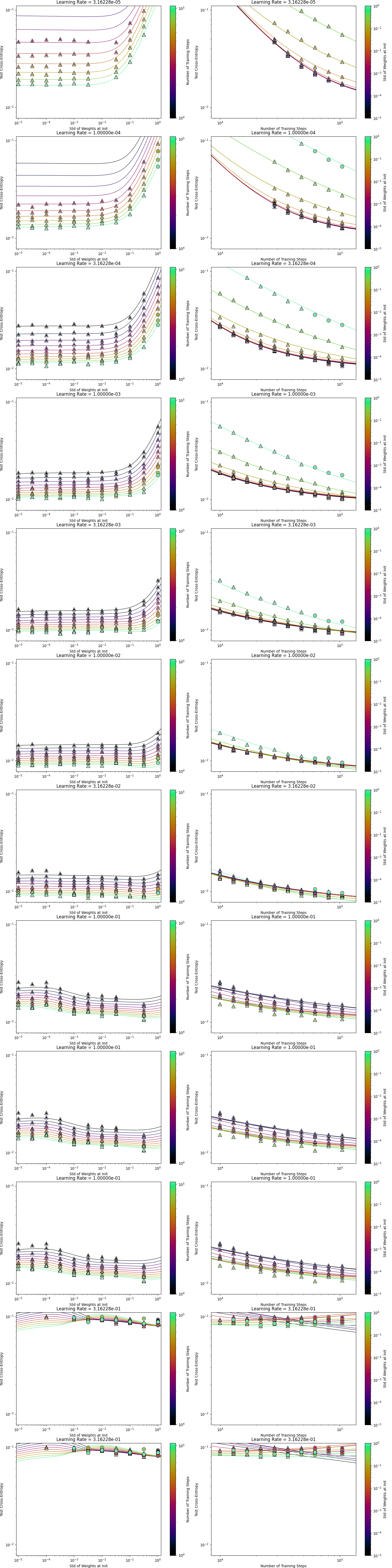}
\end{minipage}
\begin{minipage}{0.32\textwidth}
\includegraphics[width=0.7\textwidth]{figures/legend/legend2.png}
\includegraphics[width=1.0\textwidth]{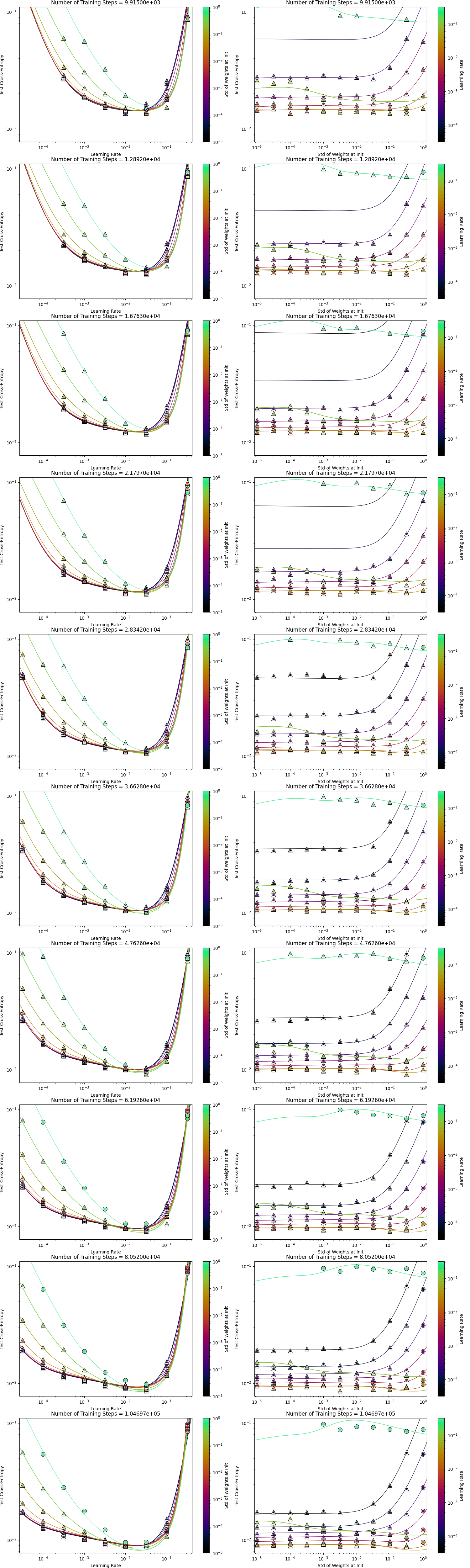}
\end{minipage}
    \caption{
    Extrapolation Results of ablation baseline of Equation \ref{eq:unsl2_ablated}. This trivariate scaling behavior is that of an MLP trained for a single epoch on dataset of \cite{greydanus2024scaling}. See Section \ref{subsection:empirical_desiderata_7} for more details.
    }
    \label{fig:baseline_additive_symmetry}
\end{figure*}

In Table \ref{table:trivariate_desiderata} which summarizes Figures \ref{fig:unsl_additive_symmetry} and \ref{fig:baseline_additive_symmetry}, Desideratum \ref{item:7} is true empirically. We obtain the trivariate scaling behavior as learning rate, standard deviation of weights at initialization, and number of training steps vary when training an MLP for a single epoch on dataset of \cite{greydanus2024scaling}. When holding the number of learned parameters of the functional forms constant, we compare the training and extrapolation RMSLE of UNSL to the following ablated functional form baseline in which the additive symmetries of Equation \ref{eq:unsl2} are removed:

\begin{equation}
\begin{split}
y = a_0 + K( \, U_0,\,\, n_{0_0},\,\, 0) \, + \, \sum_{\mathclap{ \kern 0.9mm t \in T_0}} K(\{t\},\,\, n_{0_t},\,\, t),  \hspace{3.8mm} \text{where } U_0,T_0\subseteq\{1,\ldots,m\}.
\end{split}
\label{eq:unsl2_ablated}
\end{equation}

As can be seen in Table \ref{table:trivariate_desiderata} and Figures \ref{fig:unsl_additive_symmetry} and \ref{fig:baseline_additive_symmetry}, when holding the number of learned parameters of the functional forms constant, UNSL yields fits and extrapolations with lower RMSLE than the ablated functional form baseline of Equation \ref{eq:unsl2_ablated}.

\begin{table}[hbt!]
    \centering
    \begin{tabular}{ |c|c|c| }
\hline
Set & Baseline $\downarrow$ & UNSL $\downarrow$\\
 \hline
Training & 3.80e-2 ± 1.14e-3 & \bfseries 3.49e-2 ± 1.27e-3 \\
Extrapolation & 8.09e-2 ± 5.90e-3 & \bfseries 5.11e-2 ± 4.30e-3 \\
All & 4.14e-2 ± 1.26e-3 & \bfseries 3.60e-2 ± 1.23e-3 \\
 \hline
\end{tabular}
    \caption{
    Results on trivariate scaling behavior in which Desideratum \ref{item:7} is true empirically. See Section \ref{subsection:empirical_desiderata_7} for more details.
    }
    \label{table:trivariate_desiderata}
\end{table}

\FloatBarrier

\subsection{Empirical Evidence of Desideratum \ref{item:8}}
\label{subsection:empirical_desiderata_8}

In Table \ref{table:downstream_vision}, Desideratum \ref{item:8} is true empirically because UNSL functional form outperforms A3 in the majority of instances.

Recall that A3 functional form is:

$$
y = a_0 + \left(\left(\left(\left(R(0)\right)^{-1} + {a_1}^{-1}\right)^{-1}+ \underbrace{\sum_{s=1}^{S} \left(R(s) + {a_{s+2}}^{-1}\right)^{-1}}_{\text{all oppositional forces in general}}\right)^{-1} + a_2^{-1}\right)^{-1} .
$$

meanwhile UNSL functional form (expanded out for pedagogical purposes) is:

\[
\begin{aligned}
y =\ & a_0 + \Bigg(\bigg(\left(\left(R(3)\right)^{-1} + {a_3}^{-1}\right)^{-1}
+ \underbrace{\sum_{s=1}^{S} \left(R(3+s) + {a_{3+s}}^{-1}\right)^{-1}}_{\text{oppositional force of hyperparameters}} \\
& + \underbrace{\left({a_1}^{-1} + \left(\left(R(S+4)\right)^{-1}
+ {a_{S+4}}^{-1}\right)^{-1}
+ \underbrace{\sum_{s=1}^{S} \left(R(S+4+s) + {a_{S+4+s}}^{-1}\right)^{-1}}_{\text{oppositional force of hyperparameters}}\right)^{-1}}_{\text{oppositional force of overfitting}}\bigg)^{-1} + {a_2}^{-1}\Bigg)^{-1}.
\end{aligned}
\]

As can be seen in that expansion of UNSL, oppositional force(s) of hyperparameters oppose the ``oppositional force of overfitting'' and the subset of the UNSL functional form that is not the ``oppositional force of overfitting''; meanwhile, in A3 functional form, the ``oppositional force(s) of hyperparameters'' does not oppose the ``oppositional force of overfitting''. Note that each of every ``oppositional force'' is nonnegative and that what each of every ``oppositional force'' opposes is nonnegative.

\FloatBarrier

\newpage
\FloatBarrier

\section{Plots of Extrapolation Results}
\label{section:extrapolation_benchmark}


\FloatBarrier

\subsection{Plots of Reinforcement Learning Extrapolation Results}
\label{subsection:extrapolation_reinforcement_learning}

\FloatBarrier

\begin{figure*}[h]    \centering
\includegraphics[width=0.51\textwidth]{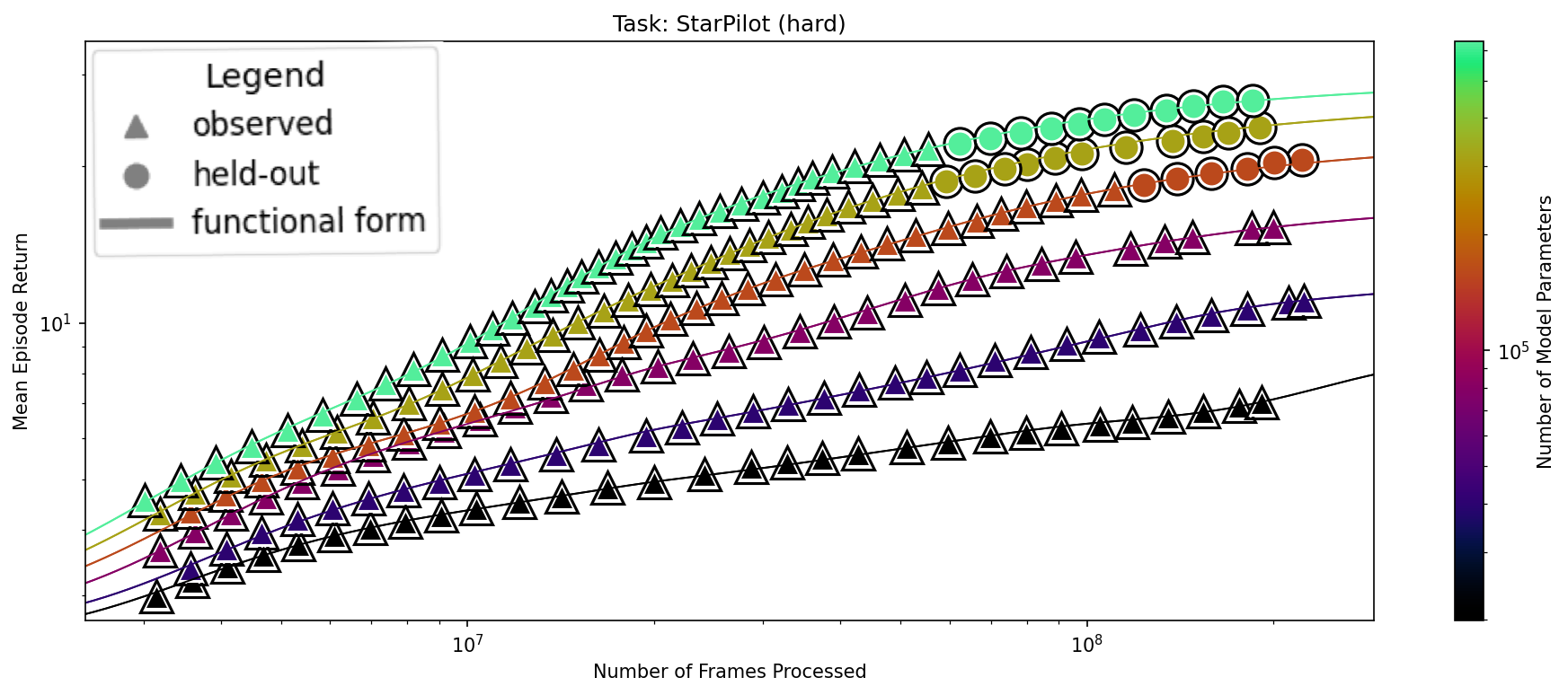}
    \caption{
    Extrapolation Results of UNSL on scaling behavior of reinforcement learning. Experimental data obtained from Figure 1a of \cite{hilton2023scaling}. Scaling behavior is that of ``StarPilot (hard)'' task of \cite{cobbe2020leveraging}. X-axis is number of frames processed (i.e. ``batch size'' multiplied by ``number of training steps'') during training. See Section \ref{subsection:extrapolation_reinforcement_learning} for more details.
    }
    \label{fig:reinforcement_learning}
\end{figure*}

In Figure \ref{fig:reinforcement_learning}, UNSL accurately extrapolates the multivariate scaling behavior of reinforcement learning.

\FloatBarrier

\subsection{Plots of Inference Scaling Extrapolation Results}
\label{subsection:extrapolation_inference_scaling}

\FloatBarrier

\begin{figure*}[h]    \centering
\includegraphics[width=0.51\textwidth]{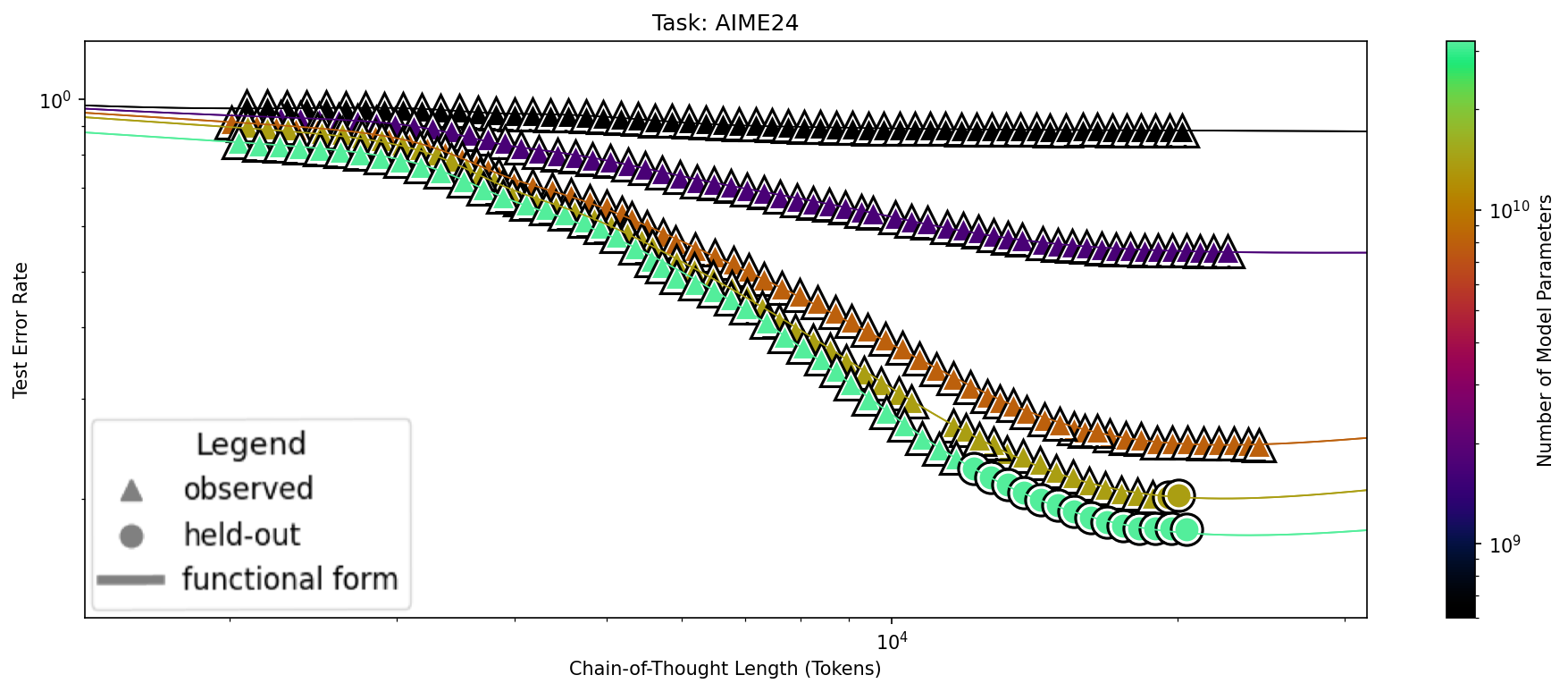}
    \caption{
    Extrapolation Results of UNSL on scaling behavior of inference scaling. Experimental data obtained from Figure 4a of \cite{sadhukhan2025kinetics}. Scaling behavior is that of test error rate on \cite{AIME2024}. X-axis is ``Chain-of-Thought Length'' (measured in number of tokens), i.e. how many ``thinking'' tokens a model outputs before outputting a final answer. See Section \ref{subsection:extrapolation_inference_scaling} for more details.
    }
    \label{fig:inference_scaling}
\end{figure*}

In Figure \ref{fig:inference_scaling}, UNSL accurately extrapolates the multivariate scaling behavior of inference (i.e. test-time) scaling.

\FloatBarrier

\subsection{Plots of ``Width vs Depth'' Extrapolation Results}
\label{subsection:extrapolation_width_depth}

\FloatBarrier

\begin{figure*}[h]    \centering
\includegraphics[width=0.99\textwidth]{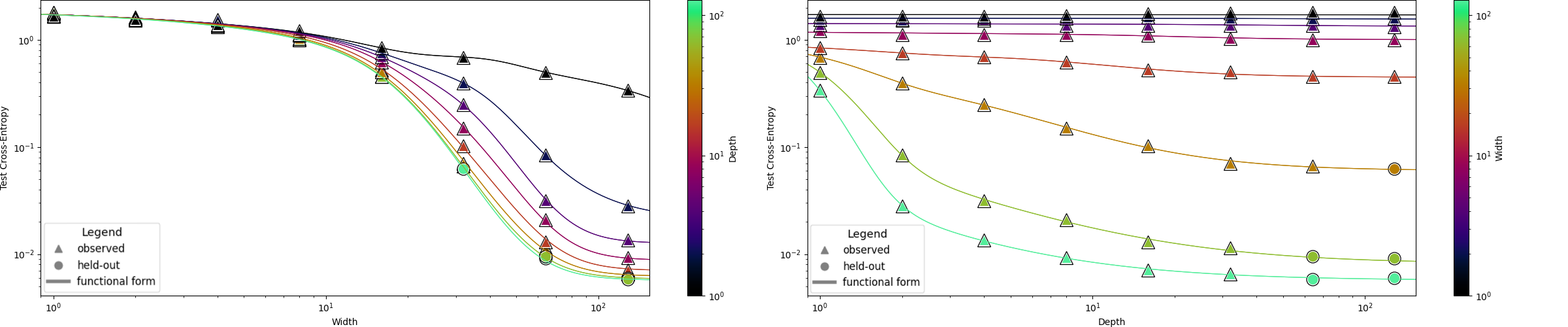}
    \caption{
    Extrapolation Results of UNSL on multivariate scaling behavior as width and depth simultaneously vary. Scaling behavior is that of an MLP trained for a single epoch on dataset of \cite{greydanus2024scaling}. See Section \ref{subsection:extrapolation_width_depth} for more details.
    }
    \label{fig:width_depth}
\end{figure*}

\FloatBarrier

In Figure \ref{fig:width_depth}, UNSL accurately extrapolates multivariate scaling behavior as width and depth vary simultaneously.

\FloatBarrier

\subsection{Plots of Multivariate ``Batch Size'' Extrapolation Results}
\label{subsection:extrapolation_batch_size}

\FloatBarrier

\begin{figure*}[h]    \centering
\includegraphics[width=0.99\textwidth]{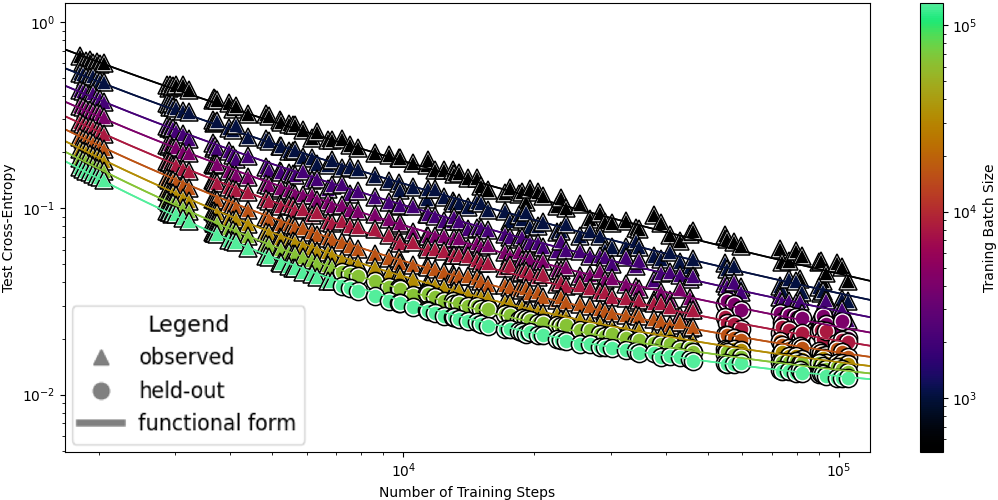}
    \caption{
    Extrapolation Results of UNSL on multivariate scaling behavior as batch size and number of training steps simultaneously vary. Scaling behavior is that of an MLP trained for a single epoch on dataset of \cite{greydanus2024scaling}. See Section \ref{subsection:extrapolation_batch_size} for more details.
    }
    \label{fig:batch_size}
\end{figure*}

\FloatBarrier


In Figure \ref{fig:batch_size}, UNSL accurately extrapolates multivariate scaling behavior when batch size is an input dimension to UNSL.


\FloatBarrier

\vspace{-2.9mm}

\newpage

\subsection{Plots of Vision Extrapolation Results}
\label{subsection:extrapolation_benchmark_vision}

\vspace{-2.9mm}

\subsubsection{Bivariate}
\label{subsubsection:extrapolation_benchmark_vision_bivariate}

\FloatBarrier

\vspace{-3.9mm}
\begin{figure*}[h]    \centering

\includegraphics[width=0.3938\textwidth]{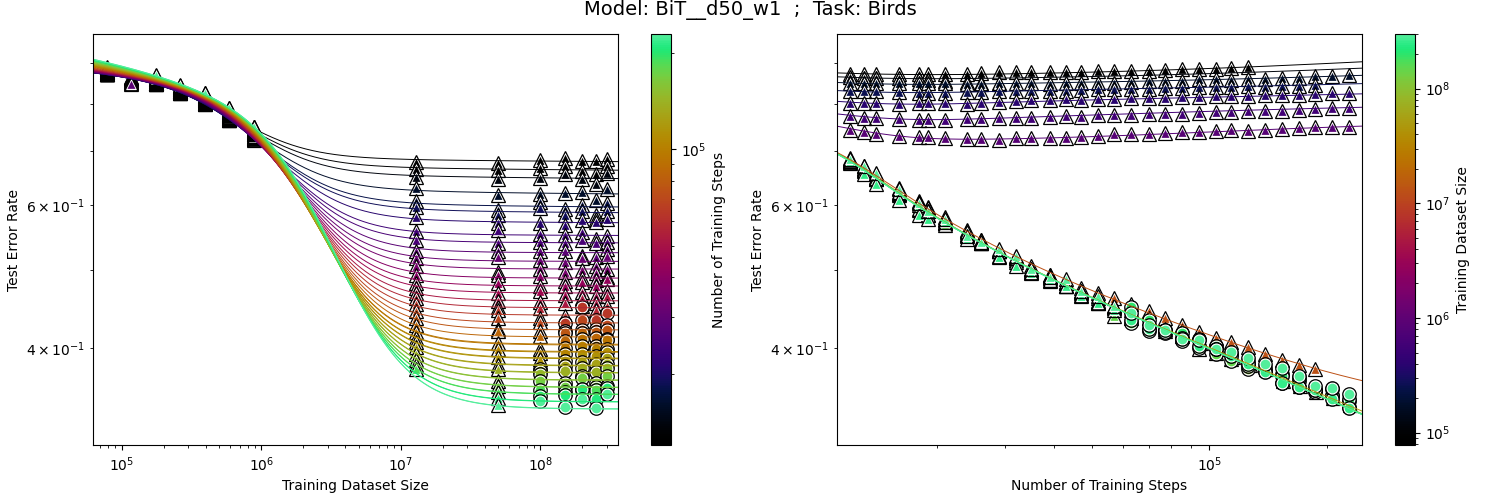}
\includegraphics[width=0.3938\textwidth]{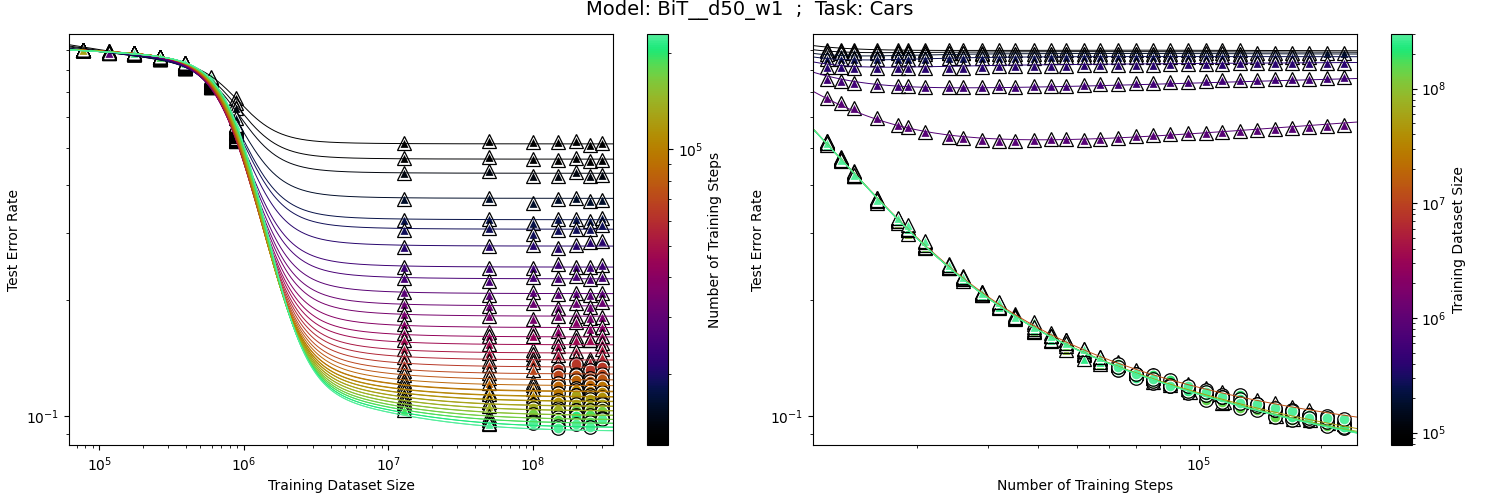}
\includegraphics[width=0.3938\textwidth]{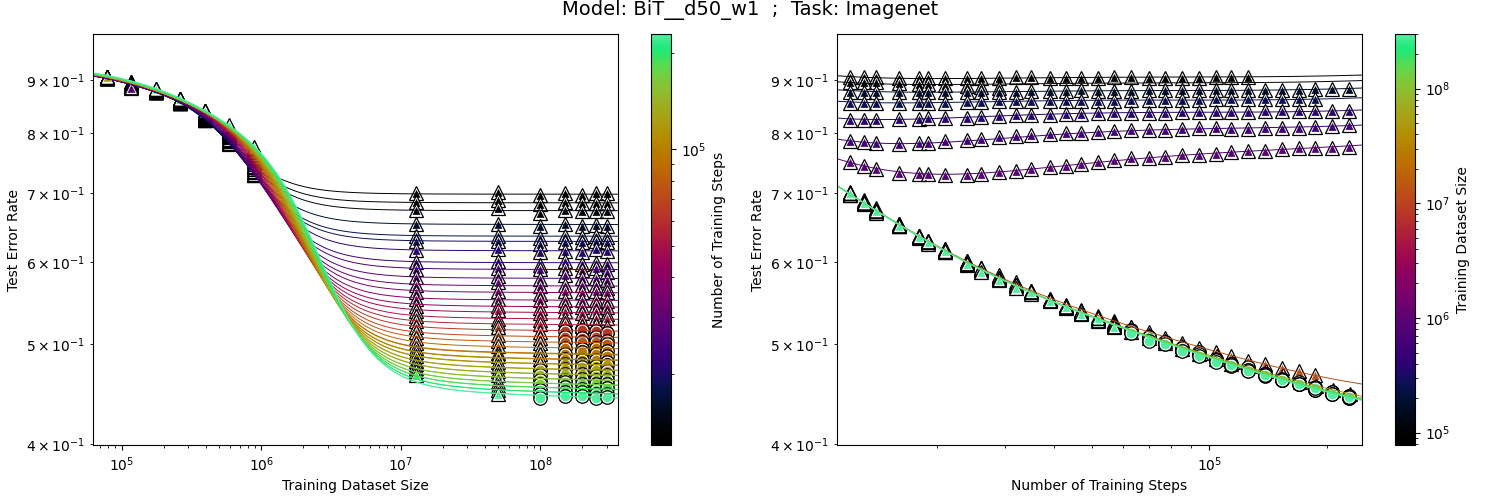}
\includegraphics[width=0.3938\textwidth]{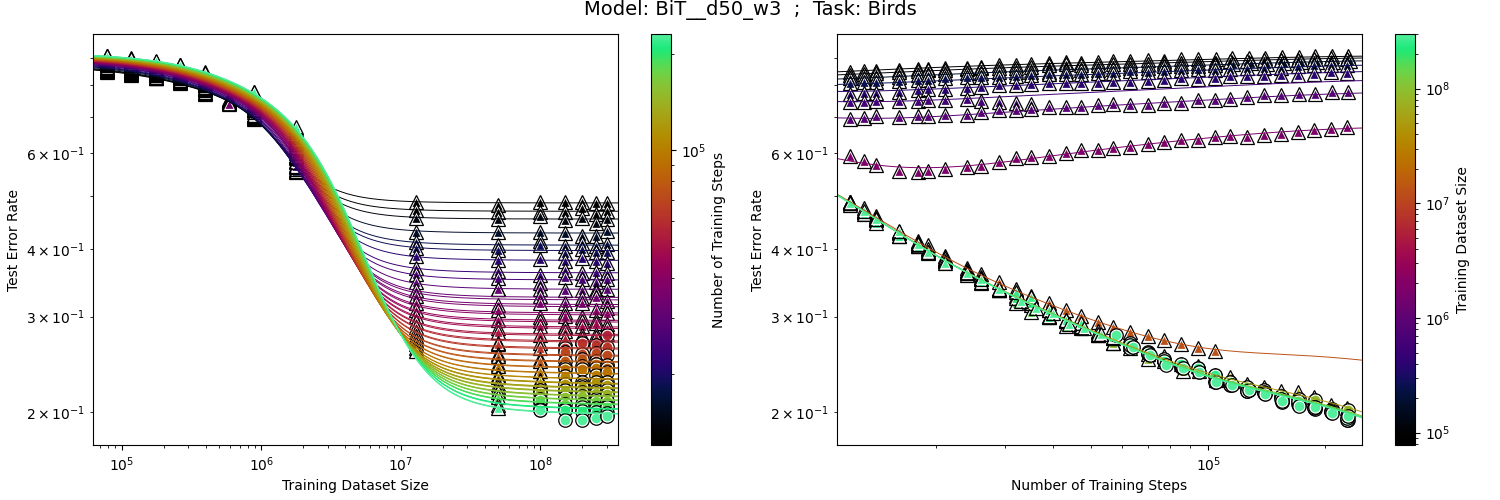}
\includegraphics[width=0.3938\textwidth]{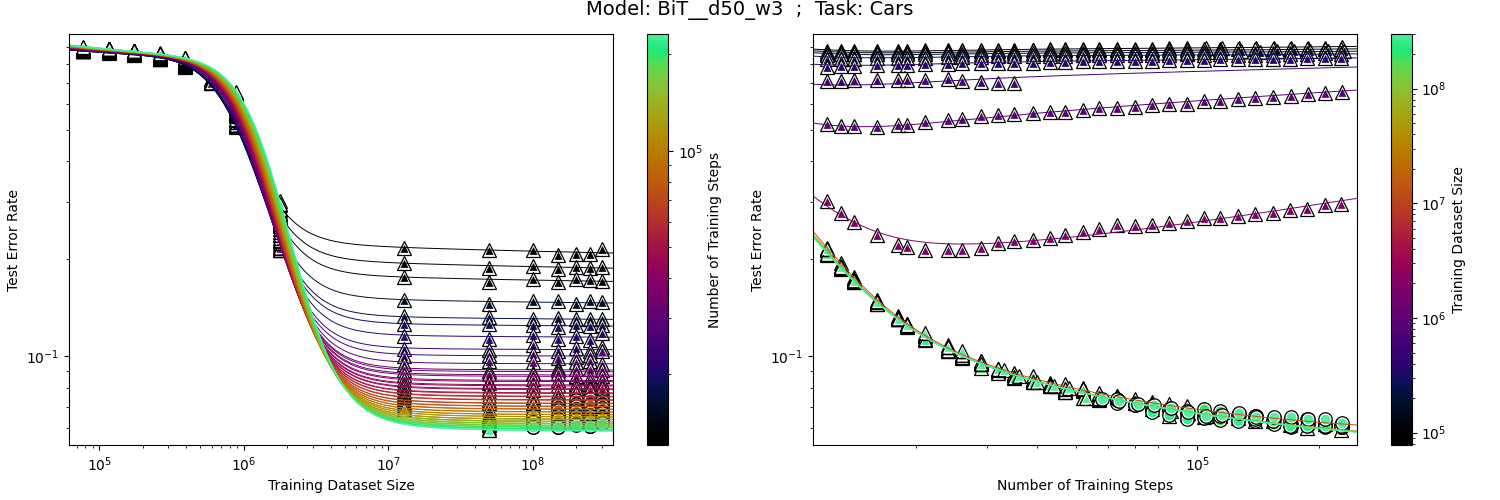}
\includegraphics[width=0.3938\textwidth]{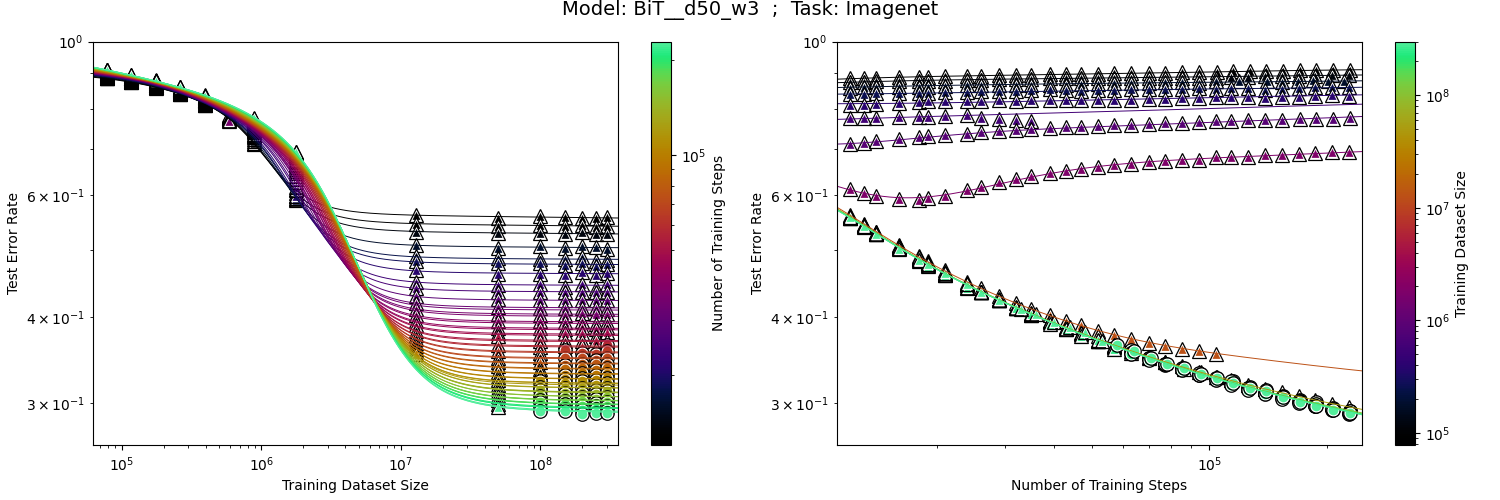}
\includegraphics[width=0.3938\textwidth]{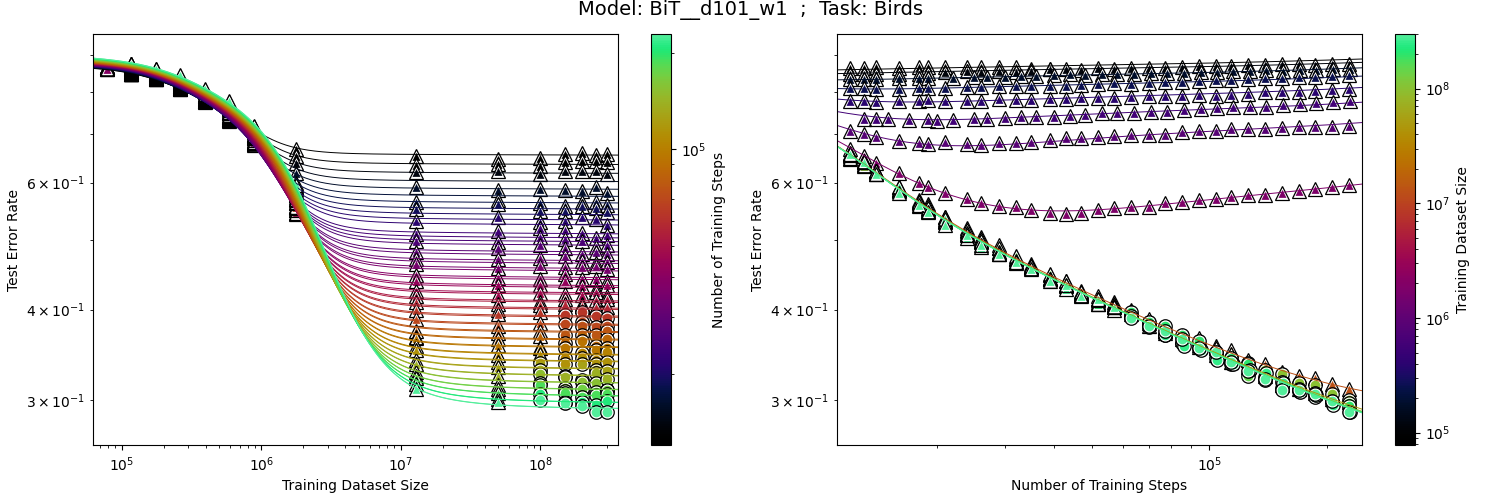}
\includegraphics[width=0.3938\textwidth]{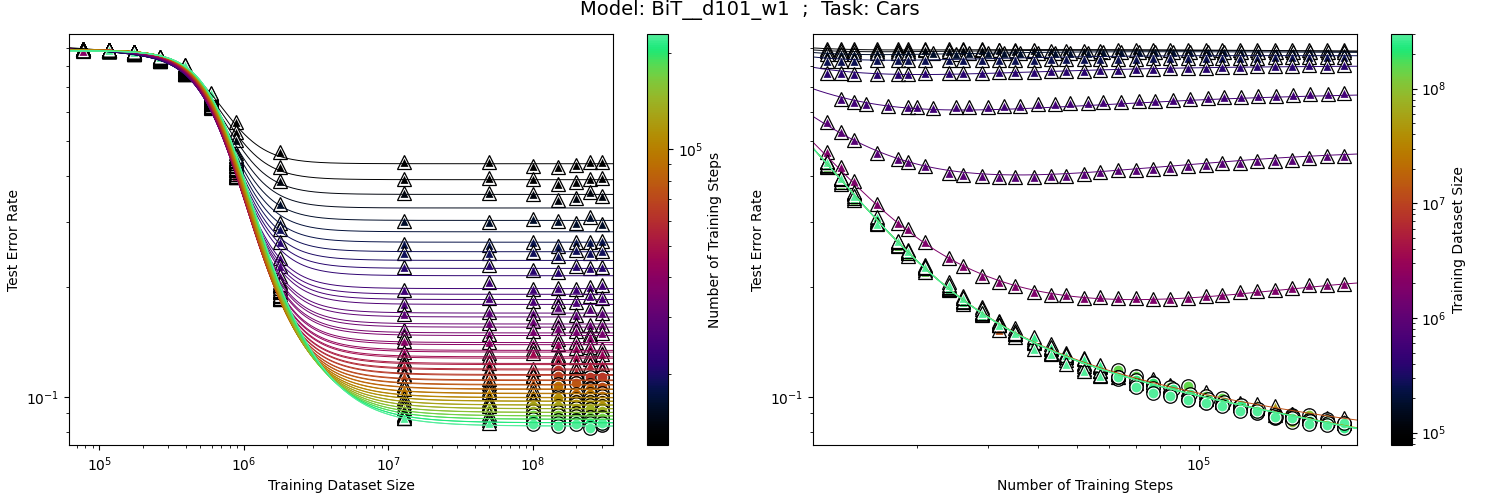}
\includegraphics[width=0.3938\textwidth]{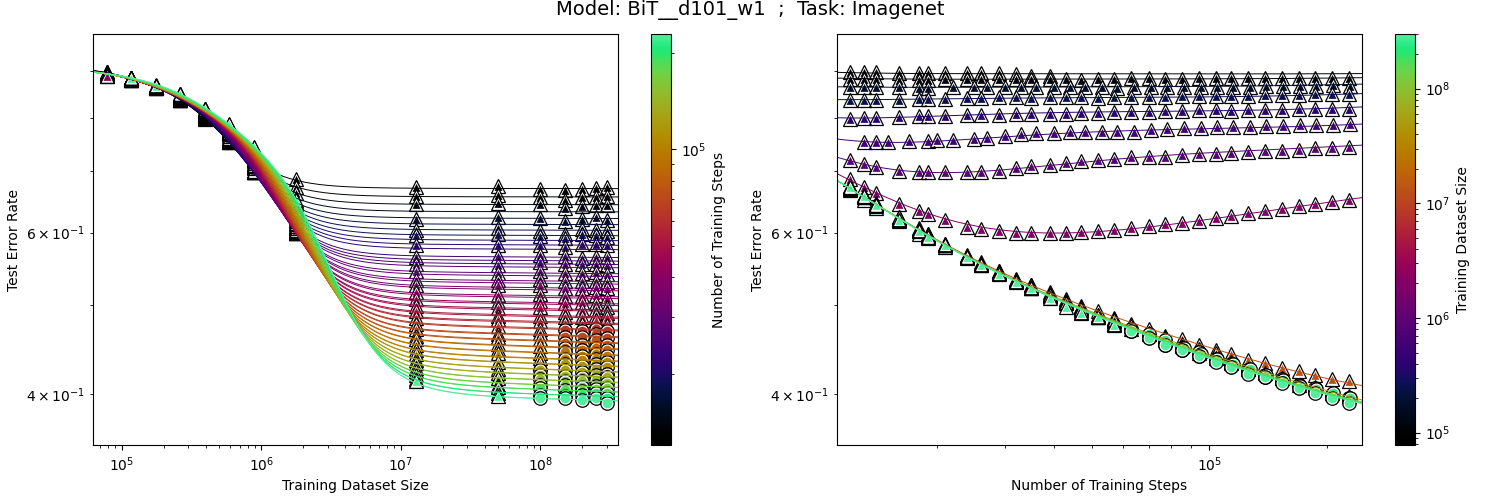}
\includegraphics[width=0.3938\textwidth]{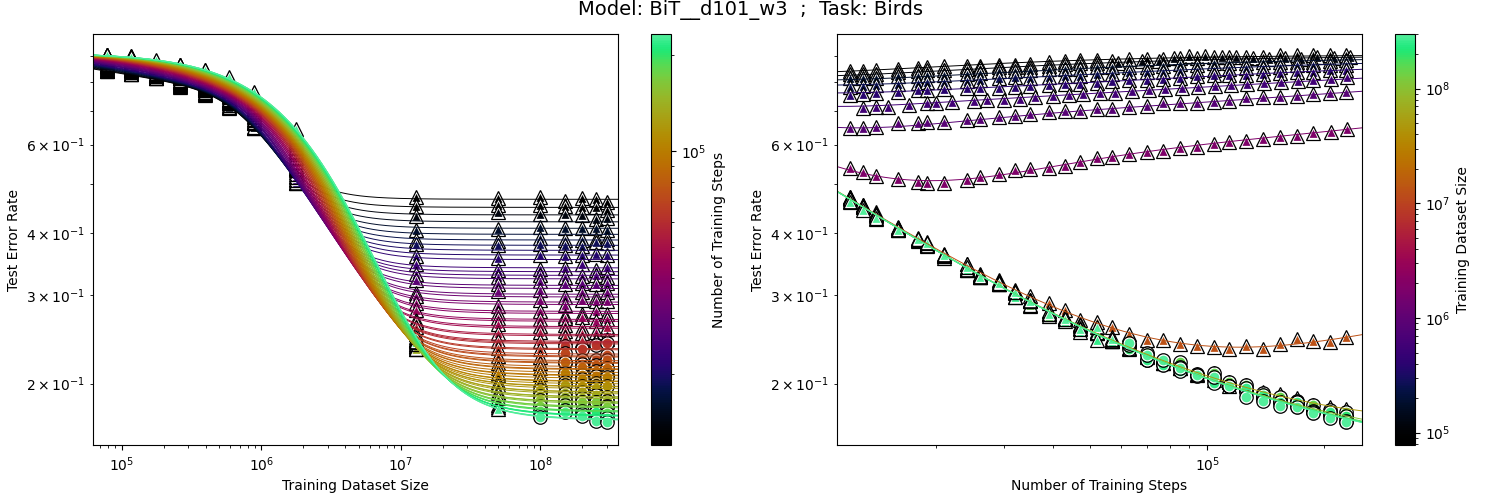}
\includegraphics[width=0.3938\textwidth]{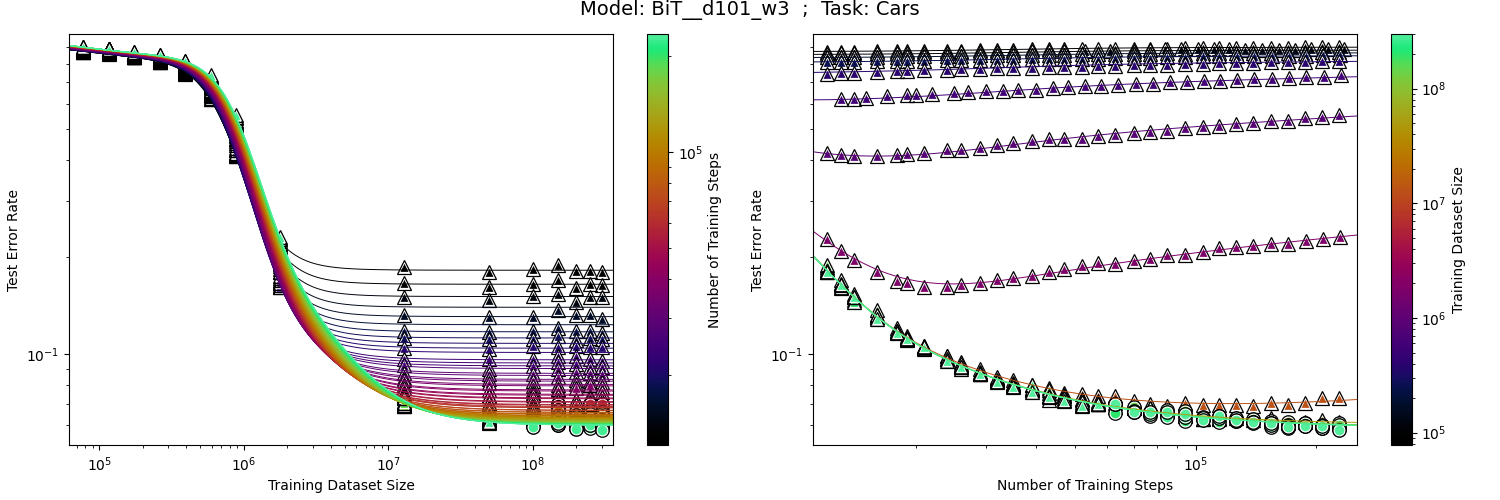}
\includegraphics[width=0.3938\textwidth]{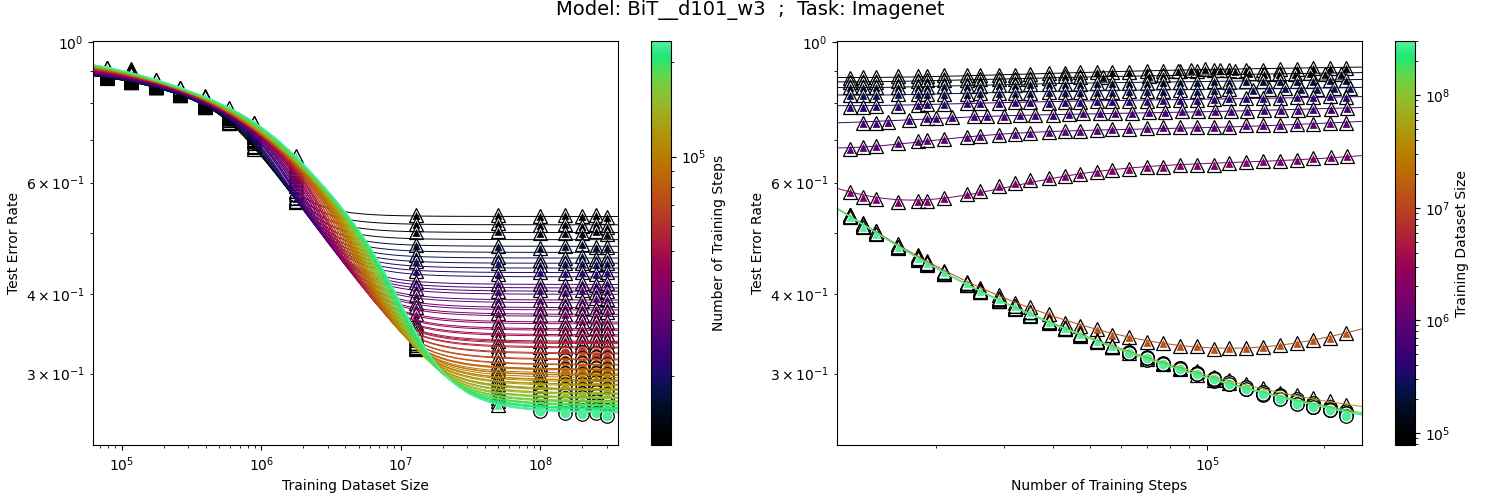}
\includegraphics[width=0.3938\textwidth]{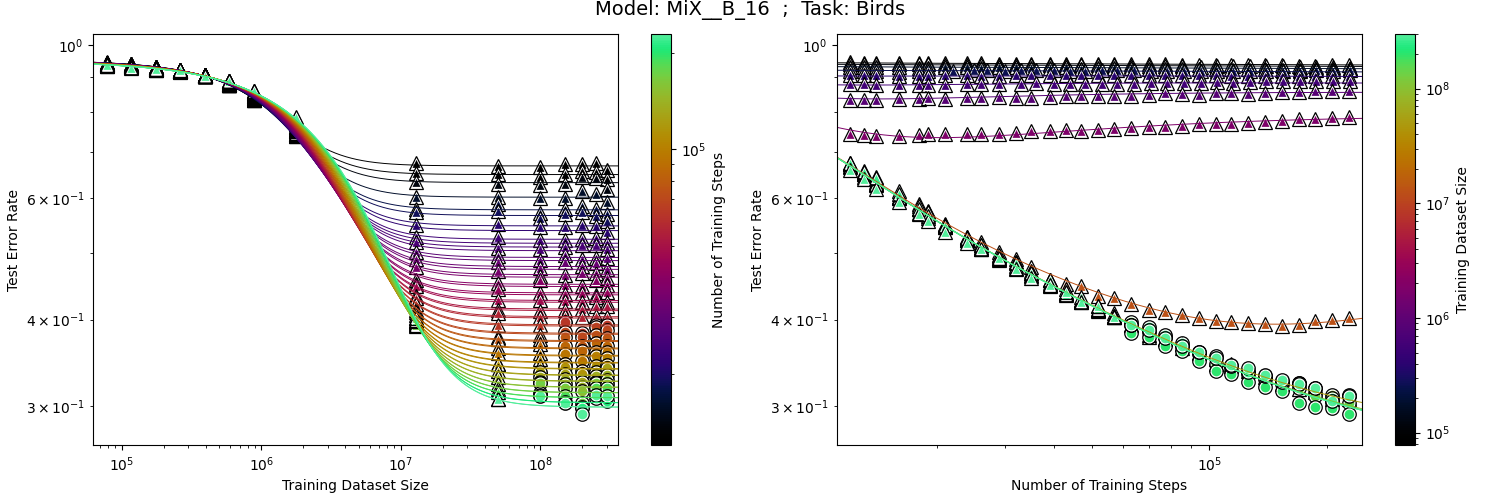}
\includegraphics[width=0.3938\textwidth]{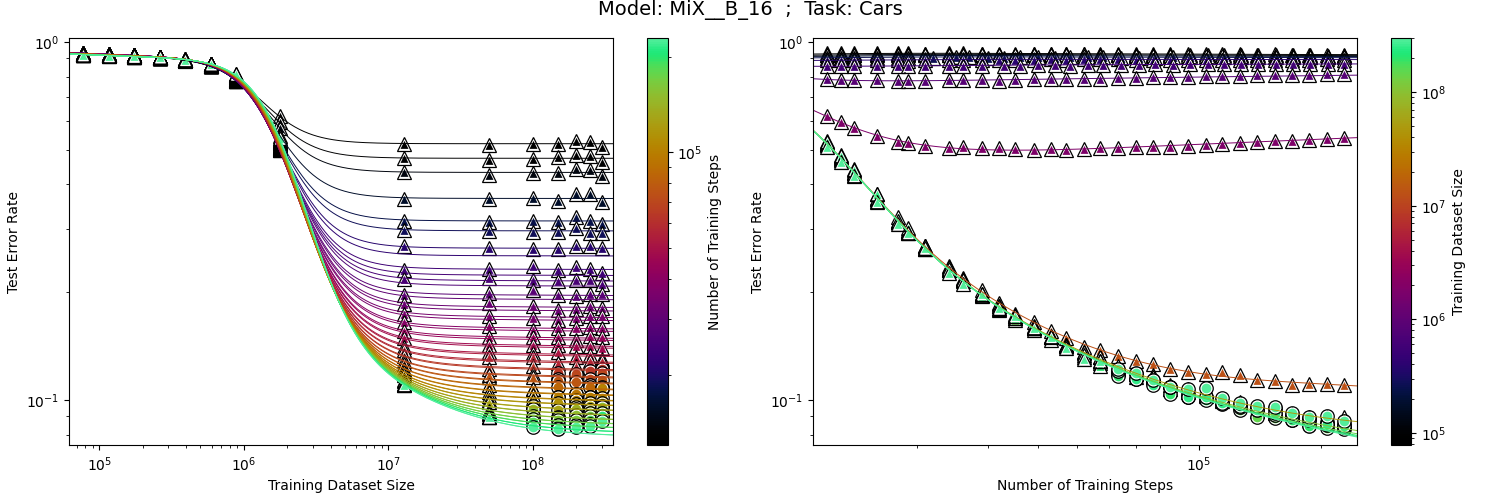}
\includegraphics[width=0.3938\textwidth]{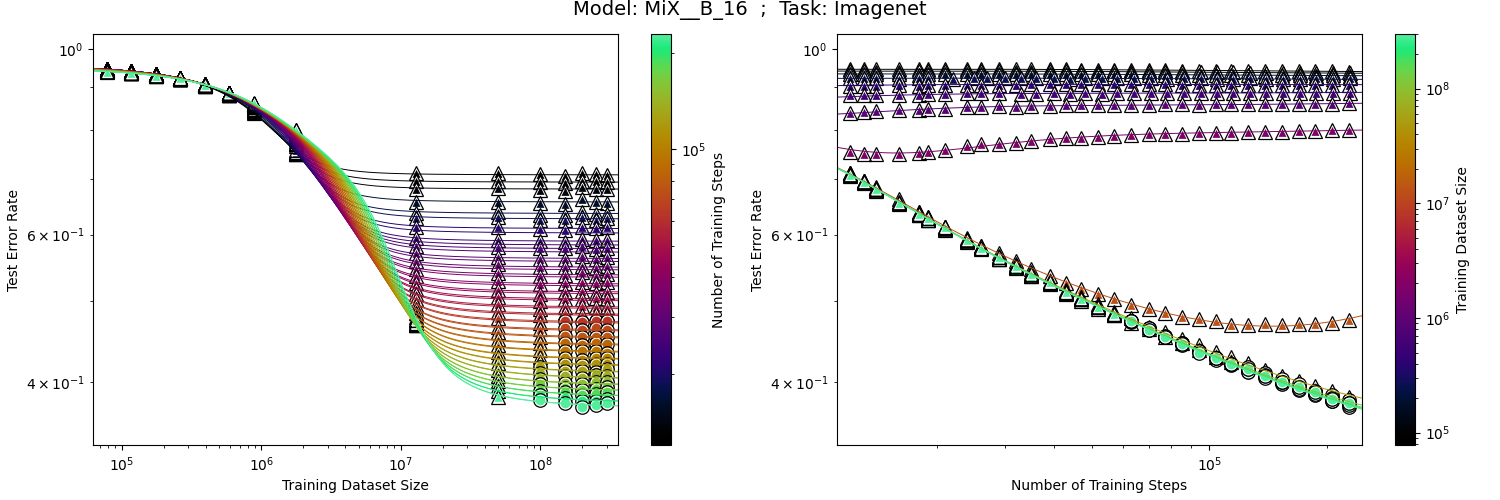}
\includegraphics[width=0.3938\textwidth]{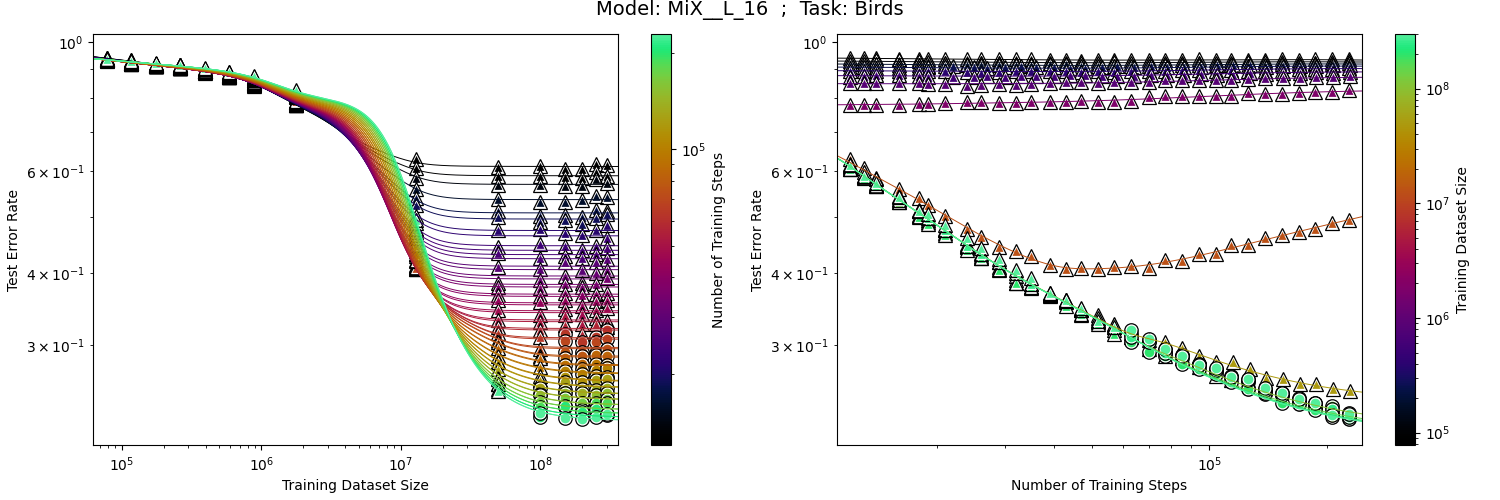}
\includegraphics[width=0.3938\textwidth]{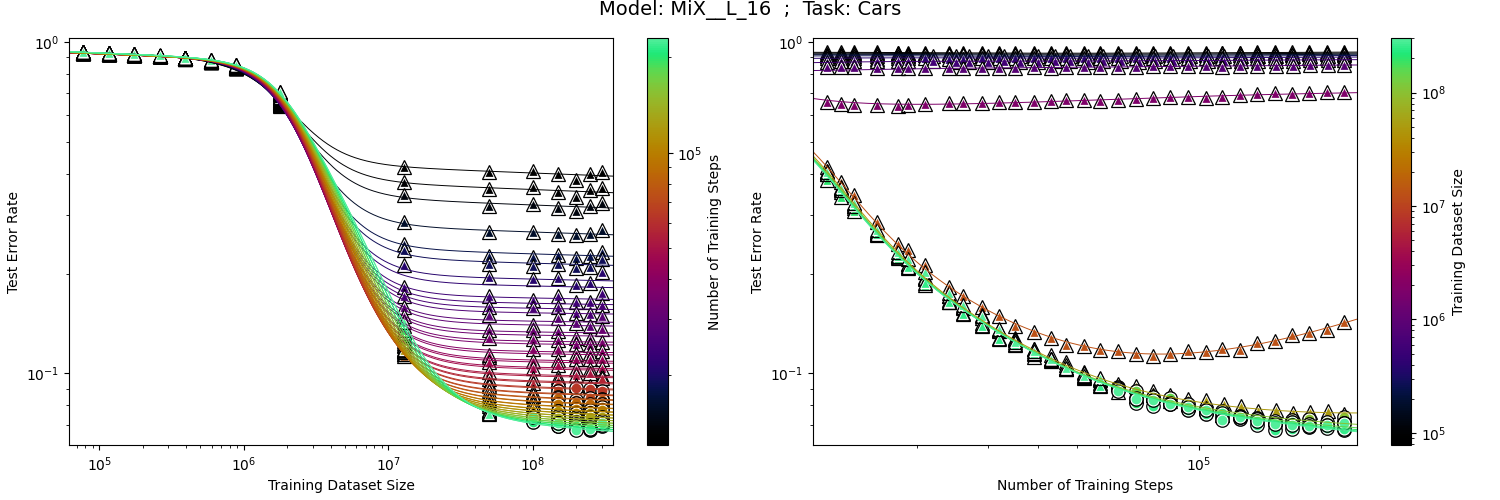}
\includegraphics[width=0.3938\textwidth]{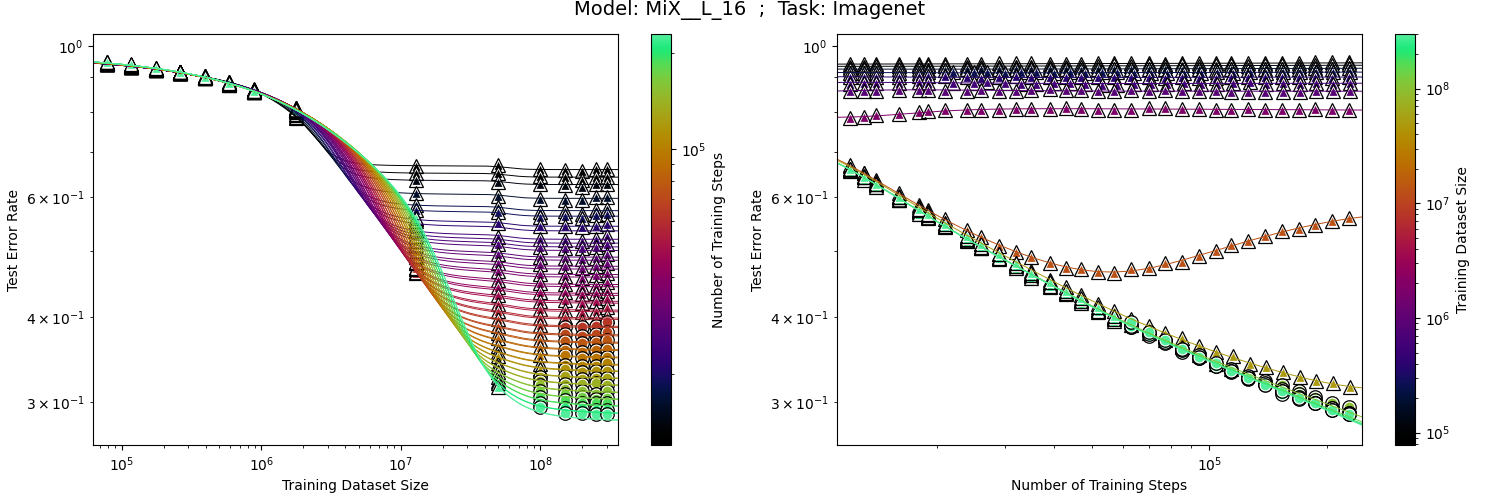}
\includegraphics[width=0.3938\textwidth]{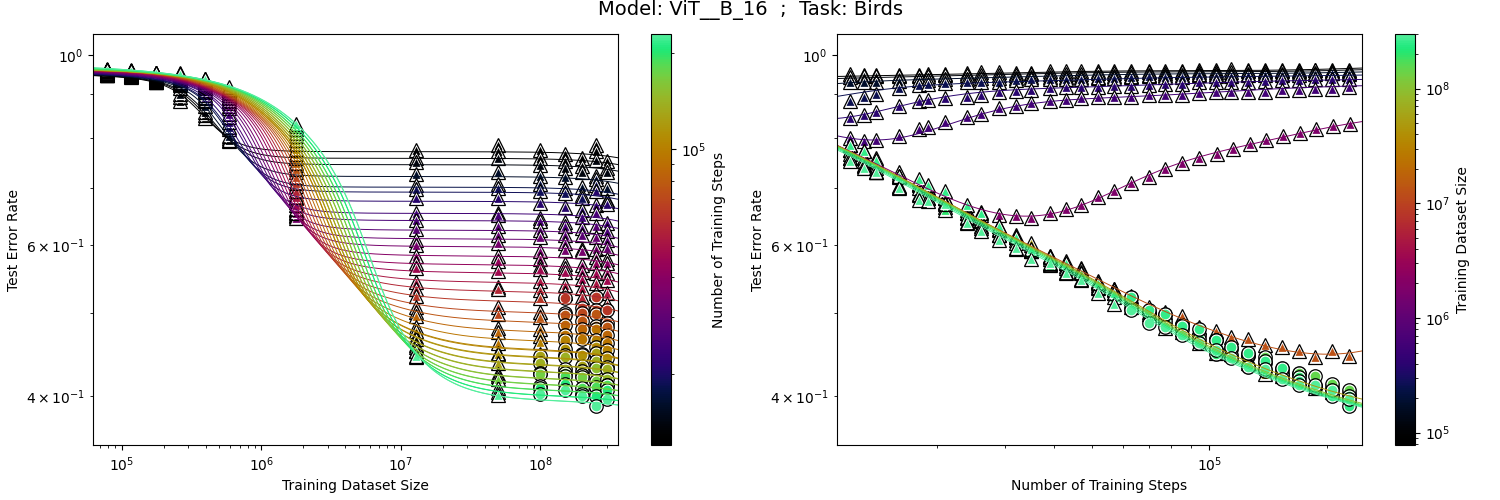}
\includegraphics[width=0.3938\textwidth]{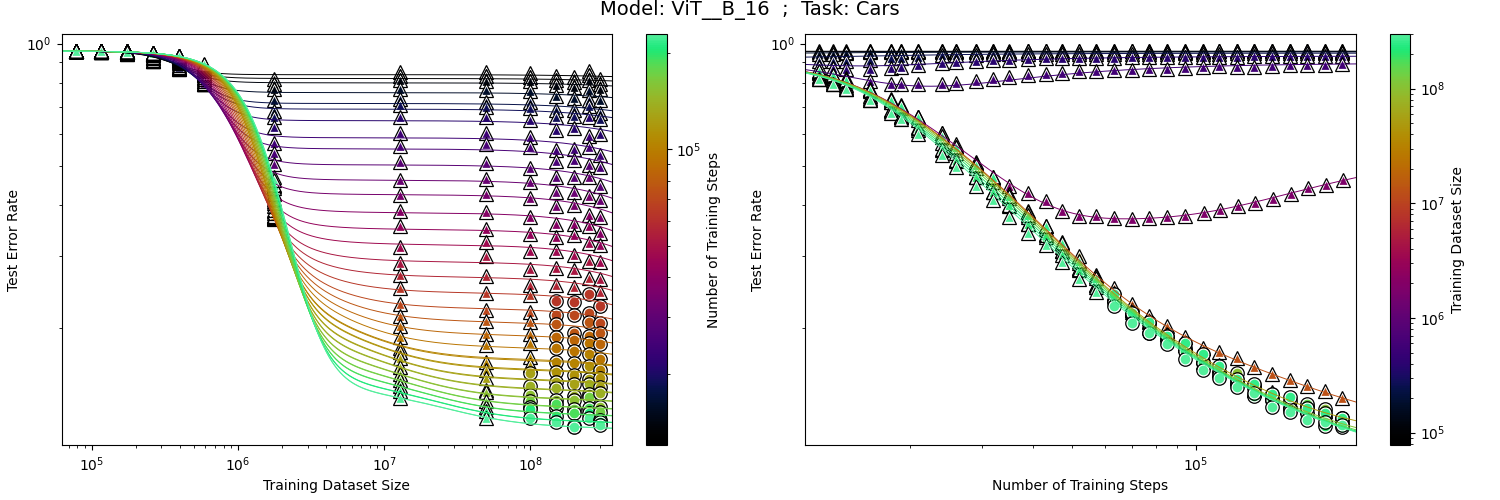}
\includegraphics[width=0.3938\textwidth]{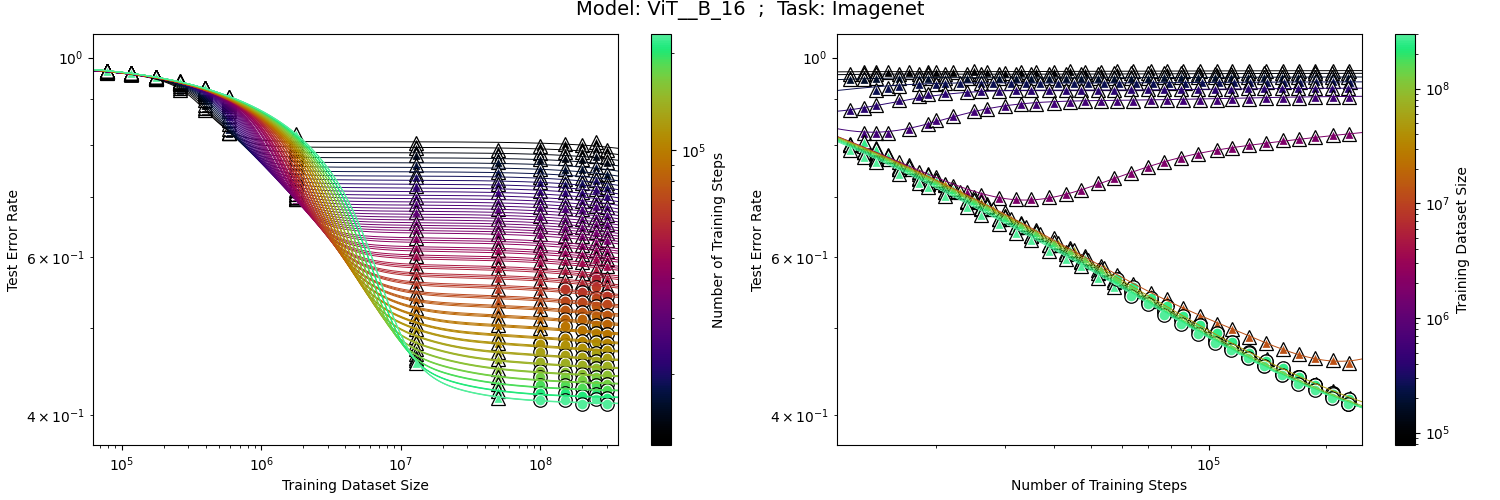}
\hspace{29.0mm}\includegraphics[width=0.179\textwidth]{figures/legend/legend2.png}
\vspace{-4.3mm}
    \caption{
    Extrapolation Results of UNSL on bivariate scaling behavior of downstream vision performance. See Section \ref{section:vision} for more details.
    }
    \label{fig:unsl_downstream_vision}
\end{figure*}

\FloatBarrier
\begin{figure*}[h]    \centering

\includegraphics[width=0.3938\textwidth]{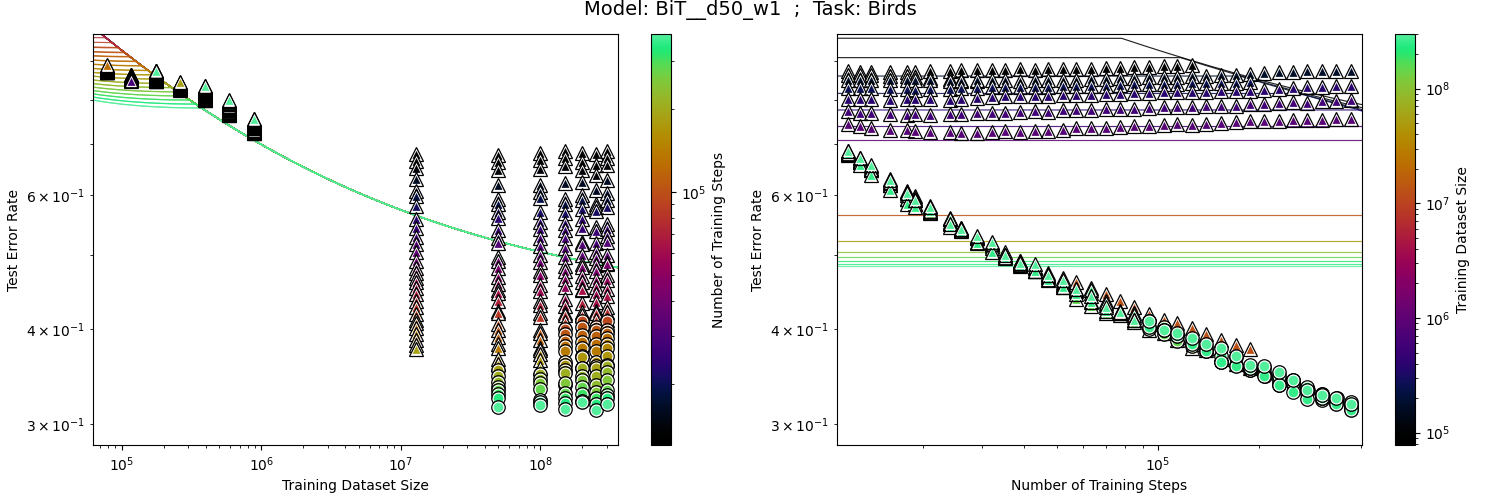}
\includegraphics[width=0.3938\textwidth]{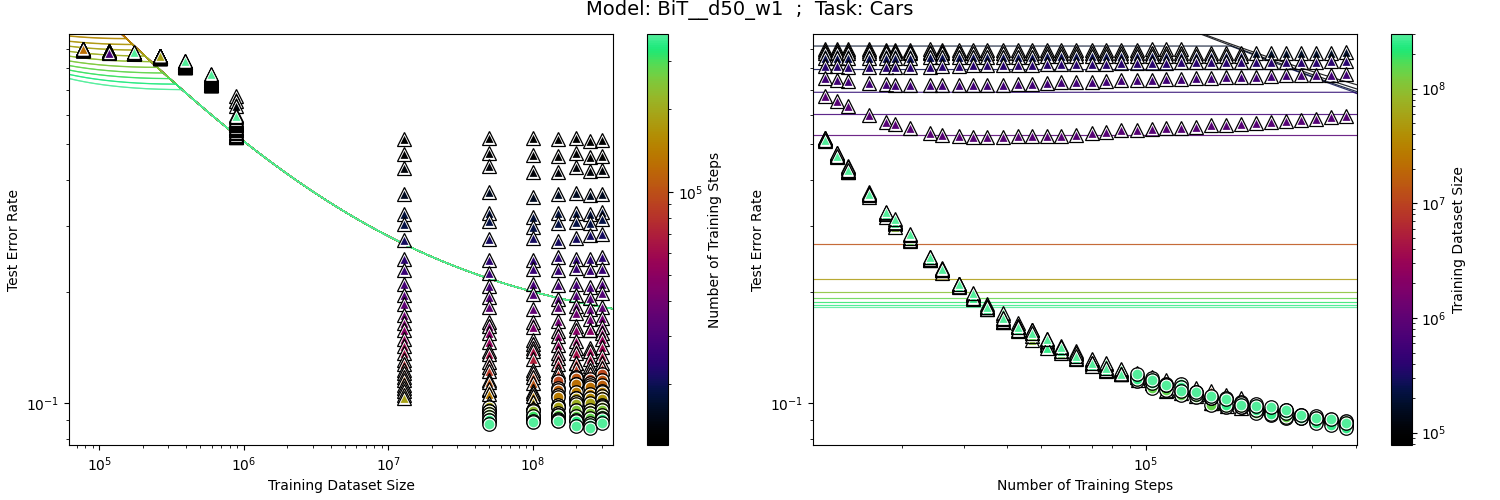}
\includegraphics[width=0.3938\textwidth]{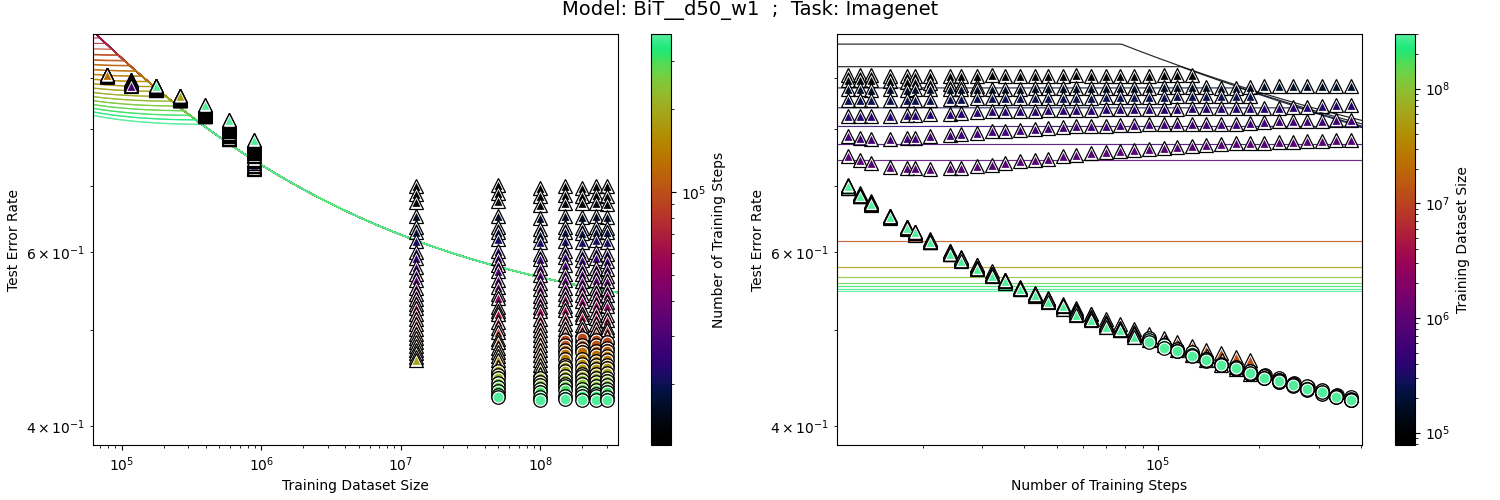}
\includegraphics[width=0.3938\textwidth]{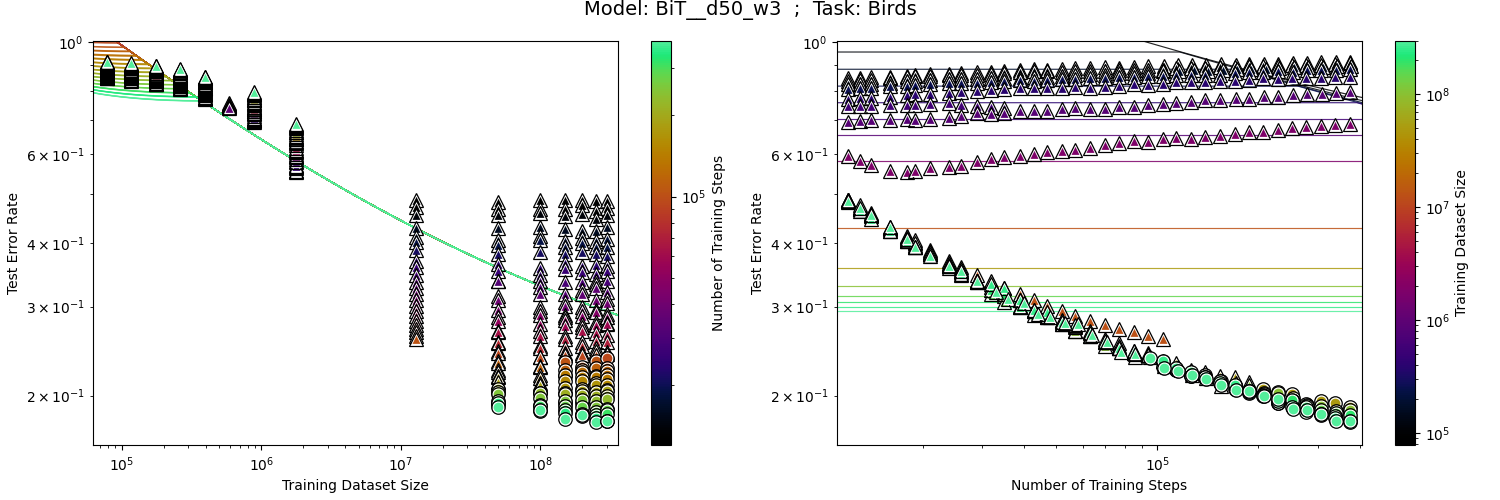}
\includegraphics[width=0.3938\textwidth]{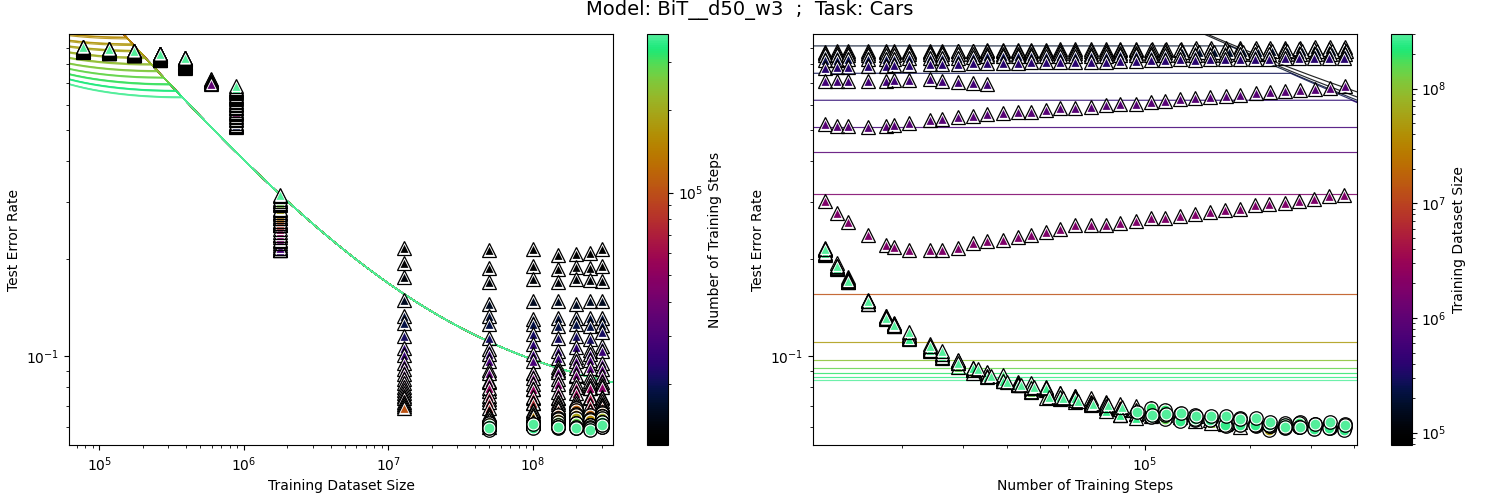}
\includegraphics[width=0.3938\textwidth]{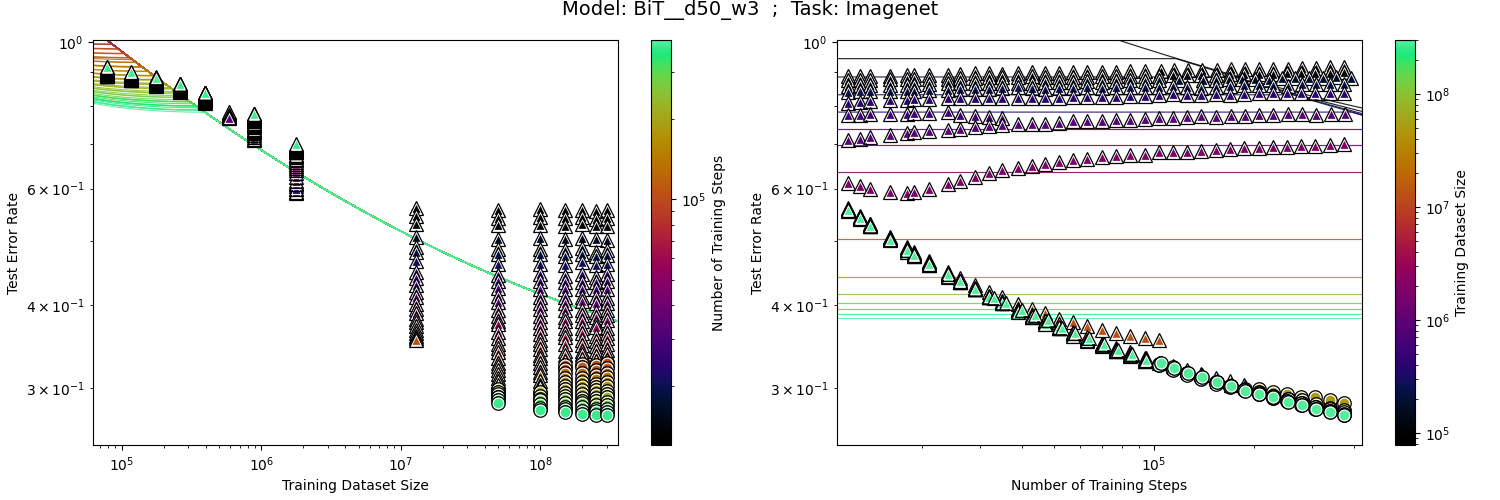}
\includegraphics[width=0.3938\textwidth]{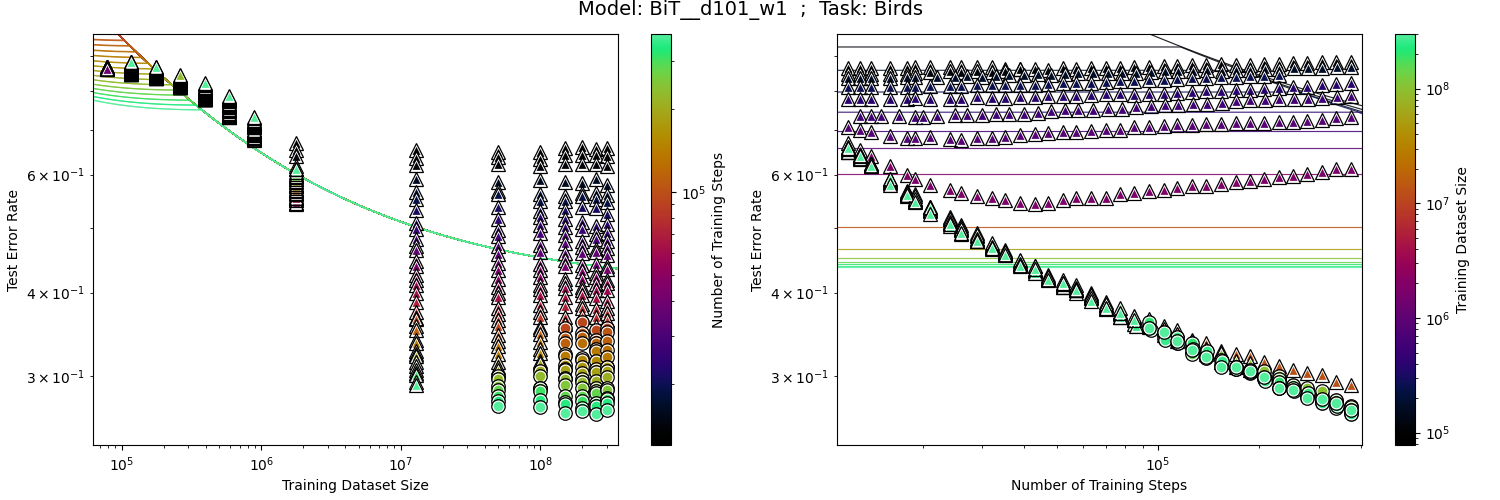}
\includegraphics[width=0.3938\textwidth]{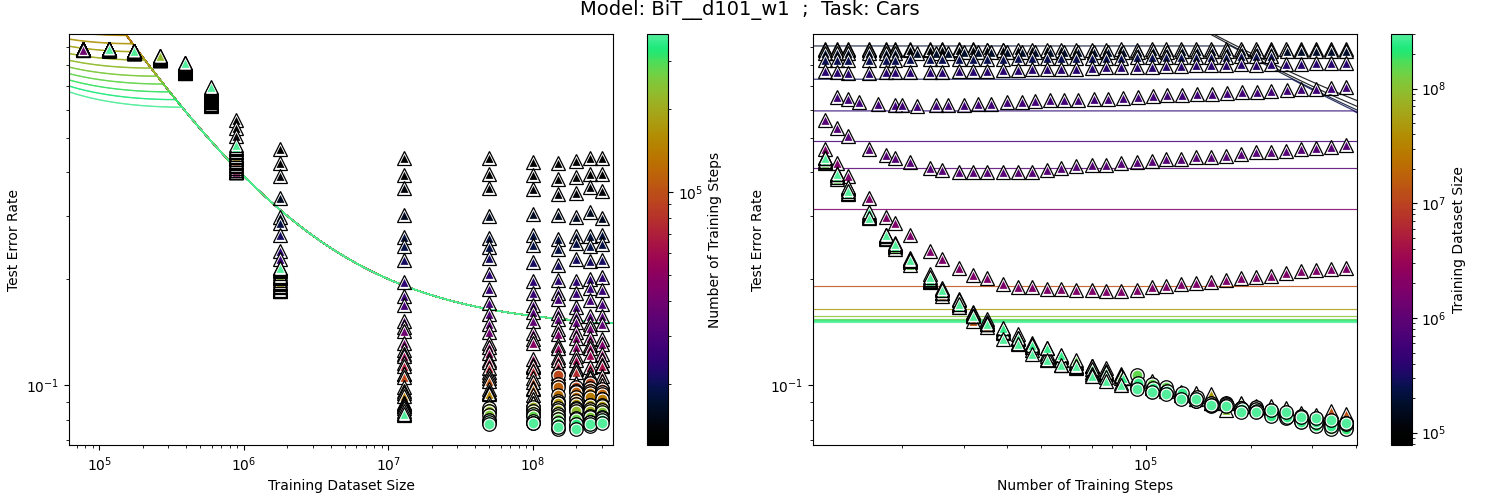}
\includegraphics[width=0.3938\textwidth]{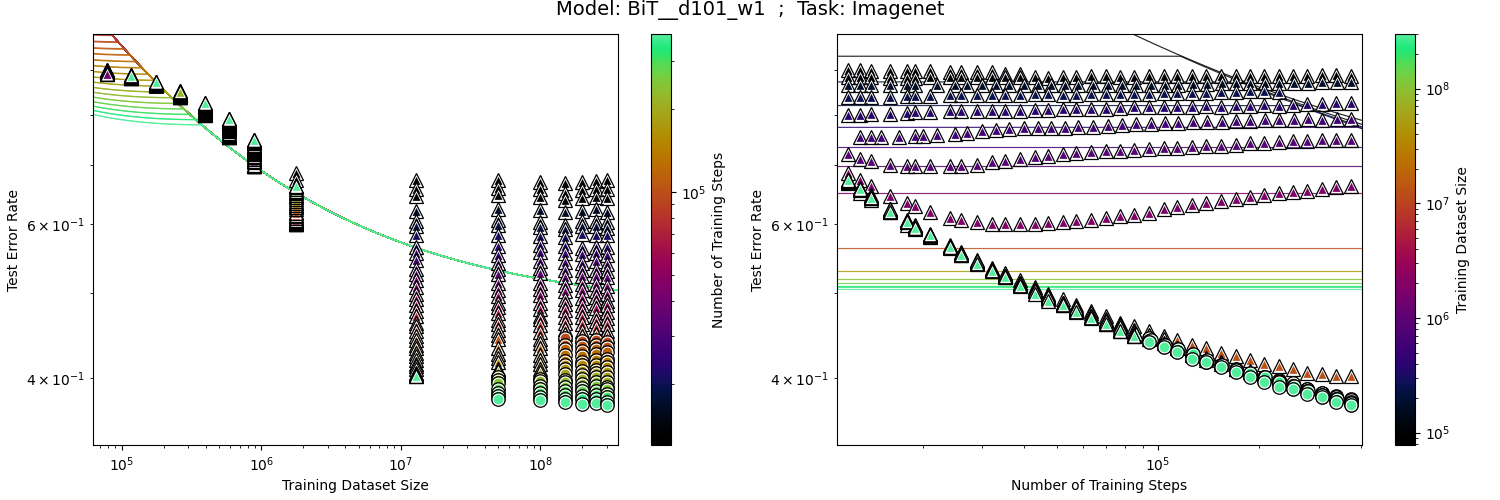}
\includegraphics[width=0.3938\textwidth]{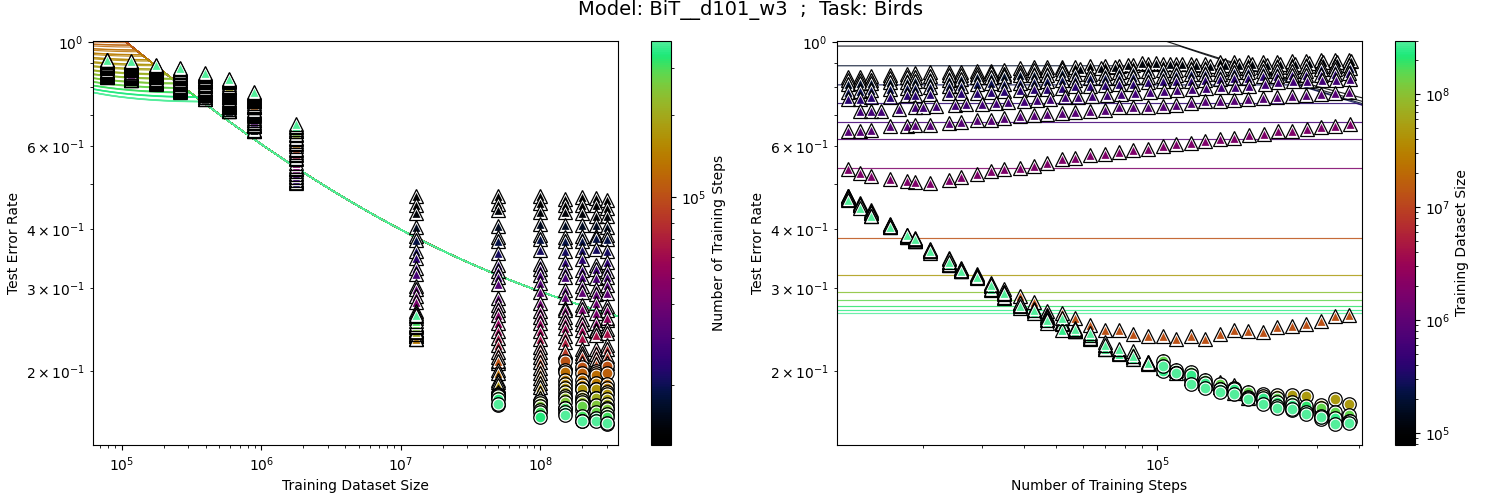}
\includegraphics[width=0.3938\textwidth]{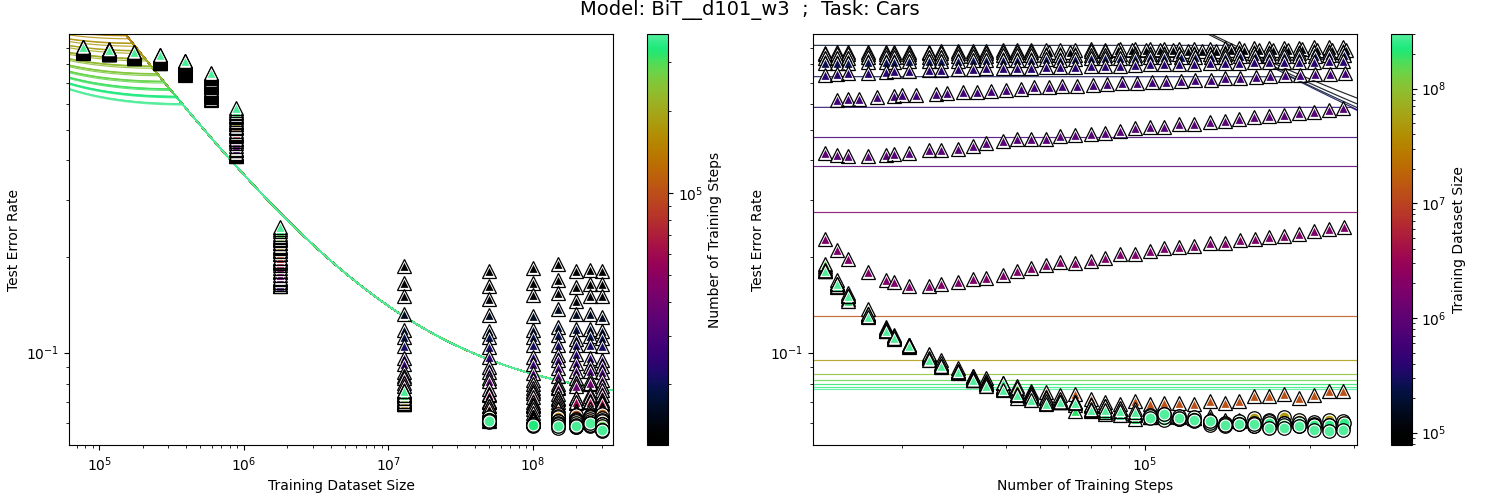}
\includegraphics[width=0.3938\textwidth]{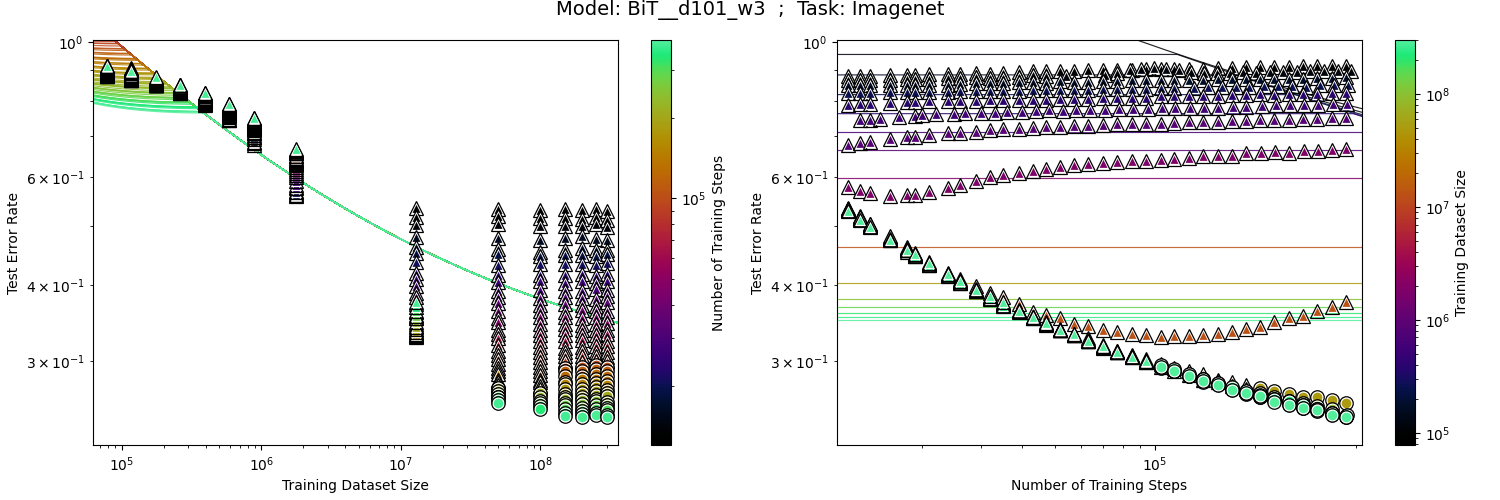}
\includegraphics[width=0.3938\textwidth]{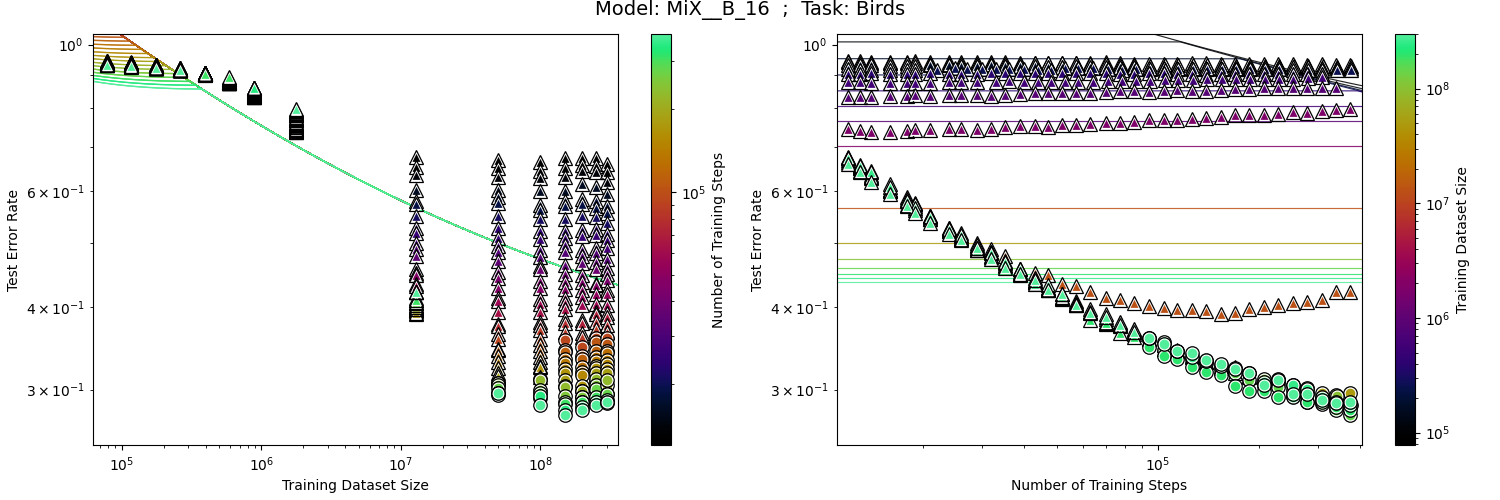}
\includegraphics[width=0.3938\textwidth]{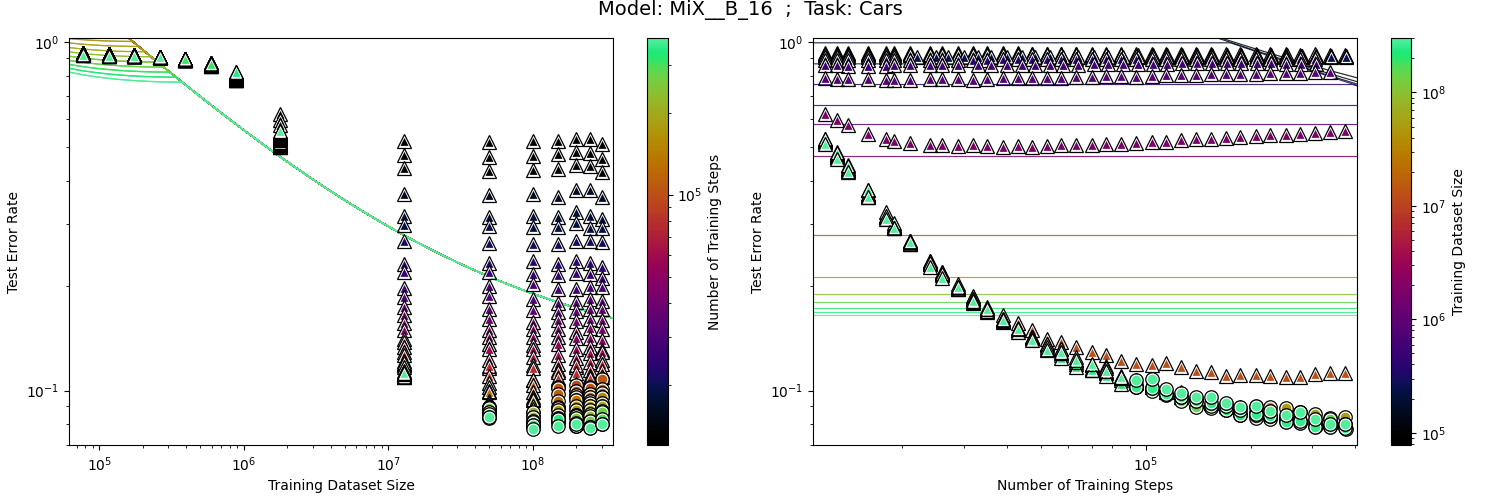}
\includegraphics[width=0.3938\textwidth]{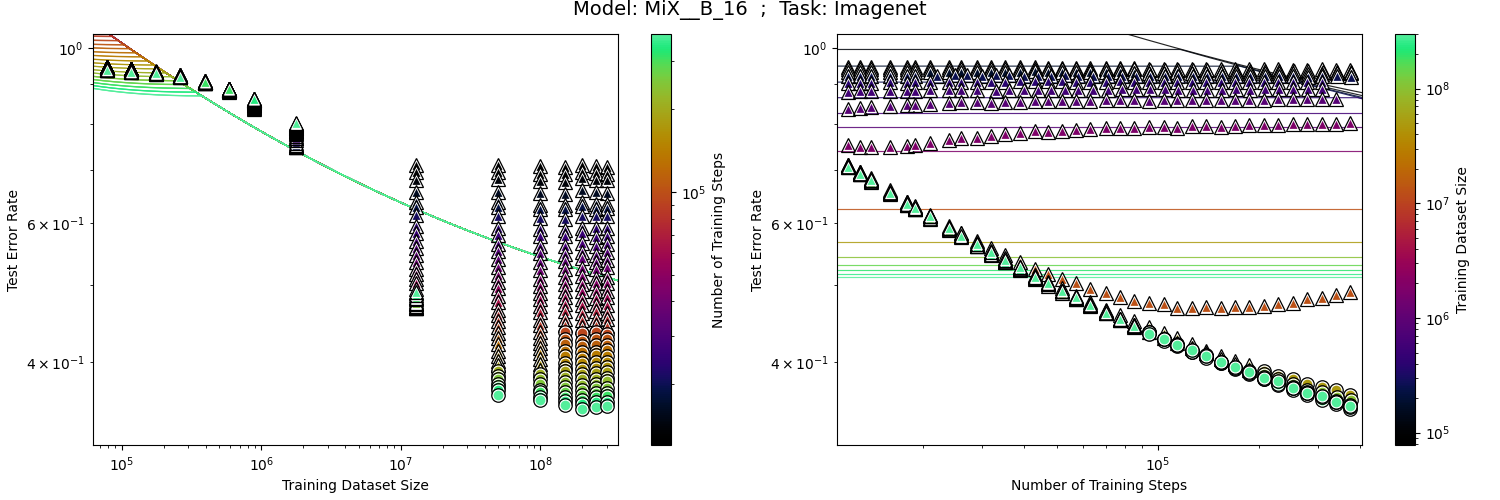}
\includegraphics[width=0.3938\textwidth]{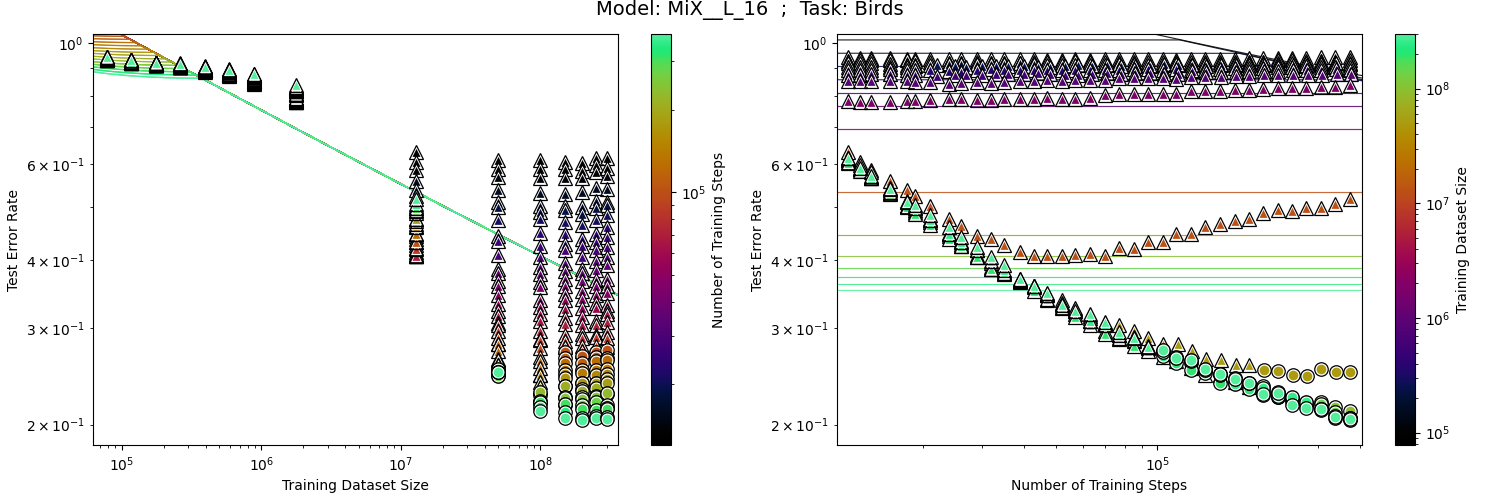}
\includegraphics[width=0.3938\textwidth]{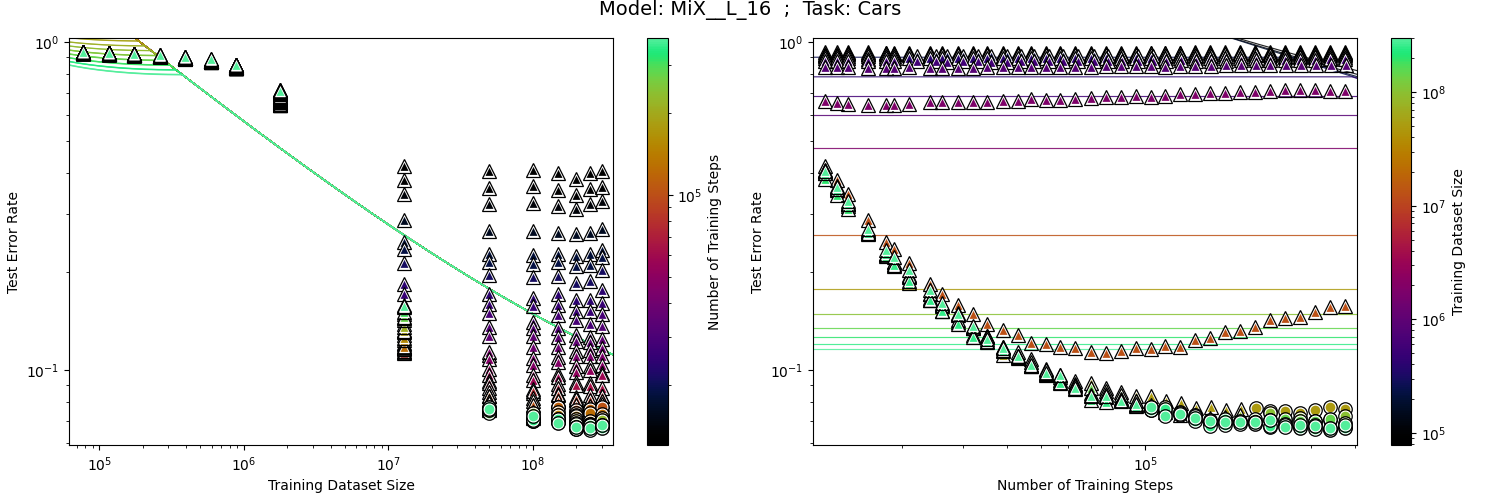}
\includegraphics[width=0.3938\textwidth]{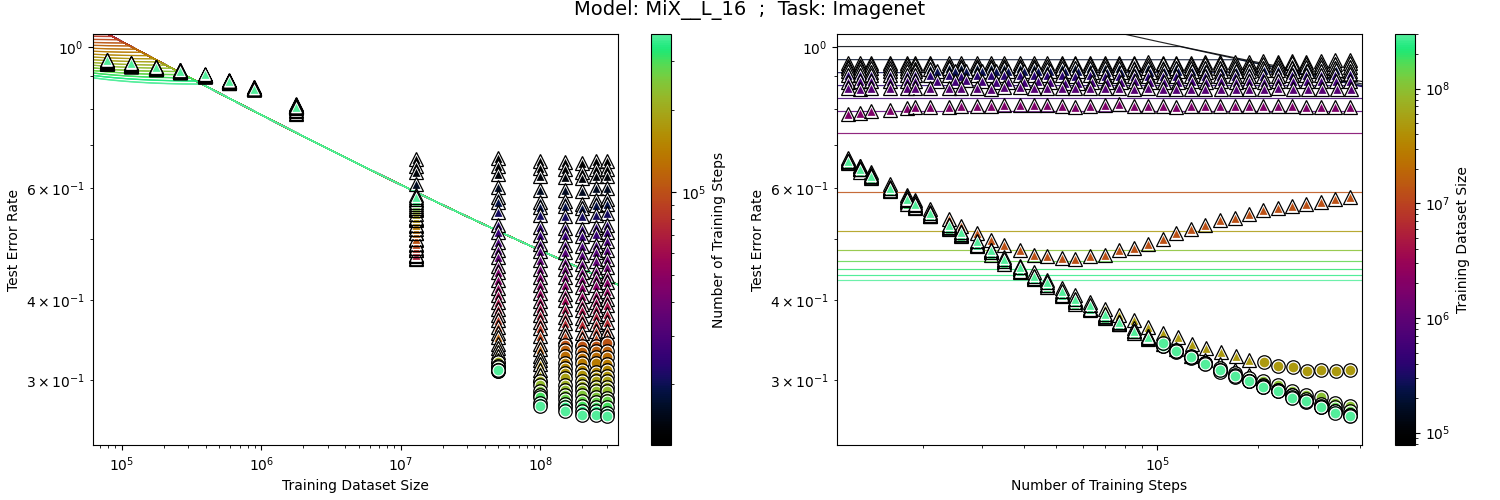}
\includegraphics[width=0.3938\textwidth]{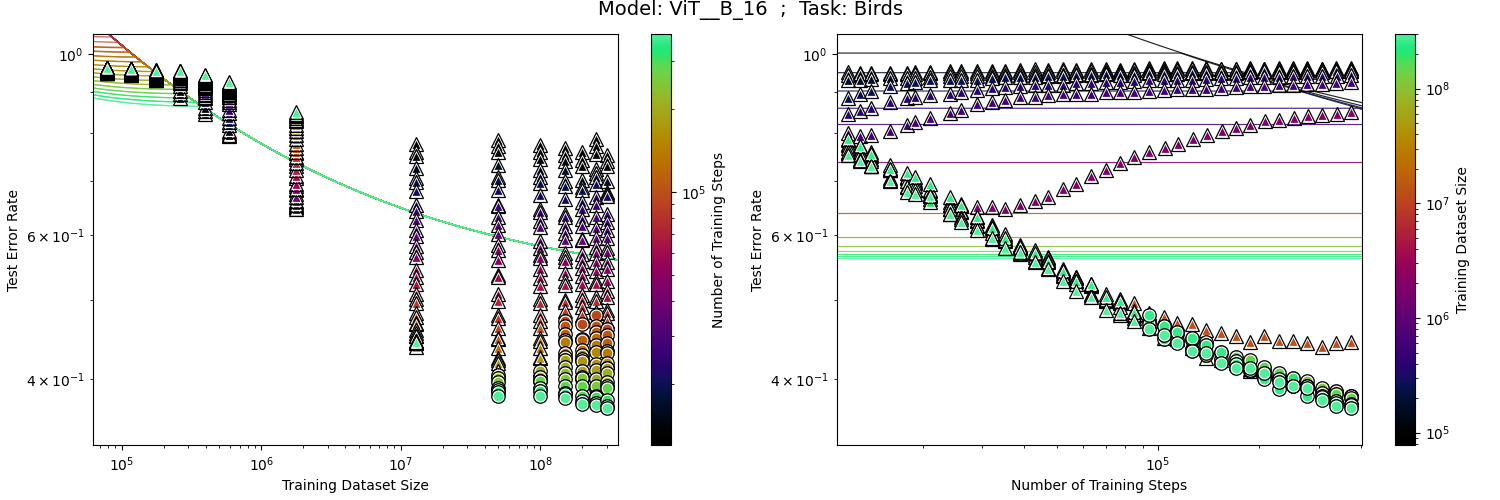}
\includegraphics[width=0.3938\textwidth]{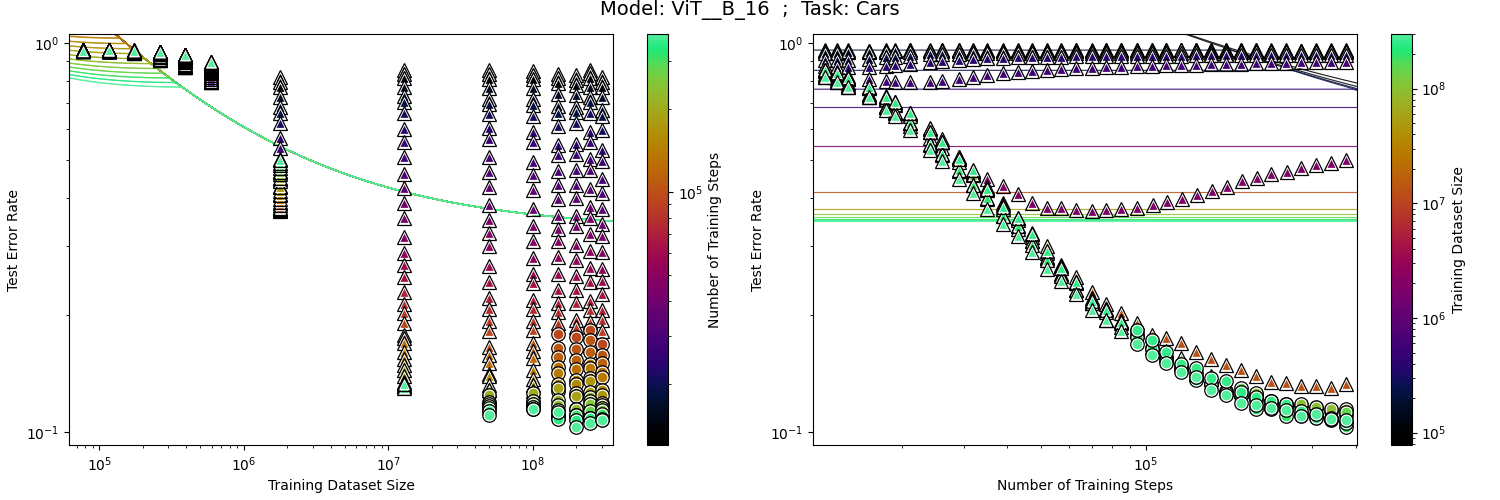}
\includegraphics[width=0.3938\textwidth]{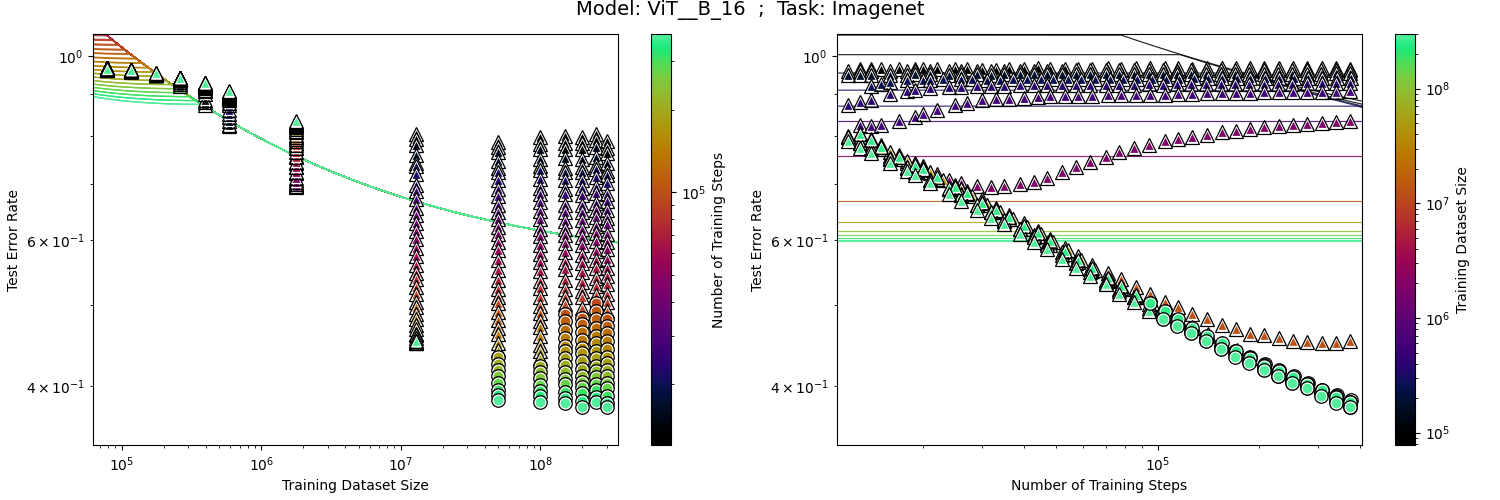}
\hspace{29.0mm}\includegraphics[width=0.179\textwidth]{figures/legend/legend2.png}
    \caption{
    Extrapolation Results of ``DC'' functional form of \citet{muennighoff2023scaling} on bivariate scaling behavior of downstream vision performance. See Section \ref{section:vision} for more details.
    }
    \label{fig:dc_downstream_vision}
\end{figure*}

\FloatBarrier
\begin{figure*}[h]    \centering

\includegraphics[width=0.3938\textwidth]{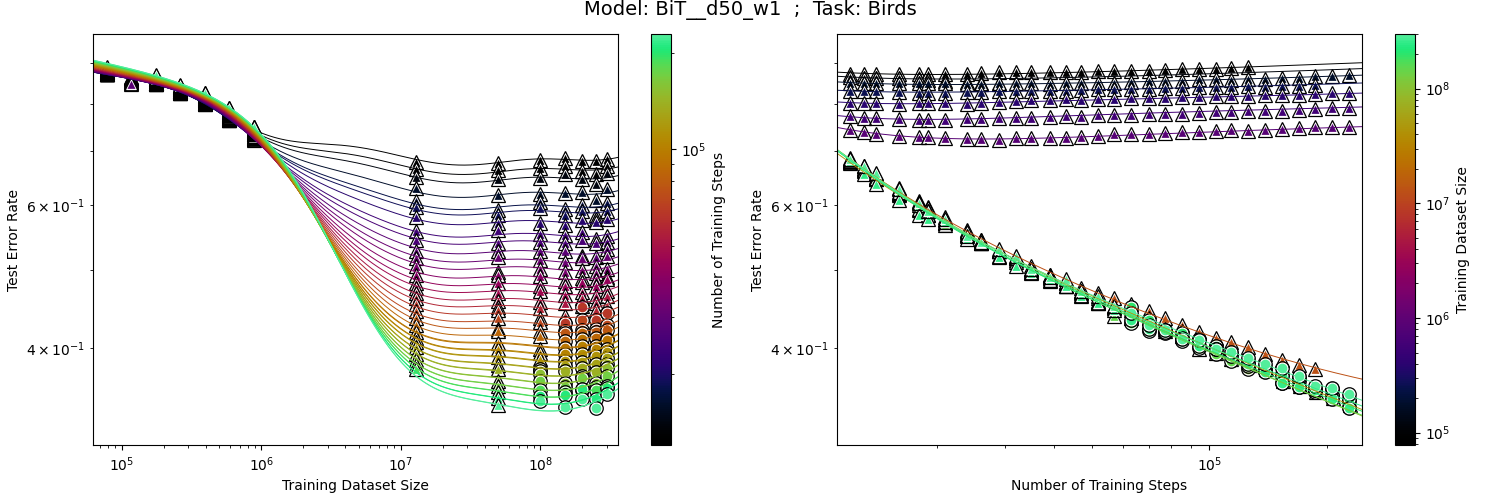}
\includegraphics[width=0.3938\textwidth]{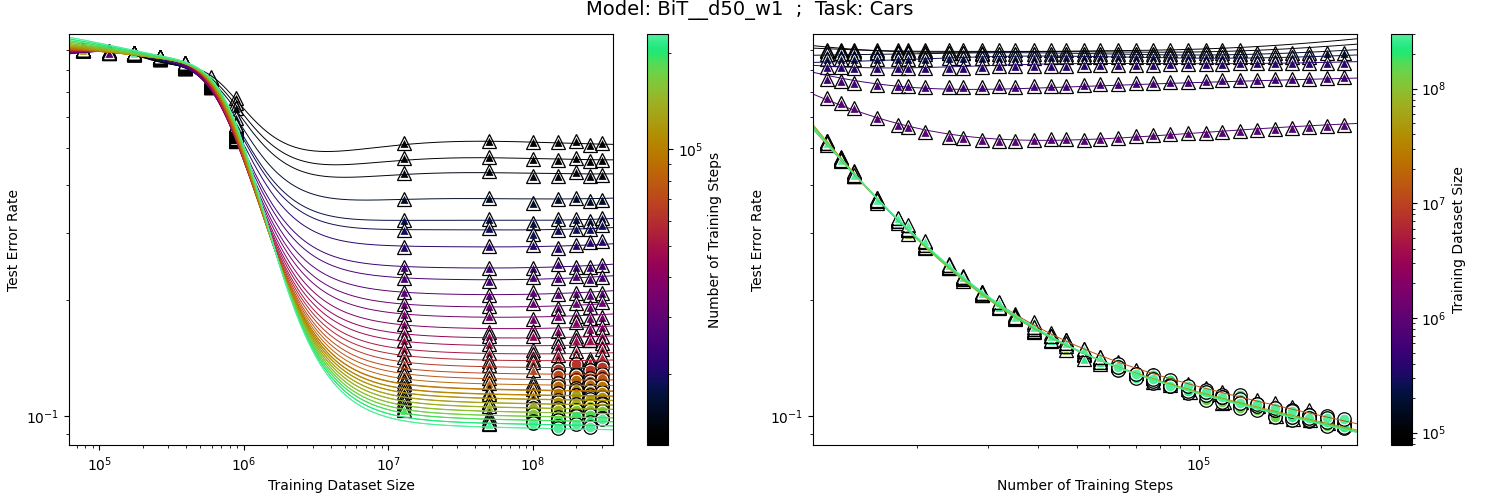}
\includegraphics[width=0.3938\textwidth]{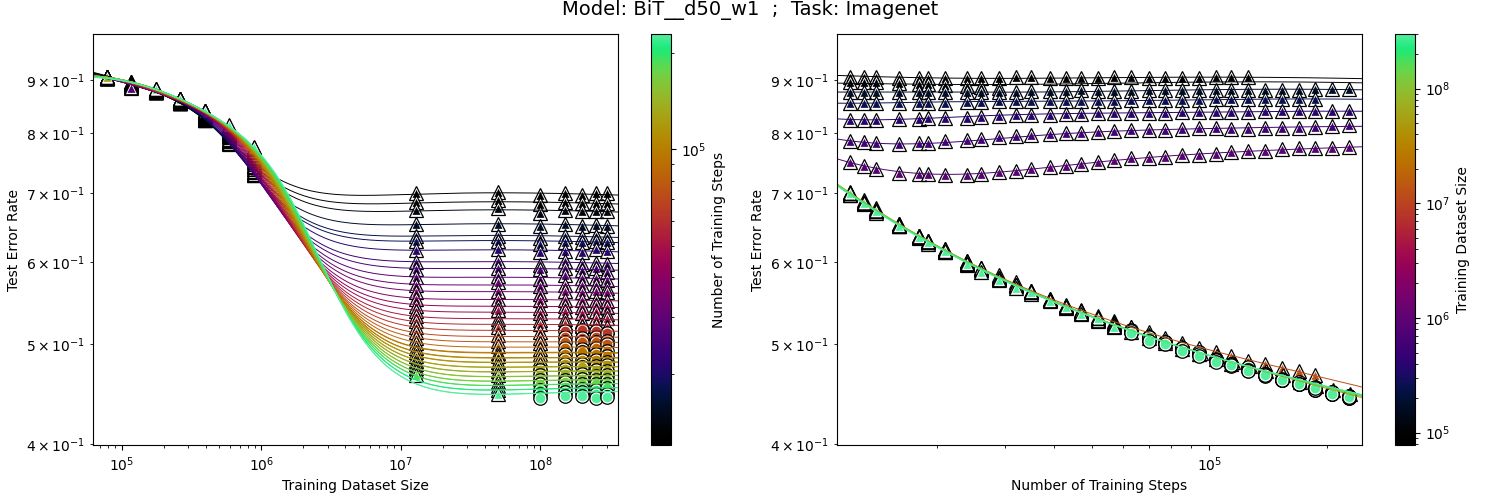}
\includegraphics[width=0.3938\textwidth]{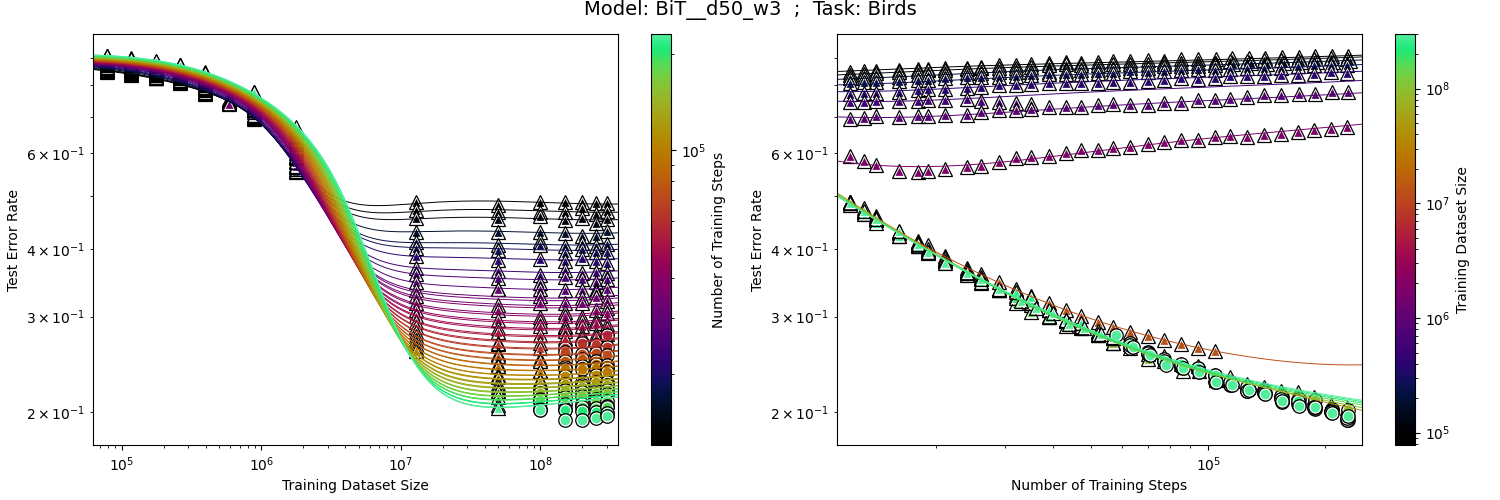}
\includegraphics[width=0.3938\textwidth]{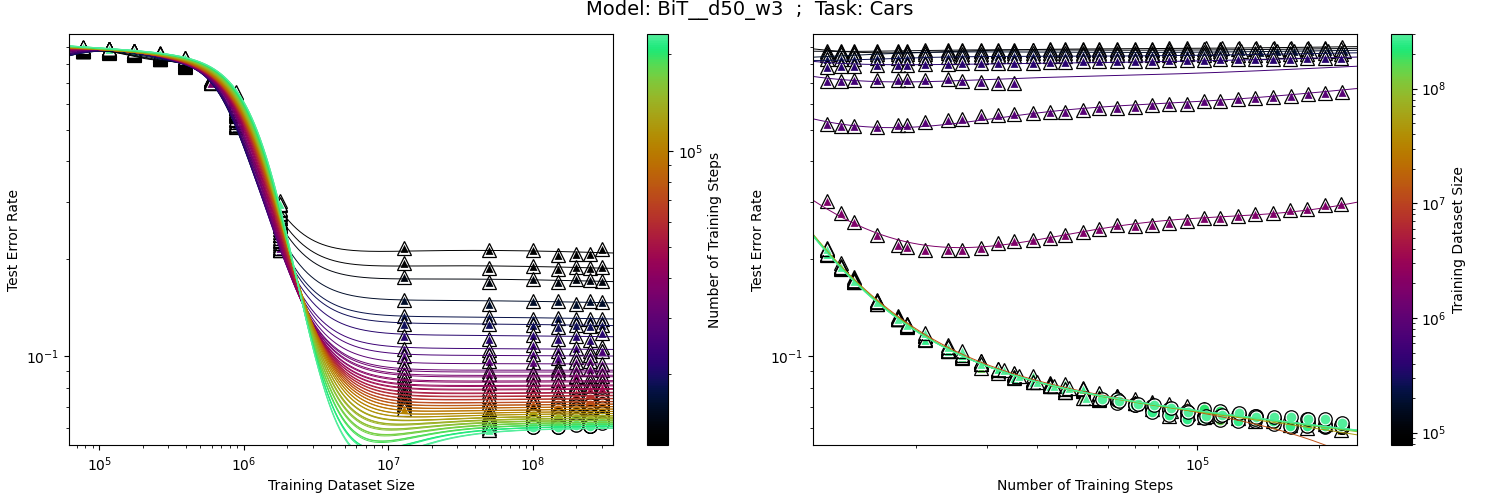}
\includegraphics[width=0.3938\textwidth]{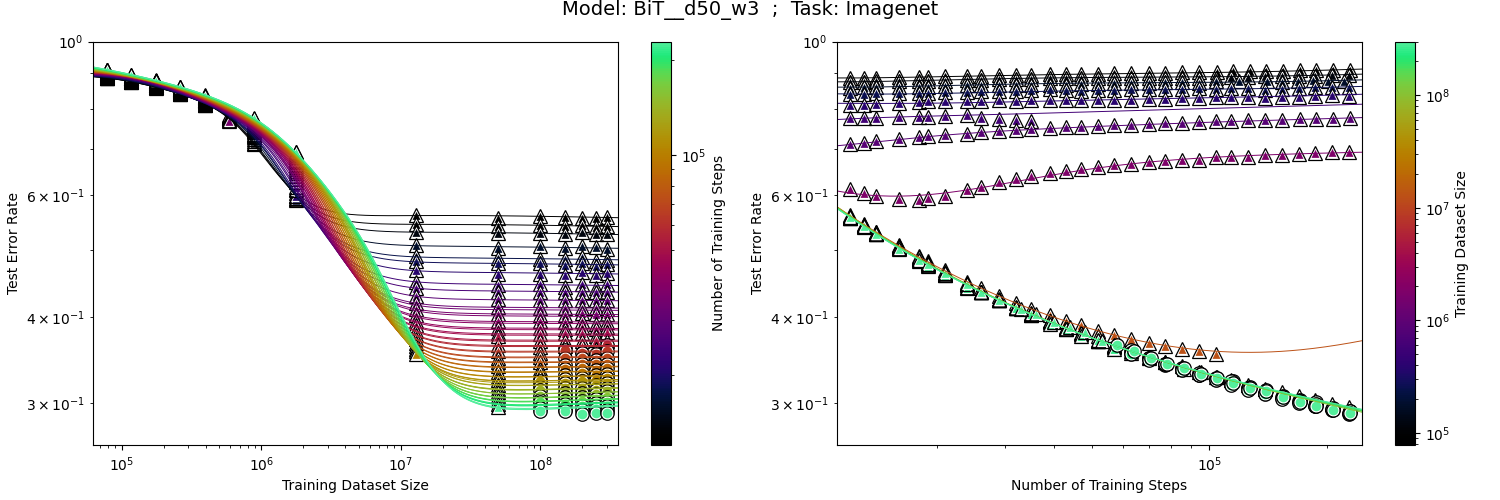}
\includegraphics[width=0.3938\textwidth]{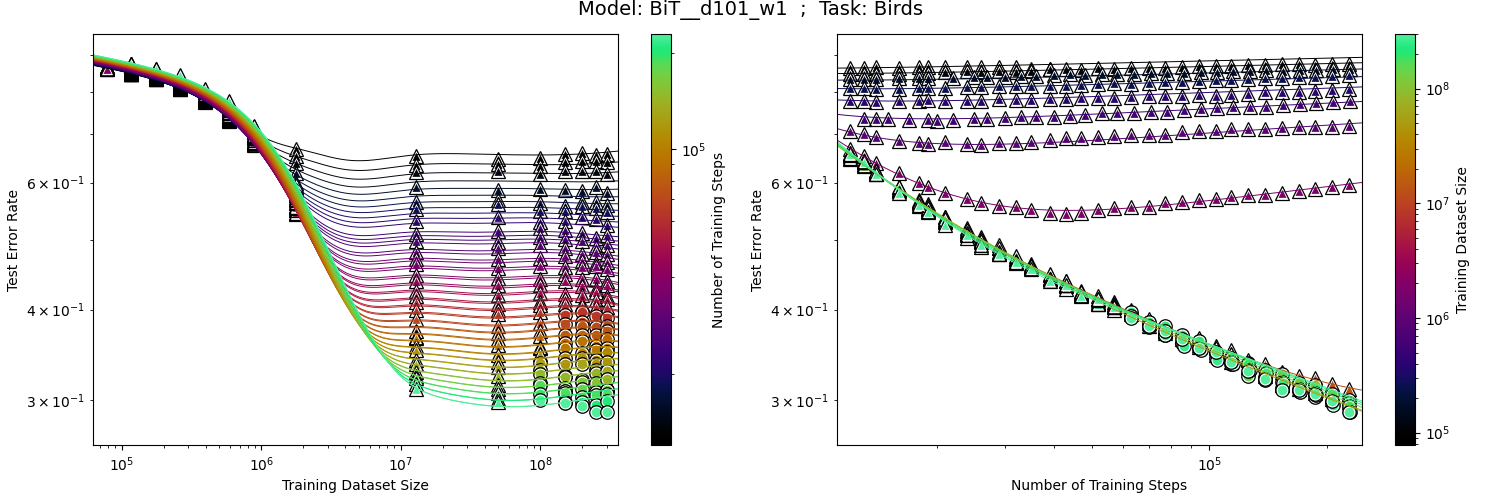}
\includegraphics[width=0.3938\textwidth]{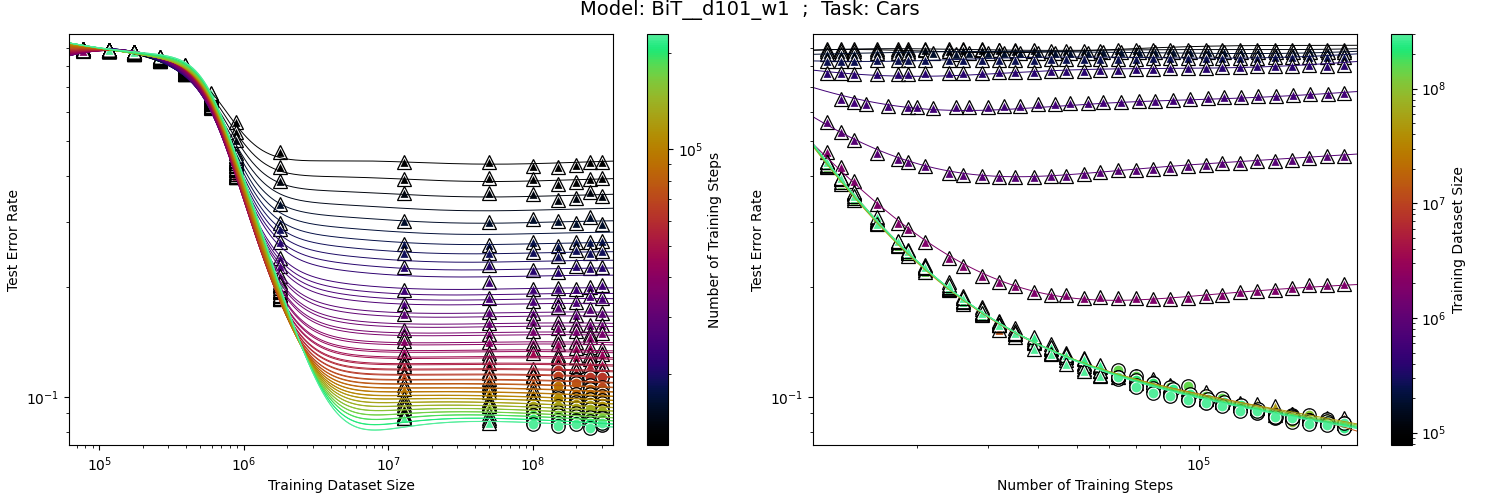}
\includegraphics[width=0.3938\textwidth]{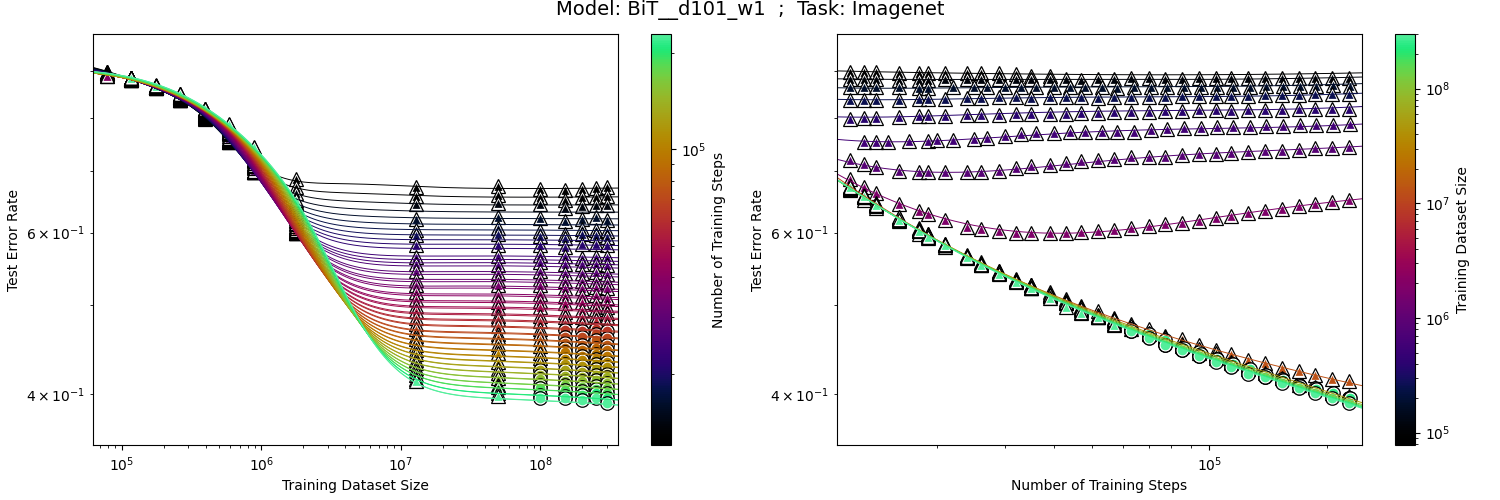}
\includegraphics[width=0.3938\textwidth]{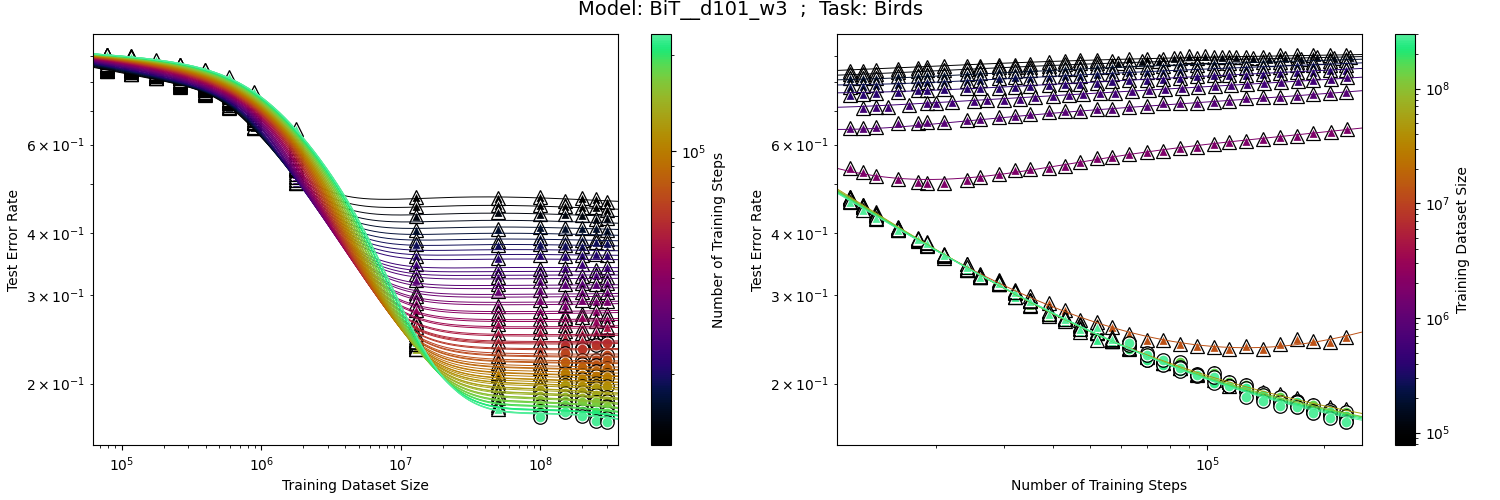}
\includegraphics[width=0.3938\textwidth]{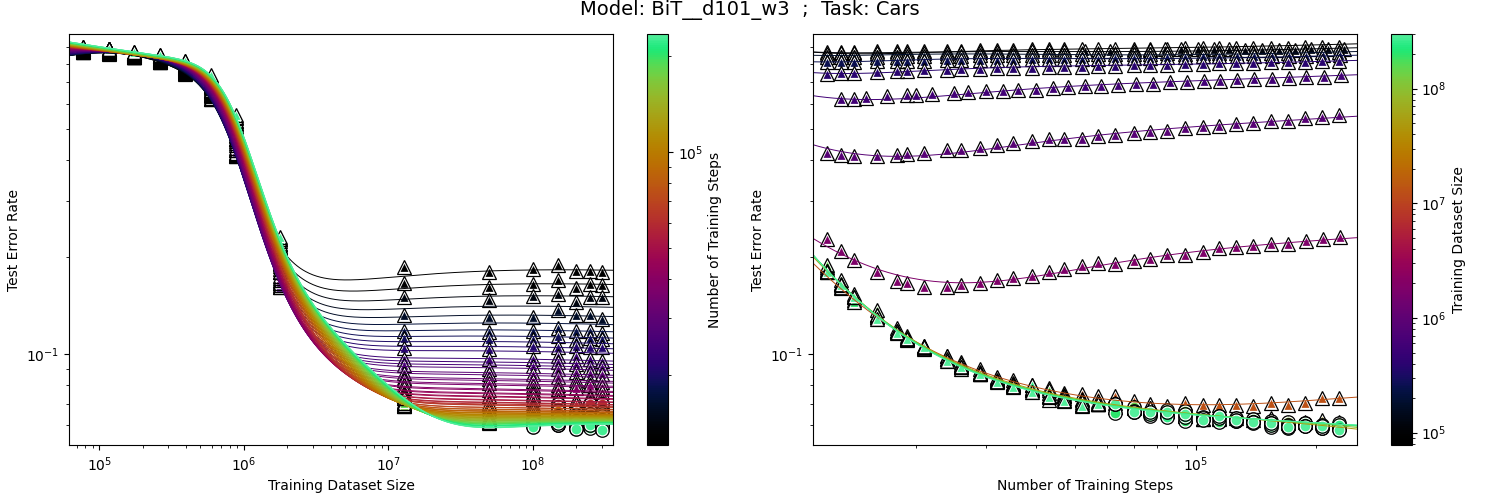}
\includegraphics[width=0.3938\textwidth]{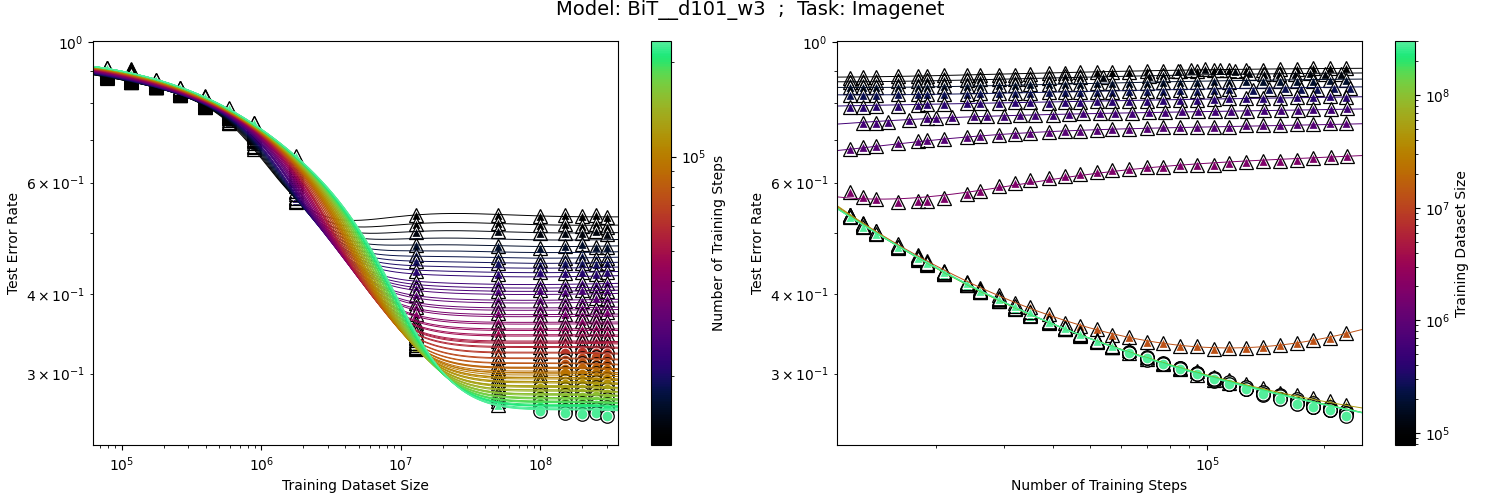}
\includegraphics[width=0.3938\textwidth]{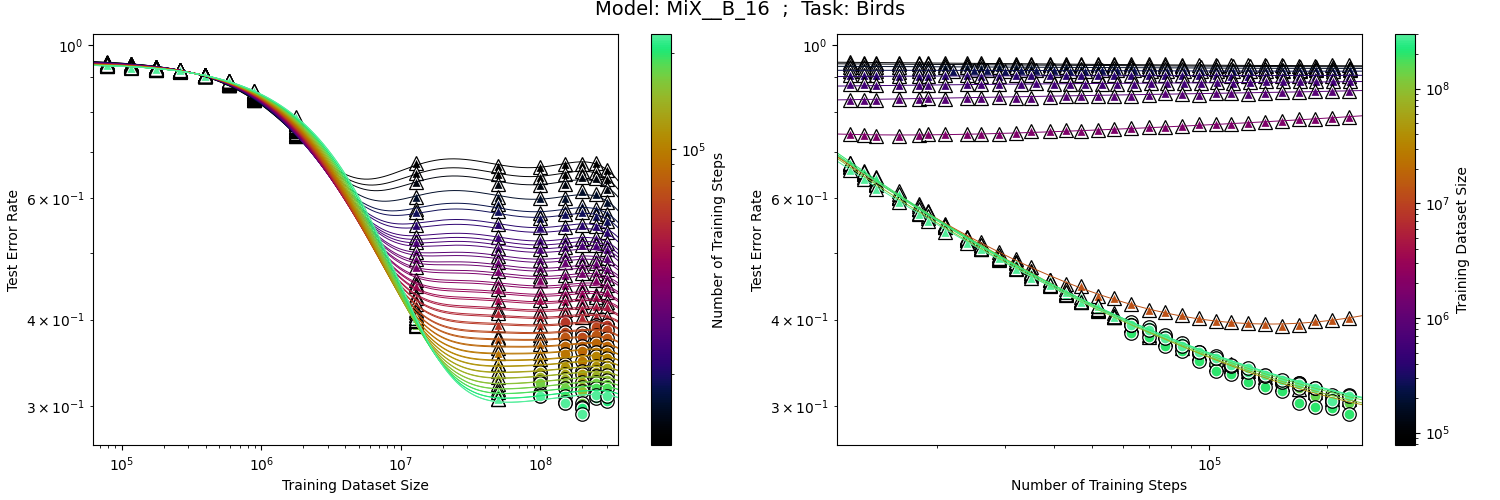}
\includegraphics[width=0.3938\textwidth]{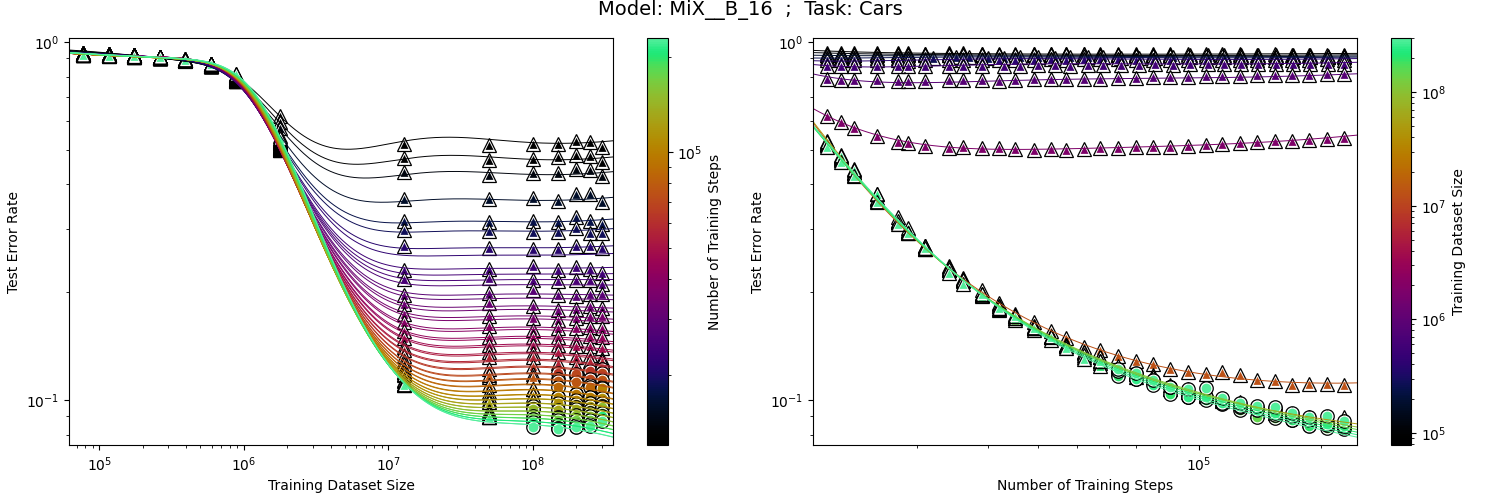}
\includegraphics[width=0.3938\textwidth]{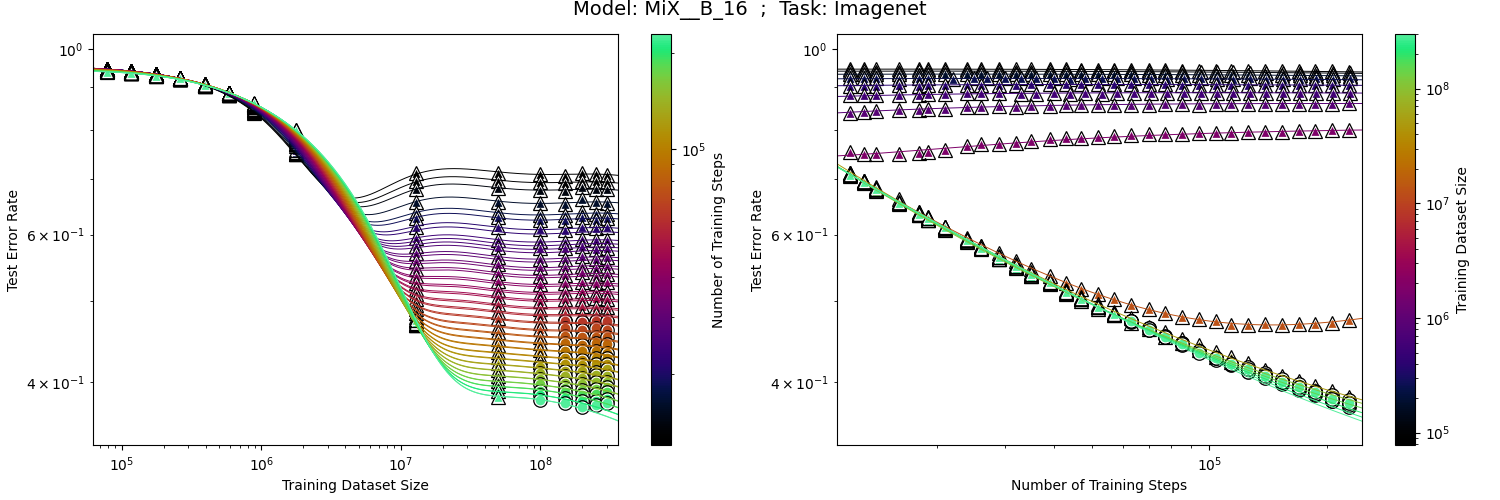}
\includegraphics[width=0.3938\textwidth]{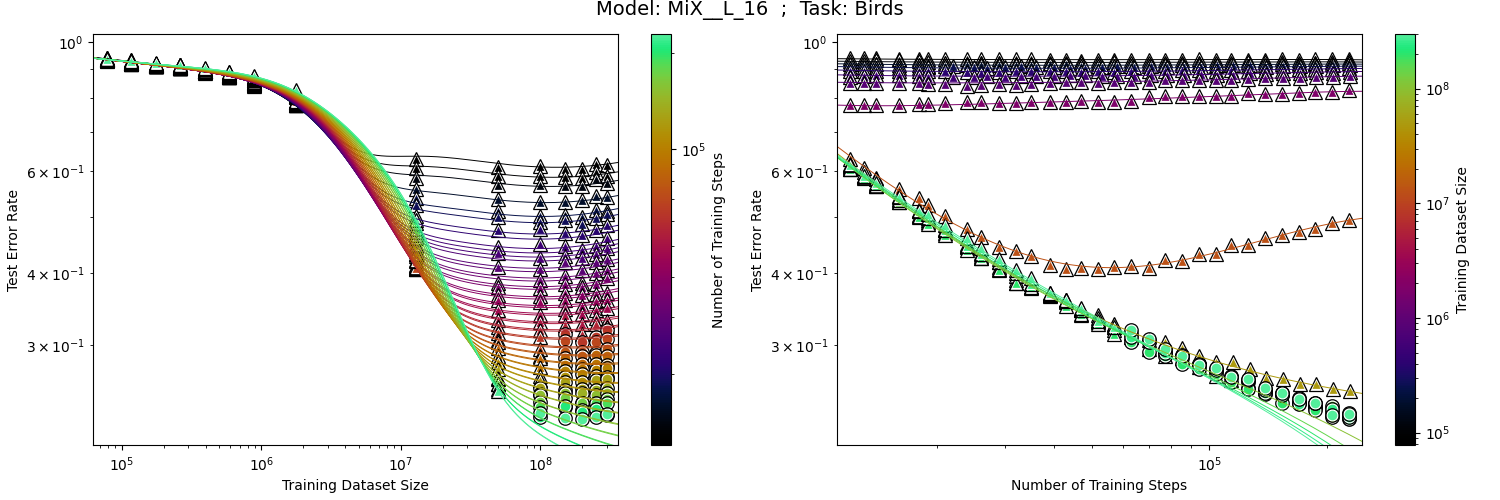}
\includegraphics[width=0.3938\textwidth]{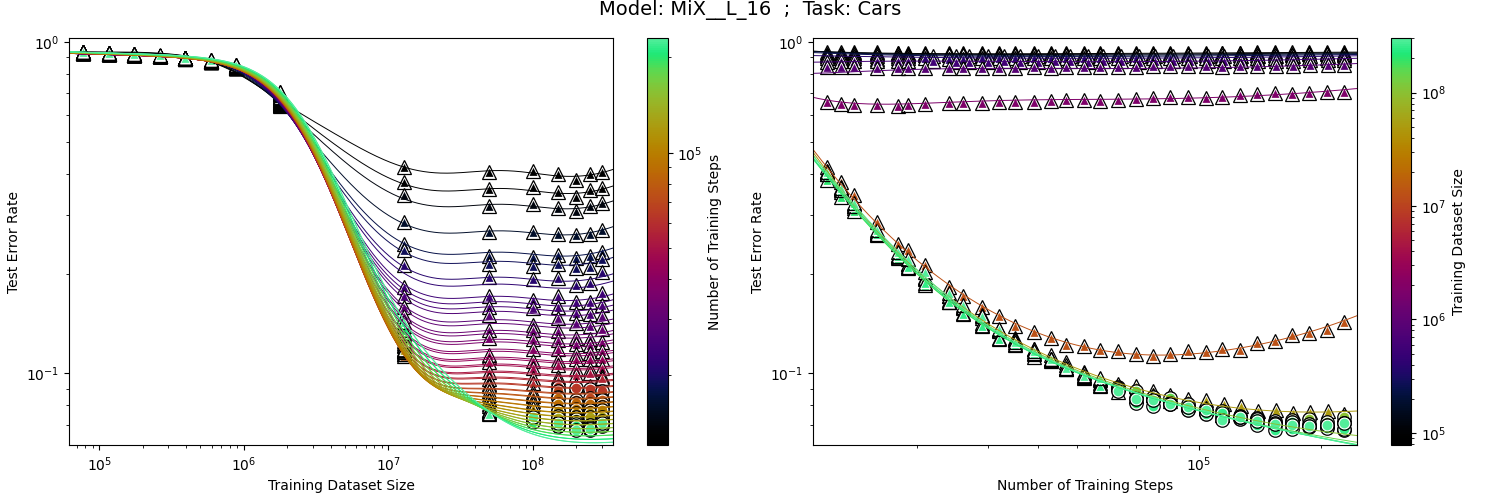}
\includegraphics[width=0.3938\textwidth]{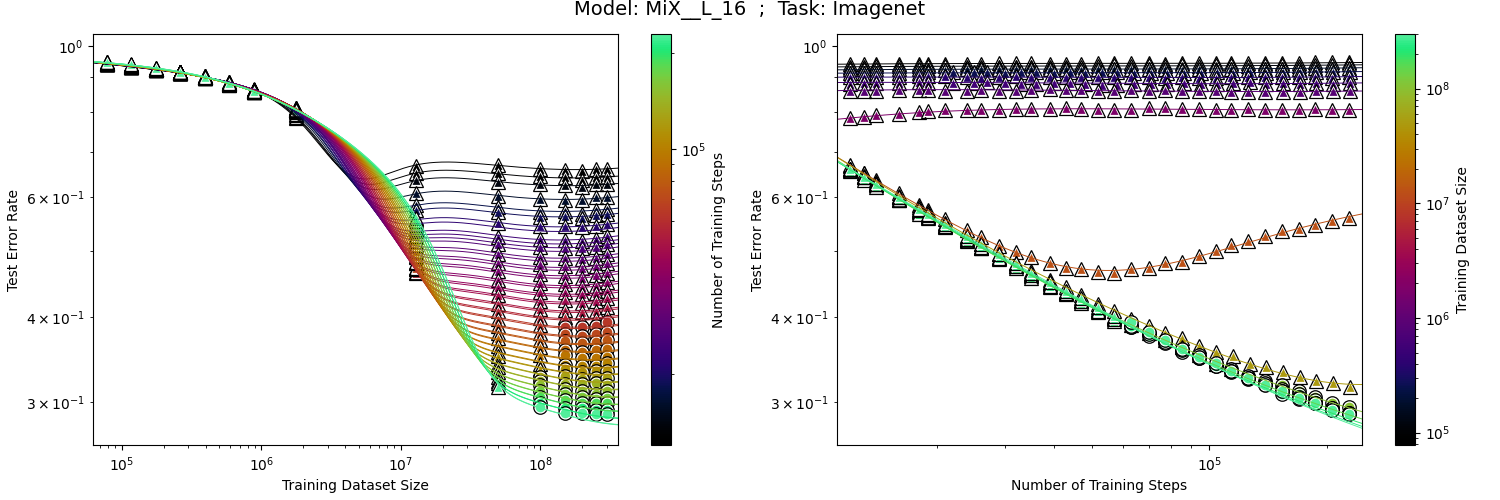}
\includegraphics[width=0.3938\textwidth]{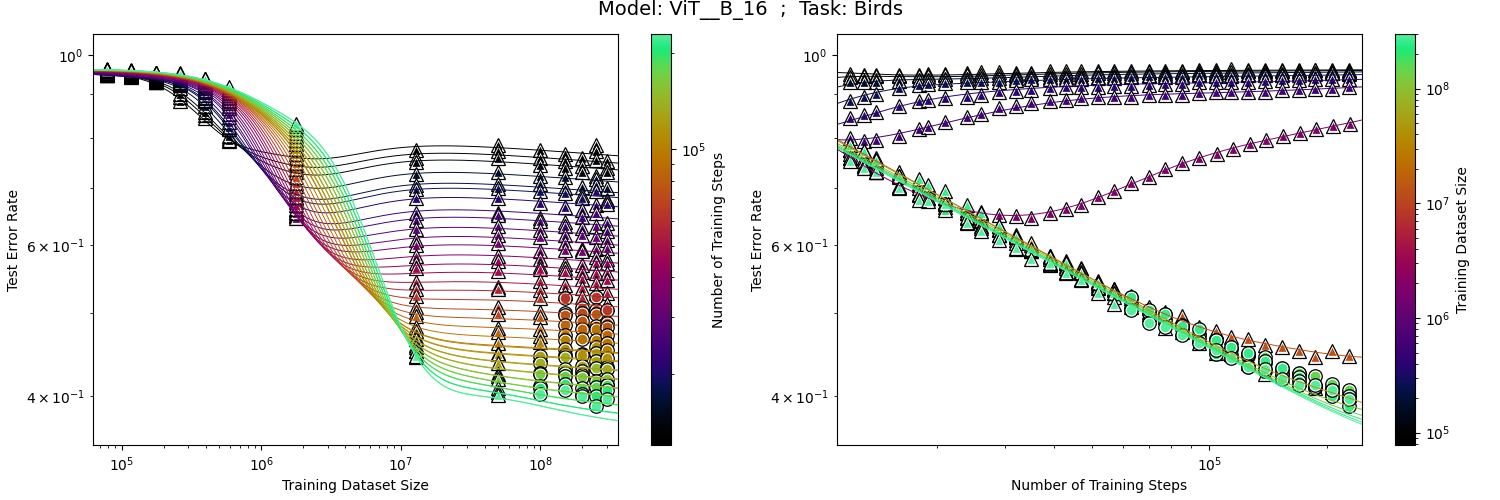}
\includegraphics[width=0.3938\textwidth]{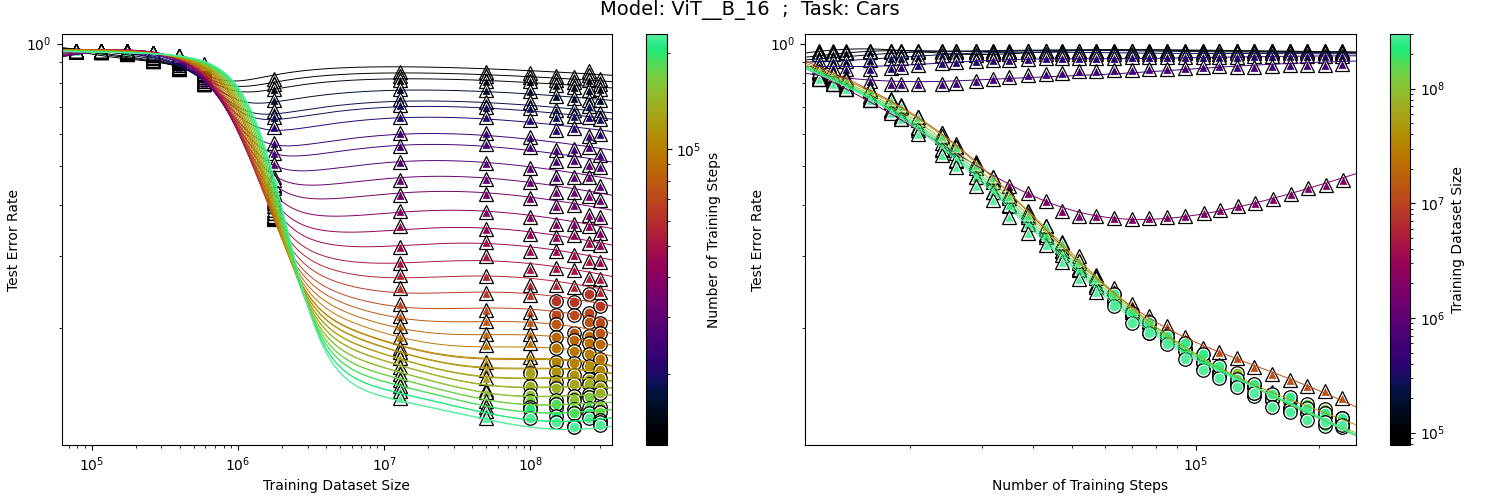}
\includegraphics[width=0.3938\textwidth]{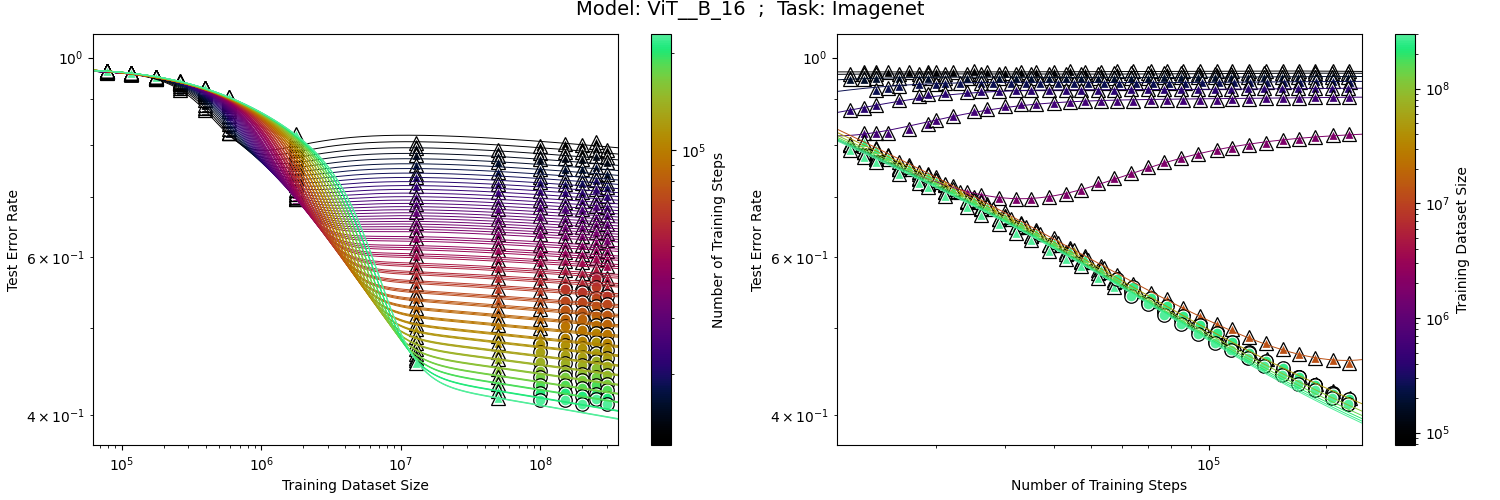}
\hspace{29.0mm}\includegraphics[width=0.179\textwidth]{figures/legend/legend2.png}
    \caption{
    Extrapolation Results of A1 functional form on bivariate scaling behavior of downstream vision performance. See Section \ref{section:vision} for more details.
    }
    \label{fig:a1_downstream_vision}
\end{figure*}

\FloatBarrier
\begin{figure*}[h]    \centering

\includegraphics[width=0.3938\textwidth]{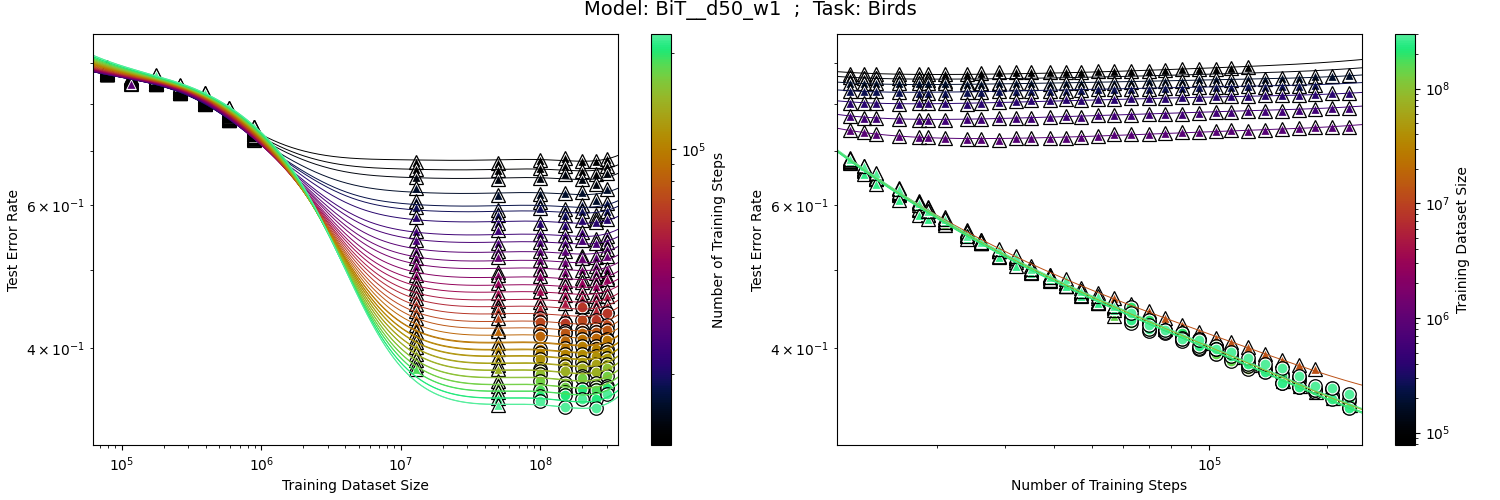}
\includegraphics[width=0.3938\textwidth]{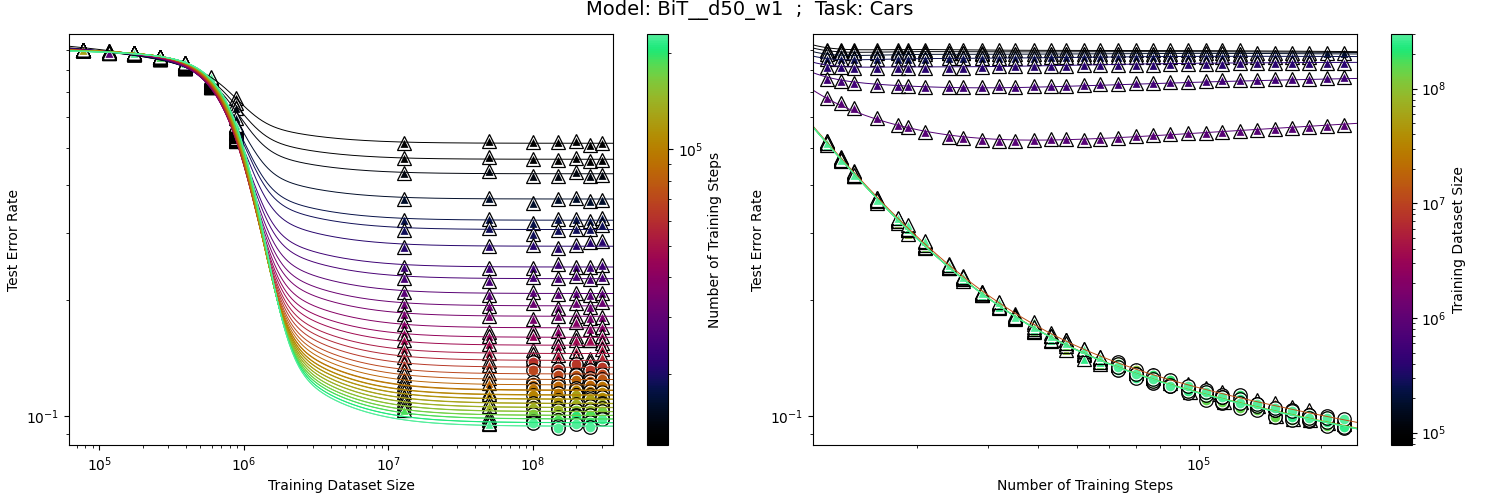}
\includegraphics[width=0.3938\textwidth]{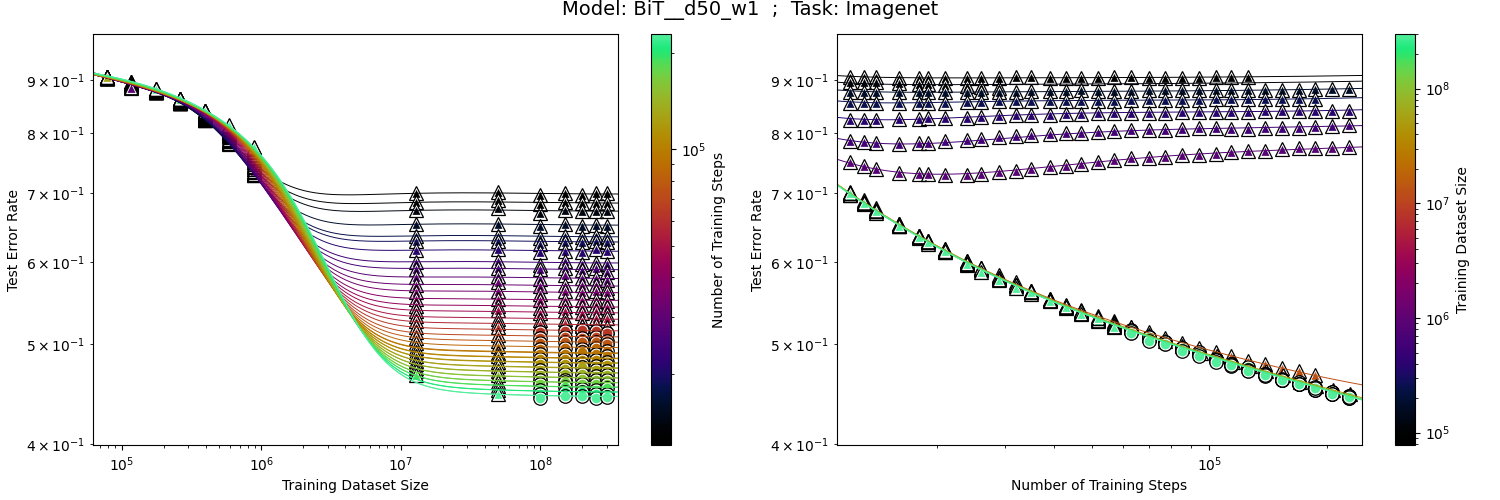}
\includegraphics[width=0.3938\textwidth]{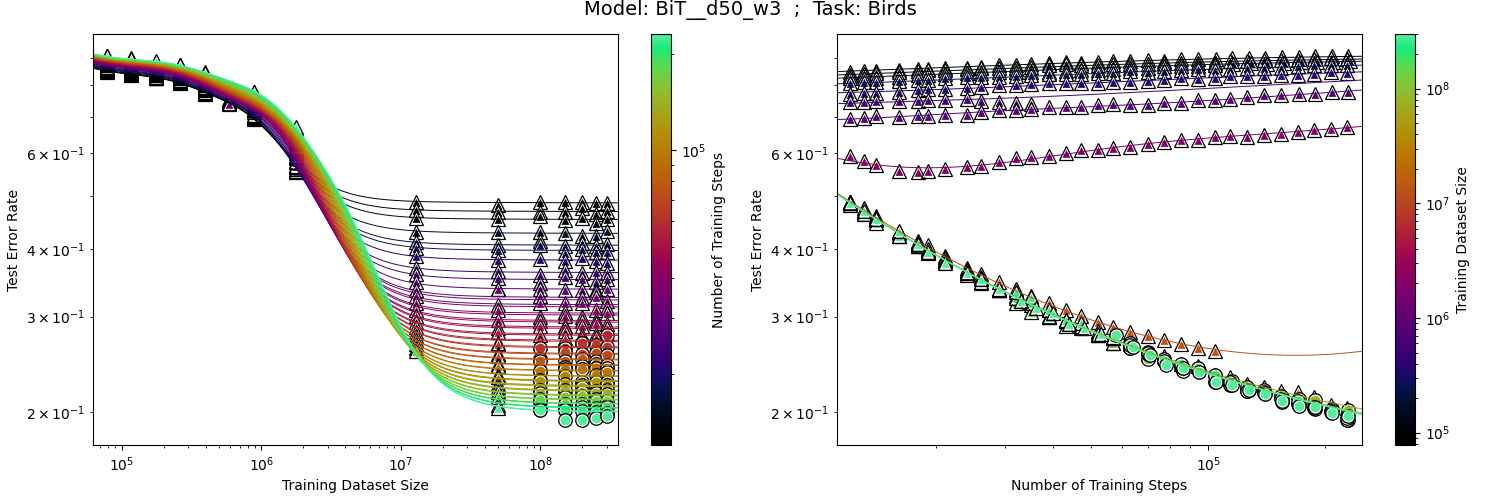}
\includegraphics[width=0.3938\textwidth]{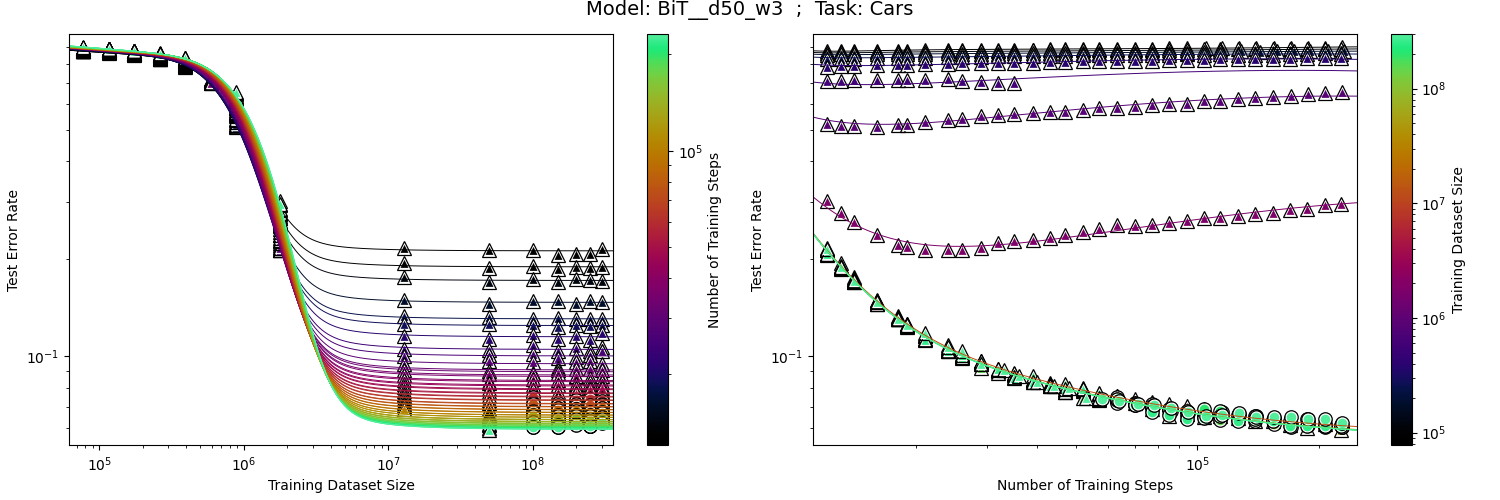}
\includegraphics[width=0.3938\textwidth]{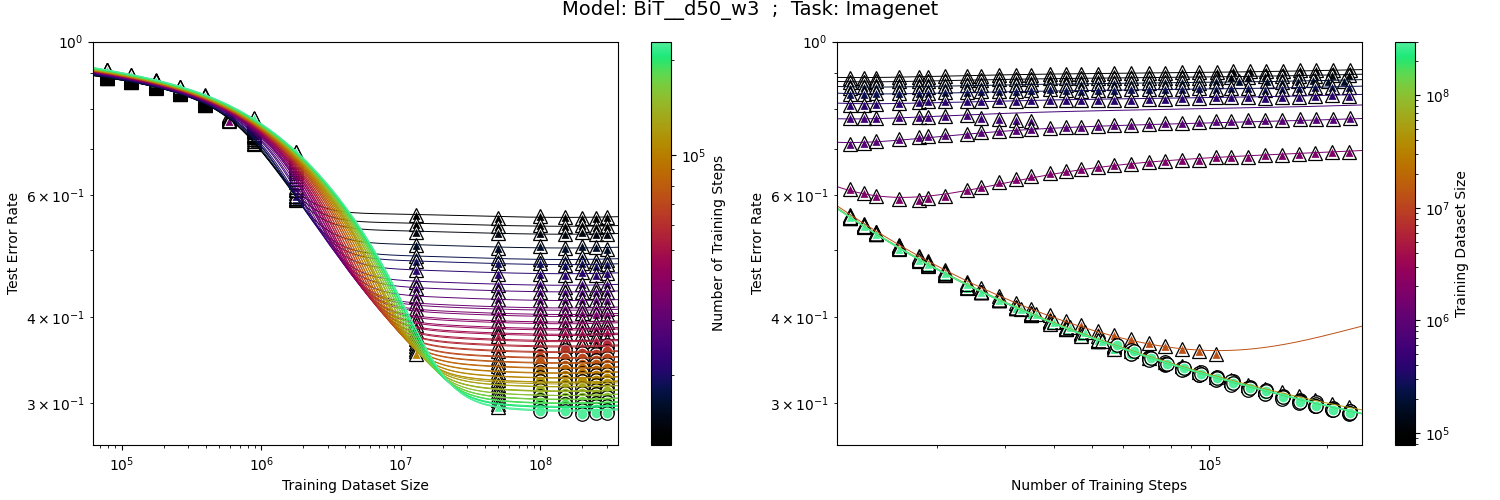}
\includegraphics[width=0.3938\textwidth]{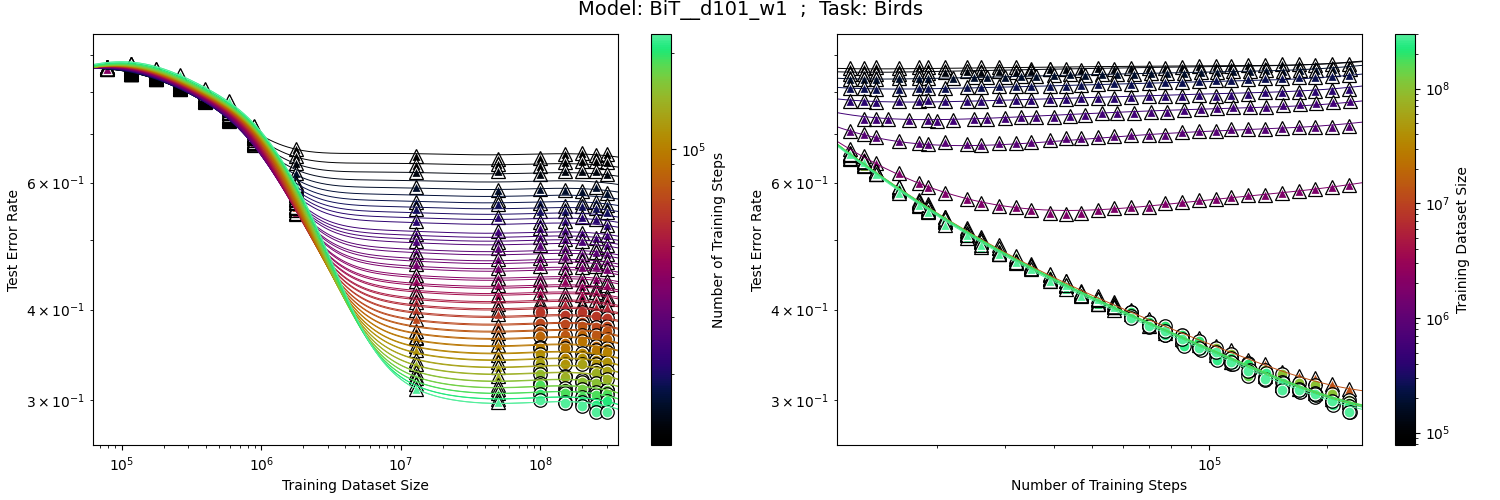}
\includegraphics[width=0.3938\textwidth]{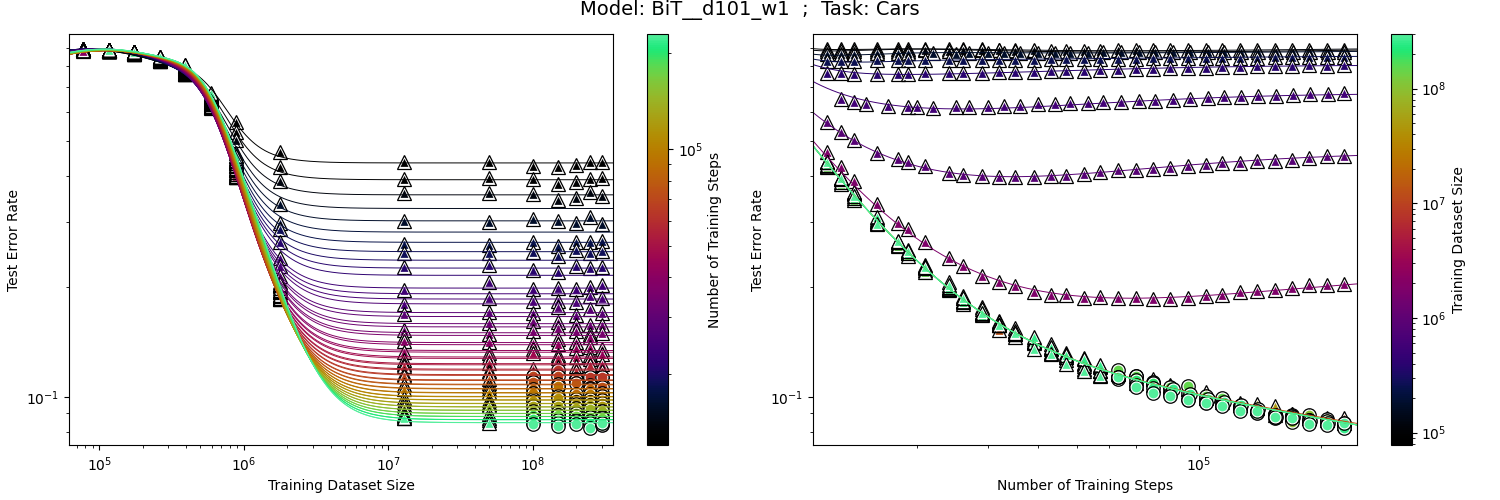}
\includegraphics[width=0.3938\textwidth]{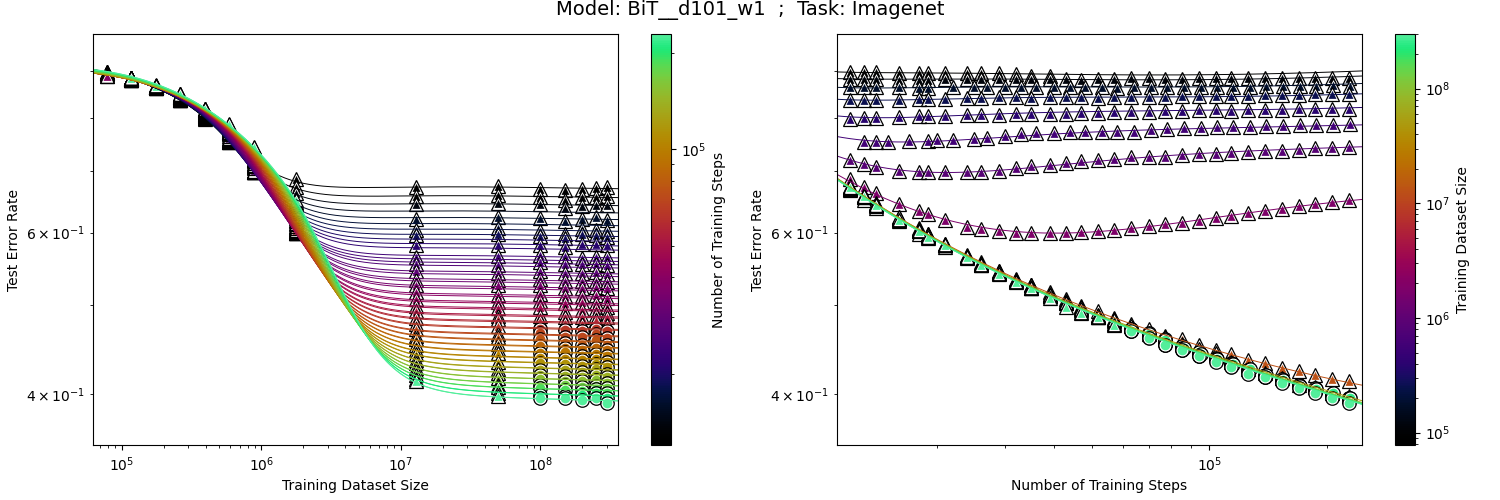}
\includegraphics[width=0.3938\textwidth]{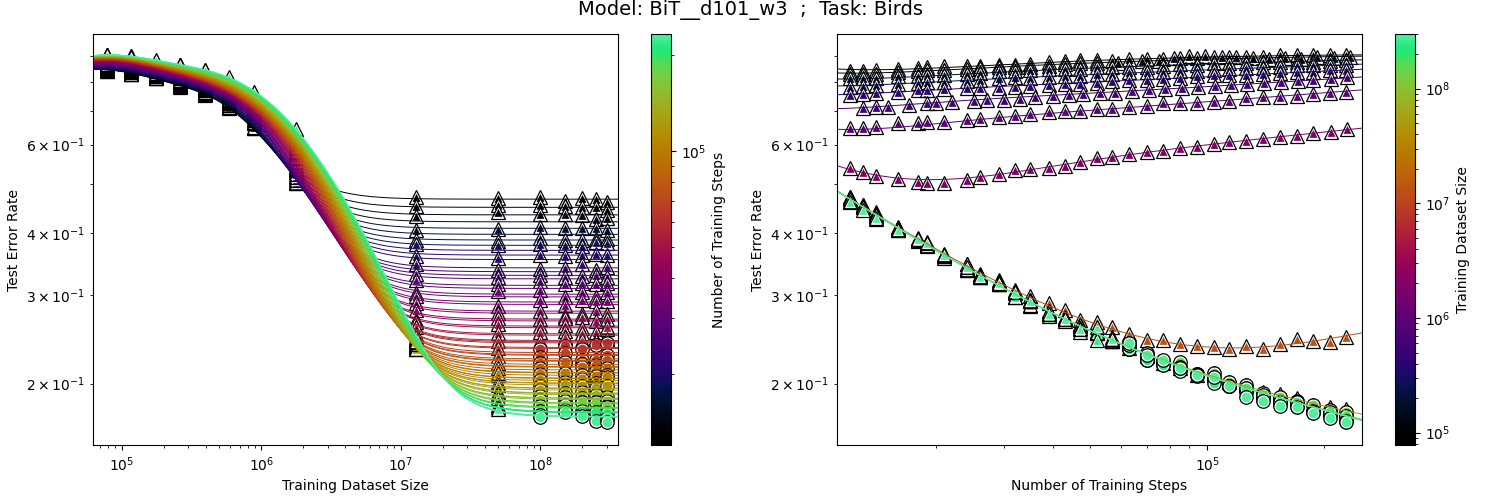}
\includegraphics[width=0.3938\textwidth]{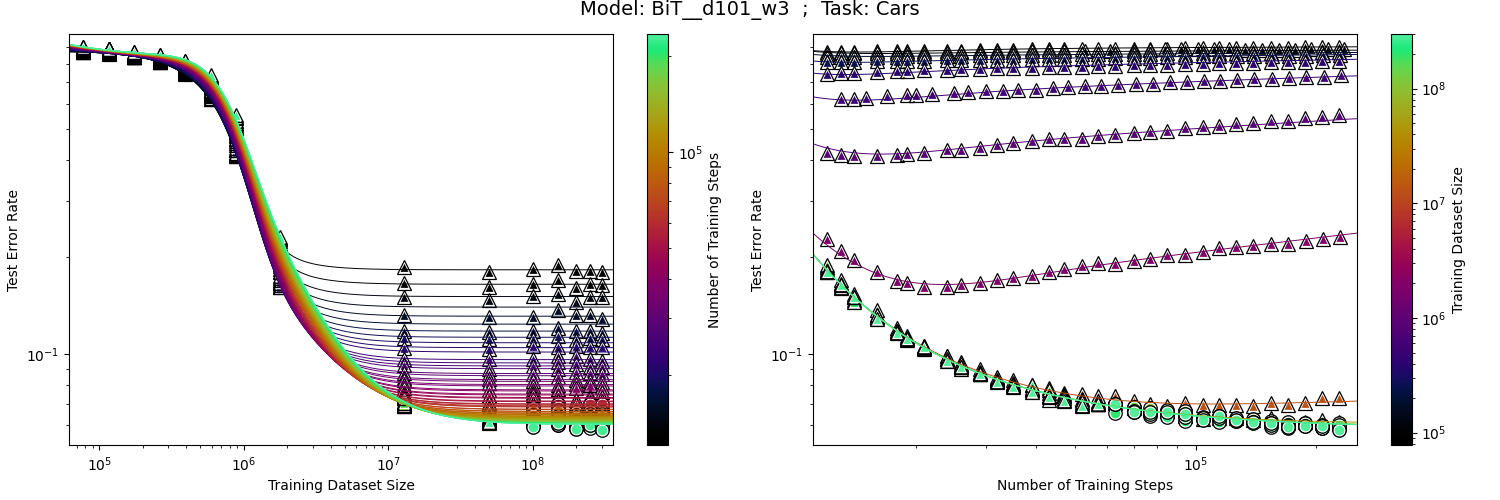}
\includegraphics[width=0.3938\textwidth]{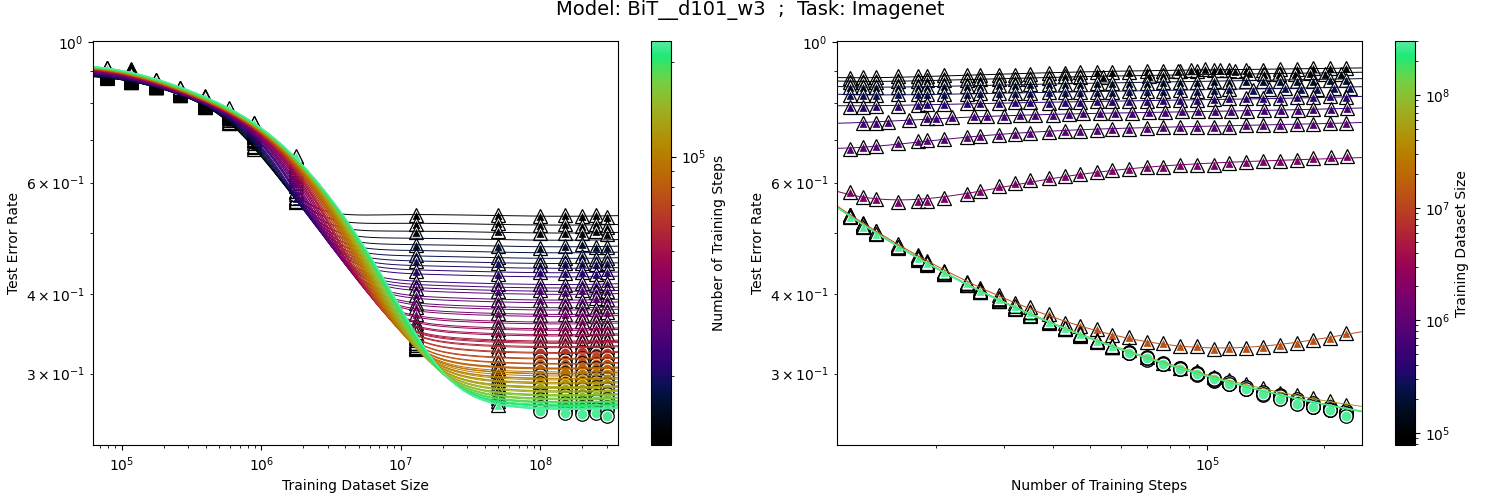}
\includegraphics[width=0.3938\textwidth]{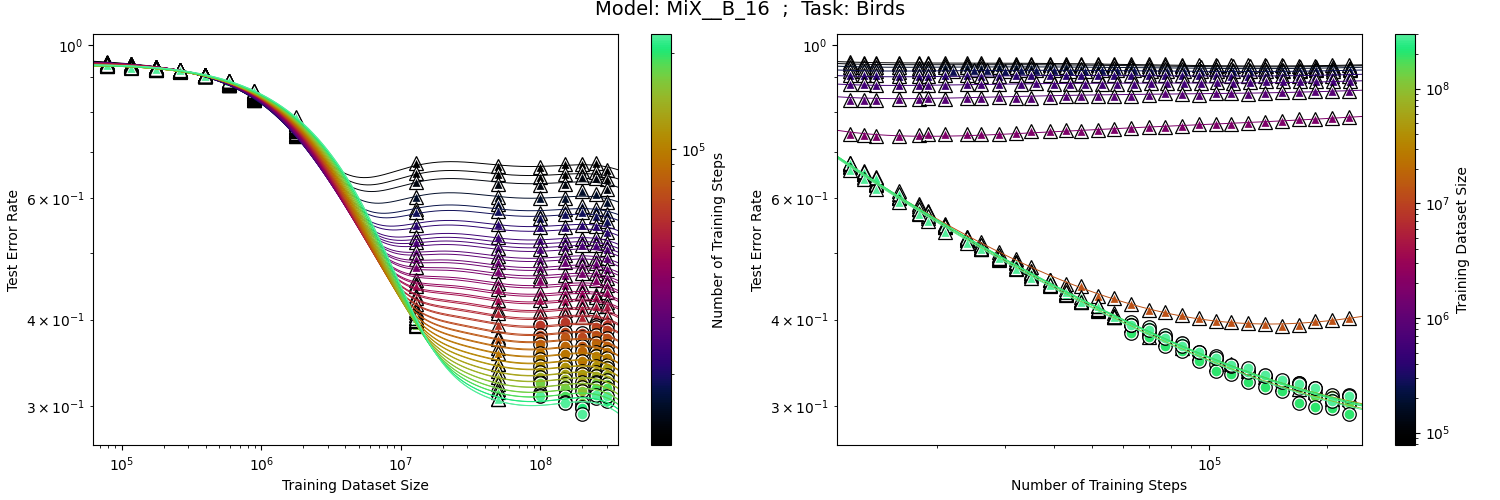}
\includegraphics[width=0.3938\textwidth]{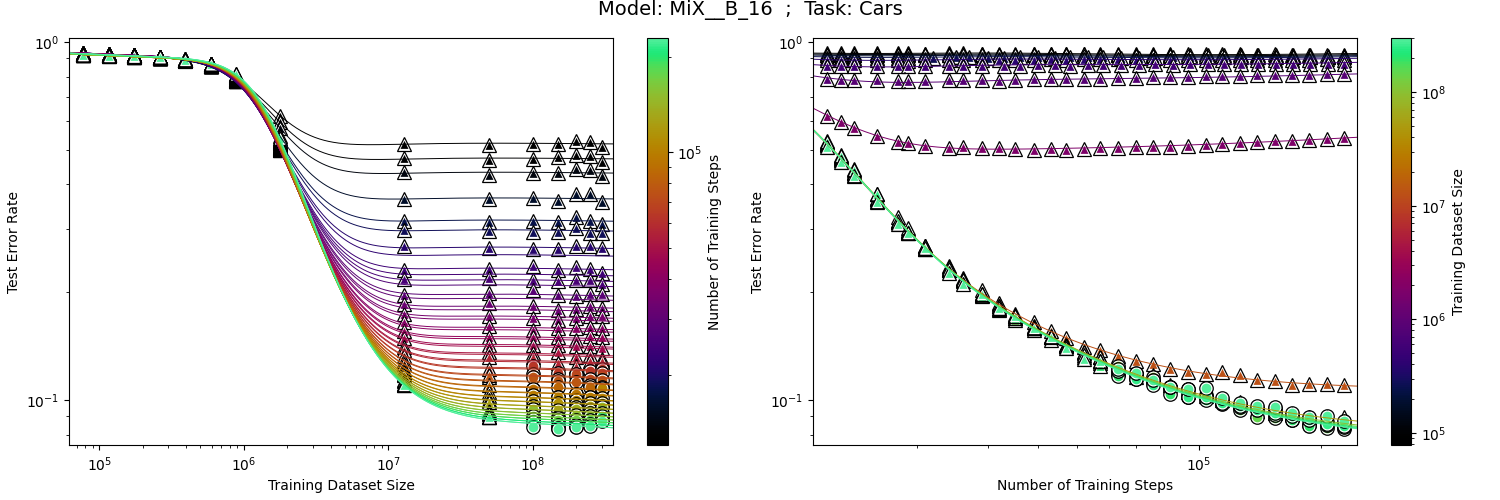}
\includegraphics[width=0.3938\textwidth]{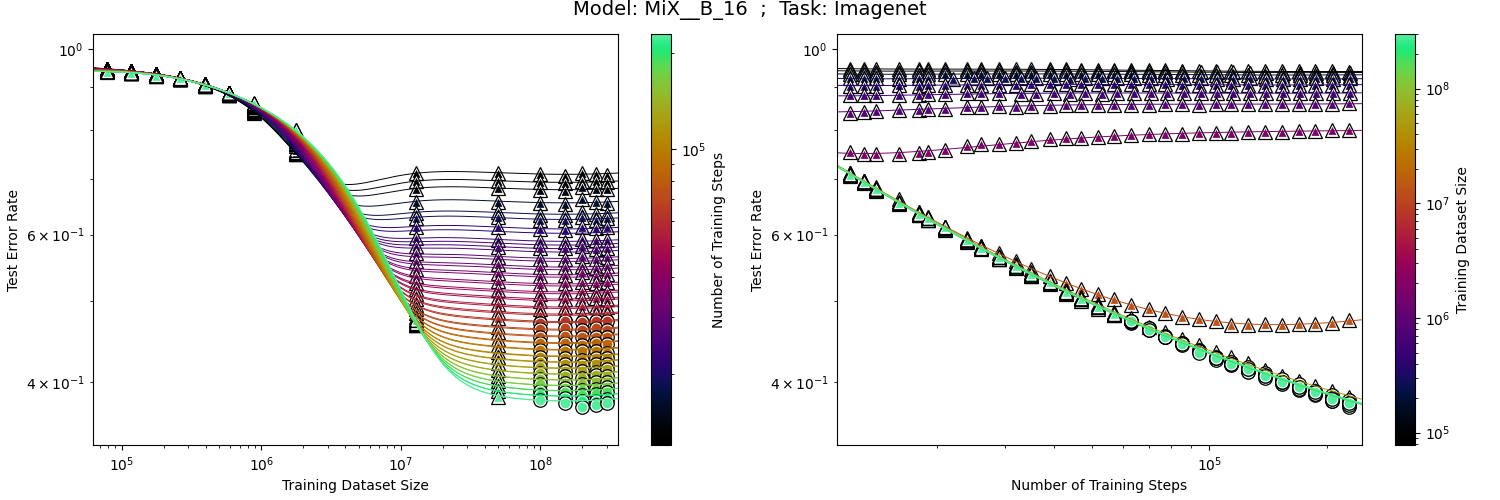}
\includegraphics[width=0.3938\textwidth]{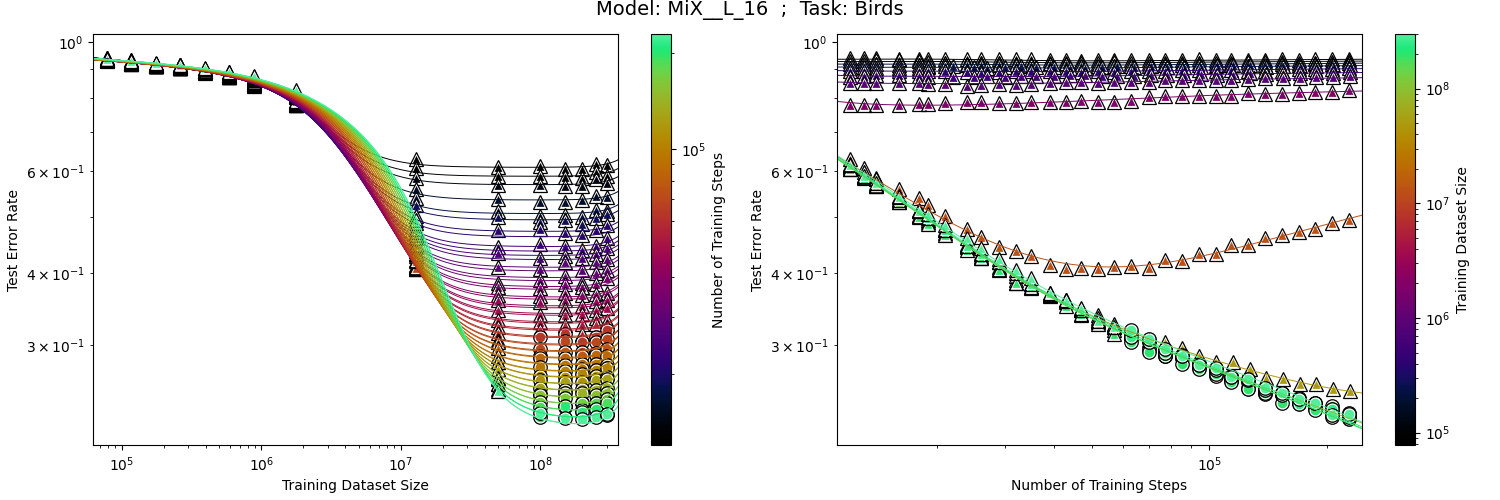}
\includegraphics[width=0.3938\textwidth]{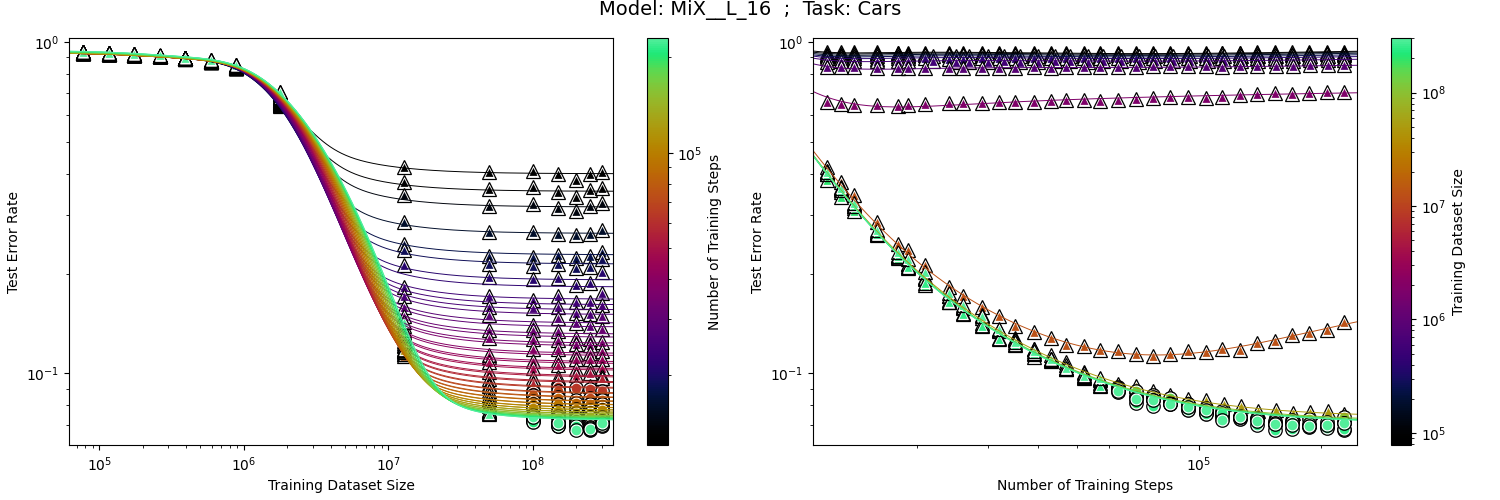}
\includegraphics[width=0.3938\textwidth]{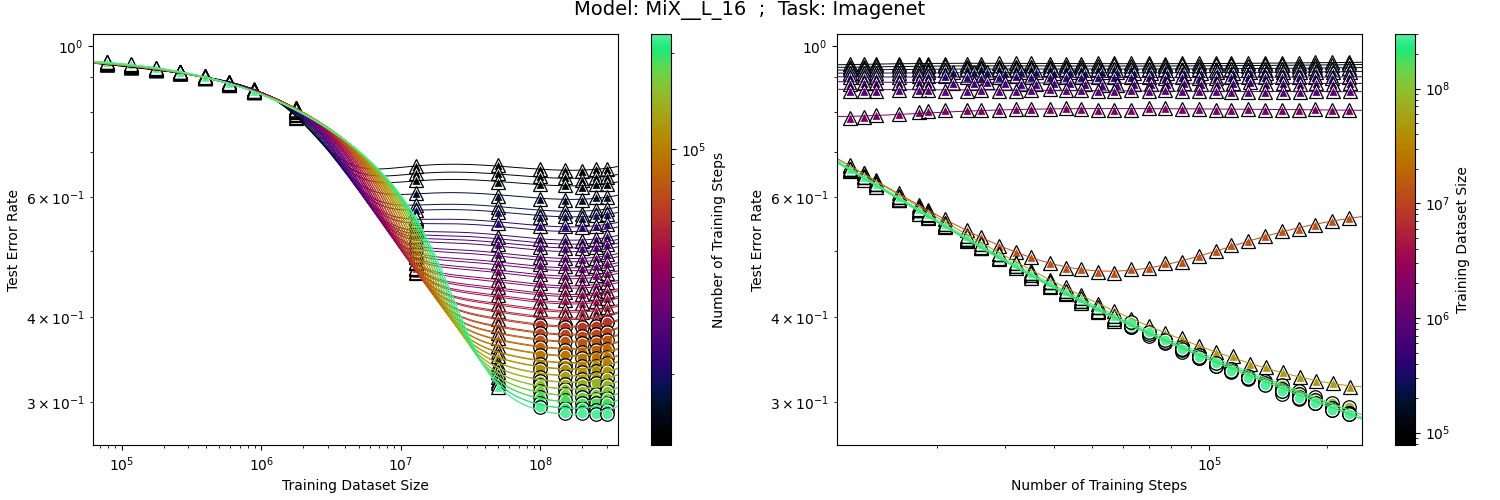}
\includegraphics[width=0.3938\textwidth]{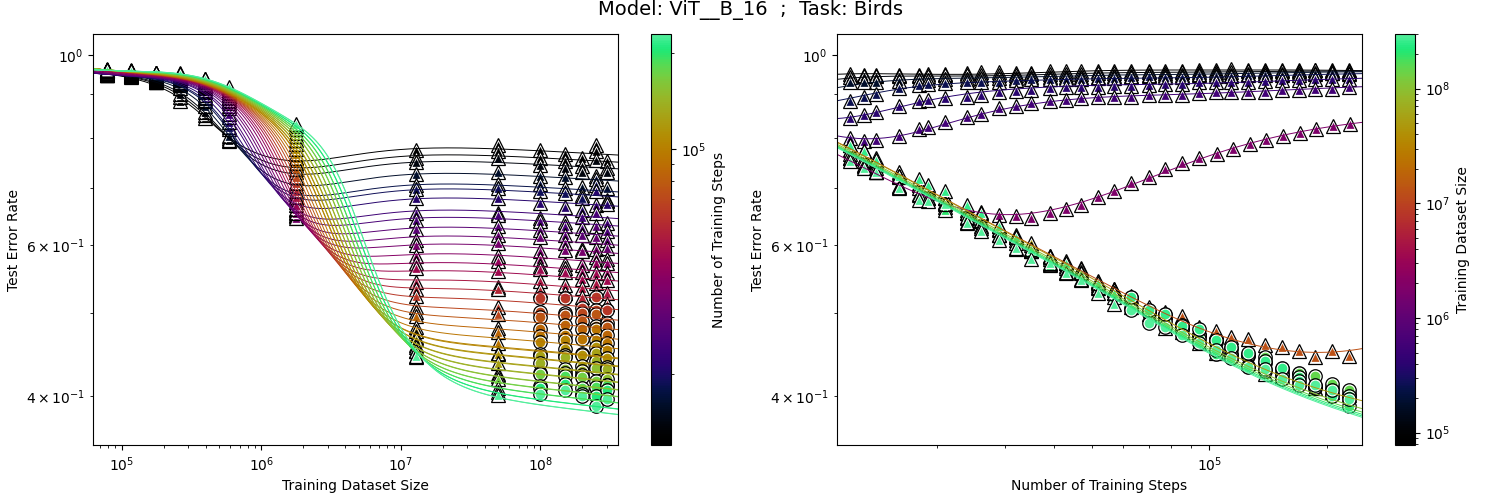}
\includegraphics[width=0.3938\textwidth]{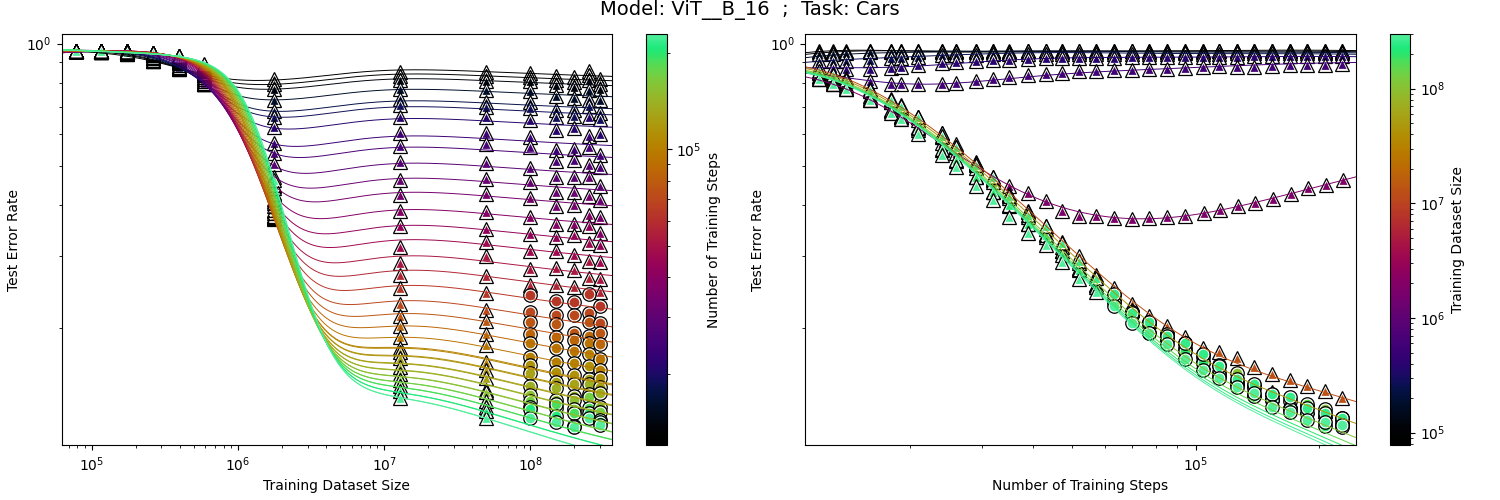}
\includegraphics[width=0.3938\textwidth]{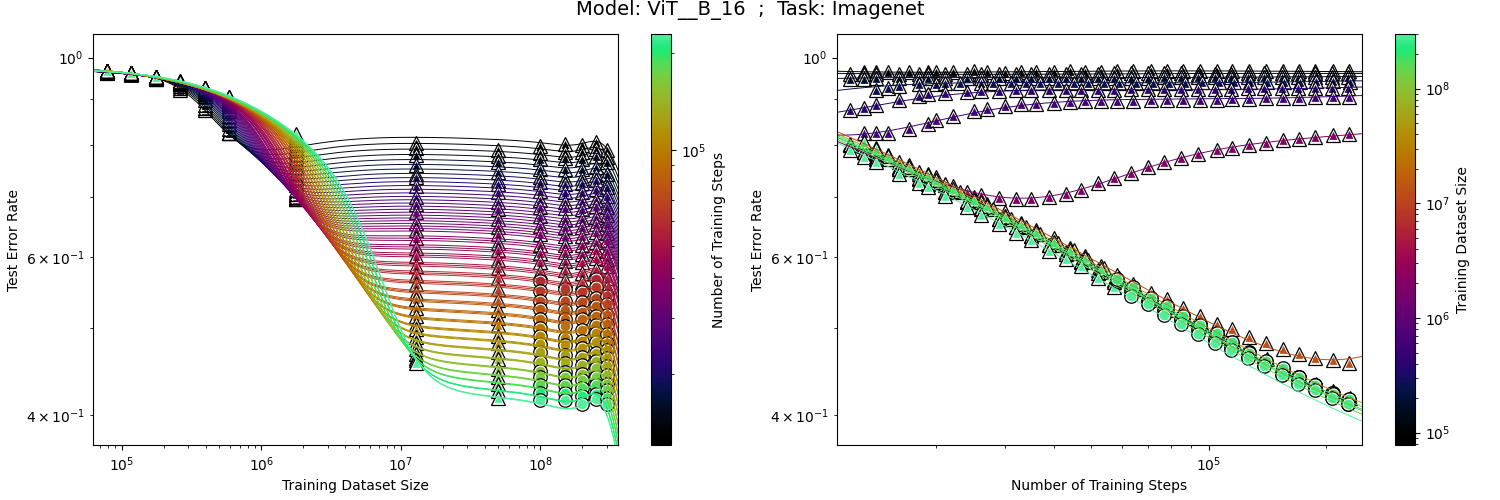}
\hspace{29.0mm}\includegraphics[width=0.179\textwidth]{figures/legend/legend2.png}
    \caption{
    Extrapolation Results of A2 functional form on bivariate scaling behavior of downstream vision performance. See Section \ref{section:vision} for more details.
    }
    \label{fig:a2_downstream_vision}
\end{figure*}

\FloatBarrier
\begin{figure*}[h]    \centering

\includegraphics[width=0.3938\textwidth]{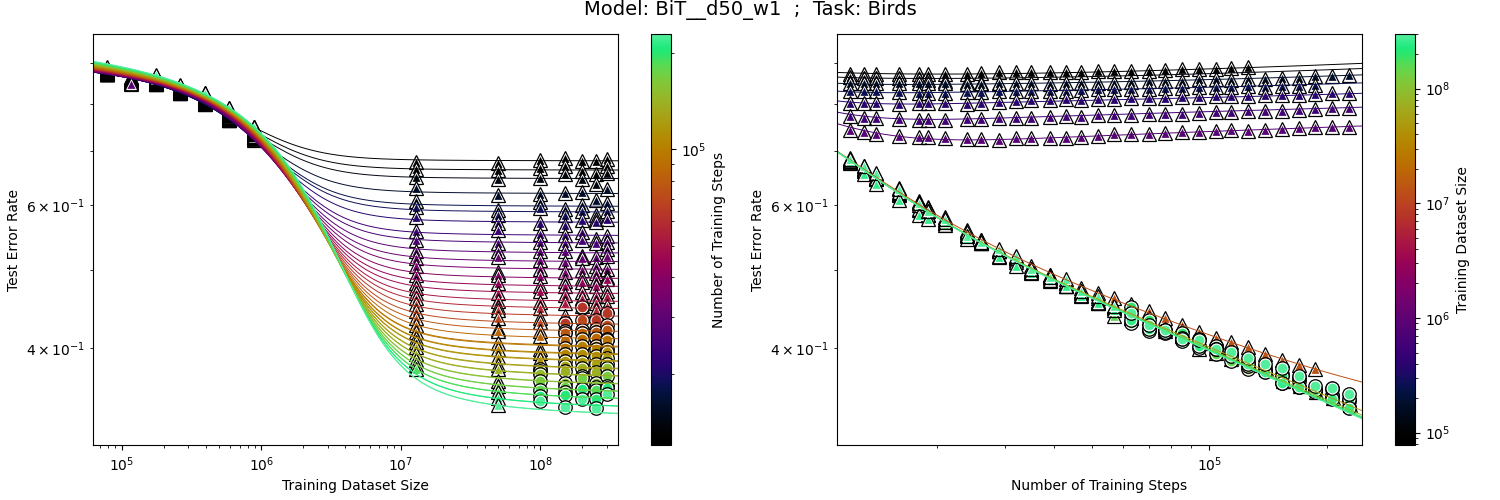}
\includegraphics[width=0.3938\textwidth]{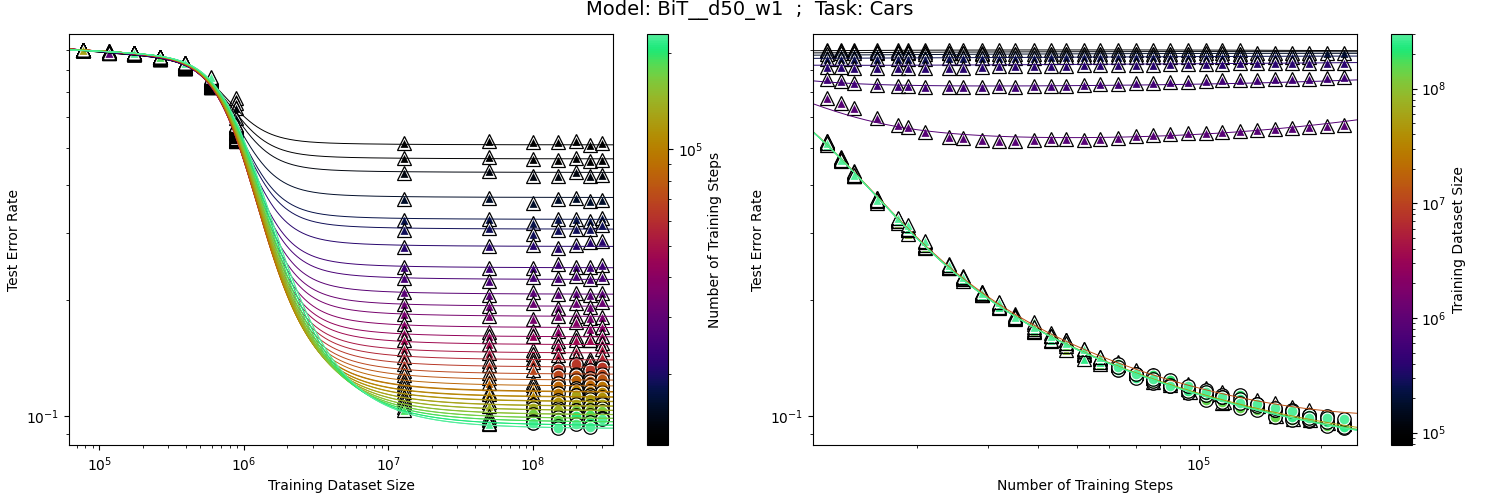}
\includegraphics[width=0.3938\textwidth]{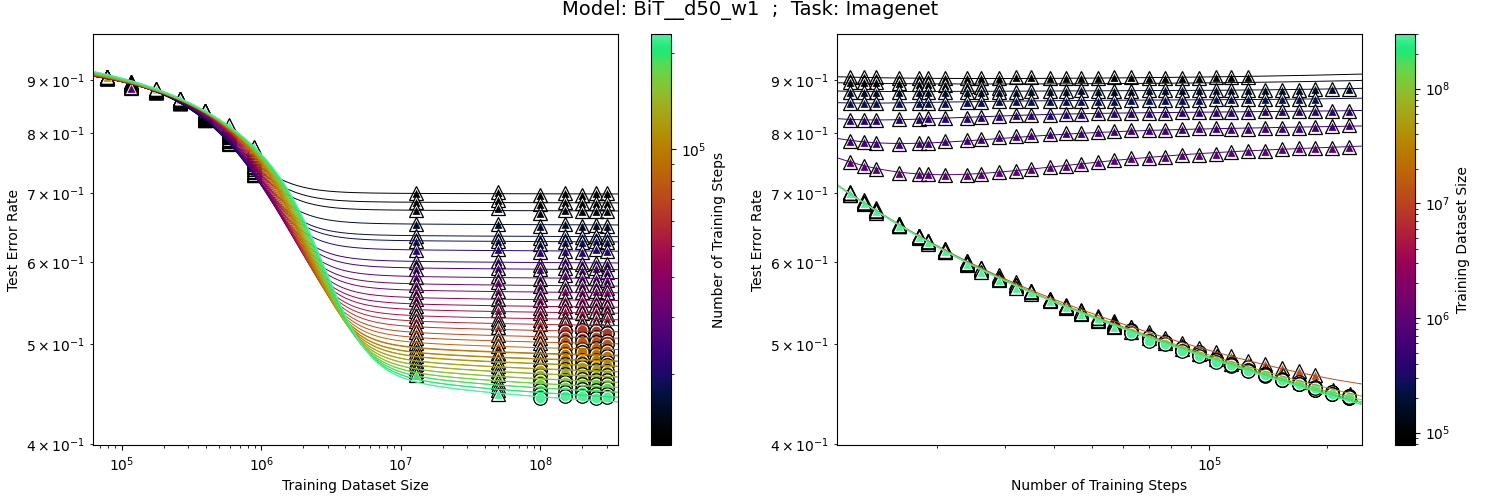}
\includegraphics[width=0.3938\textwidth]{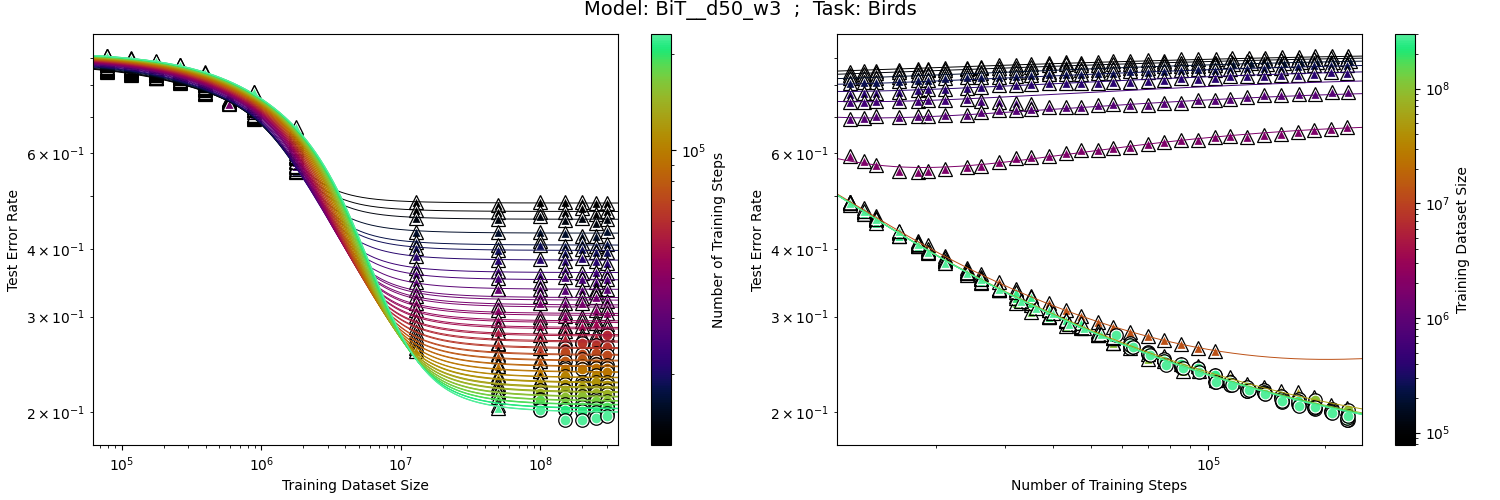}
\includegraphics[width=0.3938\textwidth]{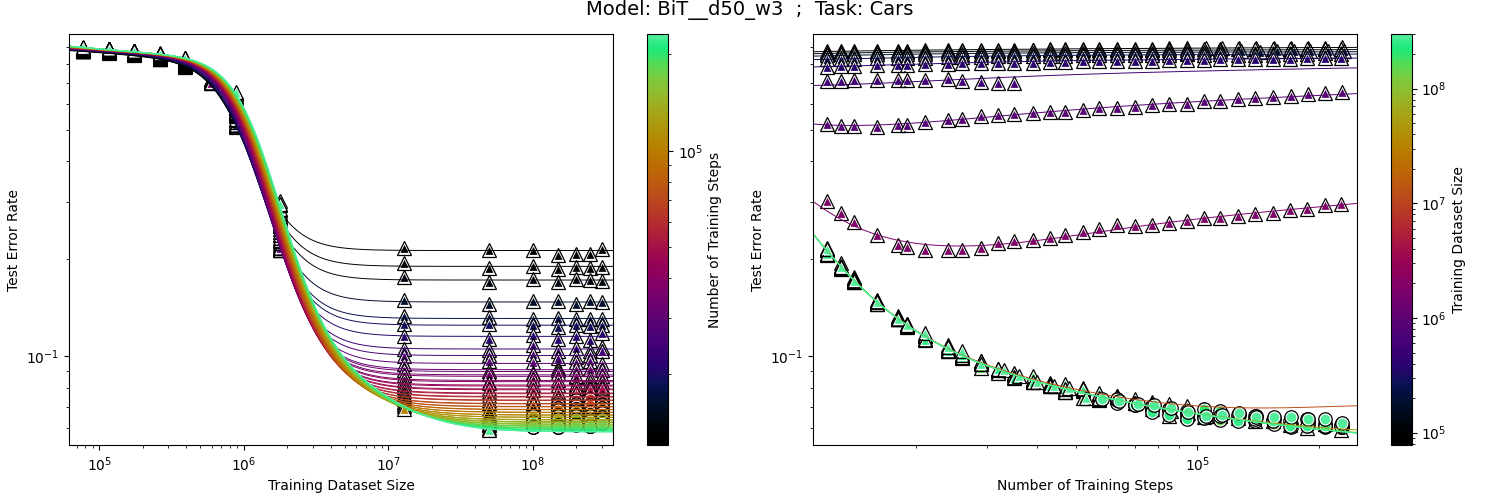}
\includegraphics[width=0.3938\textwidth]{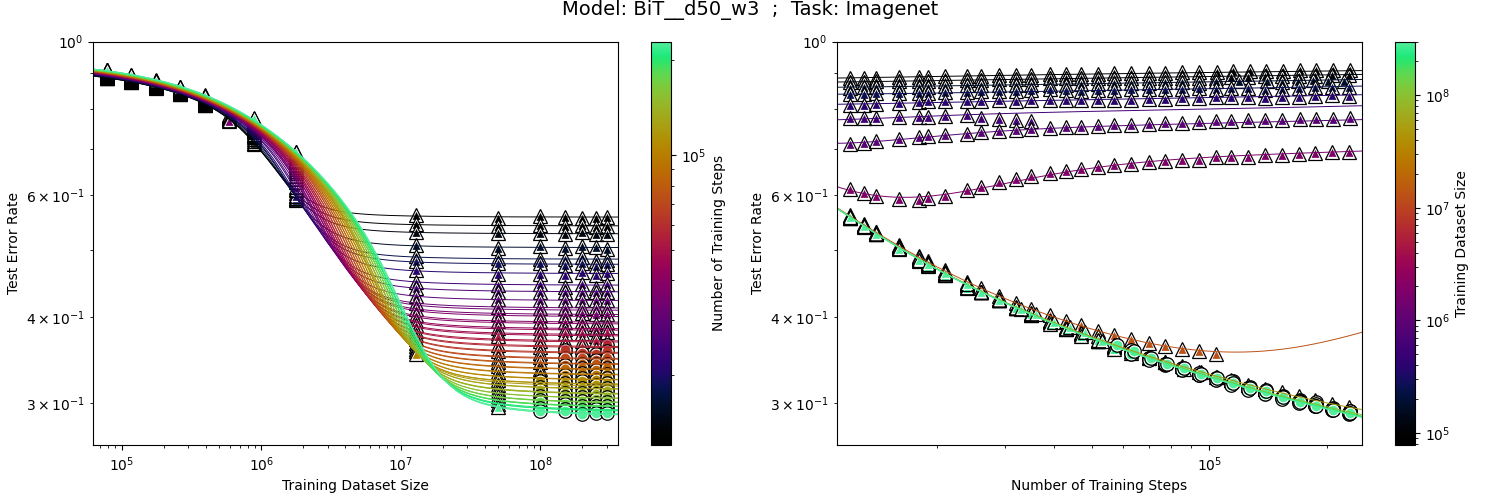}
\includegraphics[width=0.3938\textwidth]{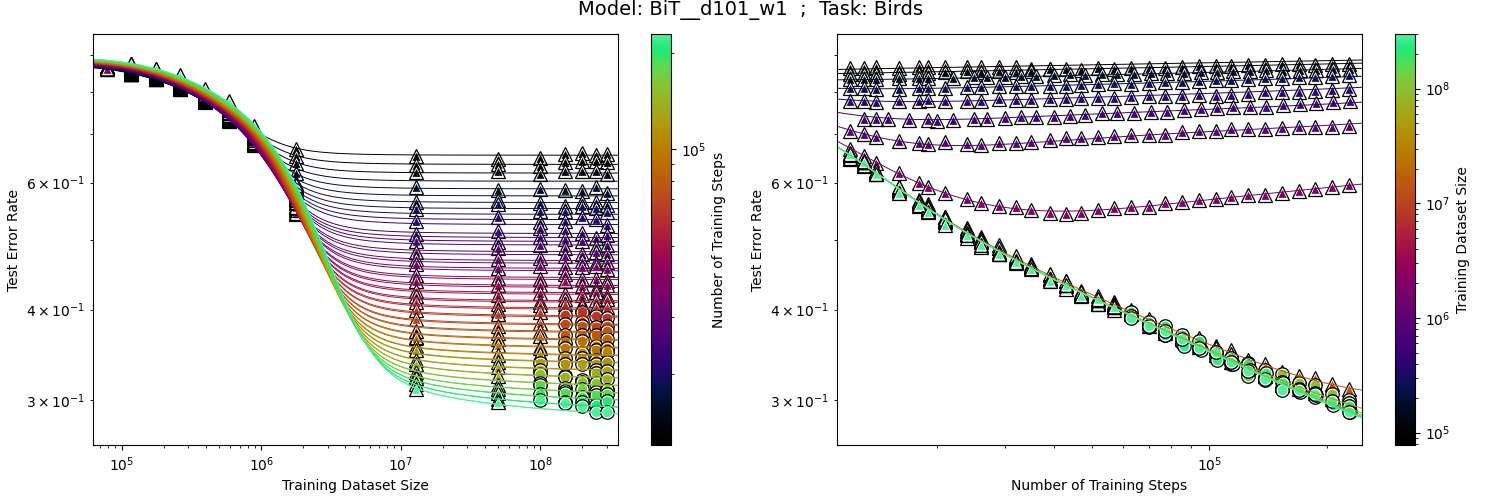}
\includegraphics[width=0.3938\textwidth]{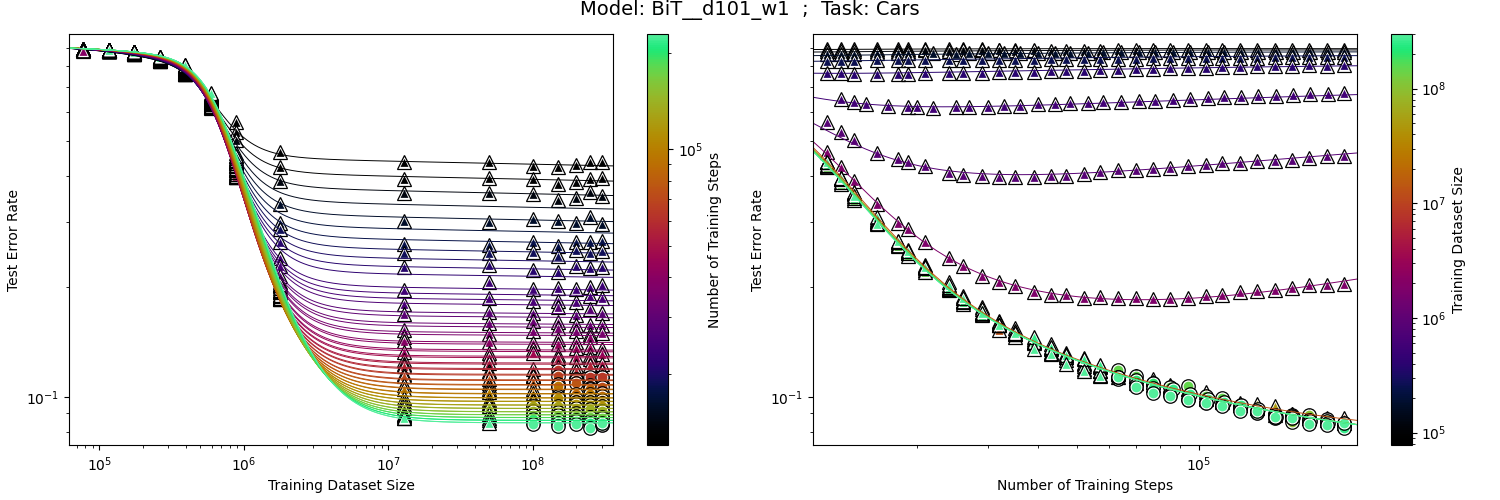}
\includegraphics[width=0.3938\textwidth]{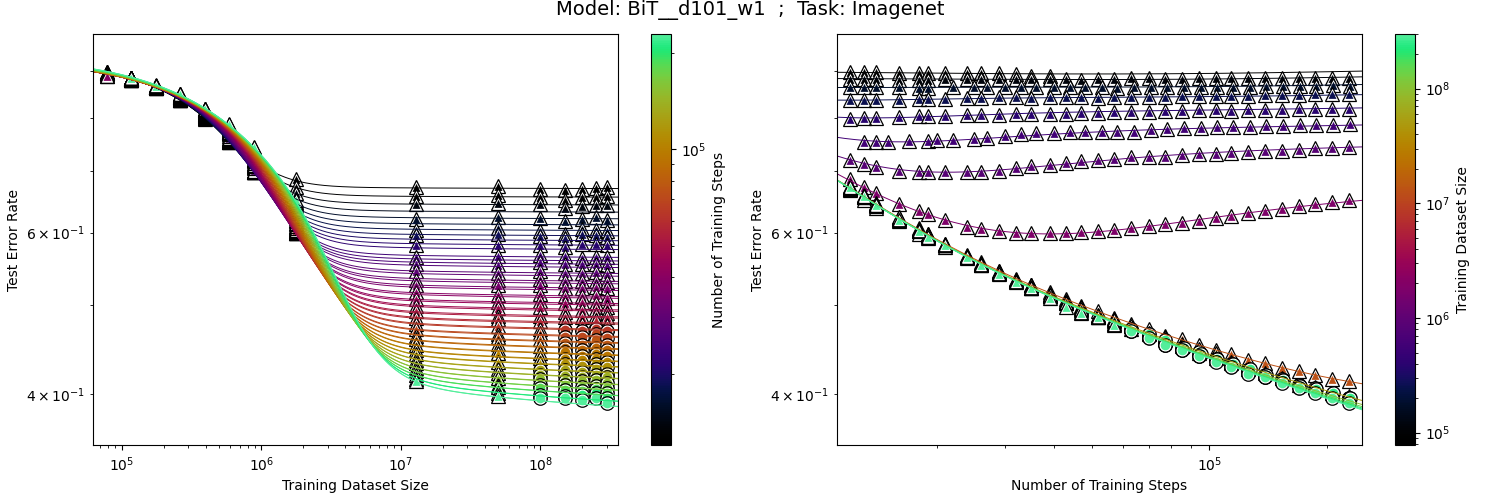}
\includegraphics[width=0.3938\textwidth]{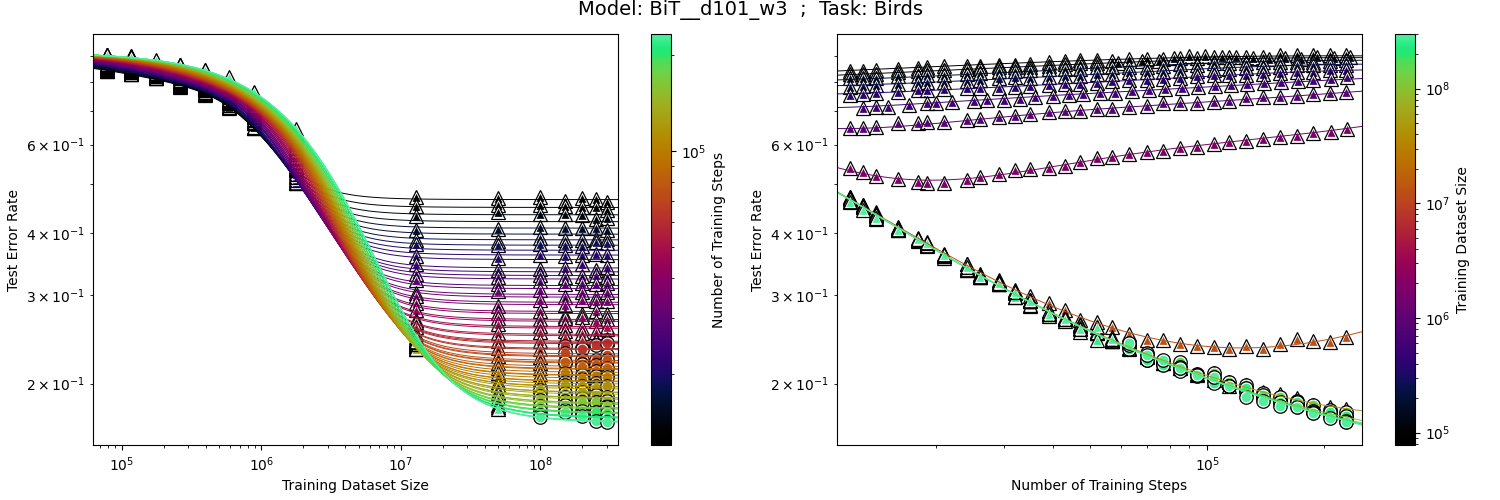}
\includegraphics[width=0.3938\textwidth]{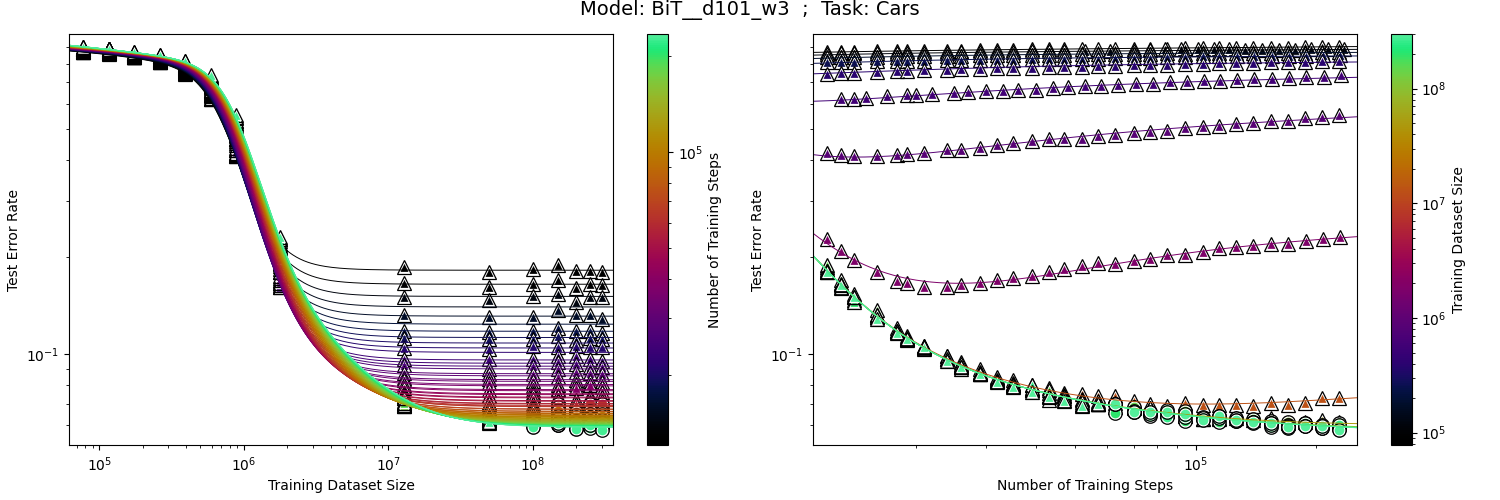}
\includegraphics[width=0.3938\textwidth]{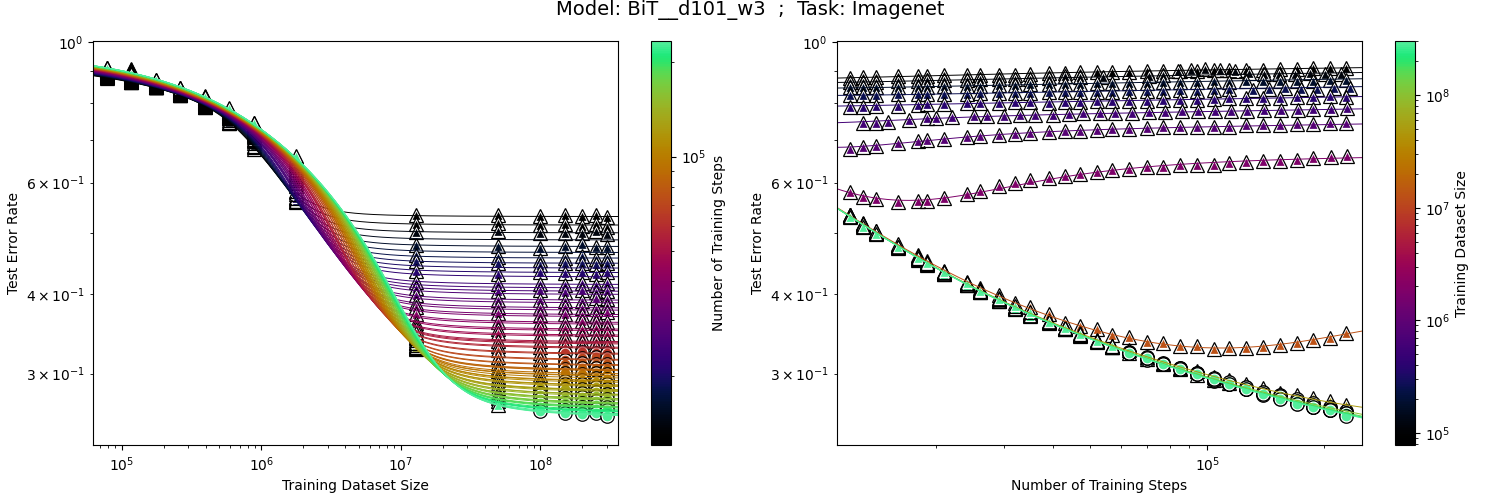}
\includegraphics[width=0.3938\textwidth]{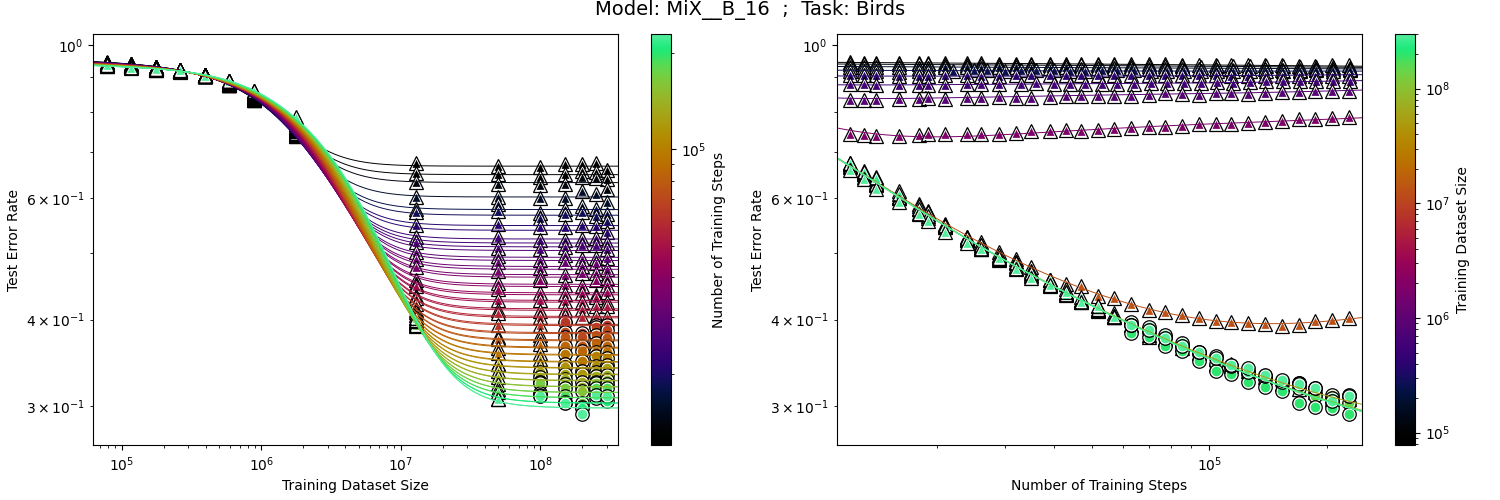}
\includegraphics[width=0.3938\textwidth]{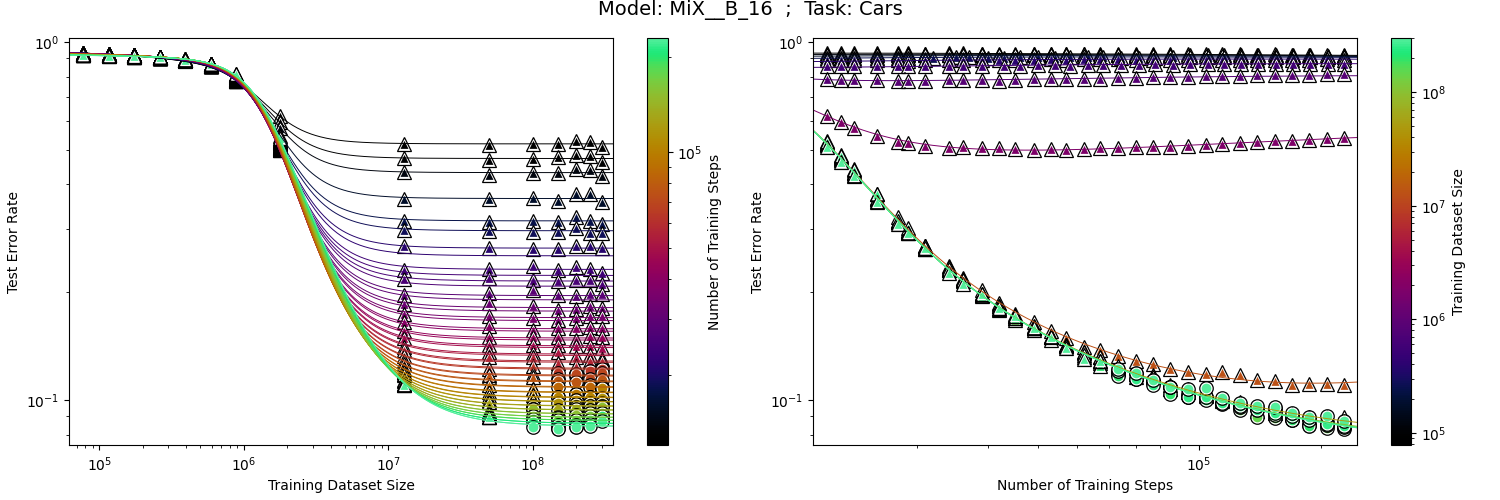}
\includegraphics[width=0.3938\textwidth]{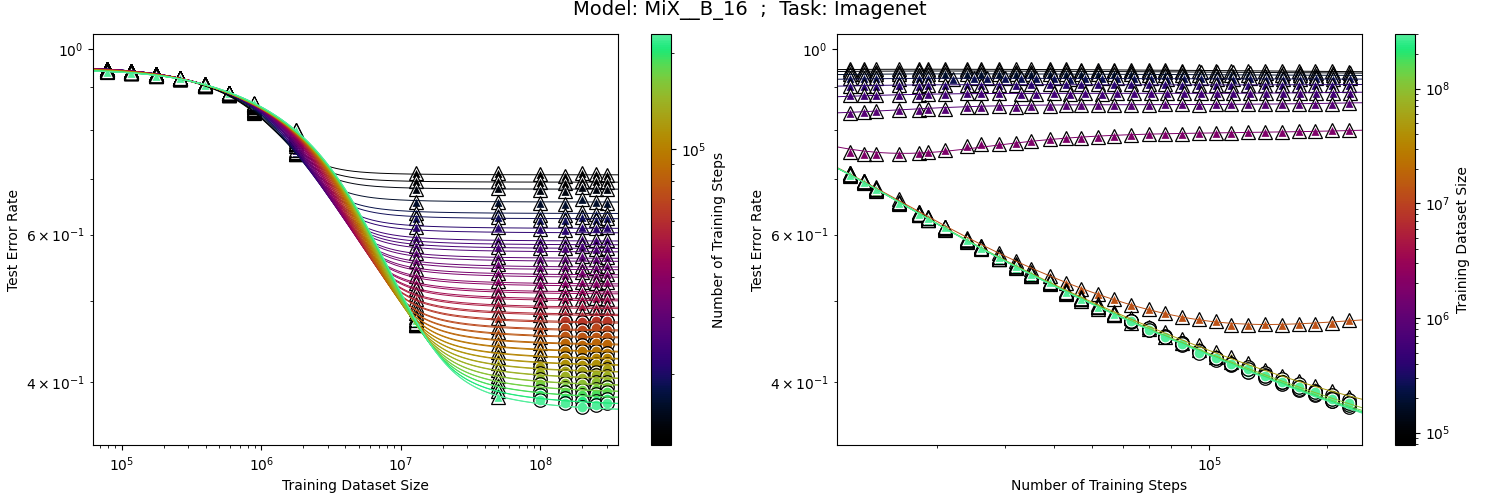}
\includegraphics[width=0.3938\textwidth]{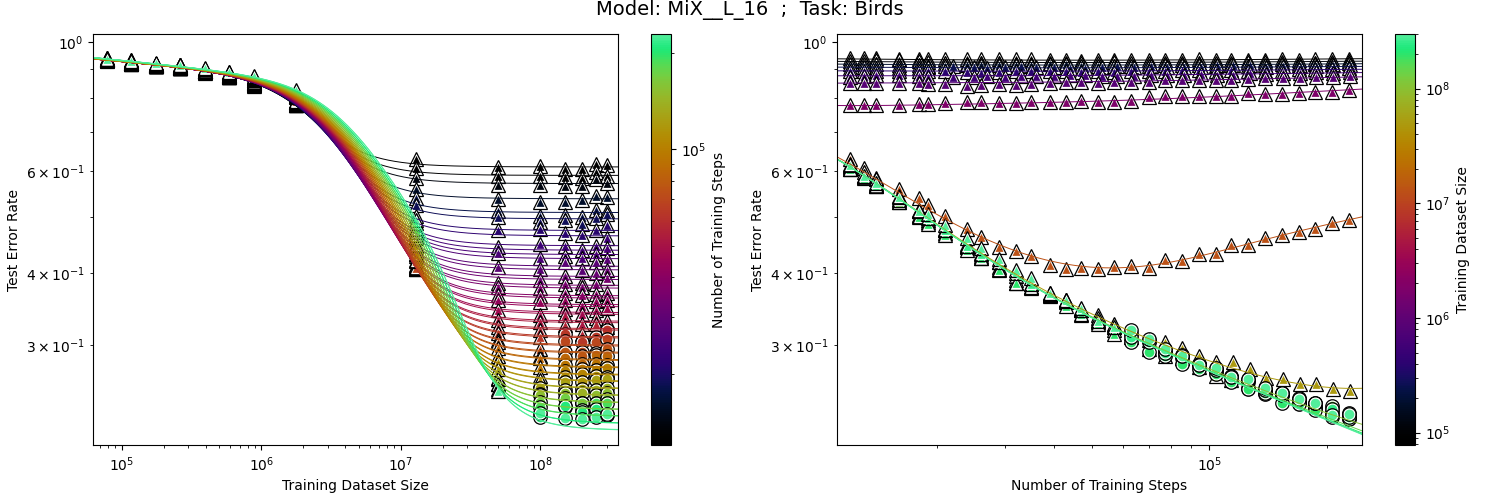}
\includegraphics[width=0.3938\textwidth]{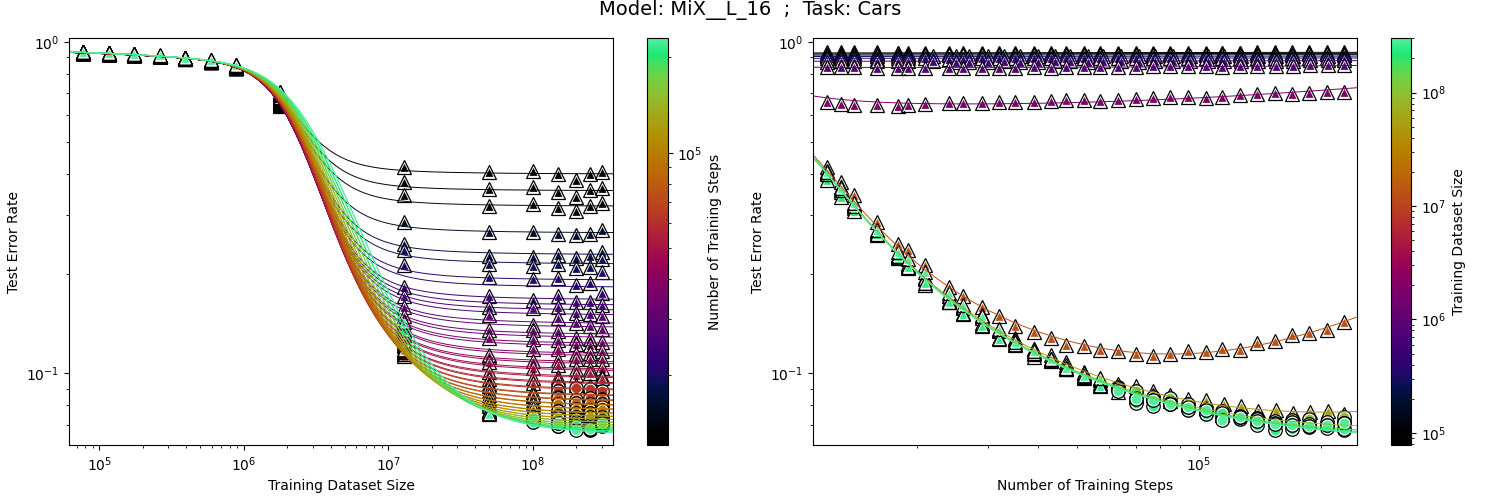}
\includegraphics[width=0.3938\textwidth]{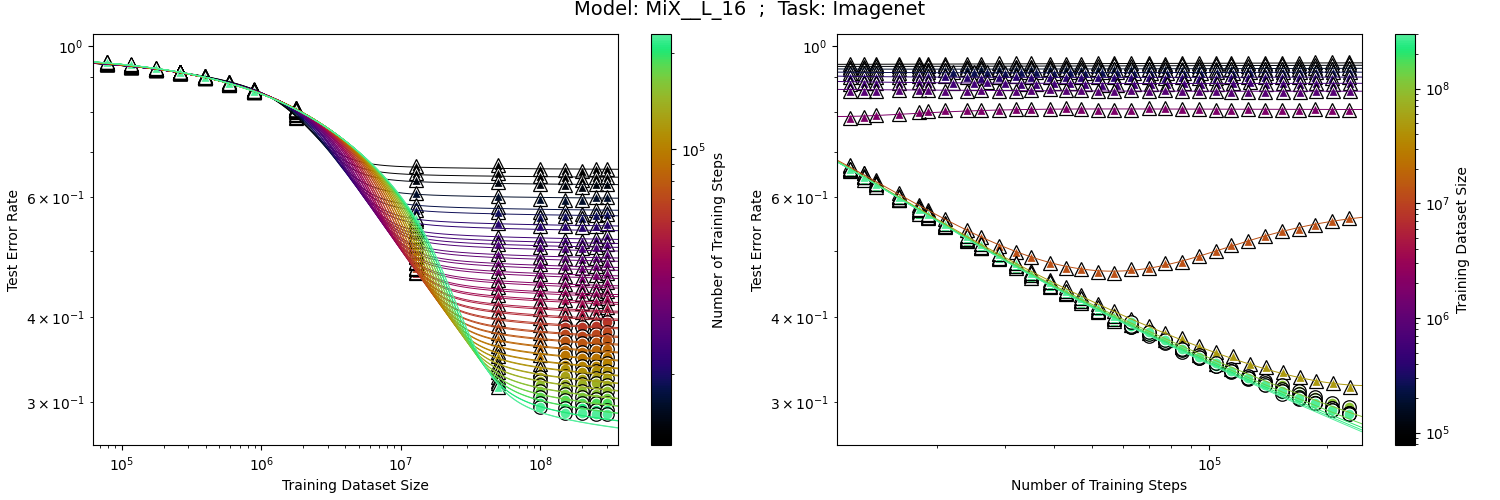}
\includegraphics[width=0.3938\textwidth]{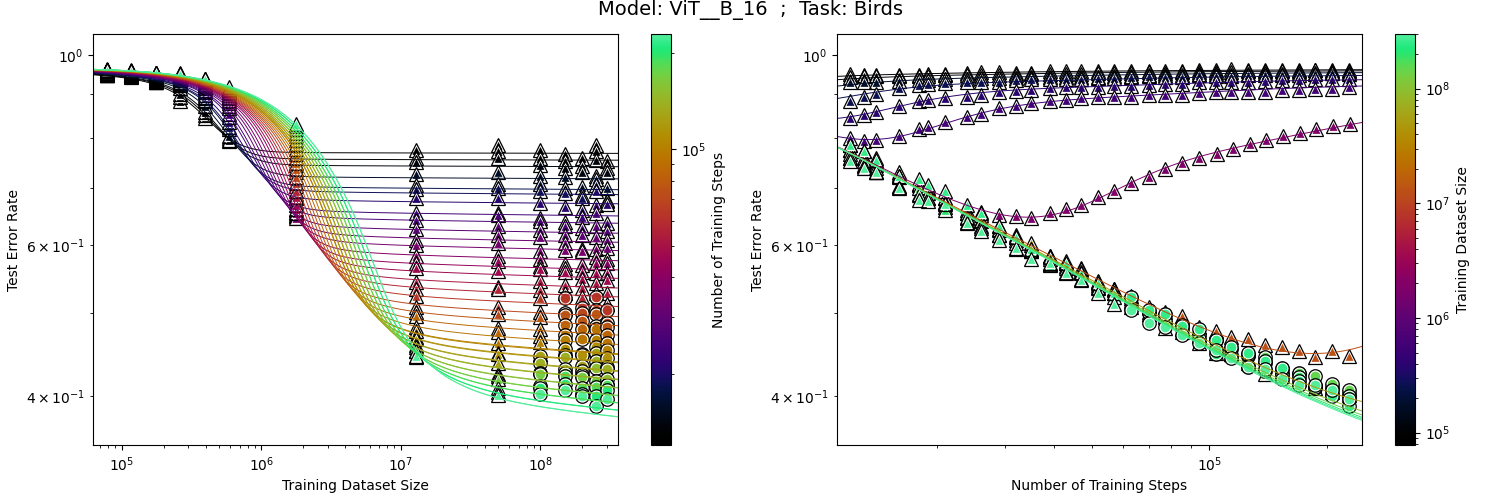}
\includegraphics[width=0.3938\textwidth]{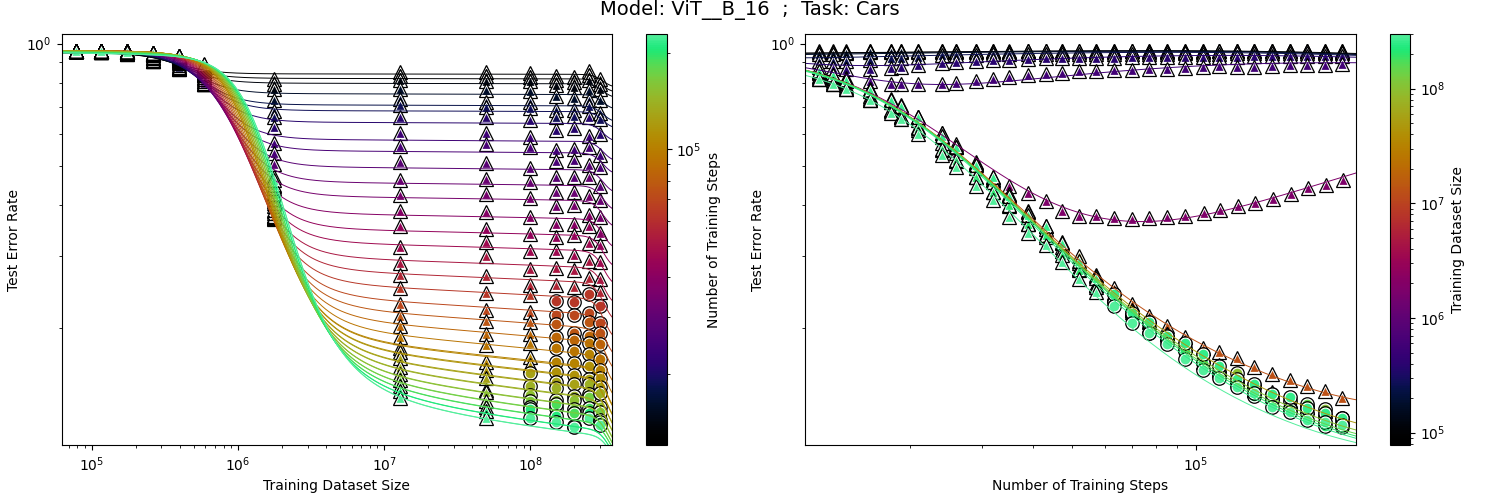}
\includegraphics[width=0.3938\textwidth]{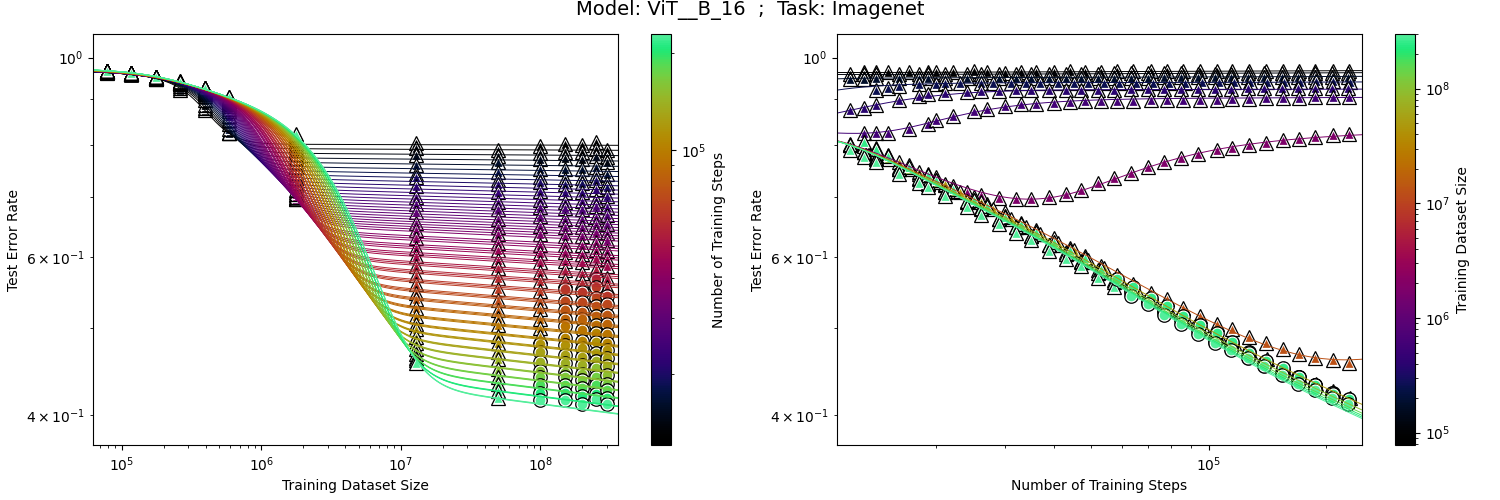}
\hspace{29.0mm}\includegraphics[width=0.179\textwidth]{figures/legend/legend2.png}
    \caption{
    Extrapolation Results of A3 functional form on bivariate scaling behavior of downstream vision performance. See Section \ref{section:vision} for more details.
    }
    \label{fig:a3_downstream_vision}
\end{figure*}

\FloatBarrier

\subsubsection{Trivariate}
\label{subsubsection:extrapolation_benchmark_vision_trivariate}

\begin{figure*}[h]    \centering
\begin{minipage}{0.75\textwidth}
\includegraphics[width=0.91\textwidth]{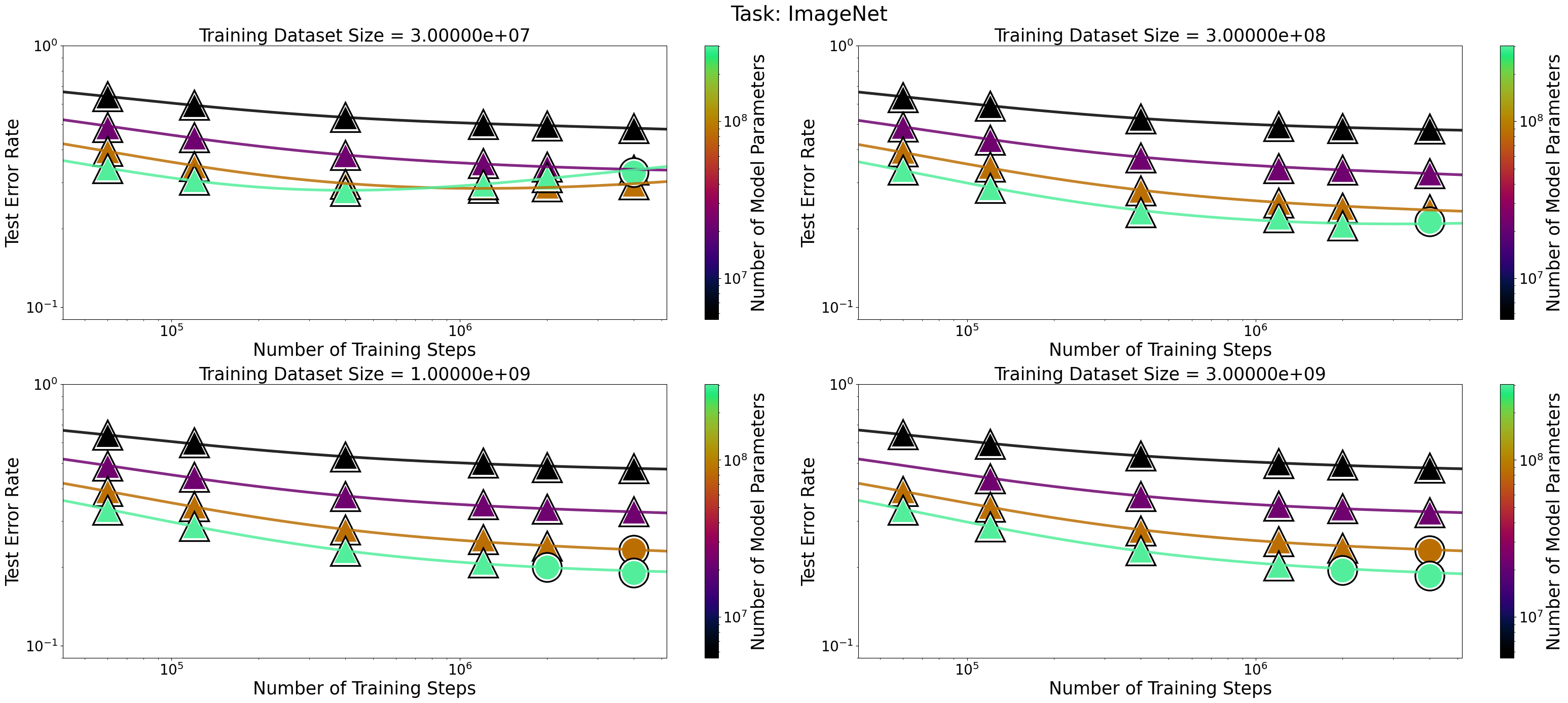}


\includegraphics[width=0.91\textwidth]{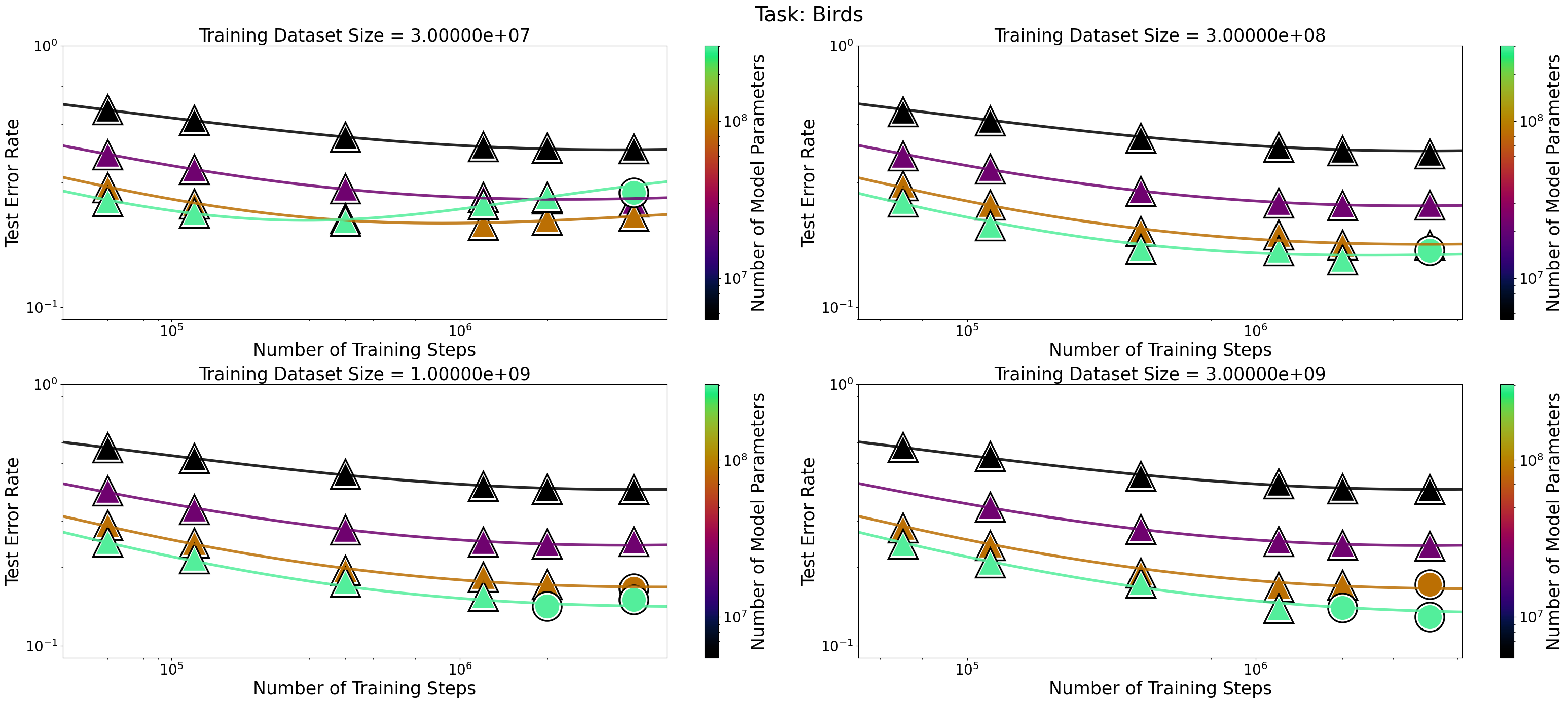}


\end{minipage}
\begin{minipage}{0.19\textwidth}\includegraphics[width=0.9\textwidth]{figures/legend/legend.png}
\end{minipage}


    \caption{
    Extrapolation Results of UNSL functional form on trivariate scaling behavior of downstream vision performance. See Section \ref{section:vision} for more details.
    }
    \label{fig:unsl_trivariate_vision}
\end{figure*}

\FloatBarrier

\begin{figure*}[h]    \centering
\begin{minipage}{0.75\textwidth}
\includegraphics[width=0.91\textwidth]{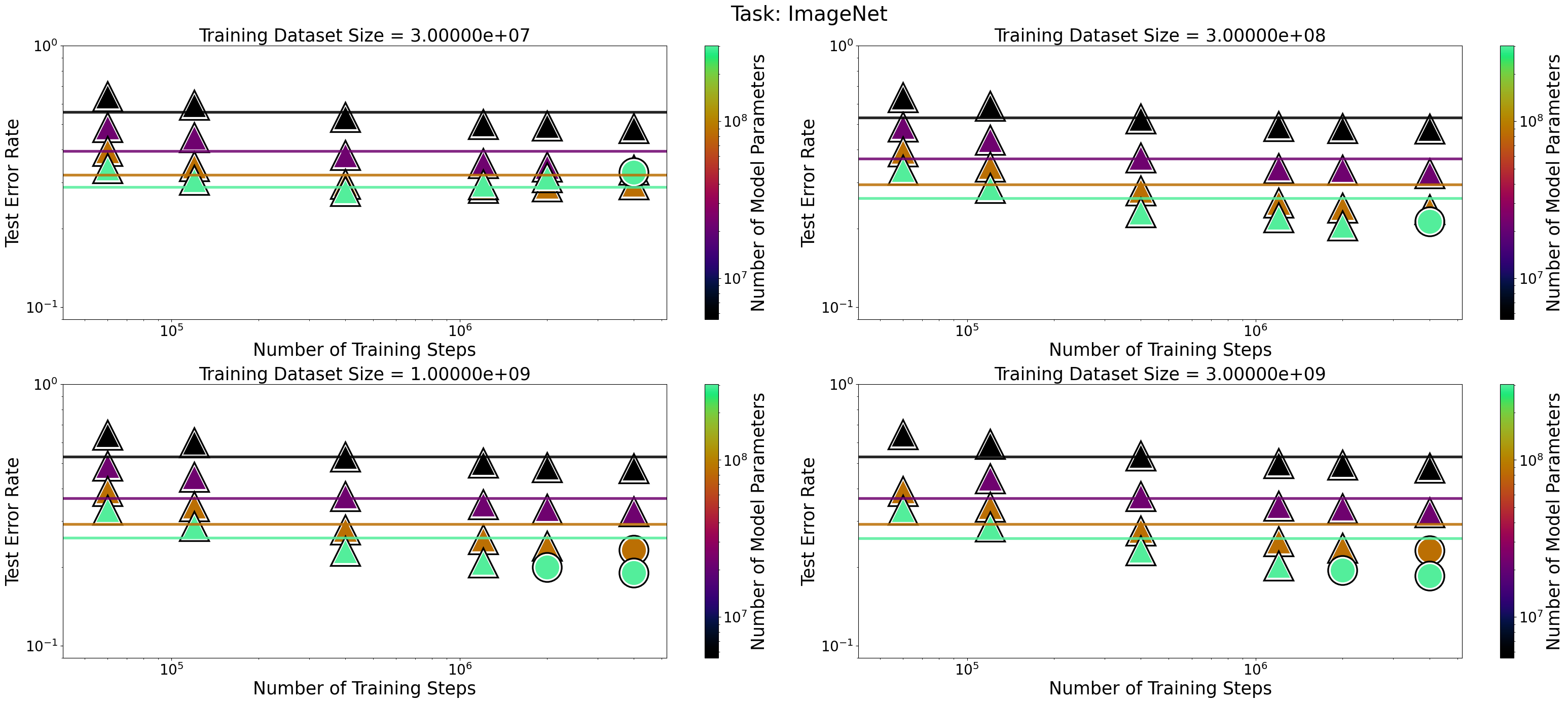}


\includegraphics[width=0.91\textwidth]{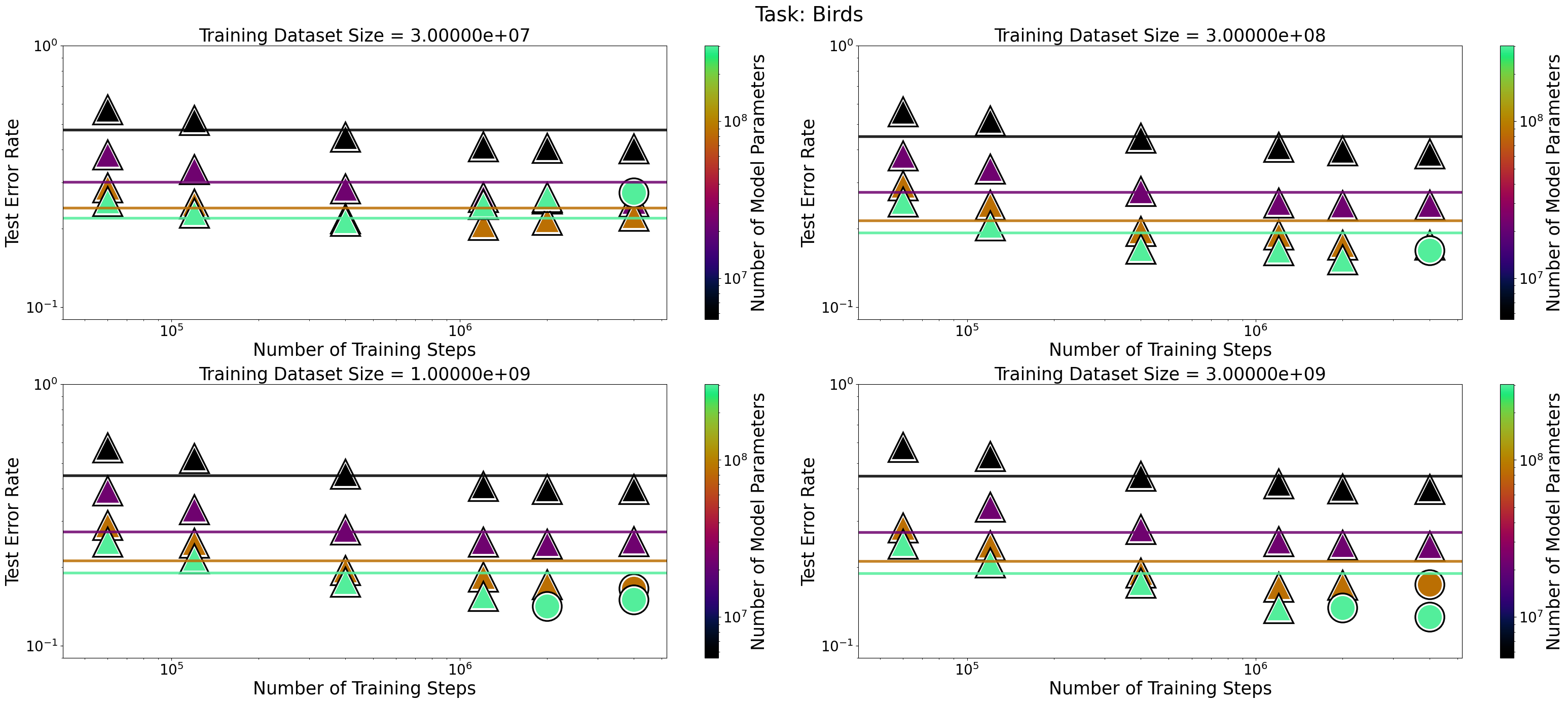}


\end{minipage}
\begin{minipage}{0.19\textwidth}\includegraphics[width=0.9\textwidth]{figures/legend/legend.png}
\end{minipage}


    \caption{
    Extrapolation Results of ``DC'' functional form of \citet{muennighoff2023scaling} on trivariate scaling behavior of downstream vision performance. See Section \ref{section:vision} for more details.
    }
    \label{fig:dc_trivariate_vision}
\end{figure*}

\FloatBarrier

\begin{figure*}[h]    \centering
\begin{minipage}{0.75\textwidth}
\includegraphics[width=0.91\textwidth]{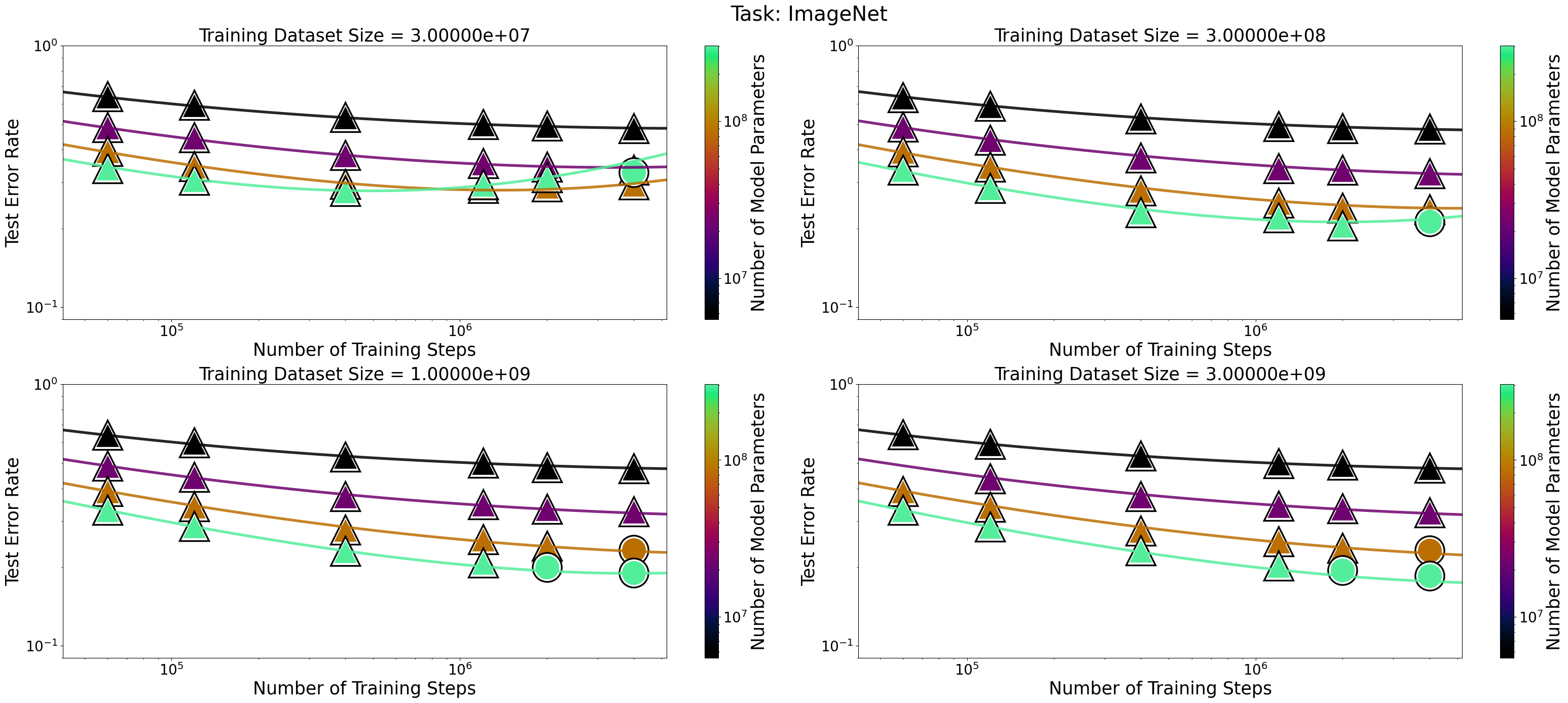}


\includegraphics[width=0.91\textwidth]{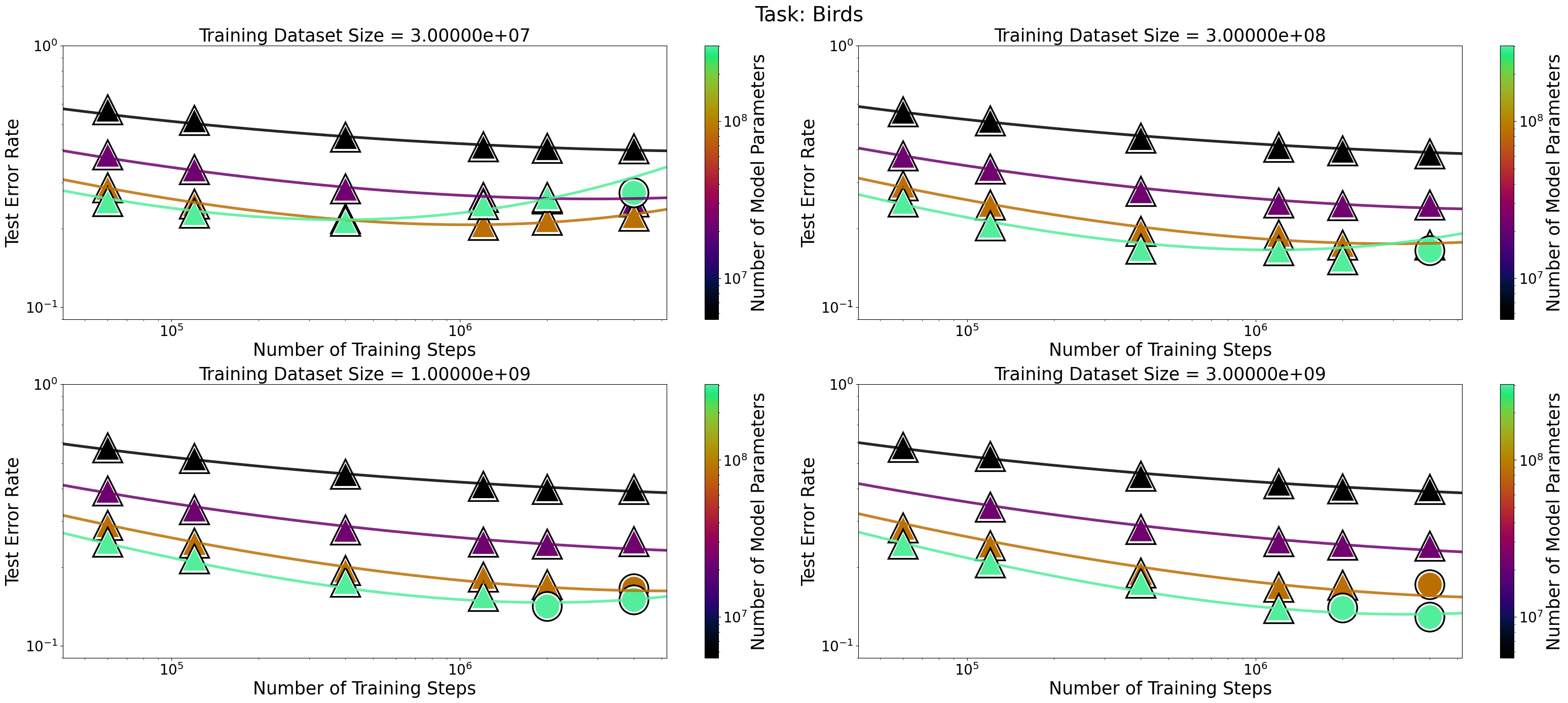}


\end{minipage}
\begin{minipage}{0.19\textwidth}\includegraphics[width=0.9\textwidth]{figures/legend/legend.png}
\end{minipage}


    \caption{
    Extrapolation Results of A1 functional form on trivariate scaling behavior of downstream vision performance. See Section \ref{section:vision} for more details.
    }
    \label{fig:a1_trivariate_vision}
\end{figure*}

\FloatBarrier

\begin{figure*}[h]    \centering
\begin{minipage}{0.75\textwidth}
\includegraphics[width=0.91\textwidth]{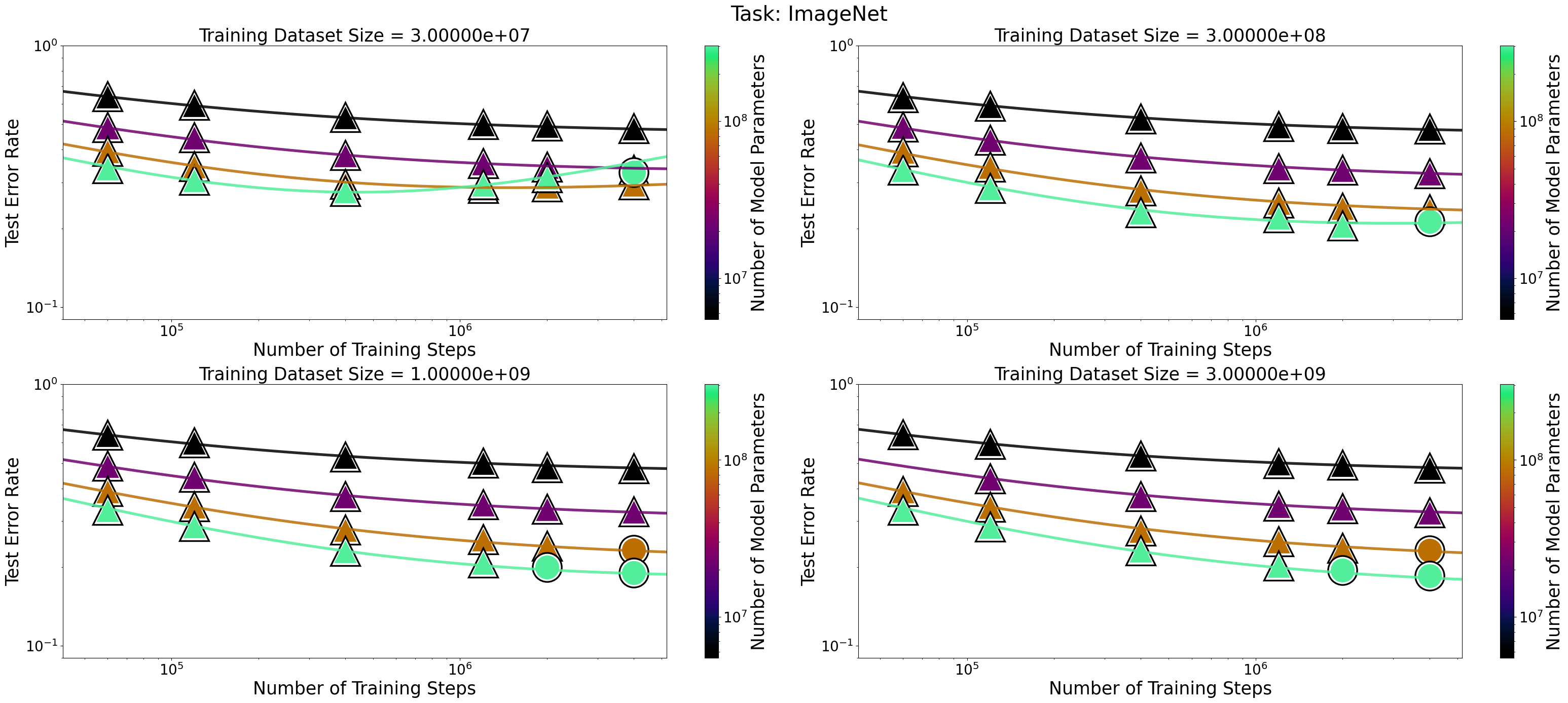}


\includegraphics[width=0.91\textwidth]{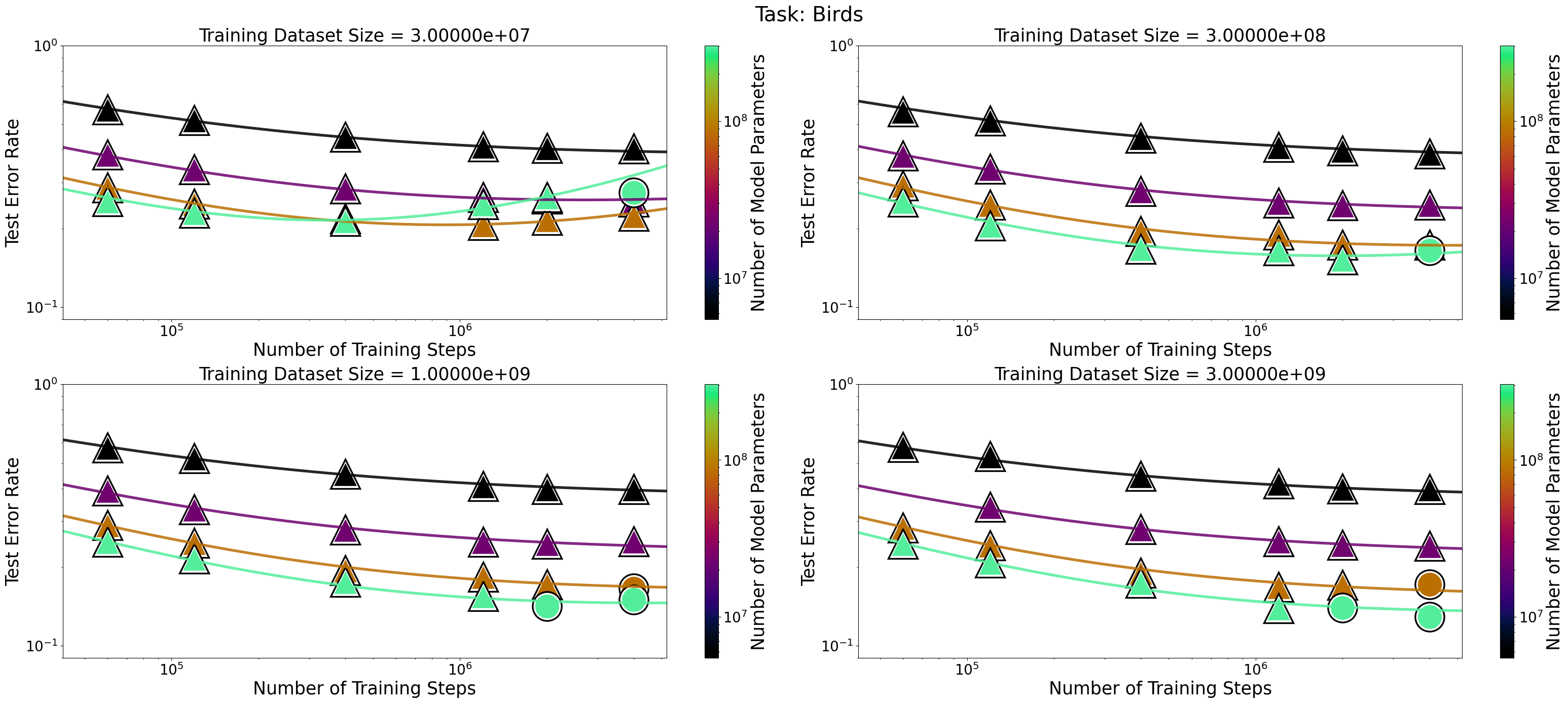}


\end{minipage}
\begin{minipage}{0.19\textwidth}\includegraphics[width=0.9\textwidth]{figures/legend/legend.png}
\end{minipage}


    \caption{
    Extrapolation Results of A2 functional form on trivariate scaling behavior of downstream vision performance. See Section \ref{section:vision} for more details.
    }
    \label{fig:a2_trivariate_vision}
\end{figure*}

\FloatBarrier

\begin{figure*}[h]    \centering
\begin{minipage}{0.75\textwidth}
\includegraphics[width=0.91\textwidth]{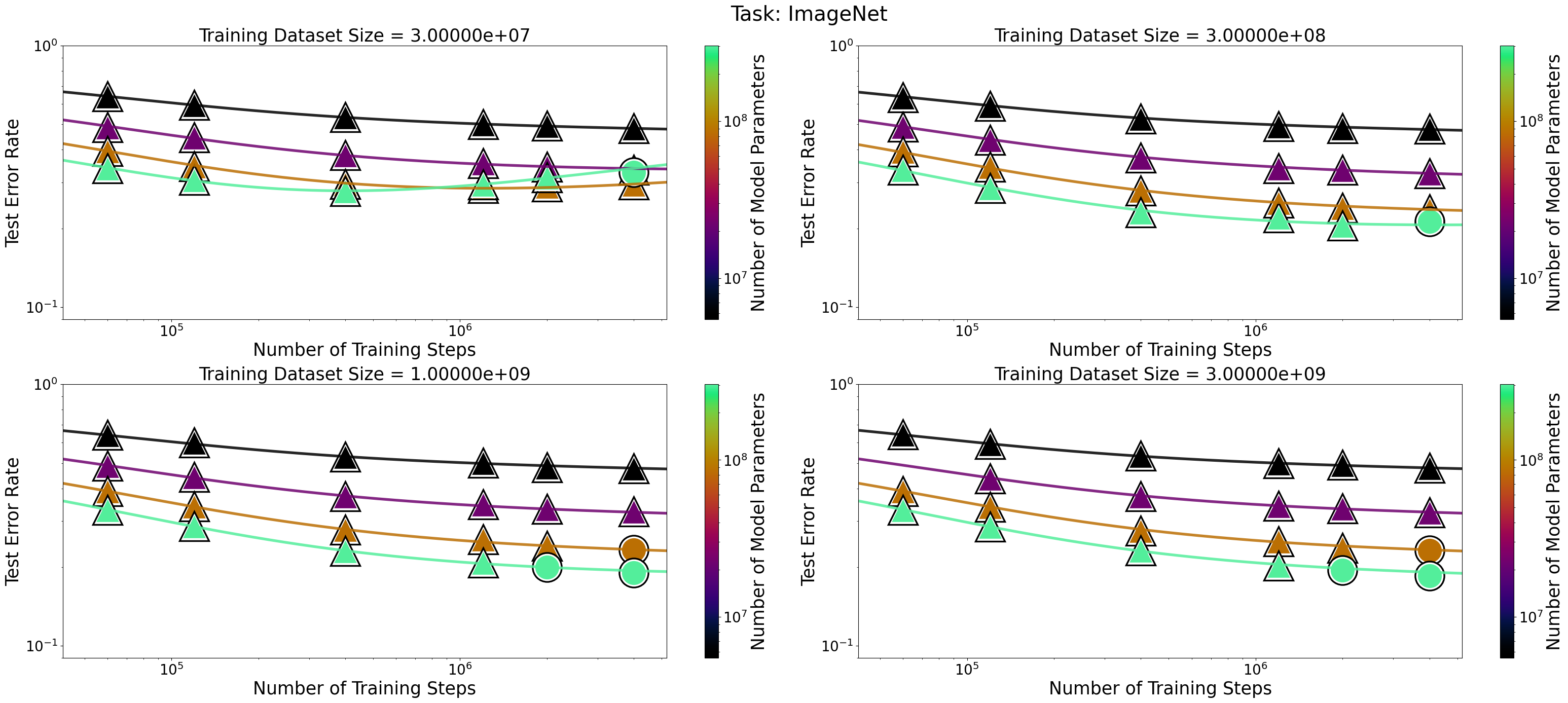}


\includegraphics[width=0.91\textwidth]{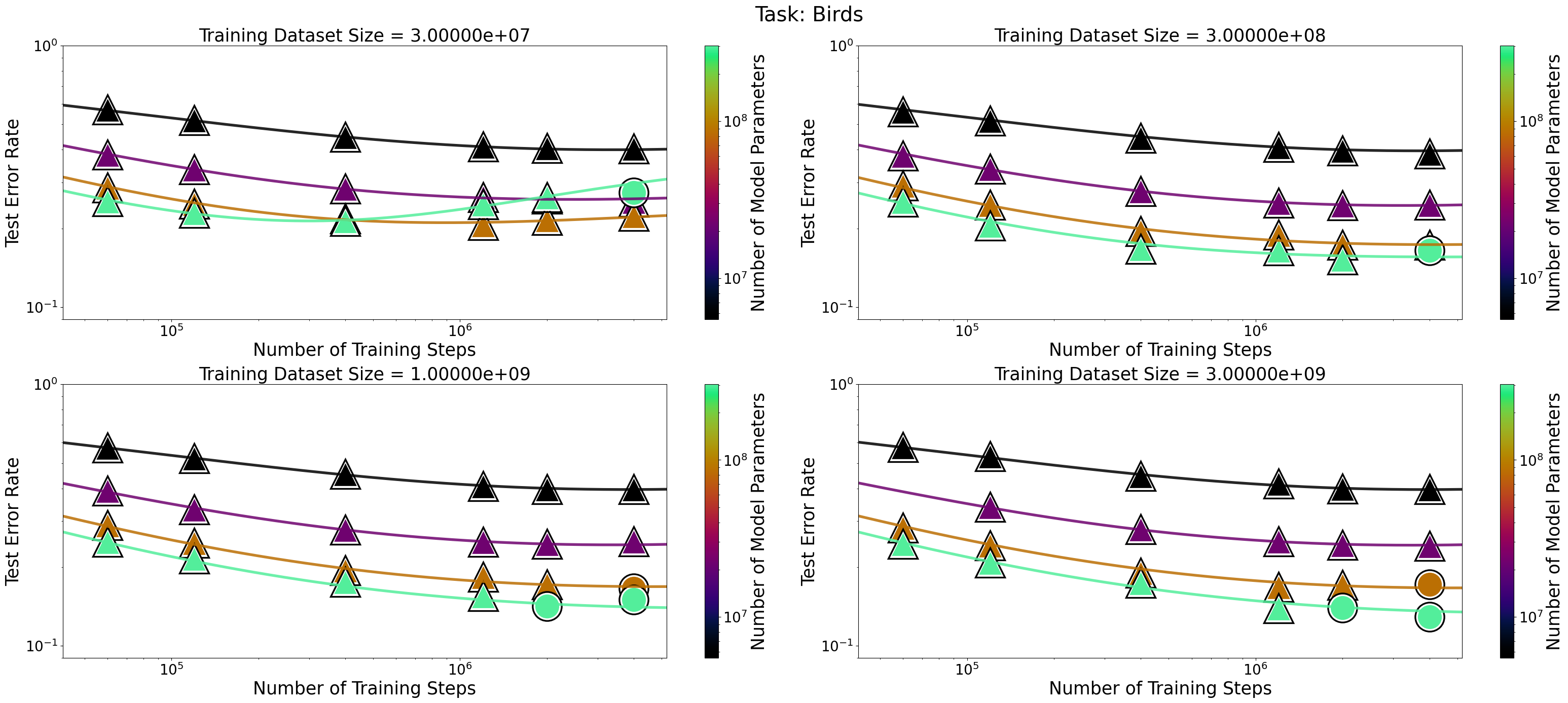}


\end{minipage}
\begin{minipage}{0.19\textwidth}\includegraphics[width=0.9\textwidth]{figures/legend/legend.png}
\end{minipage}


    \caption{
    Extrapolation Results of A3 functional form on trivariate scaling behavior of downstream vision performance. See Section \ref{section:vision} for more details.
    }
    \label{fig:a3_trivariate_vision}
\end{figure*}

\FloatBarrier

\FloatBarrier

\subsection{Plots of Language Extrapolation Results}
\label{subsection:extrapolation_benchmark_language}

\subsubsection{Trivariate}
\label{subsubsection:extrapolation_benchmark_language_trivariate}
\FloatBarrier

\FloatBarrier
\begin{figure*}[h]    \centering

\includegraphics[width=0.99\textwidth]{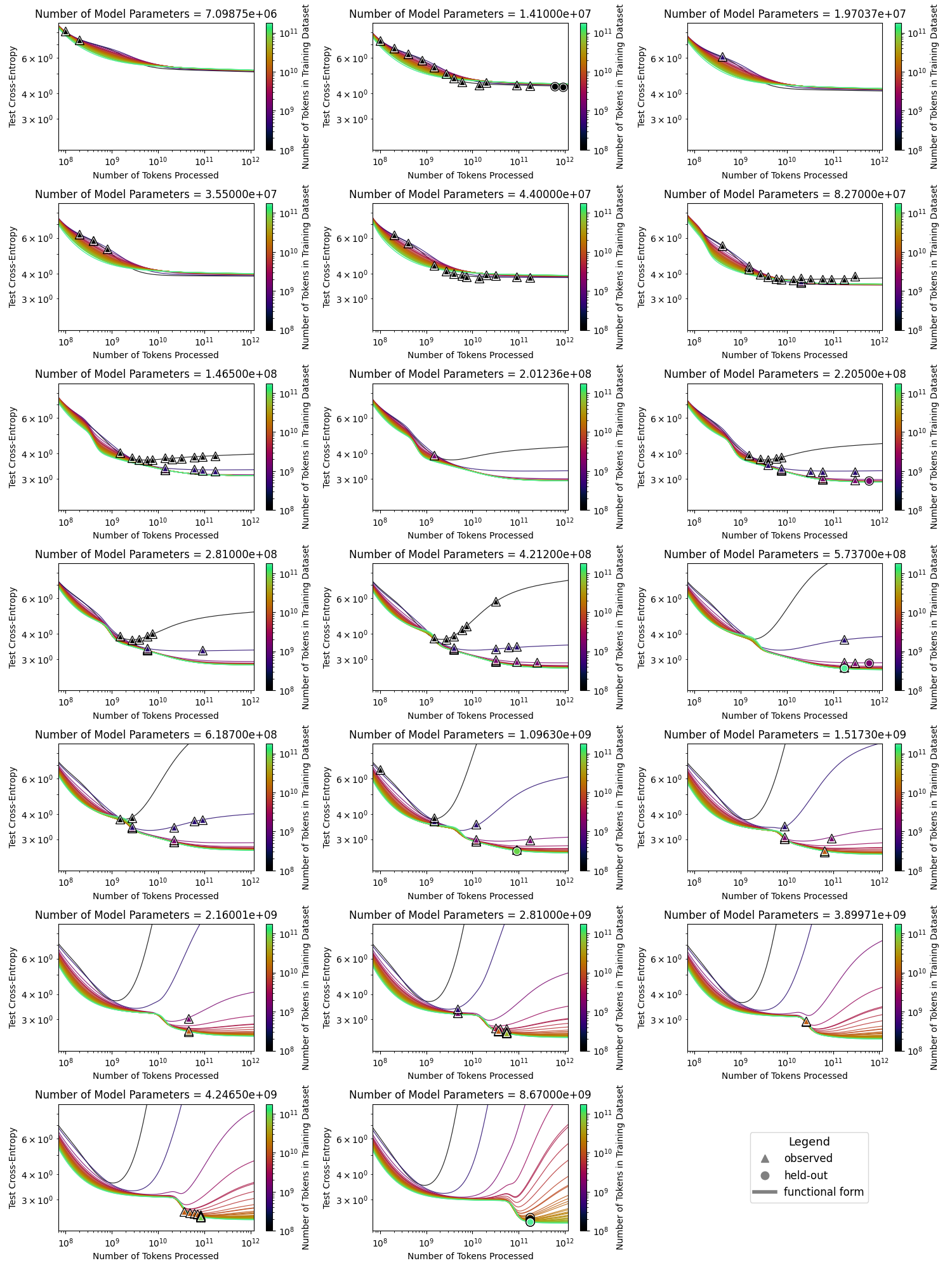}
    \caption{
    Extrapolation Results of UNSL on trivariate scaling behavior of language performance. All 20 plots are slices of single functional form fit to a single trivariate scaling behavior. The title of each plot represents the number of model parameters, the x-axis of each plot represents the number of training steps times the batch size, and the color bar of each plot represents the training dataset size. See Section \ref{section:language} for more details.
    }
    \label{fig:unsl_llm_trivariate}
\end{figure*}

\FloatBarrier
\begin{figure*}[h]    \centering

\includegraphics[width=0.99\textwidth]{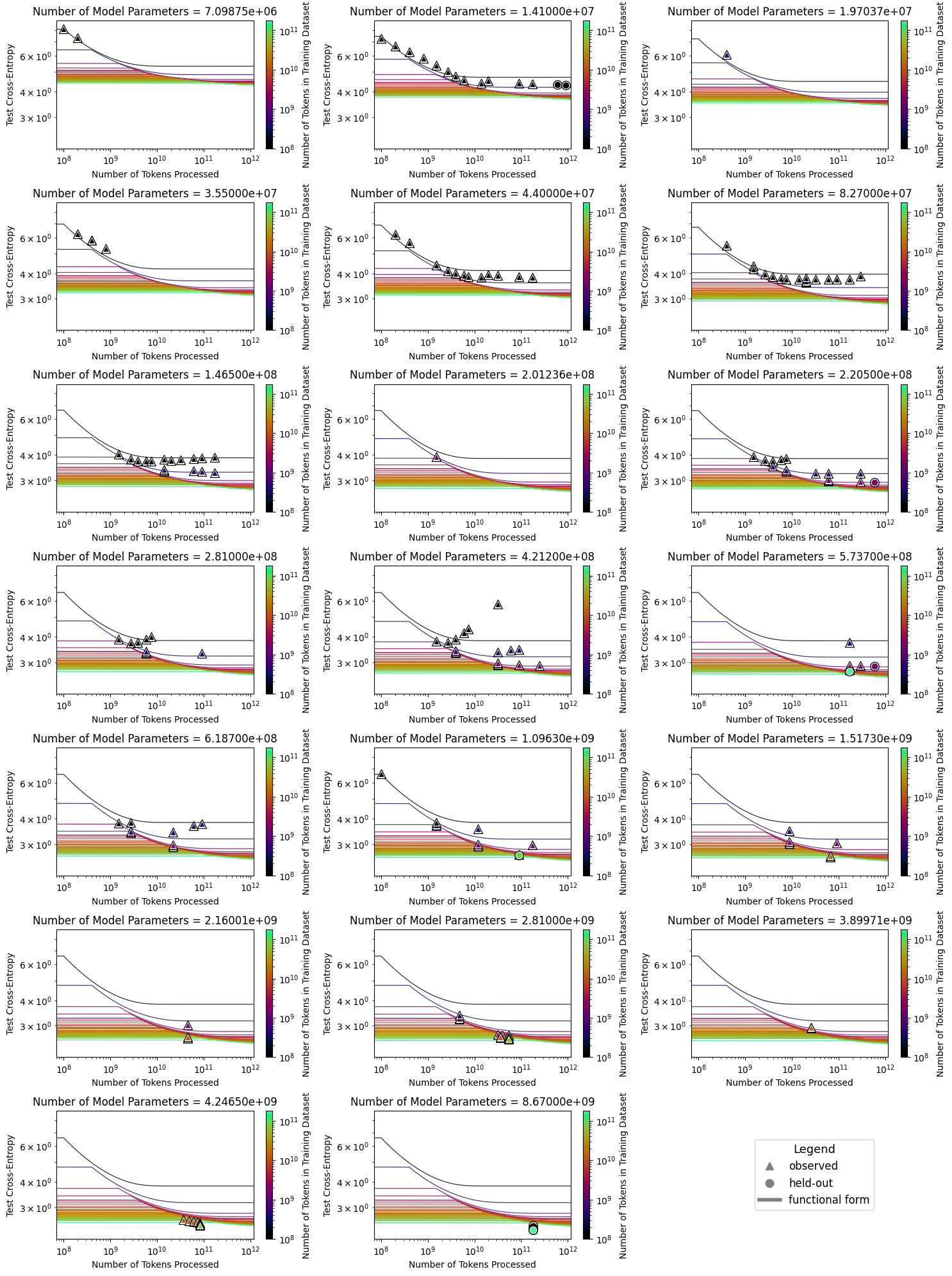}
    \caption{
    Extrapolation Results of ``DC'' functional form of \citet{muennighoff2023scaling} on trivariate scaling behavior of language performance. All 20 plots are slices of single functional form fit to a single trivariate scaling behavior. The title of each plot represents the number of model parameters, the x-axis of each plot represents the number of training steps times the batch size, and the color bar of each plot represents the training dataset size. See Section \ref{section:language} for more details.
    }
    \label{fig:dc_llm_trivariate}
\end{figure*}

\FloatBarrier
\begin{figure*}[h]    \centering

\includegraphics[width=0.99\textwidth]{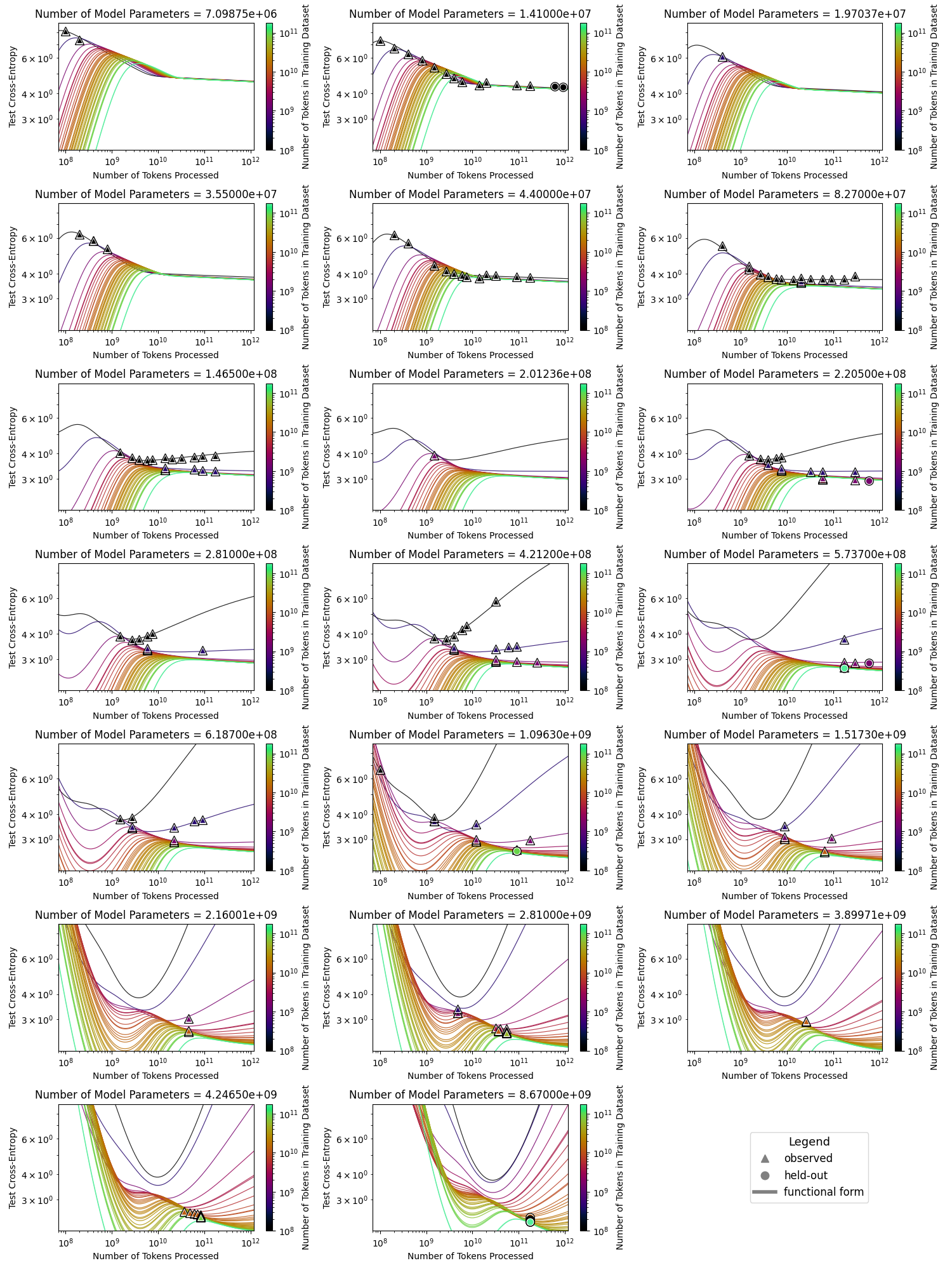}
    \caption{
    Extrapolation Results of A1 functional form on trivariate scaling behavior of language performance. All 20 plots are slices of single functional form fit to a single trivariate scaling behavior. The title of each plot represents the number of model parameters, the x-axis of each plot represents the number of training steps times the batch size, and the color bar of each plot represents the training dataset size. See Section \ref{section:language} for more details.
    }
    \label{fig:a1_llm_trivariate}
\end{figure*}

\FloatBarrier
\begin{figure*}[h]    \centering

\includegraphics[width=0.99\textwidth]{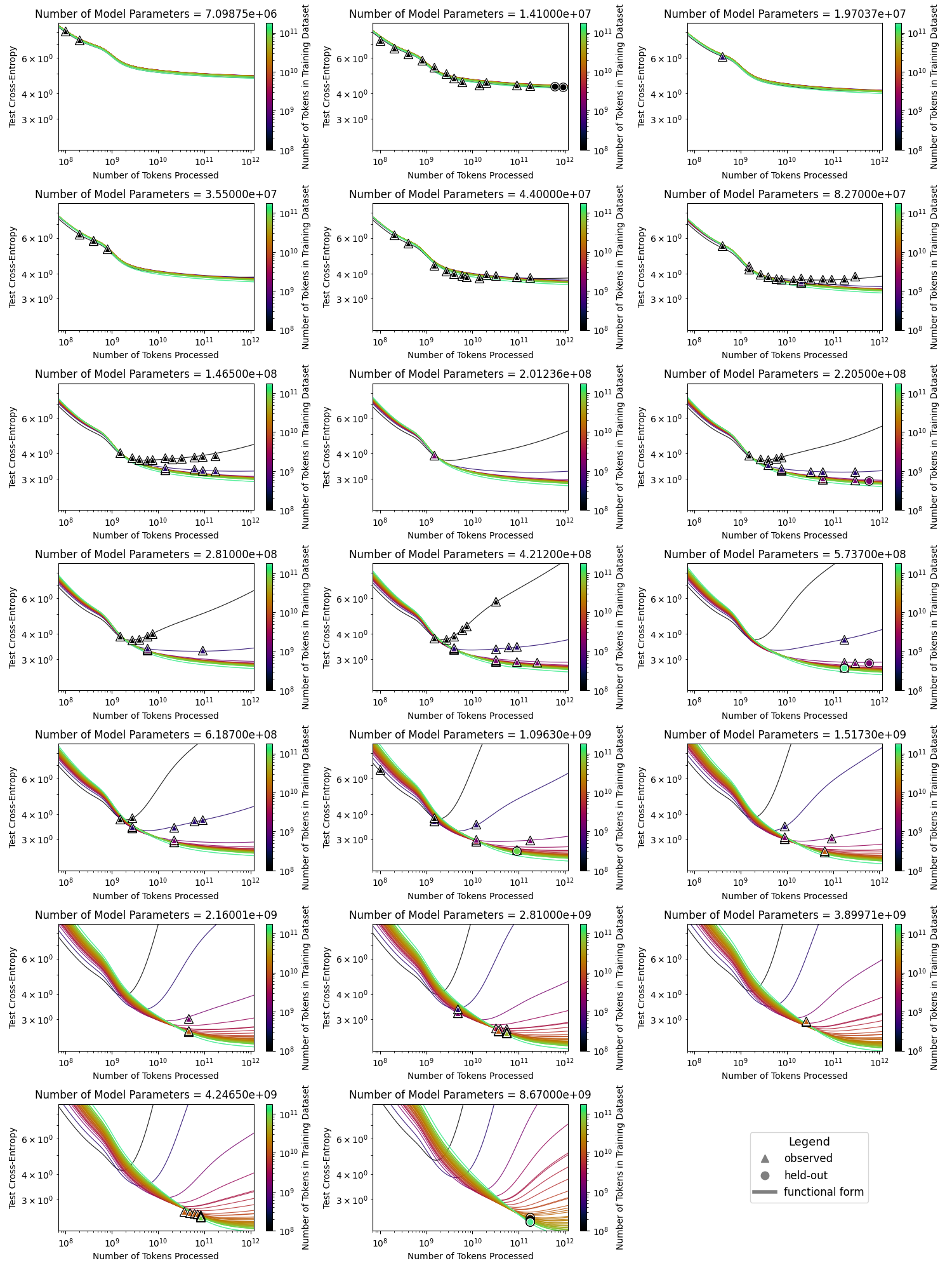}
    \caption{
    Extrapolation Results of A2 functional form on trivariate scaling behavior of language performance. All 20 plots are slices of single functional form fit to a single trivariate scaling behavior. The title of each plot represents the number of model parameters, the x-axis of each plot represents the number of training steps times the batch size, and the color bar of each plot represents the training dataset size. See Section \ref{section:language} for more details.
    }
    \label{fig:a2_llm_trivariate}
\end{figure*}

\FloatBarrier
\begin{figure*}[h]    \centering

\includegraphics[width=0.99\textwidth]{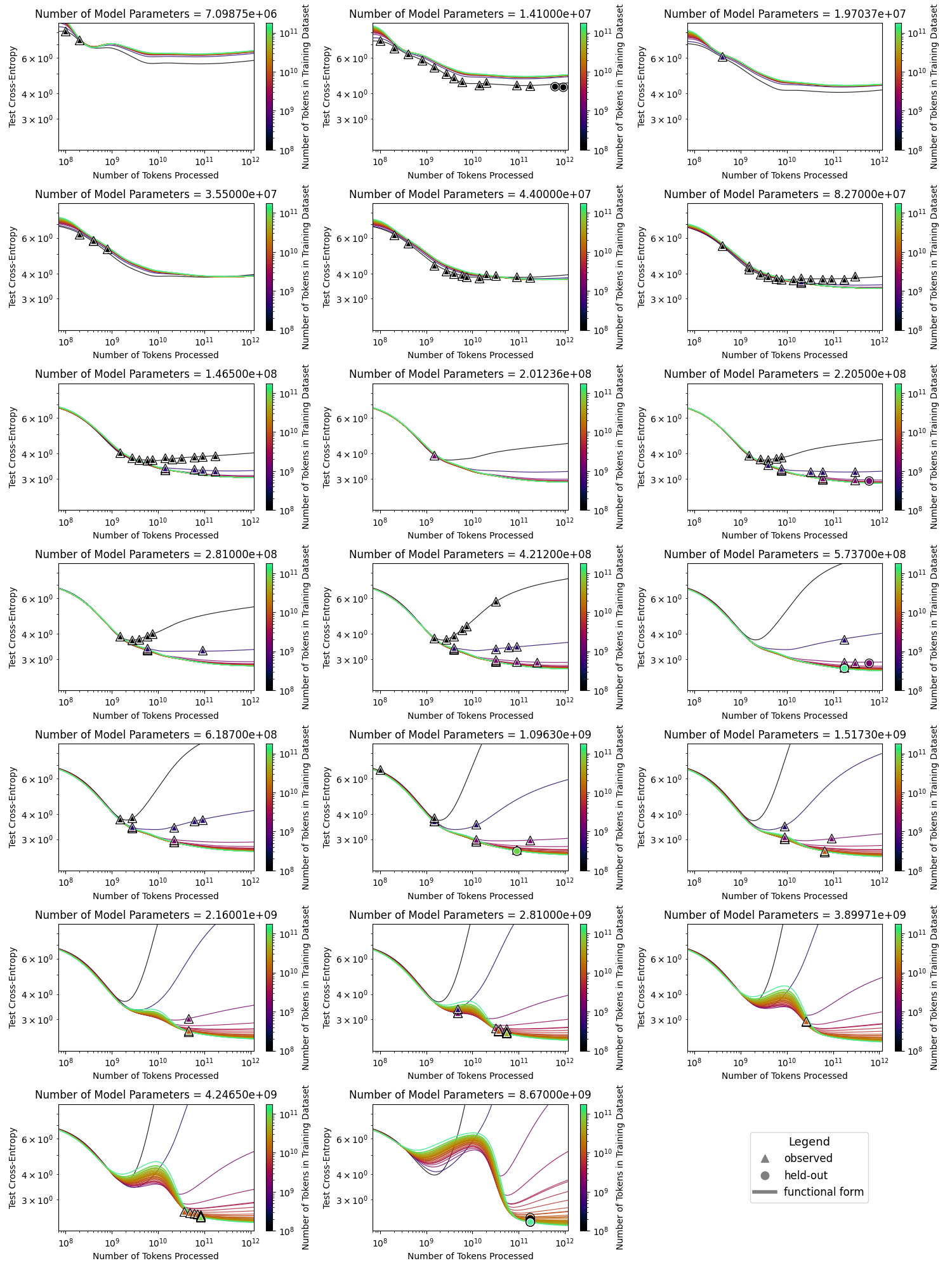}
    \caption{
    Extrapolation Results of A3 functional form on trivariate scaling behavior of language performance. All 20 plots are slices of single functional form fit to a single trivariate scaling behavior. The title of each plot represents the number of model parameters, the x-axis of each plot represents the number of training steps times the batch size, and the color bar of each plot represents the training dataset size. See Section \ref{section:language} for more details.
    }
    \label{fig:a3_llm_trivariate}
\end{figure*}

\FloatBarrier

\
\
\
\

\FloatBarrier
\vspace{-10mm}
\subsubsection{Bivariate}
\label{subsubsection:extrapolation_benchmark_language_bivariate}

\begin{figure*}[h]    \centering
\begin{minipage}{0.535\textwidth}
\includegraphics[width=1.0\textwidth]{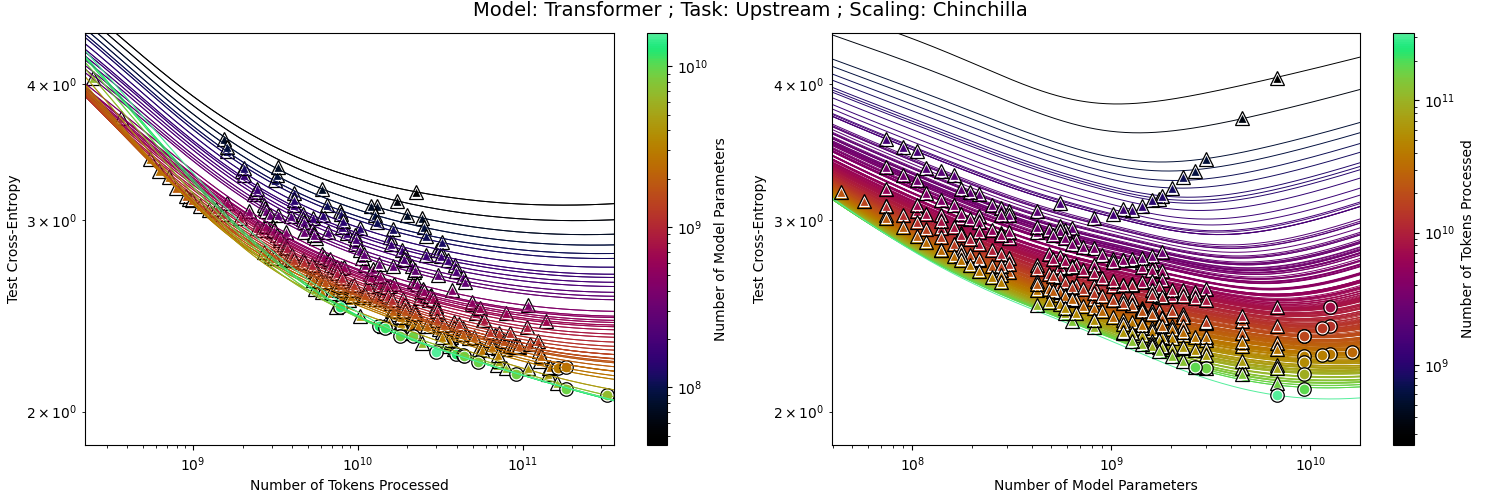}
\includegraphics[width=1.0\textwidth]{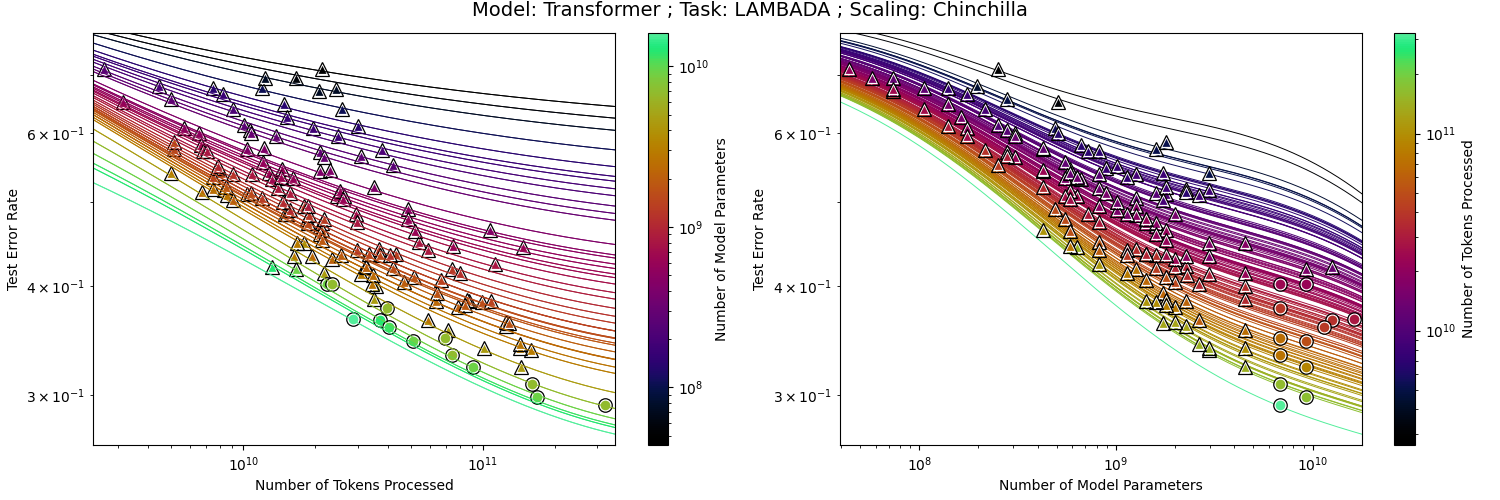}
\includegraphics[width=1.0\textwidth]{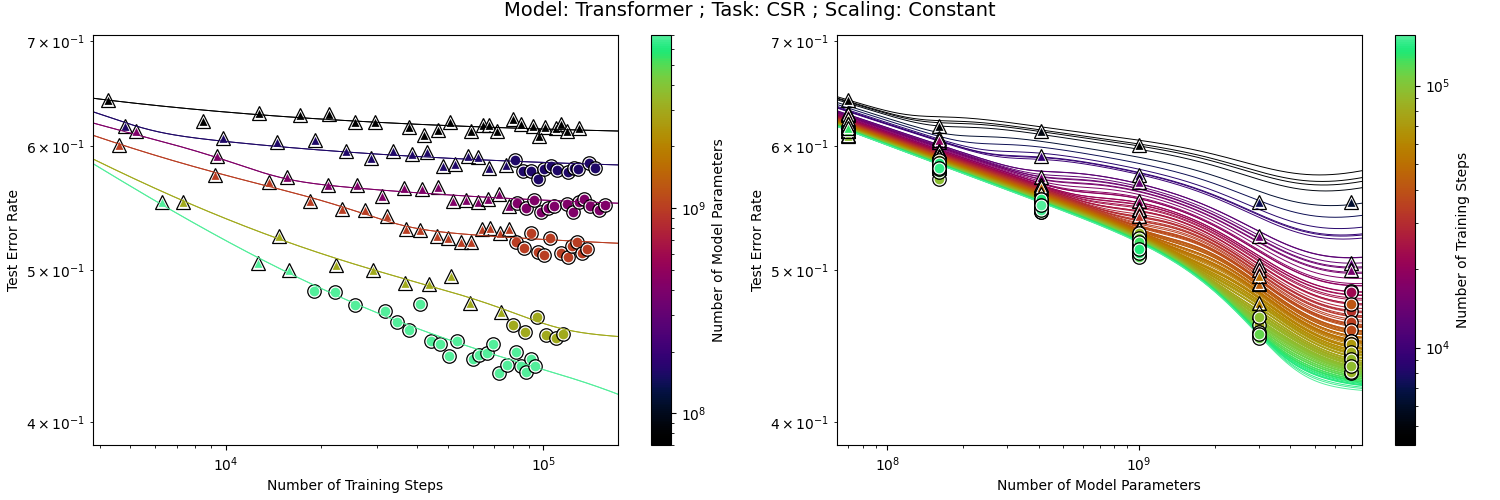}
\includegraphics[width=1.0\textwidth]{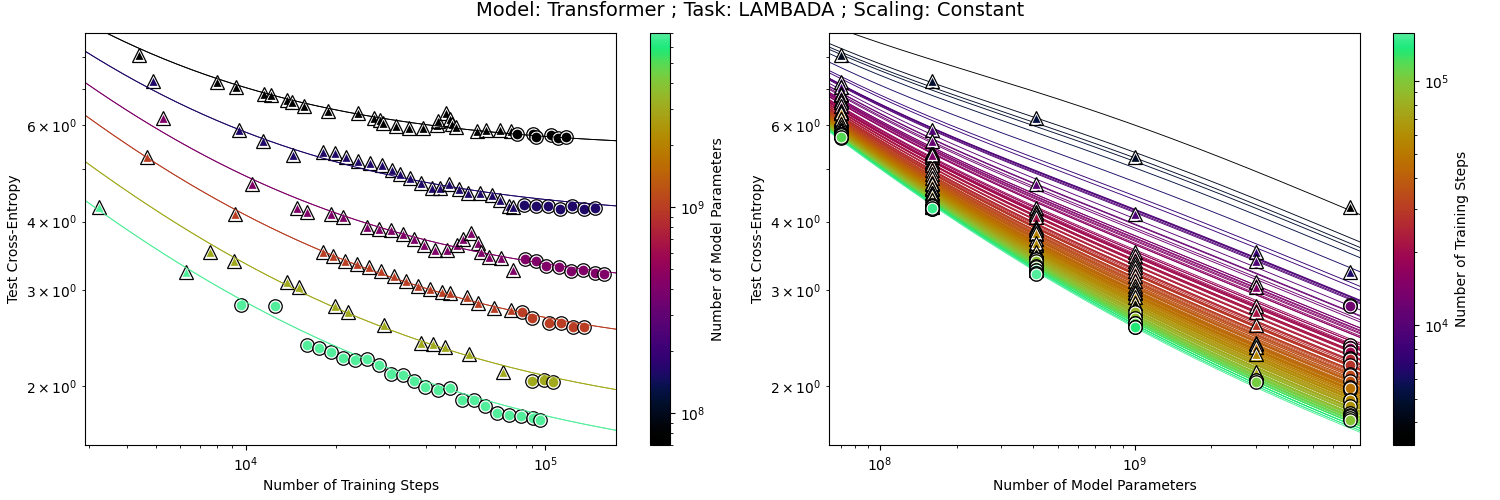}
\includegraphics[width=1.0\textwidth]{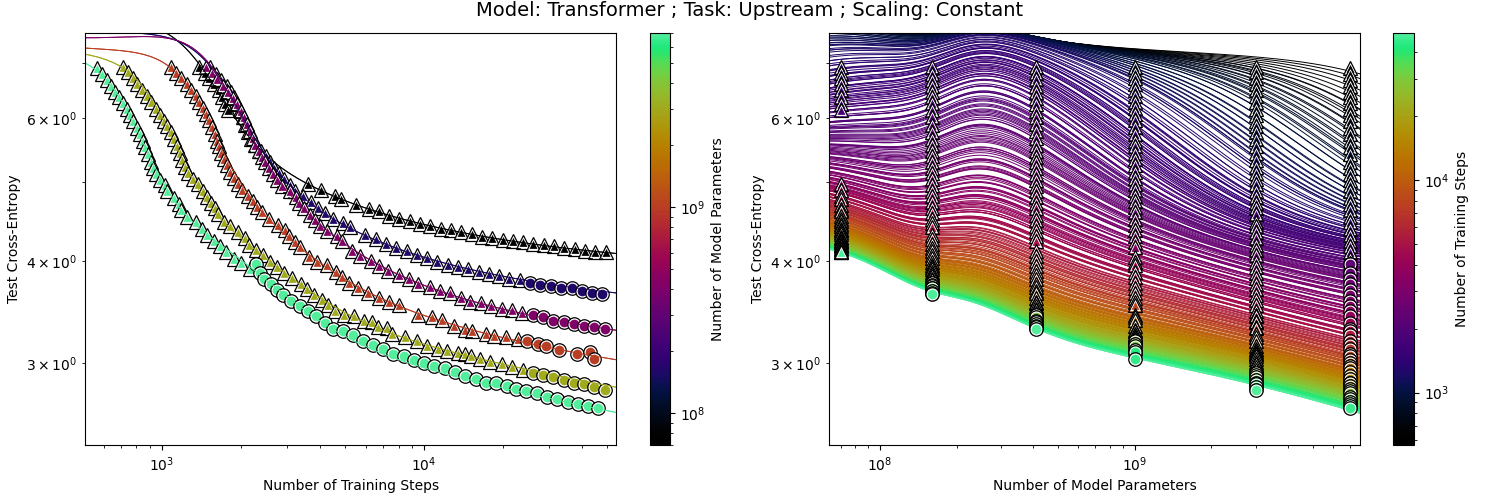}
\includegraphics[width=1.0\textwidth]{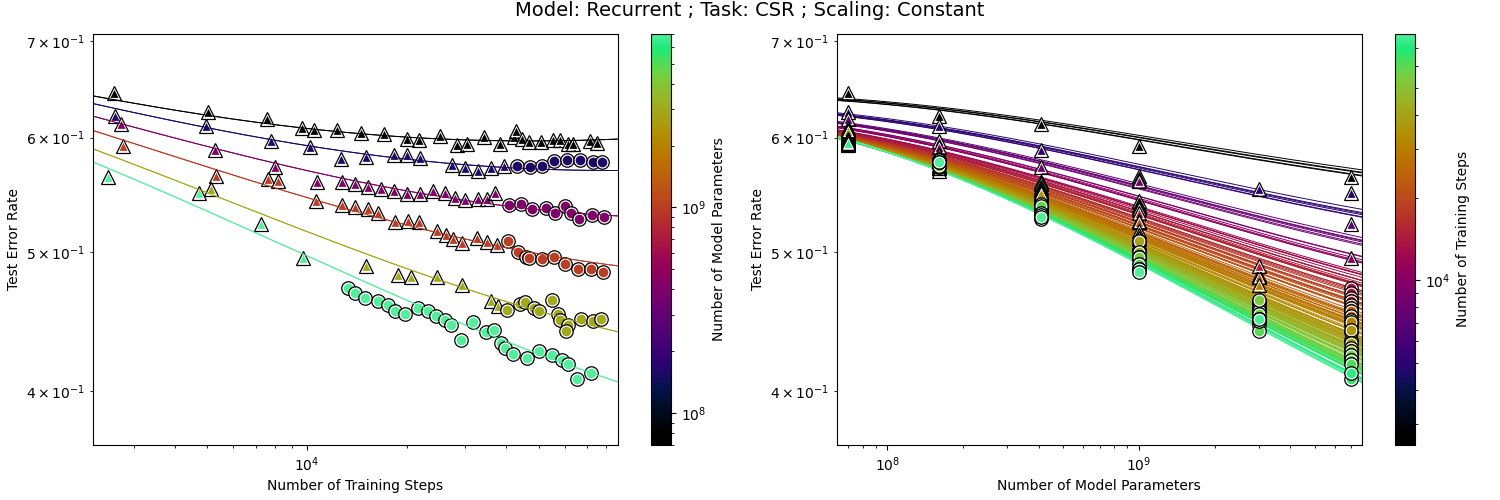}
\includegraphics[width=1.0\textwidth]{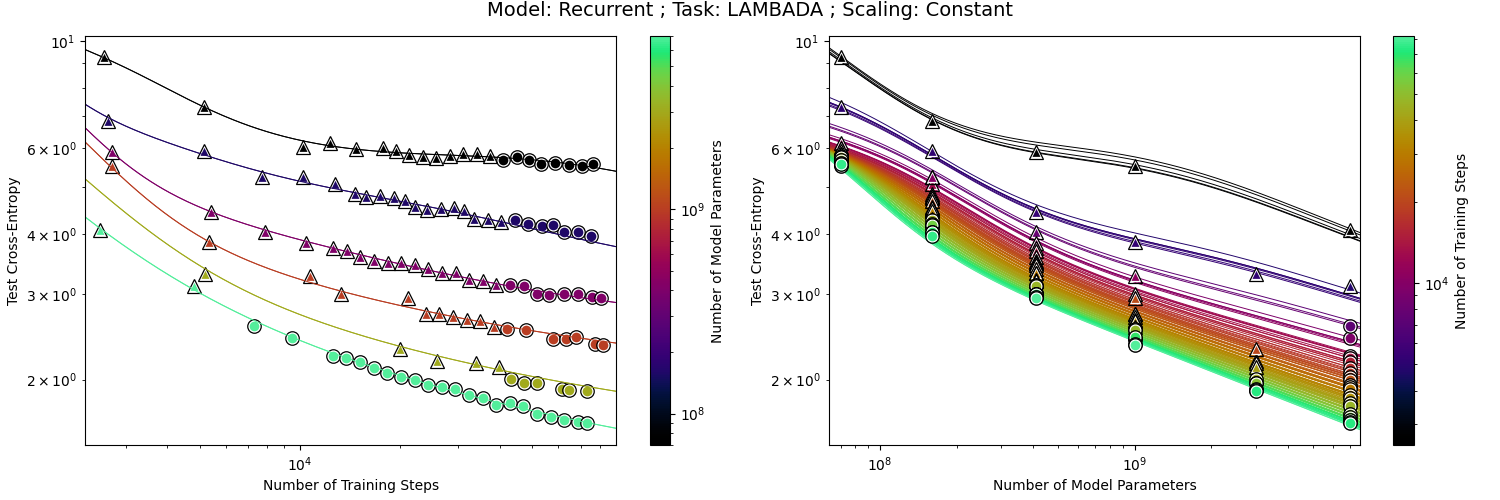}
\includegraphics[width=1.0\textwidth]{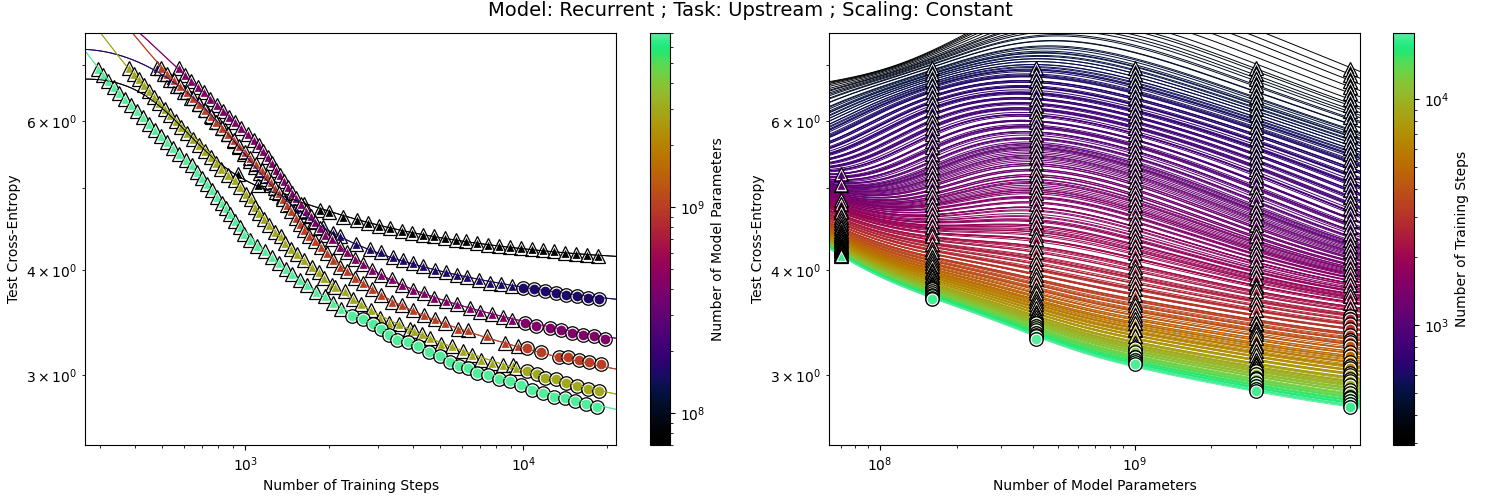}
\end{minipage}
\begin{minipage}{0.189\textwidth}
\end{minipage}
\begin{minipage}{0.19\textwidth}
\includegraphics[width=1.0\textwidth]{figures/legend/legend.png}
\end{minipage}
    \caption{
    Extrapolation Results of UNSL on bivariate scaling behavior of downstream (and upstream) language performance. See Section \ref{section:language} for more details.
    }
    \label{fig:unsl_llm_bivariate}
\end{figure*}

\FloatBarrier
\begin{figure*}[h]    \centering
\begin{minipage}{0.535\textwidth}
\includegraphics[width=1.0\textwidth]{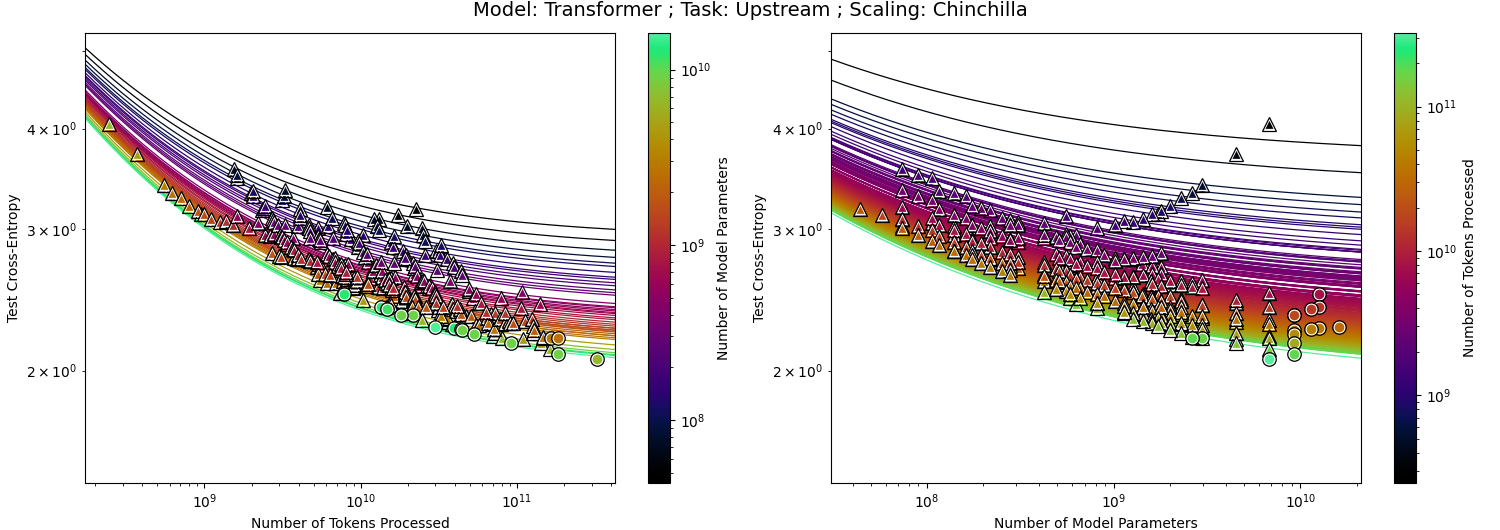}
\includegraphics[width=1.0\textwidth]{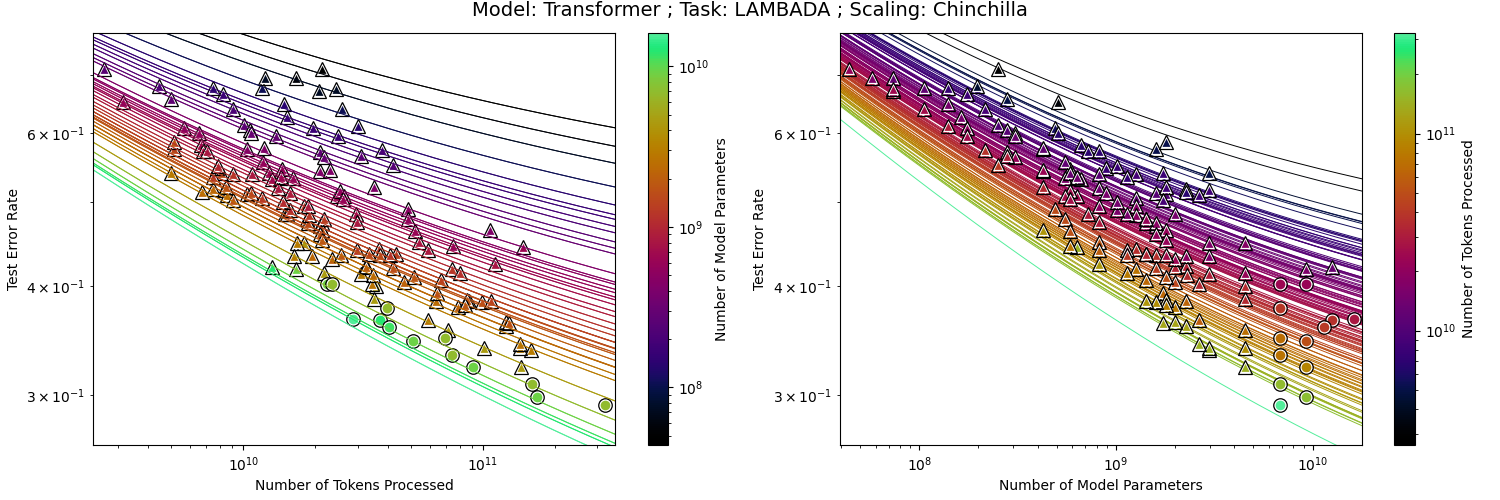}
\includegraphics[width=1.0\textwidth]{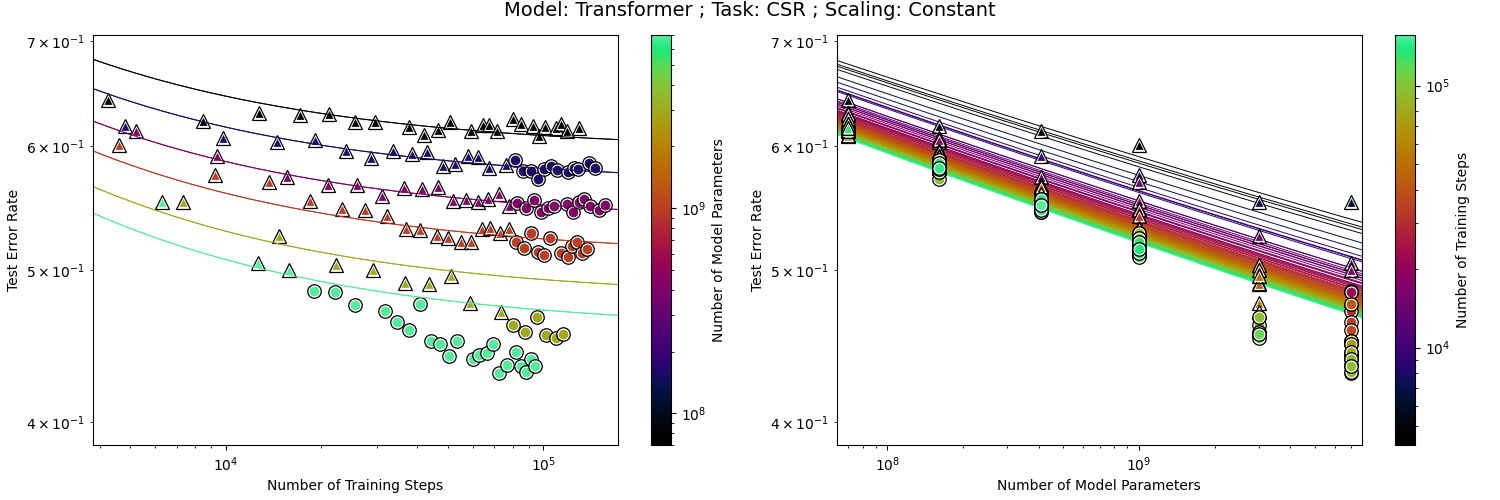}
\includegraphics[width=1.0\textwidth]{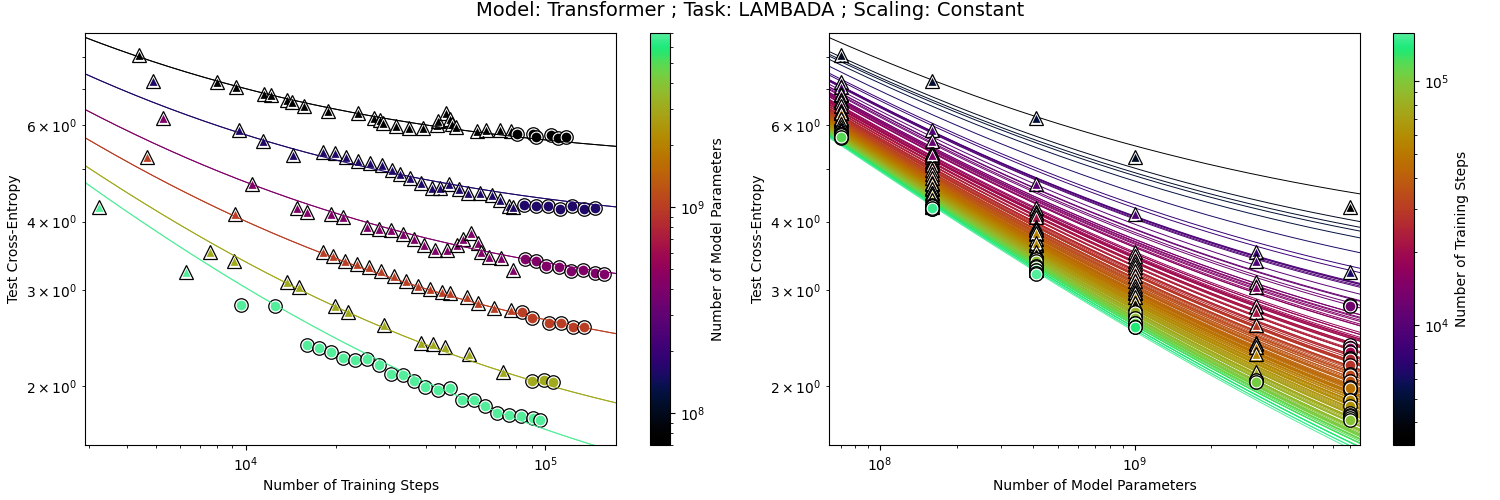}
\includegraphics[width=1.0\textwidth]{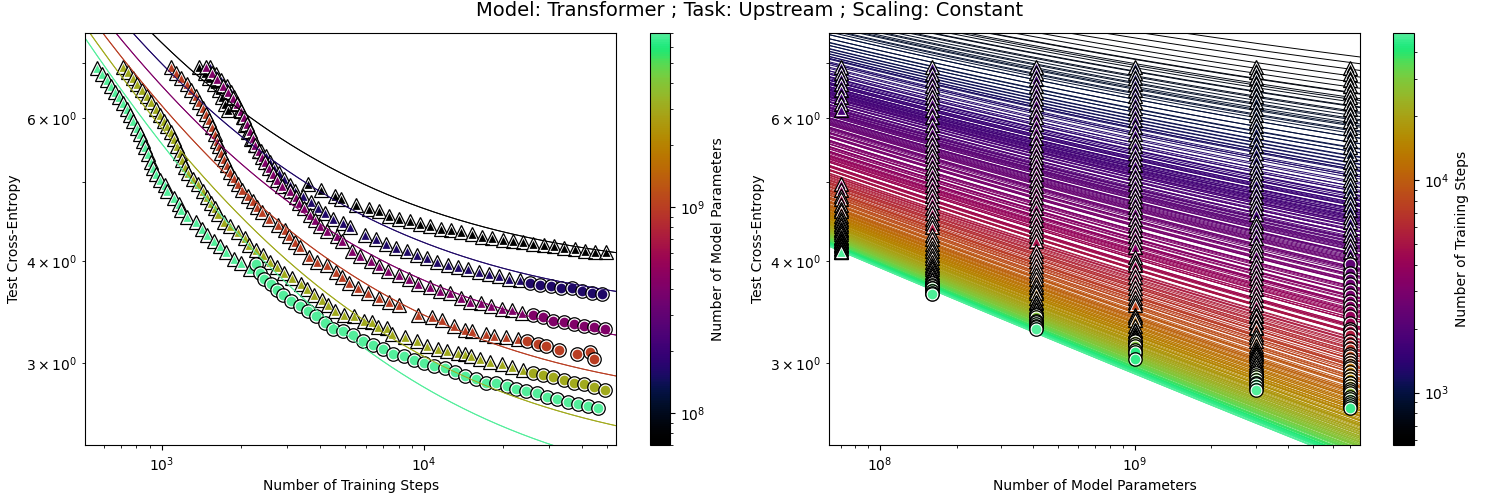}
\includegraphics[width=1.0\textwidth]{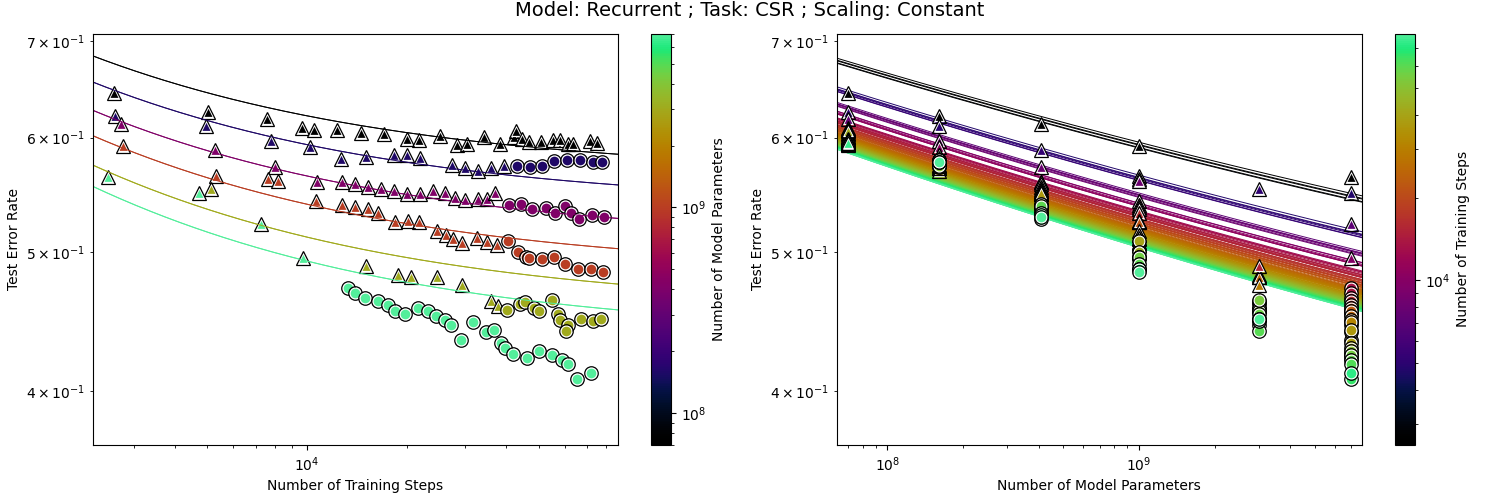}
\includegraphics[width=1.0\textwidth]{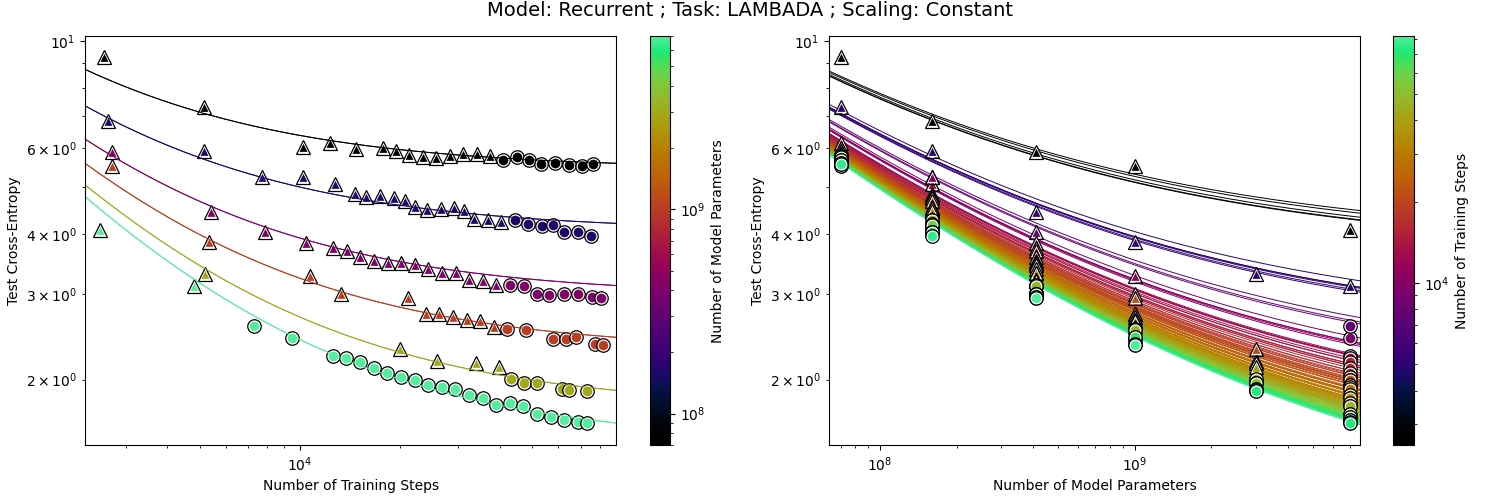}
\includegraphics[width=1.0\textwidth]{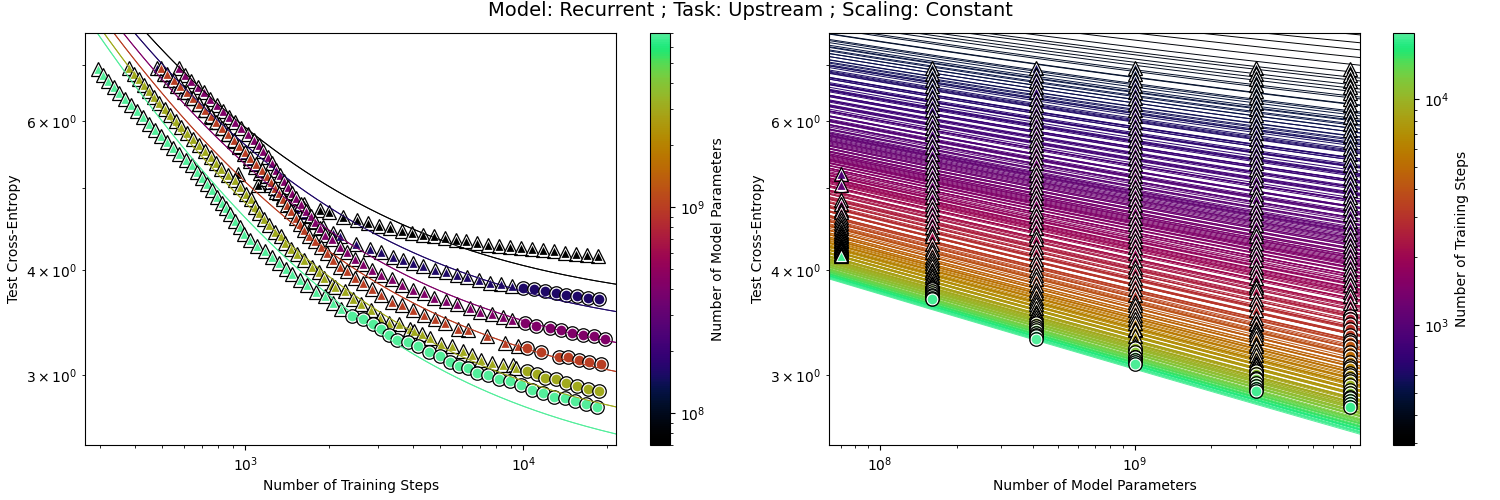}
\end{minipage}
\begin{minipage}{0.169\textwidth}
\end{minipage}
\begin{minipage}{0.19\textwidth}
\includegraphics[width=1.0\textwidth]{figures/legend/legend.png}
\end{minipage}
    \caption{
    Extrapolation Results of ``CF'' functional form of \cite{hoffmann2022training} on bivariate scaling behavior of downstream (and upstream) language performance. See Section \ref{section:language} for more details.
    }
    \label{fig:cf_llm_bivariate}
\end{figure*}

\FloatBarrier
\begin{figure*}[h]    \centering
\begin{minipage}{0.535\textwidth}
\includegraphics[width=1.0\textwidth]{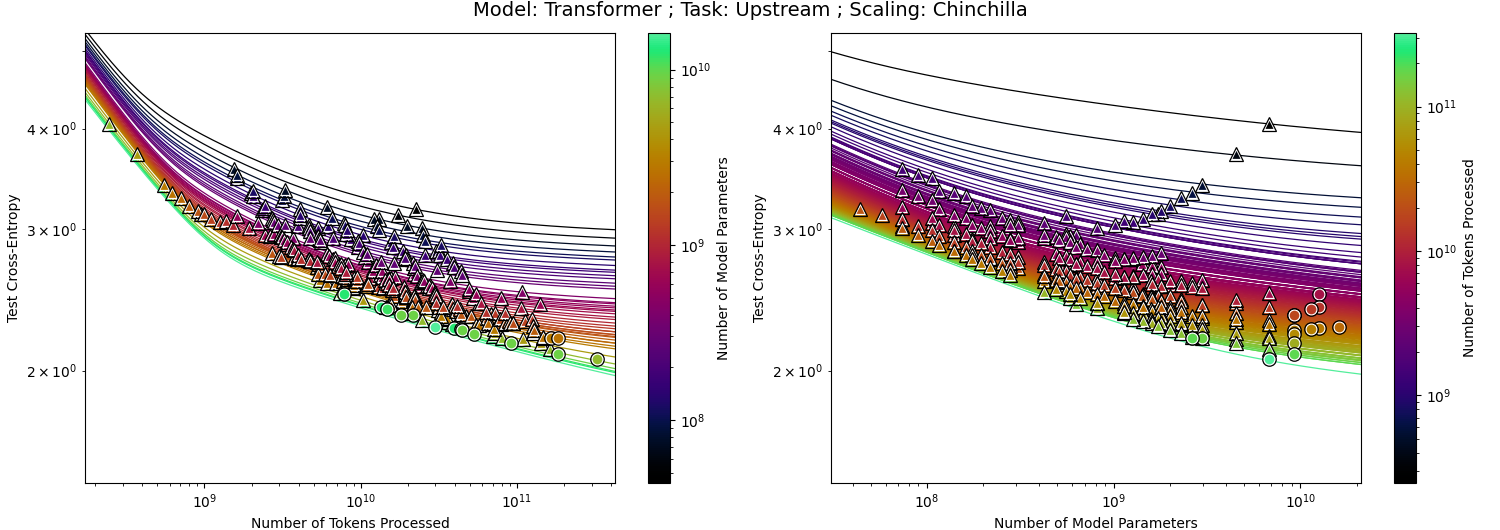}
\includegraphics[width=1.0\textwidth]{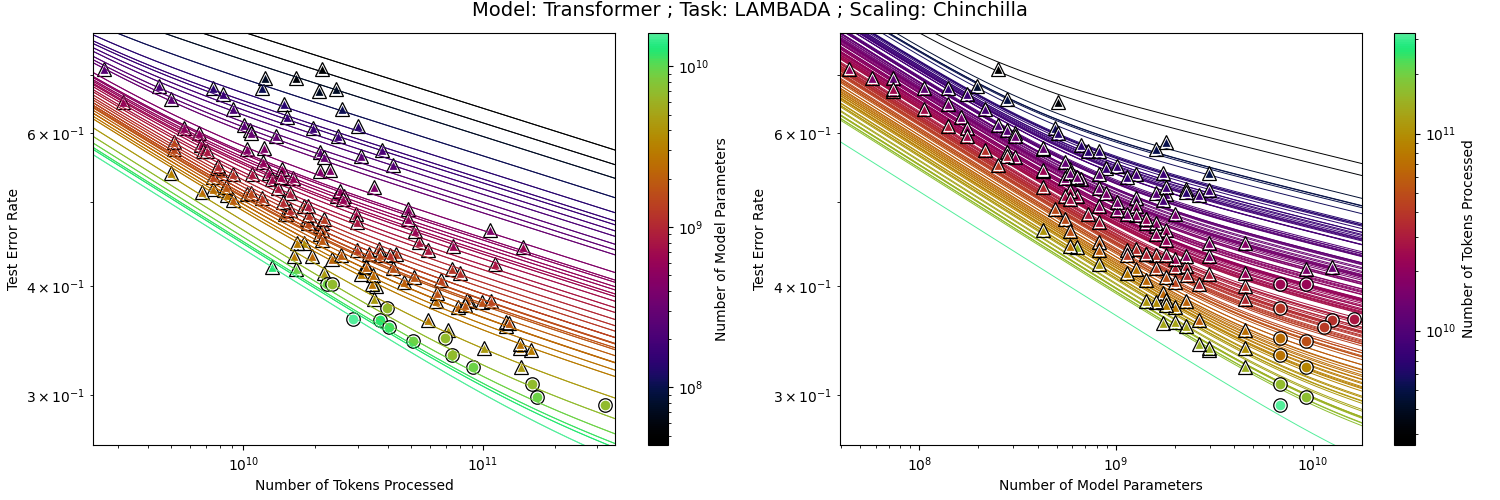}
\includegraphics[width=1.0\textwidth]{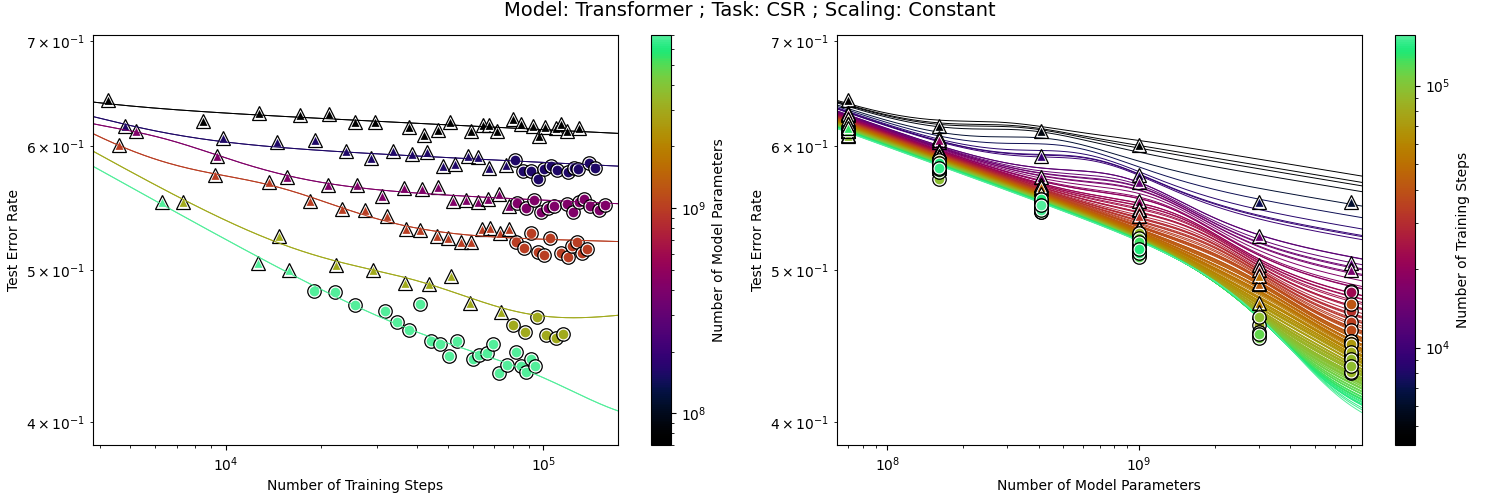}
\includegraphics[width=1.0\textwidth]{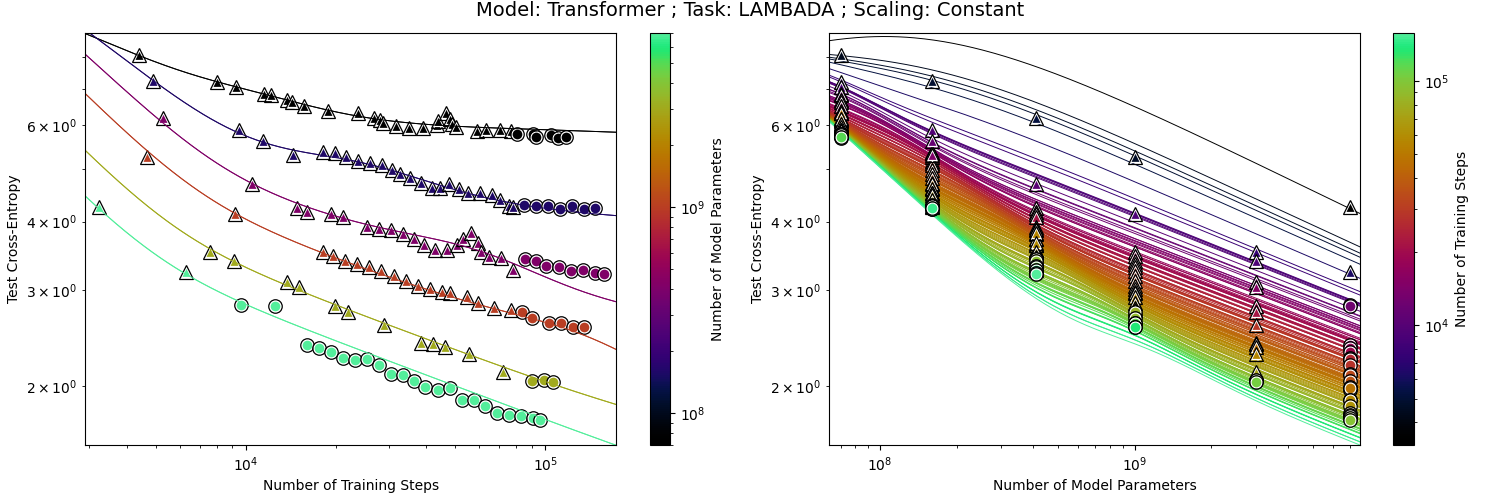}
\includegraphics[width=1.0\textwidth]{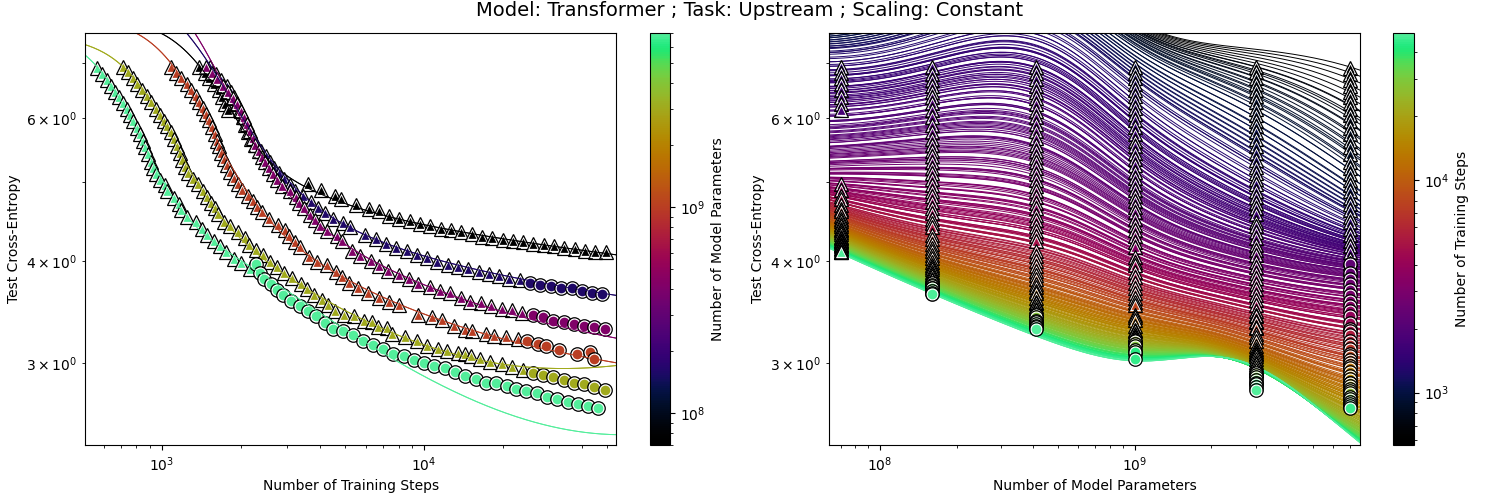}
\includegraphics[width=1.0\textwidth]{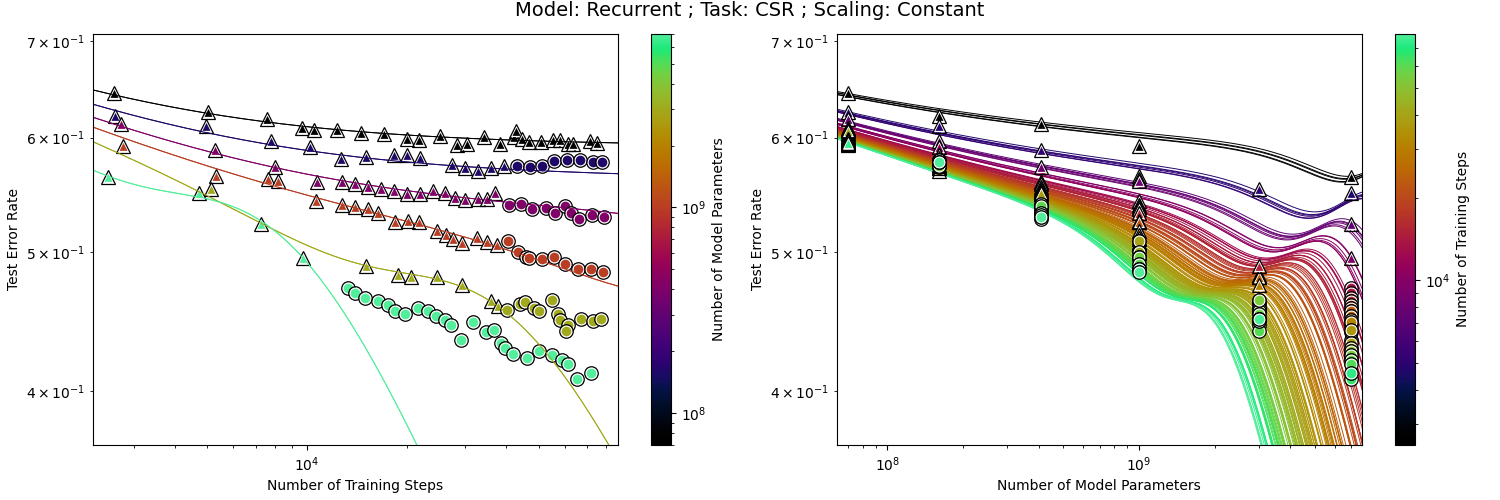}
\includegraphics[width=1.0\textwidth]{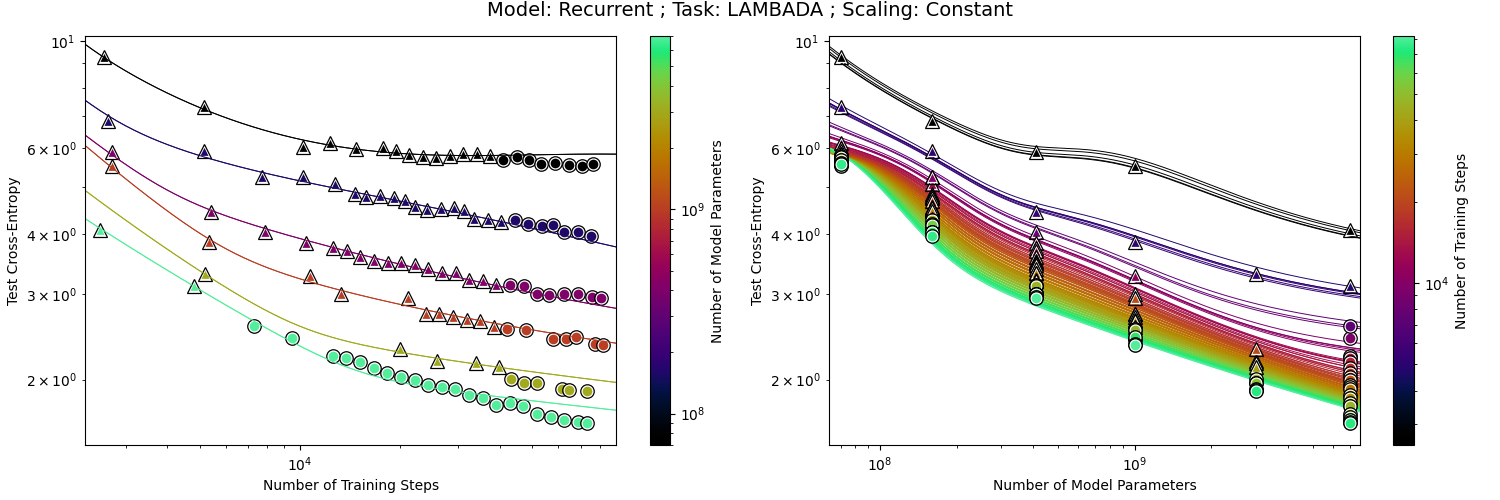}
\includegraphics[width=1.0\textwidth]{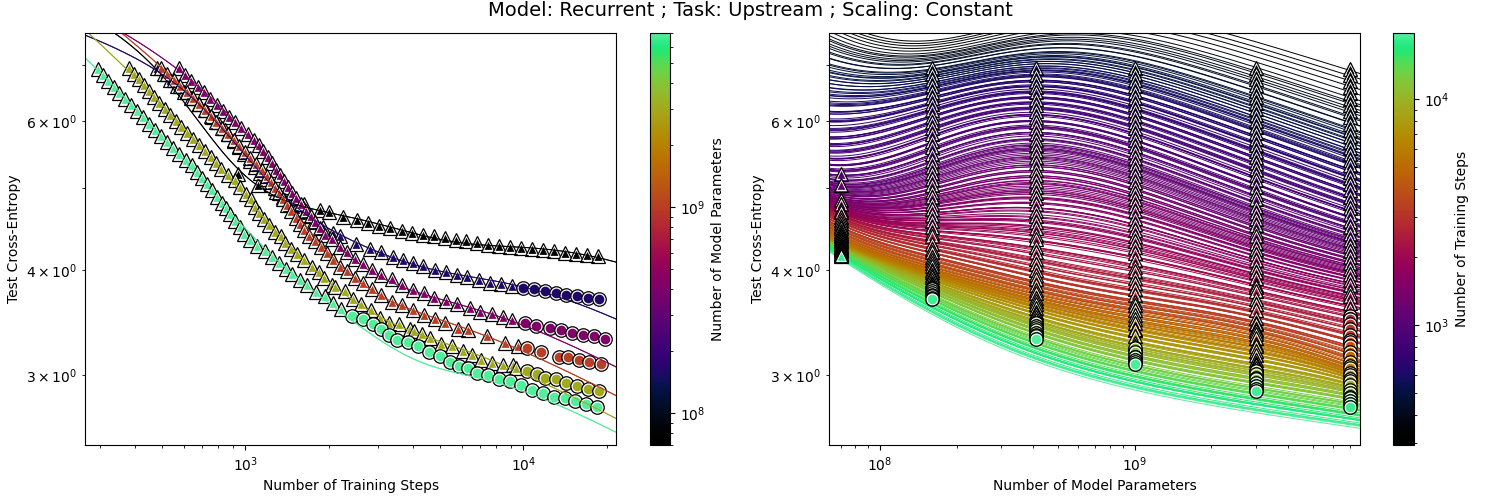}
\end{minipage}
\begin{minipage}{0.169\textwidth}
\end{minipage}
\begin{minipage}{0.19\textwidth}
\includegraphics[width=1.0\textwidth]{figures/legend/legend.png}
\end{minipage}
    \caption{
    Extrapolation Results of A1 functional form on bivariate scaling behavior of downstream (and upstream) language performance. See Section \ref{section:language} for more details.
    }
    \label{fig:a1_llm_bivariate}
\end{figure*}

\FloatBarrier
\begin{figure*}[h]    \centering
\begin{minipage}{0.535\textwidth}
\includegraphics[width=1.0\textwidth]{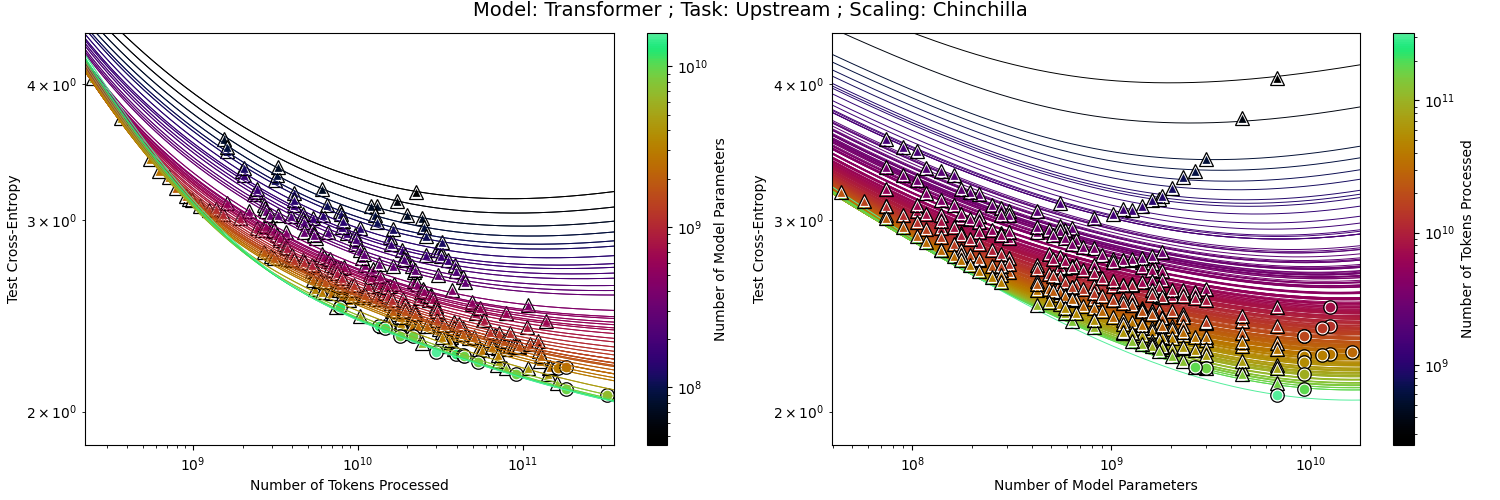}
\includegraphics[width=1.0\textwidth]{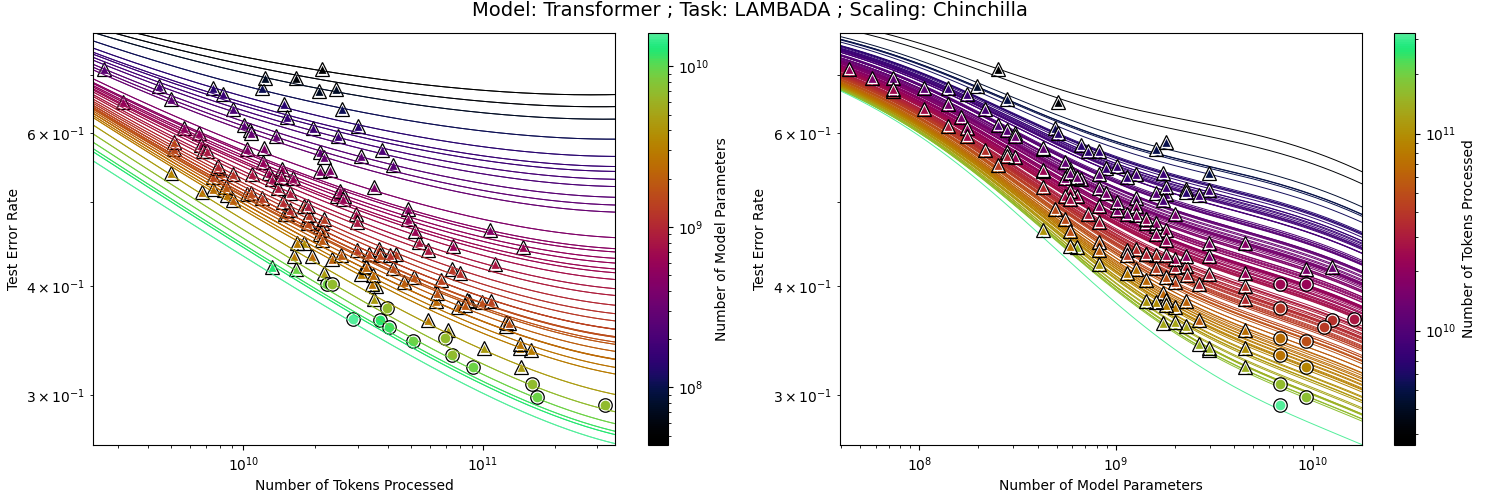}
\includegraphics[width=1.0\textwidth]{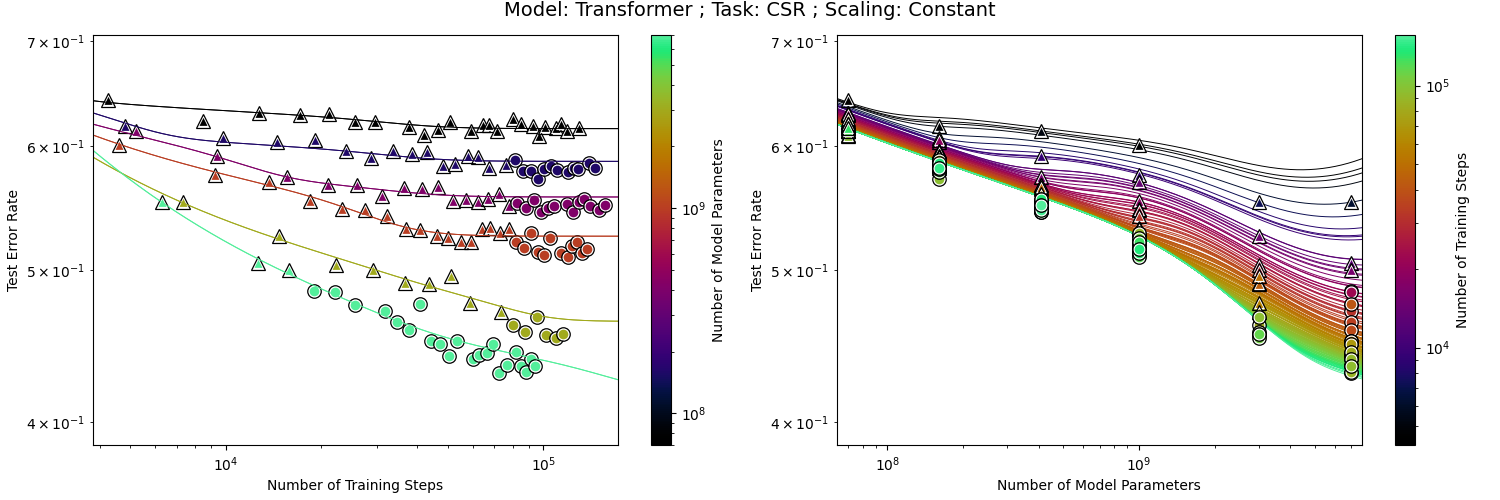}
\includegraphics[width=1.0\textwidth]{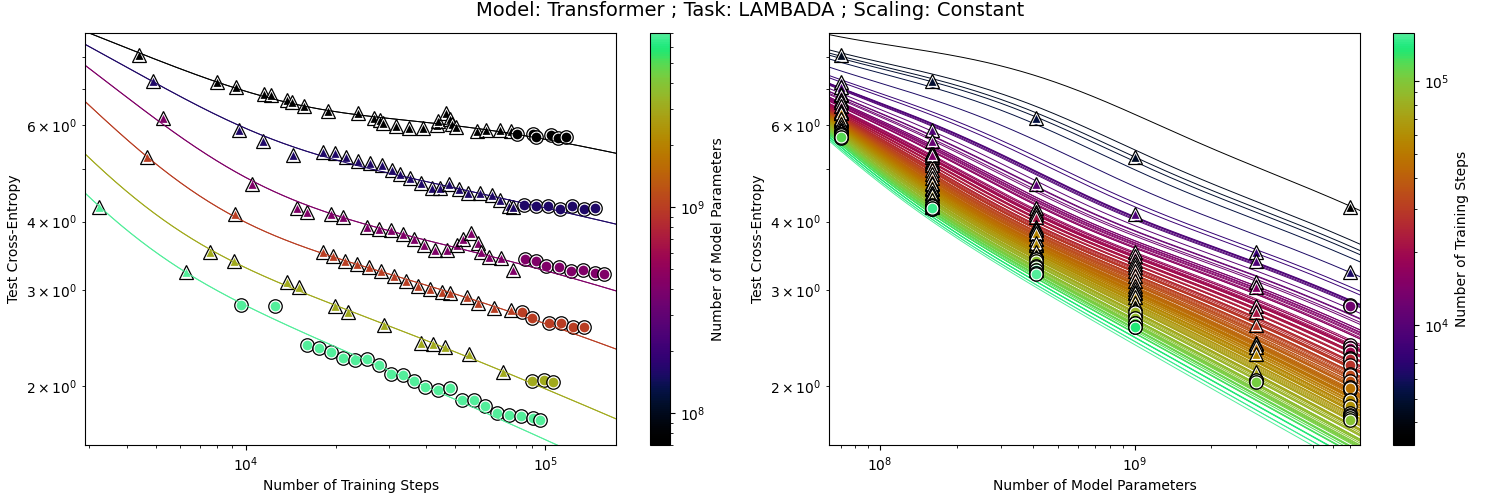}
\includegraphics[width=1.0\textwidth]{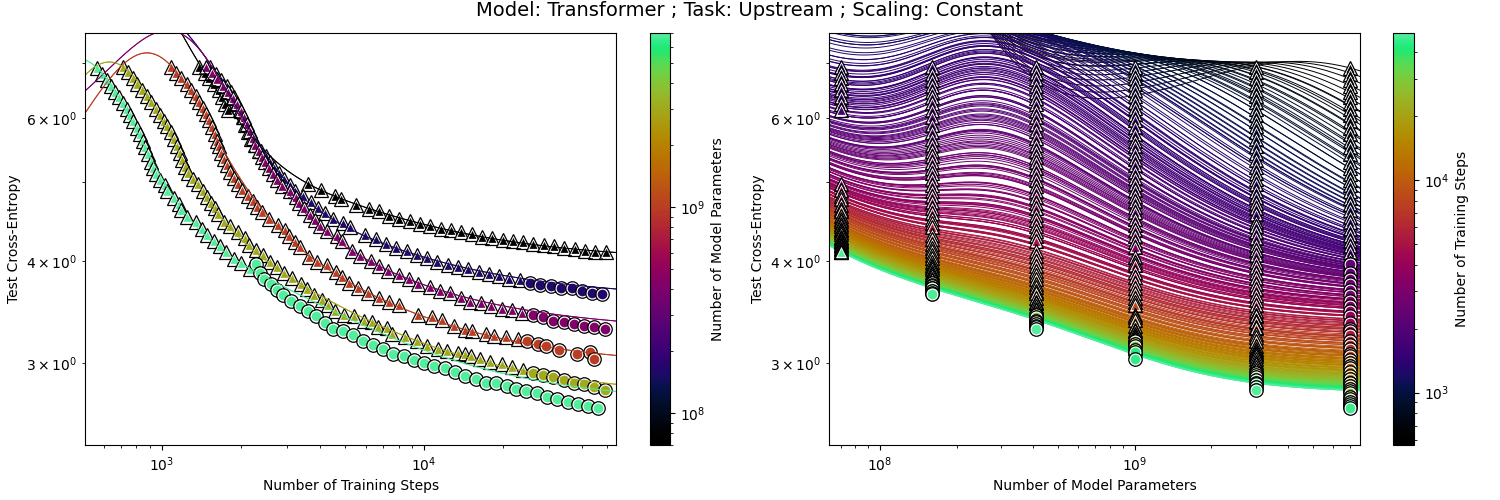}
\includegraphics[width=1.0\textwidth]{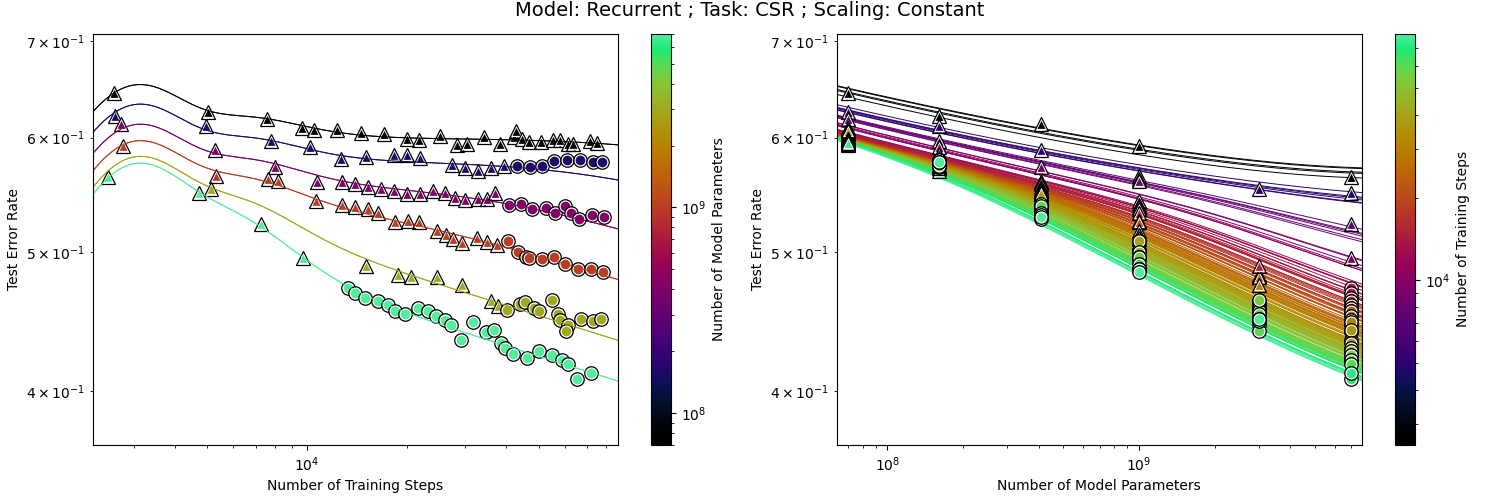}
\includegraphics[width=1.0\textwidth]{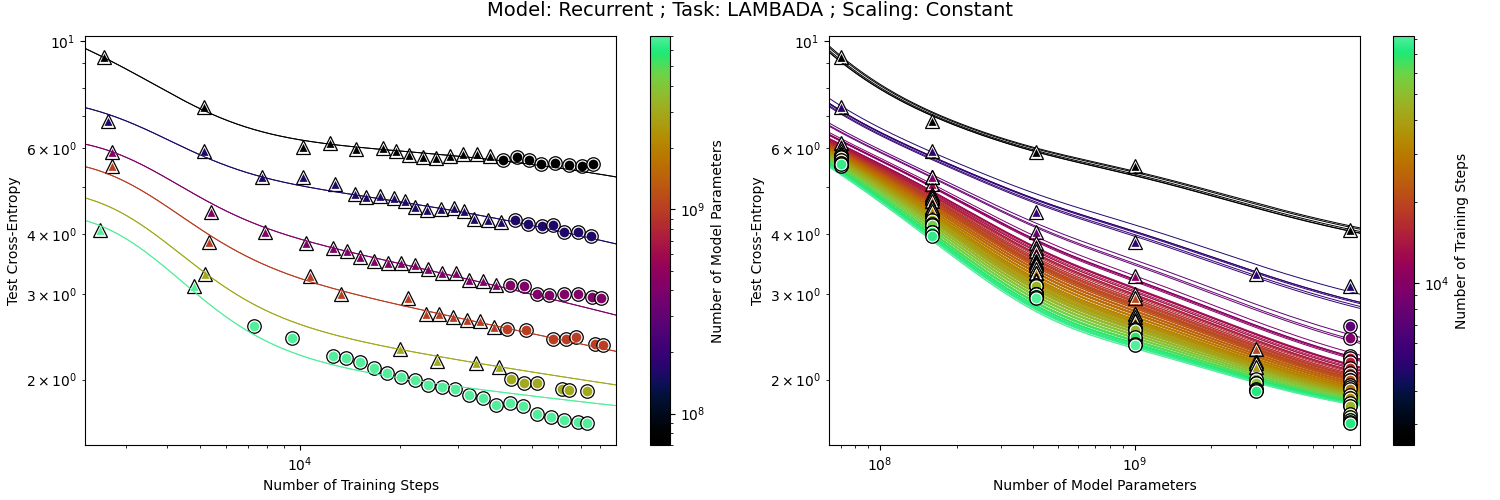}
\includegraphics[width=1.0\textwidth]{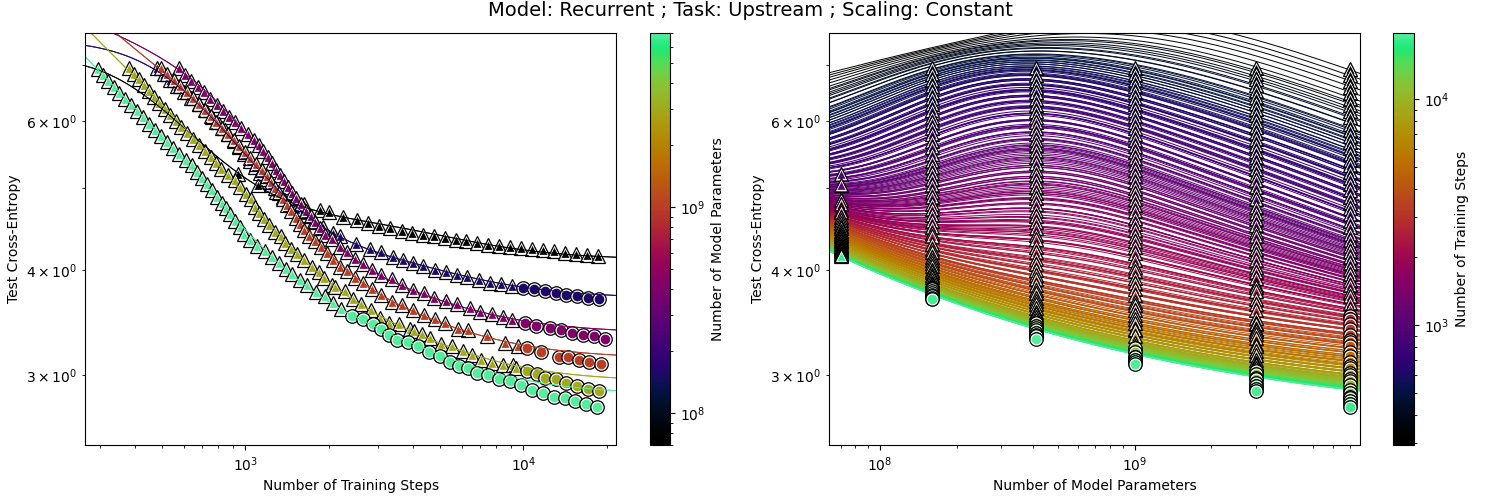}
\end{minipage}
\begin{minipage}{0.169\textwidth}
\end{minipage}
\begin{minipage}{0.19\textwidth}
\includegraphics[width=1.0\textwidth]{figures/legend/legend.png}
\end{minipage}
    \caption{
    Extrapolation Results of A2 functional form on bivariate scaling behavior of downstream (and upstream) language performance. See Section \ref{section:language} for more details.
    }
    \label{fig:a2_llm_bivariate}
\end{figure*}

\FloatBarrier



\newpage

\section{Additional Related Work}
\label{sec:additional_related_work}

There has been additional work on scaling law settings, interventions, and extensions besides those emphasized in this paper. Scaling laws have been applied in the context of autoregressive generative modeling along various axes of scale, e.g., compute/model/dataset \citep{henighan2020scalinglaws}, as well as in the regime of transfer learning/fine-tuning, i.e., the joint effect of the scale of the pre-training task and the quantity of data available for downstream fine-tuning \citep{hernandez2021scalinglawstransfer}. In addition, the influence of dataset curation/selection on scaling laws is receiving increasing attention, e.g., via pruning or data selection with the goal of optimizing scaling exponents \citep{sorscher2022beyond,ayed2023towards}. Furthermore, the problem of compute-optimal scaling, i.e., the design of scaling laws beyond the number of parameters, e.g., with respect to depth, width, and shape, has been addressed \citep{alabdulmohsin2023scalinglawscomputeoptimal}. In parallel with the problem of predictive  scaling-law fitting, a different line of work has focused on the design of explicit architectural scaling heuristics, e.g., compound scaling \citep{tan2019efficientnet}. Other learning contexts and classes of models where scaling analyses have also been pursued include reinforcement learning \citep{neumann2023scalinglawsrl} and diffusion (-transformer) generative models \citep{liang2024scalinglawsdiffusiontransformers, li2024scalabilitydiffusiontext2image}.

A related body of work has focused on the study of non-monotonic generalization properties and multi-regime phenomena with respect to scaling. Much of this work has focused on double descent phenomena that depend on model capacity, sample size, and/or training time, as seen in \citet{belkin2019reconciling, nakkiran2019deepdoubledescent, hastie2019surprises, adlam2020finegrainedbiasvariance}. Other related statistical physics viewpoints on non-monotonicity and sharp transitions in terms of phase transitions/jamming are presented in \citet{spigler2018jamming}. Another noteworthy example of non-monotonic behavior depending on optimization and regularization is ``Grokking,'' or sudden late generalization far after the start of memorization, as seen in \citet{power2022grokking, liu2022towardsgrokking, liu2022omnigrok}.

The application of scaling laws in forecasting is closely related to the problem of learning curve extrapolation, multi-fidelity hyperparameter optimization, and the scaling of hyperparameters. Because scaling laws are used operationally for forecasting on the basis of partial training runs, this problem is related to learning curve extrapolation and early termination of hyperparameter search, as discussed in \citep{domhan2015learningcurveextrapolation, swersky2014freezethaw}. It is also related to multi-fidelity/bandit and BO-based methods for HPO, as discussed in \citep{li2017hyperband, falkner2018bohb, snoek2012practicalbo, hutter2011smac}. In the area of large batch training, there are studies on scaling rules for schedules and optimizers, which analyze how optimal hyperparameters (such as the learning rate, batch size, and momentum) change with scale and batch size, as discussed in \citep{goyal2017largemini, smith2017dontdecay, you2017largebatch, mccandlish2018empiricallargebatch}. Another line of work on theory-driven parameterization of schedules aims at enabling zero-shot transfer of hyperparameters across network widths (such as in $\mu$P) as discussed in \citep{yang2021tensorprogramsv}.

Finally, the study of ``emergent abilities'' points out that the observation of clear transitions with respect to scale may actually indicate either real regime changes or the effects of nonlinear/discontinuous metrics and estimation at finite samples \citep{wei2022emergentabilities,schaeffer2023mirage}.


\section{Implementation of UNSL}
\label{sec:implementation}

We recommend using a system with at least a dozen CPUs if one wants the code to find global optima quickly.

Some notes about the code:
\begin{itemize}[leftmargin=2em, itemsep=1.5mm, parsep=0pt]
\item The training dataset always is passed in as one of the inputs to the functional form (via batch\_train argument) (e.g. even when evaluating on test dataset). This is because the inputs and outputs of functional form are z-normalized using the mean and std of the training dataset in log space. This is analogous to how batch normalization requires always using the mean and std of the training dataset (even when evaluating on test dataset). This z-normalization causes the functional form to converge to the global optimum faster and using less seeds. This z-normalization has no effect on the expressivity of the functional form.
\item The parallel ``executor.submit'' part will get stuck (and stop making progress) if any jax arrays are outside of optimize\_model function.
\item If you get an error related to parallelization, decrease ``n\_s'' variable or use a system with more CPUs.
\end{itemize}

To run this code: 
\begin{enumerate}[leftmargin=2em, itemsep=1.5mm, parsep=0pt]

\item Copy this code and replace each ␣ (that appears when pasting) with a space.

\item Run this sequence of ipython cells in order.

\end{enumerate}

\begin{ipycell}
#@title Install dependencies
␣
!pip uninstall kfac-jax -y
!pip uninstall jax -y
!pip uninstall flax -y
␣
!pip install git+https://github.com/google-deepmind/kfac-jax@85535ff524933cef2655528ad3e6931973000a19
!pip install git+https://github.com/jax-ml/jax@94233144f5469af28c065aa4263a6849338eeaa1
!pip install git+https://github.com/google/flax@405f6566014a9ebccb1b8f025f0d225bc8921beb
␣
!pip uninstall loky -y
!pip install loky==3.5.5
\end{ipycell}

\begin{ipycell}
#@title Import libraries
␣
import numpy as np
␣
import matplotlib
␣
import flax.linen as nn
import jax
import jax.numpy as jnp
from jax import random
␣
import kfac_jax
␣
import time
␣
import os
␣
import copy
␣
from loky import get_reusable_executor
␣
print(os.cpu_count())
print(len(list(os.sched_getaffinity(0))))
print(os.sched_getaffinity(0))
␣
#%matplotlib inline
␣
jax.config.update('jax_platform_name', 'cpu')
\end{ipycell}

\begin{ipycell}
#@title Select Scaling Behavior Data and Functional Form
␣
scaling_behavior = "downstream_imagenet__MiX_B/16"
#scaling_behavior = "llm_trivariate__cross_entropy__182"
␣
func_form = "unsl"
#func_form = "chinchilla"
\end{ipycell}

\begin{ipycell}
#@title Load Scaling Behavior Data
␣
␣
if scaling_behavior == "debug":
␣␣␣␣pass
␣
elif scaling_behavior == "llm_trivariate__cross_entropy__182":
␣
␣␣␣␣#region
␣␣␣␣x1 = np.array([4000000000.0, 4000000000.0, 4000000000.0, 9000000000.0, 11000000000.0, \
␣␣␣␣␣␣␣␣14000000000.0, 18000000000.0, 28000000000.0, 55000000000.0, 12000000000.0, 12000000000.0, \
␣␣␣␣␣␣␣␣12000000000.0, 12000000000.0, 12000000000.0, 17000000000.0, 21000000000.0, 28000000000.0, \
␣␣␣␣␣␣␣␣42000000000.0, 84000000000.0, 13000000000.0, 25000000000.0, 35000000000.0, 44000000000.0, \
␣␣␣␣␣␣␣␣58000000000.0, 88000000000.0, 178000000000.0, 100000000.0, 100000000.0, 100000000.0, \
␣␣␣␣␣␣␣␣100000000.0, 400000000.0, 400000000.0, 100000000.0, 100000000.0, 100000000.0, 100000000.0, \
␣␣␣␣␣␣␣␣100000000.0, 100000000.0, 100000000.0, 100000000.0, 100000000.0, 100000000.0, 100000000.0, \
␣␣␣␣␣␣␣␣100000000.0, 100000000.0, 100000000.0, 100000000.0, 100000000.0, 100000000.0, 100000000.0, \
␣␣␣␣␣␣␣␣100000000.0, 100000000.0, 100000000.0, 100000000.0, 100000000.0, 100000000.0, 100000000.0, \
␣␣␣␣␣␣␣␣100000000.0, 100000000.0, 100000000.0, 1500000000.0, 1500000000.0, 1500000000.0, \
␣␣␣␣␣␣␣␣400000000.0, 100000000.0, 2700000000.0, 1500000000.0, 400000000.0, 100000000.0, \
␣␣␣␣␣␣␣␣100000000.0, 3900000000.0, 1500000000.0, 400000000.0, 100000000.0, 100000000.0, \
␣␣␣␣␣␣␣␣100000000.0, 100000000.0, 100000000.0, 5900000000.0, 1500000000.0, 400000000.0, \
␣␣␣␣␣␣␣␣100000000.0, 100000000.0, 100000000.0, 100000000.0, 100000000.0, 400000000.0, \
␣␣␣␣␣␣␣␣7500000000.0, 1500000000.0, 400000000.0, 100000000.0, 100000000.0, 100000000.0, \
␣␣␣␣␣␣␣␣100000000.0, 100000000.0, 14000000000.0, 1500000000.0, 400000000.0, 100000000.0, \
␣␣␣␣␣␣␣␣400000000.0, 400000000.0, 400000000.0, 100000000.0, 100000000.0, 100000000.0, 100000000.0, \
␣␣␣␣␣␣␣␣100000000.0, 100000000.0, 100000000.0, 100000000.0, 100000000.0, 100000000.0, \
␣␣␣␣␣␣␣␣20000000000.0, 1500000000.0, 400000000.0, 100000000.0, 100000000.0, 100000000.0, \
␣␣␣␣␣␣␣␣100000000.0, 100000000.0, 100000000.0, 100000000.0, 100000000.0, 100000000.0, 100000000.0, \
␣␣␣␣␣␣␣␣100000000.0, 100000000.0, 100000000.0, 4800000000.0, 1500000000.0, 400000000.0, \
␣␣␣␣␣␣␣␣8800000000.0, 4800000000.0, 1500000000.0, 400000000.0, 12000000000.0, 4800000000.0, \
␣␣␣␣␣␣␣␣1500000000.0, 400000000.0, 22000000000.0, 4800000000.0, 1500000000.0, 400000000.0, \
␣␣␣␣␣␣␣␣400000000.0, 400000000.0, 32000000000.0, 4800000000.0, 1500000000.0, 400000000.0, \
␣␣␣␣␣␣␣␣100000000.0, 400000000.0, 400000000.0, 1500000000.0, 1500000000.0, 60000000000.0, \
␣␣␣␣␣␣␣␣4800000000.0, 1500000000.0, 400000000.0, 400000000.0, 400000000.0, 400000000.0, \
␣␣␣␣␣␣␣␣26000000000.0, 12000000000.0, 36000000000.0, 12000000000.0, 46000000000.0, 12000000000.0, \
␣␣␣␣␣␣␣␣1500000000.0, 66000000000.0, 12000000000.0, 91000000000.0, 12000000000.0, 174000000000.0, \
␣␣␣␣␣␣␣␣12000000000.0, 1500000000.0, 400000000.0, 1500000000.0, 1500000000.0, 1500000000.0, \
␣␣␣␣␣␣␣␣1500000000.0, 1500000000.0, 1500000000.0])
␣␣␣␣x2 = np.array([2810000000.0, 2810000000.0, 2810000000.0, 2810000000.0, 2810000000.0, \
␣␣␣␣␣␣␣␣2810000000.0, 2810000000.0, 2810000000.0, 2810000000.0, 4246500000.0, 4246500000.0, \
␣␣␣␣␣␣␣␣4246500000.0, 4246500000.0, 4246500000.0, 4246500000.0, 4246500000.0, 4246500000.0, \
␣␣␣␣␣␣␣␣4246500000.0, 4246500000.0, 8670000000.0, 8670000000.0, 8670000000.0, 8670000000.0, \
␣␣␣␣␣␣␣␣8670000000.0, 8670000000.0, 8670000000.0, 7098752.0, 7098752.0, 1096300000.0, 14100000.0, \
␣␣␣␣␣␣␣␣19703712.0, 35500000.0, 35500000.0, 35500000.0, 35500000.0, 14100000.0, 14100000.0, \
␣␣␣␣␣␣␣␣14100000.0, 14100000.0, 14100000.0, 14100000.0, 14100000.0, 14100000.0, 14100000.0, \
␣␣␣␣␣␣␣␣14100000.0, 14100000.0, 14100000.0, 14100000.0, 44000000.0, 44000000.0, 44000000.0, \
␣␣␣␣␣␣␣␣44000000.0, 44000000.0, 44000000.0, 44000000.0, 44000000.0, 44000000.0, 44000000.0, \
␣␣␣␣␣␣␣␣44000000.0, 44000000.0, 82700000.0, 201236224.0, 1096300000.0, 1096300000.0, 1096300000.0, \
␣␣␣␣␣␣␣␣618700000.0, 618700000.0, 618700000.0, 618700000.0, 618700000.0, 421200000.0, 421200000.0, \
␣␣␣␣␣␣␣␣421200000.0, 421200000.0, 421200000.0, 421200000.0, 421200000.0, 421200000.0, 281000000.0, \
␣␣␣␣␣␣␣␣281000000.0, 281000000.0, 281000000.0, 281000000.0, 281000000.0, 281000000.0, 281000000.0, \
␣␣␣␣␣␣␣␣281000000.0, 220500000.0, 220500000.0, 220500000.0, 220500000.0, 220500000.0, 220500000.0, \
␣␣␣␣␣␣␣␣220500000.0, 220500000.0, 146500000.0, 146500000.0, 146500000.0, 146500000.0, 146500000.0, \
␣␣␣␣␣␣␣␣146500000.0, 146500000.0, 146500000.0, 146500000.0, 146500000.0, 146500000.0, 146500000.0, \
␣␣␣␣␣␣␣␣146500000.0, 146500000.0, 146500000.0, 146500000.0, 146500000.0, 82700000.0, 82700000.0, \
␣␣␣␣␣␣␣␣82700000.0, 82700000.0, 82700000.0, 82700000.0, 82700000.0, 82700000.0, 82700000.0, \
␣␣␣␣␣␣␣␣82700000.0, 82700000.0, 82700000.0, 82700000.0, 82700000.0, 82700000.0, 82700000.0, \
␣␣␣␣␣␣␣␣2810000000.0, 2810000000.0, 2810000000.0, 1517300000.0, 1517300000.0, 1517300000.0, \
␣␣␣␣␣␣␣␣1517300000.0, 1096300000.0, 1096300000.0, 1096300000.0, 1096300000.0, 618700000.0, \
␣␣␣␣␣␣␣␣618700000.0, 618700000.0, 618700000.0, 618700000.0, 618700000.0, 421200000.0, 421200000.0, \
␣␣␣␣␣␣␣␣421200000.0, 421200000.0, 421200000.0, 421200000.0, 421200000.0, 421200000.0, 421200000.0, \
␣␣␣␣␣␣␣␣220500000.0, 220500000.0, 220500000.0, 220500000.0, 220500000.0, 220500000.0, 220500000.0, \
␣␣␣␣␣␣␣␣3899710720.0, 3899710720.0, 2810000000.0, 2810000000.0, 2160013824.0, 2160013824.0, \
␣␣␣␣␣␣␣␣2160013824.0, 1517300000.0, 1517300000.0, 1096300000.0, 1096300000.0, 573700000.0, \
␣␣␣␣␣␣␣␣573700000.0, 573700000.0, 573700000.0, 220500000.0, 220500000.0, 573700000.0, 573700000.0, \
␣␣␣␣␣␣␣␣1096300000.0, 1517300000.0])
␣␣␣␣x3 = np.array([32000000000.0, 40000000000.0, 55000000000.0, 55000000000.0, 55000000000.0, \
␣␣␣␣␣␣␣␣55000000000.0, 55000000000.0, 55000000000.0, 55000000000.0, 36000000000.0, 48000000000.0, \
␣␣␣␣␣␣␣␣60000000000.0, 72000000000.0, 84000000000.0, 84000000000.0, 84000000000.0, 84000000000.0, \
␣␣␣␣␣␣␣␣84000000000.0, 84000000000.0, 178000000000.0, 178000000000.0, 178000000000.0, \
␣␣␣␣␣␣␣␣178000000000.0, 178000000000.0, 178000000000.0, 178000000000.0, 100000000.0, 200000000.0, \
␣␣␣␣␣␣␣␣100000000.0, 100000000.0, 400000000.0, 400000000.0, 200000000.0, 400000000.0, 800000000.0, \
␣␣␣␣␣␣␣␣200000000.0, 400000000.0, 800000000.0, 1500000000.0, 2700000000.0, 3900000000.0, \
␣␣␣␣␣␣␣␣5900000000.0, 14000000000.0, 20000000000.0, 91000000000.0, 174000000000.0, 600000000000.0, \
␣␣␣␣␣␣␣␣900000000000.0, 200000000.0, 400000000.0, 1500000000.0, 2700000000.0, 3900000000.0, \
␣␣␣␣␣␣␣␣5900000000.0, 7500000000.0, 14000000000.0, 20000000000.0, 32000000000.0, 91000000000.0, \
␣␣␣␣␣␣␣␣174000000000.0, 1500000000.0, 1500000000.0, 1500000000.0, 1500000000.0, 1500000000.0, \
␣␣␣␣␣␣␣␣2700000000.0, 2700000000.0, 2700000000.0, 2700000000.0, 1500000000.0, 3900000000.0, \
␣␣␣␣␣␣␣␣3900000000.0, 3900000000.0, 3900000000.0, 1500000000.0, 2700000000.0, 5900000000.0, \
␣␣␣␣␣␣␣␣7500000000.0, 5900000000.0, 5900000000.0, 5900000000.0, 5900000000.0, 1500000000.0, \
␣␣␣␣␣␣␣␣2700000000.0, 3900000000.0, 7500000000.0, 91000000000.0, 7500000000.0, 7500000000.0, \
␣␣␣␣␣␣␣␣7500000000.0, 7500000000.0, 1500000000.0, 2700000000.0, 3900000000.0, 5900000000.0, \
␣␣␣␣␣␣␣␣14000000000.0, 14000000000.0, 14000000000.0, 14000000000.0, 60000000000.0, 91000000000.0, \
␣␣␣␣␣␣␣␣174000000000.0, 1500000000.0, 2700000000.0, 3900000000.0, 5900000000.0, 7500000000.0, \
␣␣␣␣␣␣␣␣20000000000.0, 32000000000.0, 60000000000.0, 91000000000.0, 174000000000.0, 20000000000.0, \
␣␣␣␣␣␣␣␣20000000000.0, 20000000000.0, 20000000000.0, 400000000.0, 1500000000.0, 2700000000.0, \
␣␣␣␣␣␣␣␣3900000000.0, 5900000000.0, 7500000000.0, 14000000000.0, 32000000000.0, 60000000000.0, \
␣␣␣␣␣␣␣␣91000000000.0, 174000000000.0, 300000000000.0, 4800000000.0, 4800000000.0, 4800000000.0, \
␣␣␣␣␣␣␣␣8800000000.0, 8800000000.0, 8800000000.0, 8800000000.0, 12000000000.0, 12000000000.0, \
␣␣␣␣␣␣␣␣12000000000.0, 12000000000.0, 22000000000.0, 22000000000.0, 22000000000.0, 22000000000.0, \
␣␣␣␣␣␣␣␣60000000000.0, 91000000000.0, 32000000000.0, 32000000000.0, 32000000000.0, 32000000000.0, \
␣␣␣␣␣␣␣␣32000000000.0, 60000000000.0, 91000000000.0, 91000000000.0, 250000000000.0, 60000000000.0, \
␣␣␣␣␣␣␣␣60000000000.0, 60000000000.0, 60000000000.0, 3900000000.0, 32000000000.0, 300000000000.0, \
␣␣␣␣␣␣␣␣26000000000.0, 26000000000.0, 36000000000.0, 36000000000.0, 46000000000.0, 46000000000.0, \
␣␣␣␣␣␣␣␣46000000000.0, 66000000000.0, 66000000000.0, 91000000000.0, 91000000000.0, 174000000000.0, \
␣␣␣␣␣␣␣␣174000000000.0, 174000000000.0, 174000000000.0, 300000000000.0, 600000000000.0, \
␣␣␣␣␣␣␣␣300000000000.0, 600000000000.0, 174000000000.0, 91000000000.0])
␣␣␣␣y = np.array([2.722962, 2.706547, 2.696432, 2.611045, 2.598793, 2.589427, 2.584592, 2.579361, \
␣␣␣␣␣␣␣␣2.574117, 2.618343, 2.577519, 2.56039, 2.539524, 2.526961, 2.48506, 2.471375, 2.467156, \
␣␣␣␣␣␣␣␣2.470016, 2.45653, 2.451606, 2.376484, 2.360029, 2.351177, 2.348107, 2.337275, 2.336741, \
␣␣␣␣␣␣␣␣8.102005, 7.36236, 6.611002, 7.278144, 6.096268, 5.79413, 6.252892, 5.799587, 5.284175, \
␣␣␣␣␣␣␣␣6.664138, 6.273184, 5.816824, 5.364749, 4.992776, 4.741132, 4.546227, 4.383392, 4.51041, \
␣␣␣␣␣␣␣␣4.396074, 4.359691, 4.331469, 4.304018, 6.18997, 5.648868, 4.387672, 4.113597, 3.988205, \
␣␣␣␣␣␣␣␣3.89434, 3.854261, 3.807672, 3.92903, 3.89224, 3.8362, 3.812315, 4.36208, 3.929866, \
␣␣␣␣␣␣␣␣3.680017, 3.704141, 3.819042, 3.415532, 3.409955, 3.449187, 3.829336, 3.784308, 3.348126, \
␣␣␣␣␣␣␣␣3.35496, 3.404755, 3.89862, 3.797898, 3.747371, 4.169935, 4.376178, 3.319542, 3.330755, \
␣␣␣␣␣␣␣␣3.386692, 3.882211, 3.877228, 3.72759, 3.754965, 3.99341, 3.312059, 3.310159, 3.325092, \
␣␣␣␣␣␣␣␣3.374362, 3.848433, 3.932167, 3.752793, 3.721901, 3.78097, 3.319729, 3.334061, 3.390927, \
␣␣␣␣␣␣␣␣3.801428, 3.34025, 3.311519, 3.277013, 4.033501, 3.80764, 3.732758, 3.718068, 3.735044, \
␣␣␣␣␣␣␣␣3.756257, 3.793078, 3.846421, 3.862952, 3.897418, 3.608018, 3.618937, 3.651057, 3.790125, \
␣␣␣␣␣␣␣␣5.488318, 4.20696, 3.953894, 3.844048, 3.774708, 3.747734, 3.725231, 3.747091, 3.747091, \
␣␣␣␣␣␣␣␣3.743505, 3.734349, 3.868704, 3.236253, 3.244249, 3.378791, 3.022525, 3.022745, 3.070052, \
␣␣␣␣␣␣␣␣3.479142, 2.925644, 2.927054, 2.996886, 3.547616, 2.900172, 2.912009, 2.973545, 3.441785, \
␣␣␣␣␣␣␣␣3.688865, 3.756154, 2.912103, 2.926206, 2.980892, 3.366502, 5.783605, 3.431953, 3.465185, \
␣␣␣␣␣␣␣␣2.917592, 2.890927, 2.981407, 2.993185, 3.032319, 3.258704, 3.525297, 3.260515, 3.257426, \
␣␣␣␣␣␣␣␣2.916993, 2.947956, 2.618932, 2.636968, 2.603919, 2.638341, 3.029574, 2.592968, 2.626113, \
␣␣␣␣␣␣␣␣2.636805, 2.665932, 2.708514, 2.738098, 2.895393, 3.752652, 2.958765, 2.937761, 2.881076, \
␣␣␣␣␣␣␣␣2.863888, 2.974257, 3.034805])
␣␣␣␣#endregion
␣
␣␣␣␣x1_new = []
␣␣␣␣x2_new = []
␣␣␣␣x3_new = []
␣␣␣␣y_new = []
␣
␣␣␣␣for _x1, _x2, _x3, _y in zip(x1, x2, x3, y):
␣␣␣␣␣␣if True:
␣␣␣␣␣␣␣␣x1_new.append(_x1)
␣␣␣␣␣␣␣␣x2_new.append(_x2)
␣␣␣␣␣␣␣␣x3_new.append(_x3)
␣␣␣␣␣␣␣␣y_new.append(_y)
␣
␣␣␣␣x1 = np.array(x1_new)
␣␣␣␣x2 = np.array(x2_new)
␣␣␣␣x3 = np.array(x3_new)
␣␣␣␣y = np.array(y_new)
␣
␣␣␣␣trivariate = True
␣
␣␣␣␣overfit = True
␣
␣␣␣␣nonmon_hparams = True
␣
␣␣␣␣eval_metric_upper_limit = False
␣
␣␣␣␣x1_name = "Training Dataset Size"
␣␣␣␣x2_name = "Number of Model Parameters"
␣␣␣␣x3_name = "Number of Training Steps"
␣␣␣␣y_name = "Test Cross-Entropy"
␣
elif scaling_behavior == "downstream_imagenet__MiX_B/16":
␣
␣␣␣␣#region
␣␣␣␣x1 = [12000, 13000, 14000, 16000, 18000, 19000, 21000, 24000, 26000, 29000, 32000, 35000, \
␣␣␣␣␣␣␣␣39000, 43000, 47000, 52000, 57000, 63000, 70000, 77000, 85000, 94000, 104000, 114000, \
␣␣␣␣␣␣␣␣126000, 139000, 154000, 170000, 187000, 207000, 228000, 252000, 278000, 307000, 339000, \
␣␣␣␣␣␣␣␣374000, 12000, 13000, 14000, 16000, 18000, 19000, 21000, 24000, 26000, 29000, 32000, \
␣␣␣␣␣␣␣␣35000, 39000, 43000, 47000, 52000, 57000, 63000, 70000, 77000, 85000, 95000, 105000, \
␣␣␣␣␣␣␣␣115000, 127000, 140000, 155000, 171000, 188000, 208000, 229000, 254000, 280000, 309000, \
␣␣␣␣␣␣␣␣341000, 376000, 12000, 13000, 14000, 16000, 18000, 19000, 21000, 24000, 26000, 29000, \
␣␣␣␣␣␣␣␣32000, 35000, 39000, 43000, 47000, 52000, 57000, 63000, 70000, 77000, 85000, 94000, \
␣␣␣␣␣␣␣␣104000, 114000, 126000, 139000, 154000, 171000, 188000, 208000, 229000, 253000, 279000, \
␣␣␣␣␣␣␣␣308000, 340000, 375000, 12000, 13000, 14000, 16000, 18000, 19000, 22000, 25000, 27000, \
␣␣␣␣␣␣␣␣30000, 33000, 36000, 40000, 44000, 48000, 53000, 58000, 64000, 71000, 78000, 86000, 95000, \
␣␣␣␣␣␣␣␣105000, 115000, 127000, 140000, 155000, 171000, 188000, 208000, 229000, 253000, 279000, \
␣␣␣␣␣␣␣␣308000, 340000, 375000, 12000, 13000, 14000, 16000, 18000, 19000, 21000, 24000, 26000, \
␣␣␣␣␣␣␣␣29000, 32000, 35000, 39000, 43000, 47000, 52000, 57000, 63000, 70000, 77000, 85000, 94000, \
␣␣␣␣␣␣␣␣104000, 114000, 126000, 139000, 154000, 170000, 187000, 207000, 228000, 252000, 278000, \
␣␣␣␣␣␣␣␣307000, 12000, 13000, 14000, 16000, 18000, 19000, 21000, 24000, 26000, 29000, 33000, \
␣␣␣␣␣␣␣␣36000, 40000, 44000, 48000, 53000, 58000, 64000, 71000, 78000, 86000, 95000, 105000, \
␣␣␣␣␣␣␣␣115000, 127000, 140000, 155000, 171000, 188000, 208000, 229000, 253000, 279000, 308000, \
␣␣␣␣␣␣␣␣12000, 13000, 14000, 16000, 18000, 19000, 21000, 24000, 26000, 29000, 32000, 35000, 39000, \
␣␣␣␣␣␣␣␣43000, 47000, 52000, 57000, 63000, 70000, 77000, 85000, 94000, 104000, 114000, 126000, \
␣␣␣␣␣␣␣␣139000, 154000, 170000, 187000, 207000, 228000, 252000, 278000, 307000, 339000, 12000, \
␣␣␣␣␣␣␣␣13000, 14000, 16000, 18000, 19000, 21000, 24000, 26000, 29000, 32000, 35000, 39000, 43000, \
␣␣␣␣␣␣␣␣47000, 52000, 57000, 63000, 70000, 77000, 85000, 94000, 104000, 114000, 126000, 139000, \
␣␣␣␣␣␣␣␣154000, 170000, 187000, 207000, 228000, 252000, 278000, 307000, 339000, 374000, 12000, \
␣␣␣␣␣␣␣␣13000, 14000, 16000, 18000, 19000, 21000, 24000, 26000, 29000, 32000, 35000, 39000, 43000, \
␣␣␣␣␣␣␣␣47000, 52000, 57000, 63000, 70000, 77000, 85000, 94000, 104000, 114000, 126000, 139000, \
␣␣␣␣␣␣␣␣154000, 170000, 187000, 207000, 228000, 252000, 278000, 307000, 339000, 374000, 12000, \
␣␣␣␣␣␣␣␣13000, 14000, 16000, 18000, 19000, 21000, 24000, 26000, 29000, 32000, 35000, 39000, 43000, \
␣␣␣␣␣␣␣␣47000, 52000, 57000, 63000, 70000, 77000, 85000, 94000, 104000, 114000, 126000, 139000, \
␣␣␣␣␣␣␣␣154000, 170000, 187000, 207000, 228000, 252000, 278000, 307000, 339000, 374000, 12000, \
␣␣␣␣␣␣␣␣13000, 14000, 16000, 18000, 19000, 21000, 24000, 26000, 29000, 32000, 35000, 39000, 43000, \
␣␣␣␣␣␣␣␣47000, 52000, 57000, 63000, 70000, 77000, 85000, 94000, 104000, 114000, 126000, 139000, \
␣␣␣␣␣␣␣␣154000, 170000, 187000, 207000, 228000, 252000, 278000, 308000, 340000, 375000, 12000, \
␣␣␣␣␣␣␣␣13000, 14000, 16000, 18000, 19000, 21000, 24000, 26000, 29000, 32000, 35000, 39000, 43000, \
␣␣␣␣␣␣␣␣47000, 52000, 57000, 63000, 70000, 77000, 85000, 94000, 104000, 114000, 126000, 139000, \
␣␣␣␣␣␣␣␣154000, 170000, 187000, 207000, 228000, 252000, 278000, 307000, 339000, 374000, 12000, \
␣␣␣␣␣␣␣␣13000, 14000, 16000, 18000, 19000, 21000, 24000, 26000, 29000, 32000, 35000, 39000, 43000, \
␣␣␣␣␣␣␣␣47000, 52000, 57000, 63000, 70000, 77000, 85000, 94000, 104000, 114000, 126000, 139000, \
␣␣␣␣␣␣␣␣154000, 170000, 187000, 207000, 228000, 252000, 278000, 307000, 339000, 374000, 12000, \
␣␣␣␣␣␣␣␣13000, 14000, 16000, 18000, 19000, 21000, 24000, 26000, 29000, 32000, 35000, 39000, 43000, \
␣␣␣␣␣␣␣␣47000, 52000, 57000, 63000, 70000, 77000, 85000, 94000, 104000, 114000, 126000, 139000, \
␣␣␣␣␣␣␣␣154000, 170000, 187000, 207000, 228000, 252000, 278000, 307000, 339000, 374000, 12000, \
␣␣␣␣␣␣␣␣13000, 14000, 16000, 18000, 19000, 21000, 24000, 26000, 29000, 32000, 35000, 39000, 43000, \
␣␣␣␣␣␣␣␣47000, 52000, 57000, 63000, 70000, 77000, 85000, 94000, 104000, 114000, 126000, 139000, \
␣␣␣␣␣␣␣␣154000, 170000, 187000, 207000, 228000, 252000, 278000, 307000, 339000, 374000]
␣␣␣␣x2 =  [77824, 77824, 77824, 77824, 77824, 77824, 77824, 77824, 77824, 77824, 77824, 77824, \
␣␣␣␣␣␣␣␣77824, 77824, 77824, 77824, 77824, 77824, 77824, 77824, 77824, 77824, 77824, 77824, 77824, \
␣␣␣␣␣␣␣␣77824, 77824, 77824, 77824, 77824, 77824, 77824, 77824, 77824, 77824, 77824, 116736, \
␣␣␣␣␣␣␣␣116736, 116736, 116736, 116736, 116736, 116736, 116736, 116736, 116736, 116736, 116736, \
␣␣␣␣␣␣␣␣116736, 116736, 116736, 116736, 116736, 116736, 116736, 116736, 116736, 116736, 116736, \
␣␣␣␣␣␣␣␣116736, 116736, 116736, 116736, 116736, 116736, 116736, 116736, 116736, 116736, 116736, \
␣␣␣␣␣␣␣␣116736, 116736, 175104, 175104, 175104, 175104, 175104, 175104, 175104, 175104, 175104, \
␣␣␣␣␣␣␣␣175104, 175104, 175104, 175104, 175104, 175104, 175104, 175104, 175104, 175104, 175104, \
␣␣␣␣␣␣␣␣175104, 175104, 175104, 175104, 175104, 175104, 175104, 175104, 175104, 175104, 175104, \
␣␣␣␣␣␣␣␣175104, 175104, 175104, 175104, 175104, 262144, 262144, 262144, 262144, 262144, 262144, \
␣␣␣␣␣␣␣␣262144, 262144, 262144, 262144, 262144, 262144, 262144, 262144, 262144, 262144, 262144, \
␣␣␣␣␣␣␣␣262144, 262144, 262144, 262144, 262144, 262144, 262144, 262144, 262144, 262144, 262144, \
␣␣␣␣␣␣␣␣262144, 262144, 262144, 262144, 262144, 262144, 262144, 262144, 393216, 393216, 393216, \
␣␣␣␣␣␣␣␣393216, 393216, 393216, 393216, 393216, 393216, 393216, 393216, 393216, 393216, 393216, \
␣␣␣␣␣␣␣␣393216, 393216, 393216, 393216, 393216, 393216, 393216, 393216, 393216, 393216, 393216, \
␣␣␣␣␣␣␣␣393216, 393216, 393216, 393216, 393216, 393216, 393216, 393216, 393216, 589824, 589824, \
␣␣␣␣␣␣␣␣589824, 589824, 589824, 589824, 589824, 589824, 589824, 589824, 589824, 589824, 589824, \
␣␣␣␣␣␣␣␣589824, 589824, 589824, 589824, 589824, 589824, 589824, 589824, 589824, 589824, 589824, \
␣␣␣␣␣␣␣␣589824, 589824, 589824, 589824, 589824, 589824, 589824, 589824, 589824, 589824, 884736, \
␣␣␣␣␣␣␣␣884736, 884736, 884736, 884736, 884736, 884736, 884736, 884736, 884736, 884736, 884736, \
␣␣␣␣␣␣␣␣884736, 884736, 884736, 884736, 884736, 884736, 884736, 884736, 884736, 884736, 884736, \
␣␣␣␣␣␣␣␣884736, 884736, 884736, 884736, 884736, 884736, 884736, 884736, 884736, 884736, 884736, \
␣␣␣␣␣␣␣␣884736, 1769472, 1769472, 1769472, 1769472, 1769472, 1769472, 1769472, 1769472, 1769472, \
␣␣␣␣␣␣␣␣1769472, 1769472, 1769472, 1769472, 1769472, 1769472, 1769472, 1769472, 1769472, 1769472, \
␣␣␣␣␣␣␣␣1769472, 1769472, 1769472, 1769472, 1769472, 1769472, 1769472, 1769472, 1769472, 1769472, \
␣␣␣␣␣␣␣␣1769472, 1769472, 1769472, 1769472, 1769472, 1769472, 1769472, 12845721, 12845721, \
␣␣␣␣␣␣␣␣12845721, 12845721, 12845721, 12845721, 12845721, 12845721, 12845721, 12845721, 12845721, \
␣␣␣␣␣␣␣␣12845721, 12845721, 12845721, 12845721, 12845721, 12845721, 12845721, 12845721, 12845721, \
␣␣␣␣␣␣␣␣12845721, 12845721, 12845721, 12845721, 12845721, 12845721, 12845721, 12845721, 12845721, \
␣␣␣␣␣␣␣␣12845721, 12845721, 12845721, 12845721, 12845721, 12845721, 12845721, 50000000, 50000000, \
␣␣␣␣␣␣␣␣50000000, 50000000, 50000000, 50000000, 50000000, 50000000, 50000000, 50000000, 50000000, \
␣␣␣␣␣␣␣␣50000000, 50000000, 50000000, 50000000, 50000000, 50000000, 50000000, 50000000, 50000000, \
␣␣␣␣␣␣␣␣50000000, 50000000, 50000000, 50000000, 50000000, 50000000, 50000000, 50000000, 50000000, \
␣␣␣␣␣␣␣␣50000000, 50000000, 50000000, 50000000, 50000000, 50000000, 50000000, 100000000, \
␣␣␣␣␣␣␣␣100000000, 100000000, 100000000, 100000000, 100000000, 100000000, 100000000, 100000000, \
␣␣␣␣␣␣␣␣100000000, 100000000, 100000000, 100000000, 100000000, 100000000, 100000000, 100000000, \
␣␣␣␣␣␣␣␣100000000, 100000000, 100000000, 100000000, 100000000, 100000000, 100000000, 100000000, \
␣␣␣␣␣␣␣␣100000000, 100000000, 100000000, 100000000, 100000000, 100000000, 100000000, 100000000, \
␣␣␣␣␣␣␣␣100000000, 100000000, 100000000, 150000000, 150000000, 150000000, 150000000, 150000000, \
␣␣␣␣␣␣␣␣150000000, 150000000, 150000000, 150000000, 150000000, 150000000, 150000000, 150000000, \
␣␣␣␣␣␣␣␣150000000, 150000000, 150000000, 150000000, 150000000, 150000000, 150000000, 150000000, \
␣␣␣␣␣␣␣␣150000000, 150000000, 150000000, 150000000, 150000000, 150000000, 150000000, 150000000, \
␣␣␣␣␣␣␣␣150000000, 150000000, 150000000, 150000000, 150000000, 150000000, 150000000, 200000000, \
␣␣␣␣␣␣␣␣200000000, 200000000, 200000000, 200000000, 200000000, 200000000, 200000000, 200000000, \
␣␣␣␣␣␣␣␣200000000, 200000000, 200000000, 200000000, 200000000, 200000000, 200000000, 200000000, \
␣␣␣␣␣␣␣␣200000000, 200000000, 200000000, 200000000, 200000000, 200000000, 200000000, 200000000, \
␣␣␣␣␣␣␣␣200000000, 200000000, 200000000, 200000000, 200000000, 200000000, 200000000, 200000000, \
␣␣␣␣␣␣␣␣200000000, 200000000, 200000000, 250000000, 250000000, 250000000, 250000000, 250000000, \
␣␣␣␣␣␣␣␣250000000, 250000000, 250000000, 250000000, 250000000, 250000000, 250000000, 250000000, \
␣␣␣␣␣␣␣␣250000000, 250000000, 250000000, 250000000, 250000000, 250000000, 250000000, 250000000, \
␣␣␣␣␣␣␣␣250000000, 250000000, 250000000, 250000000, 250000000, 250000000, 250000000, 250000000, \
␣␣␣␣␣␣␣␣250000000, 250000000, 250000000, 250000000, 250000000, 250000000, 250000000, 300000000, \
␣␣␣␣␣␣␣␣300000000, 300000000, 300000000, 300000000, 300000000, 300000000, 300000000, 300000000, \
␣␣␣␣␣␣␣␣300000000, 300000000, 300000000, 300000000, 300000000, 300000000, 300000000, 300000000, \
␣␣␣␣␣␣␣␣300000000, 300000000, 300000000, 300000000, 300000000, 300000000, 300000000, 300000000, \
␣␣␣␣␣␣␣␣300000000, 300000000, 300000000, 300000000, 300000000, 300000000, 300000000, 300000000, \
␣␣␣␣␣␣␣␣300000000, 300000000, 300000000]
␣␣␣␣y = [0.9468800016000001, 0.9453600012000001, 0.9475400001000001, 0.9465200007000001, \
␣␣␣␣␣␣␣␣0.9446600005, 0.9470600002, 0.9458799995, 0.9472000003000001, 0.9476400018000001, \
␣␣␣␣␣␣␣␣0.9453600012000001, 0.9438400008000001, 0.9482400008, 0.9420800023, 0.9439400025, \
␣␣␣␣␣␣␣␣0.9453400001000001, 0.9397000000000001, 0.9431800023, 0.9402200021, 0.9398600012, \
␣␣␣␣␣␣␣␣0.9428599998, 0.9400399998000001, 0.9401600026, 0.9425800033, 0.9422800019, \
␣␣␣␣␣␣␣␣0.9384599999000001, 0.9385600016000001, 0.9422800019, 0.9387600012, 0.9414000027, \
␣␣␣␣␣␣␣␣0.9428400025, 0.940200001, 0.9409400001, 0.9390000030000001, 0.9394000024, \
␣␣␣␣␣␣␣␣0.9381600022000001, 0.938500002, 0.9412200004000001, 0.9402800016, 0.9406400025, \
␣␣␣␣␣␣␣␣0.9359600022000001, 0.9404600002, 0.9421600029, 0.9387200028, 0.9411600009000001, \
␣␣␣␣␣␣␣␣0.9392800033, 0.939600002, 0.9386200011, 0.9384599999000001, 0.9372600019, 0.938000001, \
␣␣␣␣␣␣␣␣0.9384599999000001, 0.9390400015, 0.9384599999000001, 0.9412400015000001, 0.9410800003, \
␣␣␣␣␣␣␣␣0.9353200048, 0.9393400028000001, 0.9349400029, 0.9375200011, 0.9381800033000001, \
␣␣␣␣␣␣␣␣0.9349400029, 0.9357400015, 0.934360005, 0.9357200041, 0.9415600002000001, \
␣␣␣␣␣␣␣␣0.9343400002000001, 0.9341600016, 0.931280002, 0.9336000010000001, 0.9288799986, \
␣␣␣␣␣␣␣␣0.9322800040000001, 0.9340400025000001, 0.9314000010000001, 0.9337200001, 0.9328000024, \
␣␣␣␣␣␣␣␣0.9313800037000001, 0.9328400046, 0.9345400035, 0.9332000017000001, 0.9345600009, \
␣␣␣␣␣␣␣␣0.9301199988000001, 0.9338000044, 0.9345199987, 0.9351399988, 0.9323400036, 0.9357400015, \
␣␣␣␣␣␣␣␣0.9326599985, 0.9314800054000001, 0.9366200045, 0.9322400019, 0.9291200042000001, \
␣␣␣␣␣␣␣␣0.9354400039, 0.9301000014, 0.9310600013, 0.9294800013000001, 0.9336199984, \
␣␣␣␣␣␣␣␣0.9295200035000001, 0.9301600009000001, 0.9289000034, 0.9311399981, 0.928960003, \
␣␣␣␣␣␣␣␣0.9290800020000001, 0.9261600003, 0.9294600040000001, 0.9258400053, 0.9280200005, \
␣␣␣␣␣␣␣␣0.9291599989, 0.9315799996, 0.9232200012, 0.9225400016, 0.920979999, 0.9200199991, \
␣␣␣␣␣␣␣␣0.9213199988, 0.9239000008, 0.9186400026, 0.9236200005, 0.9224600047, 0.920660004, \
␣␣␣␣␣␣␣␣0.9222199991000001, 0.9222400039, 0.9224200025, 0.9214800000000001, 0.9199400023000001, \
␣␣␣␣␣␣␣␣0.9254800007, 0.9210000038, 0.9220200032, 0.9230000004000001, 0.9205400050000001, \
␣␣␣␣␣␣␣␣0.9240000024, 0.9231199995, 0.9207400009000001, 0.9209199995, 0.9238400012, 0.9201200008, \
␣␣␣␣␣␣␣␣0.9234599993, 0.9219000041000001, 0.9211400002000001, 0.9200400040000001, \
␣␣␣␣␣␣␣␣0.9219600037000001, 0.9220200032, 0.9228200018, 0.9209400043, 0.9152199998, \
␣␣␣␣␣␣␣␣0.9221599996000001, 0.9020600021, 0.9046600014, 0.9043600038, 0.9024000019, 0.9066600055, \
␣␣␣␣␣␣␣␣0.9039399996, 0.9049400017, 0.9057000056000001, 0.9045400023, 0.9051000029, 0.9068600014, \
␣␣␣␣␣␣␣␣0.9075800031000001, 0.9068400040000001, 0.9095200002, 0.9119400010000001, \
␣␣␣␣␣␣␣␣0.9071800038000001, 0.9082600027000001, 0.9073200002, 0.9054200053, 0.9087800011, \
␣␣␣␣␣␣␣␣0.9059600011000001, 0.908040002, 0.9099600017, 0.90766, 0.9071800038000001, \
␣␣␣␣␣␣␣␣0.9059600011000001, 0.9056400061000001, 0.9077199996, 0.9101200029000001, \
␣␣␣␣␣␣␣␣0.9048600048000001, 0.9028000012, 0.9075600058000001, 0.905840002, 0.9104400054, \
␣␣␣␣␣␣␣␣0.8812400028, 0.8769000024, 0.8821400031000001, 0.8797200024, 0.8786400035, 0.8856000006, \
␣␣␣␣␣␣␣␣0.8814600036, 0.8819999993000001, 0.8852200061000001, 0.8850600049, 0.8815600052, \
␣␣␣␣␣␣␣␣0.8803800046, 0.8845200017, 0.8832399994000001, 0.8825200051000001, 0.8854399994000001, \
␣␣␣␣␣␣␣␣0.8817600012000001, 0.8866799995, 0.8840800002, 0.8875400051000001, 0.8882599995, \
␣␣␣␣␣␣␣␣0.8919799998, 0.8845200017, 0.8862400055, 0.8871600032, 0.8825599998, 0.8845800012, \
␣␣␣␣␣␣␣␣0.8882599995, 0.8860200047000001, 0.8884000033, 0.8887000009, 0.8853400052, 0.8845600039, \
␣␣␣␣␣␣␣␣0.8883400038, 0.8384200037, 0.8409800082000001, 0.8425600082, 0.8434400111, 0.8466999978, \
␣␣␣␣␣␣␣␣0.8451199979, 0.8513600081, 0.8542800099000001, 0.8516000062, 0.8549000025, 0.8536999971, \
␣␣␣␣␣␣␣␣0.8490599990000001, 0.8544000089, 0.8593399972, 0.8602200001, 0.8616200089, 0.8561200052, \
␣␣␣␣␣␣␣␣0.8550599962000001, 0.855640009, 0.8580000103000001, 0.8622599989, 0.8556600064000001, \
␣␣␣␣␣␣␣␣0.8591600060000001, 0.8602399975, 0.8613000065, 0.8557400107, 0.8571999967, \
␣␣␣␣␣␣␣␣0.8624600023000001, 0.8606600016, 0.8594800085000001, 0.8612000048, 0.8569800109, \
␣␣␣␣␣␣␣␣0.8653199971000001, 0.8593600094, 0.8611800075, 0.752700001, 0.745480001, \
␣␣␣␣␣␣␣␣0.7470999956000001, 0.7475599945, 0.7498199940000001, 0.7548400015000001, 0.7580000013, \
␣␣␣␣␣␣␣␣0.7639600039000001, 0.7731600106000001, 0.7685400099, 0.7747200131, 0.780520007, \
␣␣␣␣␣␣␣␣0.7807999998, 0.7841200083000001, 0.7819800079, 0.7829000056000001, 0.7864800096, \
␣␣␣␣␣␣␣␣0.7843400091, 0.7910600007, 0.7888600081, 0.7888000011, 0.7928600013, 0.7980800122, \
␣␣␣␣␣␣␣␣0.7940400094000001, 0.7908200026000001, 0.7940800041, 0.7927000076, 0.7947200090000001, \
␣␣␣␣␣␣␣␣0.7947000116, 0.8002000004, 0.8047400117, 0.8068599999, 0.8052800000000001, 0.8003600091, \
␣␣␣␣␣␣␣␣0.7996000051000001, 0.8032599986000001, 0.7100999951, 0.6930000186, 0.6851000190000001, \
␣␣␣␣␣␣␣␣0.6584399939000001, 0.6373400092, 0.6297399998000001, 0.6123400033, 0.5938400030000001, \
␣␣␣␣␣␣␣␣0.5833000243, 0.5700800121, 0.5558000207, 0.5479600132, 0.5351400077, 0.5248399973, \
␣␣␣␣␣␣␣␣0.5177200139, 0.510800004, 0.5033000112, 0.49437999730000004, 0.49106001850000003, \
␣␣␣␣␣␣␣␣0.4827600121, 0.4779199958, 0.4746000171, 0.47251999380000004, 0.46707999710000003, \
␣␣␣␣␣␣␣␣0.4676200151, 0.47034001350000004, 0.4693400264, 0.4700199962, 0.4710000157, 0.4719600081, \
␣␣␣␣␣␣␣␣0.4747000337, 0.4745799899, 0.4787200093, 0.48155999180000003, 0.48655998710000004, \
␣␣␣␣␣␣␣␣0.49511998890000003, 0.7111999989000001, 0.6953400075, 0.6830800176, 0.6536999941, \
␣␣␣␣␣␣␣␣0.6360400021, 0.6323200166, 0.6112399995000001, 0.5901600122, 0.5810400248, 0.5657800138, \
␣␣␣␣␣␣␣␣0.5536600053, 0.5403999984, 0.5280800164, 0.5174800158, 0.5052400231, 0.4957399964, \
␣␣␣␣␣␣␣␣0.48536002640000003, 0.47584003210000003, 0.4645400047, 0.4597600102, 0.451480031, \
␣␣␣␣␣␣␣␣0.4430199862, 0.4357200265, 0.4305999875, 0.42192000150000003, 0.41430002450000003, \
␣␣␣␣␣␣␣␣0.40454000230000003, 0.4001800418, 0.39741998910000004, 0.39076000450000004, 0.3865799904, \
␣␣␣␣␣␣␣␣0.380580008, 0.3752599955, 0.3713999987, 0.36746001240000004, 0.36250001190000003, \
␣␣␣␣␣␣␣␣0.7072400153, 0.6927399933, 0.6792800128, 0.6573400199, 0.6339600086, 0.6228200197, \
␣␣␣␣␣␣␣␣0.6073600054, 0.5886999965, 0.5793800056, 0.5607199967000001, 0.5493200123, 0.5397199988, \
␣␣␣␣␣␣␣␣0.5246400237000001, 0.5096000135000001, 0.4985200167, 0.4890400171, 0.483520031, \
␣␣␣␣␣␣␣␣0.4693000317, 0.46398001910000003, 0.451660037, 0.446960032, 0.43704003100000005, \
␣␣␣␣␣␣␣␣0.4294199944, 0.4234400392, 0.41809999940000003, 0.41122001410000003, 0.4042000175, \
␣␣␣␣␣␣␣␣0.3986800313, 0.389959991, 0.3852199912, 0.3820199966, 0.374720037, 0.36915999650000003, \
␣␣␣␣␣␣␣␣0.3685200214, 0.3593200445, 0.3581399918, 0.7078000009000001, 0.6887000203, 0.6784799993, \
␣␣␣␣␣␣␣␣0.6534200013, 0.6328400075, 0.6237600148, 0.6072600186, 0.5881000161000001, 0.5783400238, \
␣␣␣␣␣␣␣␣0.5637000203, 0.5498000085, 0.5362000167000001, 0.5257800221, 0.5147000253, \
␣␣␣␣␣␣␣␣0.5023800135000001, 0.4920400381, 0.4848200083, 0.47548002, 0.4647600055, 0.4529399872, \
␣␣␣␣␣␣␣␣0.44517999890000004, 0.43494004010000004, 0.4260799885, 0.41953998800000003, \
␣␣␣␣␣␣␣␣0.41040003300000005, 0.4063200355, 0.3979800344, 0.3915799856, 0.3887000084, 0.3842000365, \
␣␣␣␣␣␣␣␣0.37696003910000003, 0.3715000153, 0.3680400252, 0.3602400422, 0.3571000099, \
␣␣␣␣␣␣␣␣0.35206002000000003, 0.7112600207, 0.6957200170000001, 0.6827999949, 0.6604399979000001, \
␣␣␣␣␣␣␣␣0.6374600232000001, 0.629640013, 0.6124400198000001, 0.5917400122, 0.5812399983000001, \
␣␣␣␣␣␣␣␣0.5640600026, 0.5519600213, 0.542840004, 0.5240800083, 0.5133600235, 0.5011800230000001, \
␣␣␣␣␣␣␣␣0.49340003730000004, 0.48006004100000005, 0.4709799886, 0.46202003960000004, 0.452639997, \
␣␣␣␣␣␣␣␣0.4438800216, 0.4360800385, 0.42448002100000004, 0.4201200008, 0.4143599868, 0.404399991, \
␣␣␣␣␣␣␣␣0.3996800184, 0.3899800181, 0.3867999911, 0.37814003230000004, 0.37507998940000004, \
␣␣␣␣␣␣␣␣0.3676400185, 0.3636000156, 0.3574399948, 0.35269999500000004, 0.3483000398, \
␣␣␣␣␣␣␣␣0.7124000192000001, 0.6934400201, 0.6806000173, 0.6572400033, 0.6354400218, 0.6240200102, \
␣␣␣␣␣␣␣␣0.6135199964, 0.5918200016, 0.575819999, 0.5584000051, 0.5481400192, 0.5354200006000001, \
␣␣␣␣␣␣␣␣0.5264800191, 0.5113800168, 0.5032000244, 0.4897199869, 0.4806799889, 0.47324001790000003, \
␣␣␣␣␣␣␣␣0.46420001980000003, 0.454760015, 0.4474800229, 0.4367600083, 0.4273200035, 0.4189800024, \
␣␣␣␣␣␣␣␣0.41190004350000003, 0.4091799855, 0.40097999570000004, 0.3931000233, 0.3876200318, \
␣␣␣␣␣␣␣␣0.3788599968, 0.3740000129, 0.36981999870000004, 0.3649600148, 0.360740006, 0.3551999927, \
␣␣␣␣␣␣␣␣0.3528000116, 0.7074200213, 0.6914200187, 0.6810400188, 0.6536999941, 0.6334000230000001, \
␣␣␣␣␣␣␣␣0.6261599958, 0.6121000051000001, 0.5911599994, 0.5768200159, 0.563560009, 0.5514599979, \
␣␣␣␣␣␣␣␣0.5377400219, 0.5260000229, 0.5130000114000001, 0.5023000240000001, 0.491320014, \
␣␣␣␣␣␣␣␣0.48294001820000004, 0.47220003600000005, 0.4616200328, 0.4522399902, 0.44446003440000004, \
␣␣␣␣␣␣␣␣0.4335800409, 0.4249200225, 0.41997998950000004, 0.41408002380000003, 0.40904003380000004, \
␣␣␣␣␣␣␣␣0.40096002820000004, 0.3950999975, 0.3882800341, 0.38148003820000004, 0.37819999460000003, \
␣␣␣␣␣␣␣␣0.37128001450000003, 0.3662599921, 0.36228001120000003, 0.35747998950000004, 0.3524399996]
␣␣␣␣#endregion
␣
␣␣␣␣x1 = np.array(x1)
␣␣␣␣x2 = np.array(x2)
␣␣␣␣y = np.array(y)
␣
␣␣␣␣x3 = np.ones_like(x1)
␣
␣␣␣␣overfit = True
␣
␣␣␣␣trivariate = False
␣
␣␣␣␣nonmon_hparams = True
␣
␣␣␣␣eval_metric_upper_limit = True
␣
␣␣␣␣x1_name = "Number of Training Steps"
␣␣␣␣x2_name = "Training Dataset Size"
␣␣␣␣x3_name = "x3"
␣␣␣␣y_name = "Test Error Rate"
␣
else:
␣␣␣␣raise ValueError("scaling_behavior "+str(scaling_behavior)+" not available")
␣
␣
␣
x1_u = np.unique(x1)
x2_u = np.unique(x2)
x3_u = np.unique(x3)
␣
print("len(x1_u)", len(x1_u))
print("len(x2_u)", len(x2_u))
print("len(x3_u)", len(x3_u))
\end{ipycell}

\begin{ipycell}
#@title Create/Configure Train Val Test Split
␣
# no split
"""
x1_val_split = 1e30
x2_val_split = 1e30
x3_val_split = 1e30
␣
x1_test_split = 1e30
x2_test_split = 1e30
x3_test_split = 1e30
#"""
␣
#"""
x1_val_split = x1.max()/2
x2_val_split = x2.max()/2
x3_val_split = x3.max()/2
␣
x1_test_split = x1.max()/2
x2_test_split = x2.max()/2
x3_test_split = x3.max()/2
#"""
␣
␣
if not trivariate:
␣␣␣␣x3_val_split = 1e30
␣␣␣␣x3_test_split = 1e30
␣
␣
x1_train = []
x2_train = []
x3_train = []
y_train = []
␣
x1_val = []
x2_val = []
x3_val = []
y_val = []
␣
x1_test = []
x2_test = []
x3_test = []
y_test = []
␣
for _x1, _x2, _x3, _y in zip(x1, x2, x3, y):
␣␣␣␣if ((_x1 < x1_val_split) and (_x2 < x2_val_split) and (_x3 < x3_val_split)):
␣␣␣␣␣␣␣␣x1_train.append(_x1)
␣␣␣␣␣␣␣␣x2_train.append(_x2)
␣␣␣␣␣␣␣␣x3_train.append(_x3)
␣␣␣␣␣␣␣␣y_train.append(_y)
␣␣␣␣elif ((_x1 < x1_test_split) and (_x2 < x2_test_split) and (_x3 < x3_test_split)):
␣␣␣␣␣␣␣␣x1_val.append(_x1)
␣␣␣␣␣␣␣␣x2_val.append(_x2)
␣␣␣␣␣␣␣␣x3_val.append(_x3)
␣␣␣␣␣␣␣␣y_val.append(_y)
␣␣␣␣else:
␣␣␣␣␣␣␣␣x1_test.append(_x1)
␣␣␣␣␣␣␣␣x2_test.append(_x2)
␣␣␣␣␣␣␣␣x3_test.append(_x3)
␣␣␣␣␣␣␣␣y_test.append(_y)
␣
x1_train = np.array(x1_train)
x2_train = np.array(x2_train)
x3_train = np.array(x3_train)
y_train = np.array(y_train)
␣
x1_val = np.array(x1_val)
x2_val = np.array(x2_val)
x3_val = np.array(x3_val)
y_val = np.array(y_val)
␣
x1_test = np.array(x1_test)
x2_test = np.array(x2_test)
x3_test = np.array(x3_test)
y_test = np.array(y_test)
␣
x_train = np.stack([x1_train, x2_train, x3_train], -1)
x_val = np.stack([x1_val, x2_val, x3_val], -1)
x_test = np.stack([x1_test, x2_test, x3_test], -1)
␣
print("y_train.shape", y_train.shape)
print("y_val.shape", y_val.shape)
print("y_test.shape", y_test.shape)
␣
assert y_train.shape[0] > 0
\end{ipycell}

\begin{ipycell}
#@title The functional forms
␣
class BiasOnlyDense(nn.Module):
␣␣␣␣features: int
␣
␣␣␣␣@nn.compact
␣␣␣␣def __call__(self, x):
␣␣␣␣␣␣␣␣bias = self.param("bias", nn.initializers.zeros, (self.features,))
␣␣␣␣␣␣␣␣return x + bias
␣
␣
def mbnsl(x, n_breaks, name):
␣␣␣␣if n_breaks == 0:
␣␣␣␣␣␣␣␣out = nn.Dense(1)(x)
␣␣␣␣else:
␣␣␣␣␣␣␣␣out = nn.Dense(1)(x) + nn.Dense(1, use_bias=False)(
␣␣␣␣␣␣␣␣␣␣␣␣nn.softplus(type(name, (nn.Dense,), {})(n_breaks)(x))
␣␣␣␣␣␣␣␣)
␣␣␣␣return out
␣
␣
def upper_limit(m, l, eps_array):
␣␣␣␣out = m - jnp.log(1.0 + jnp.exp(m + l))
␣␣␣␣return out
␣
␣
class FunctionalForm(nn.Module):
␣␣␣␣func_form: str
␣␣␣␣n_breaks: int
␣␣␣␣trivariate: bool
␣␣␣␣overfit: bool           # turns overfitting component on/off
␣␣␣␣nonmon_hparams: bool    # turns nonmonotonic hyperparameters component on/off
␣␣␣␣eval_metric_upper_limit: bool    # turns a_2 on/off
␣
␣␣␣␣@nn.compact
␣␣␣␣def __call__(self, x, batch_train):
␣␣␣␣␣␣␣␣n = self.n_breaks
␣
␣␣␣␣␣␣␣␣x = jnp.log(x)
␣
␣␣␣␣␣␣␣␣offset = 1e-20
␣␣␣␣␣␣␣␣x_train, y_train = batch_train
␣␣␣␣␣␣␣␣x_train, y_train = jnp.log(x_train), jnp.log(y_train + offset)
␣
␣␣␣␣␣␣␣␣x_mean = jnp.mean(x_train, axis=0, keepdims=True)
␣␣␣␣␣␣␣␣x_std = jnp.std(x_train, axis=0, keepdims=True)
␣
␣␣␣␣␣␣␣␣# normalize input
␣␣␣␣␣␣␣␣x = (x - x_mean) / x_std
␣
␣␣␣␣␣␣␣␣x_U = x[:, [0, 1, 2]] if self.trivariate else x[:, [0, 1]]
␣␣␣␣␣␣␣␣x_1 = x[:, [0]]
␣␣␣␣␣␣␣␣x_2 = x[:, [1]]
␣␣␣␣␣␣␣␣x_3 = x[:, [2]]
␣
␣␣␣␣␣␣␣␣y_mean = jnp.mean(y_train)
␣␣␣␣␣␣␣␣u = jnp.ones_like(x_1[:, 0]) * y_mean
␣
␣␣␣␣␣␣␣␣eps = jnp.log(offset)
␣␣␣␣␣␣␣␣eps_array = jnp.ones_like(x_1) * eps   # this is used to prevent dividing by zero
␣␣␣␣␣␣␣␣x0 = jnp.zeros_like(x_1)
␣
␣␣␣␣␣␣␣␣def make_a(tag):
␣␣␣␣␣␣␣␣␣␣␣␣return type(tag, (BiasOnlyDense,), {})(1)(x0)
␣
␣␣␣␣␣␣␣␣def cat(xs):
␣␣␣␣␣␣␣␣␣␣␣␣return jnp.concatenate(xs, axis=-1)
␣
␣␣␣␣␣␣␣␣def R_components(suffix):
␣␣␣␣␣␣␣␣␣␣␣␣if self.trivariate:
␣␣␣␣␣␣␣␣␣␣␣␣␣␣␣␣return [
␣␣␣␣␣␣␣␣␣␣␣␣␣␣␣␣␣␣␣␣mbnsl(x_U, n, f"x_123{suffix}"),
␣␣␣␣␣␣␣␣␣␣␣␣␣␣␣␣␣␣␣␣mbnsl(x_1, n, f"x_1{suffix}"),
␣␣␣␣␣␣␣␣␣␣␣␣␣␣␣␣␣␣␣␣mbnsl(x_2, n, f"x_2{suffix}"),
␣␣␣␣␣␣␣␣␣␣␣␣␣␣␣␣␣␣␣␣mbnsl(x_3, n, f"x_3{suffix}"),
␣␣␣␣␣␣␣␣␣␣␣␣␣␣␣␣]
␣␣␣␣␣␣␣␣␣␣␣␣return [
␣␣␣␣␣␣␣␣␣␣␣␣␣␣␣␣mbnsl(x_U, n, f"x_12{suffix}"),
␣␣␣␣␣␣␣␣␣␣␣␣␣␣␣␣mbnsl(x_1, n, f"x_1{suffix}"),
␣␣␣␣␣␣␣␣␣␣␣␣␣␣␣␣mbnsl(x_2, n, f"x_2{suffix}"),
␣␣␣␣␣␣␣␣␣␣␣␣]
␣
␣␣␣␣␣␣␣␣a_0 = make_a("a_0")
␣␣␣␣␣␣␣␣if eval_metric_upper_limit:
␣␣␣␣␣␣␣␣␣␣␣␣a_2 = make_a("a_2")
␣␣␣␣␣␣␣␣a_3 = make_a("a_3")
␣
␣␣␣␣␣␣␣␣if self.func_form == "unsl":
␣
␣␣␣␣␣␣␣␣␣␣␣␣# ---- overfitting component ----
␣␣␣␣␣␣␣␣␣␣␣␣if self.overfit:
␣␣␣␣␣␣␣␣␣␣␣␣␣␣␣␣a_1 = make_a("a_1")
␣␣␣␣␣␣␣␣␣␣␣␣␣␣␣␣a_5 = make_a("a_5")
␣
␣␣␣␣␣␣␣␣␣␣␣␣␣␣␣␣if self.nonmon_hparams:
␣␣␣␣␣␣␣␣␣␣␣␣␣␣␣␣␣␣␣␣a_6 = make_a("a_6")
␣␣␣␣␣␣␣␣␣␣␣␣␣␣␣␣␣␣␣␣log_Q5_hparam_force = cat(R_components("_r_o") + [eps_array, a_6, eps_array])
␣␣␣␣␣␣␣␣␣␣␣␣␣␣␣␣␣␣␣␣log_Q5_hparam_force = -jax.scipy.special.logsumexp(
␣␣␣␣␣␣␣␣␣␣␣␣␣␣␣␣␣␣␣␣␣␣␣␣log_Q5_hparam_force, axis=-1, keepdims=True
␣␣␣␣␣␣␣␣␣␣␣␣␣␣␣␣␣␣␣␣)
␣␣␣␣␣␣␣␣␣␣␣␣␣␣␣␣else:
␣␣␣␣␣␣␣␣␣␣␣␣␣␣␣␣␣␣␣␣log_Q5_hparam_force = eps_array
␣
␣␣␣␣␣␣␣␣␣␣␣␣␣␣␣␣log_R5 = cat(R_components("_o") + [eps_array])
␣␣␣␣␣␣␣␣␣␣␣␣␣␣␣␣log_R5 = jax.scipy.special.logsumexp(log_R5, axis=-1, keepdims=True)
␣␣␣␣␣␣␣␣␣␣␣␣␣␣␣␣log_Q5_first = upper_limit(log_R5, a_5, eps_array)
␣
␣␣␣␣␣␣␣␣␣␣␣␣␣␣␣␣log_overfit_force = cat([log_Q5_first, log_Q5_hparam_force, a_1, eps_array])
␣␣␣␣␣␣␣␣␣␣␣␣␣␣␣␣log_overfit_force = -jax.scipy.special.logsumexp(
␣␣␣␣␣␣␣␣␣␣␣␣␣␣␣␣␣␣␣␣log_overfit_force, axis=-1, keepdims=True
␣␣␣␣␣␣␣␣␣␣␣␣␣␣␣␣)
␣␣␣␣␣␣␣␣␣␣␣␣else:
␣␣␣␣␣␣␣␣␣␣␣␣␣␣␣␣log_overfit_force = eps_array
␣
␣␣␣␣␣␣␣␣␣␣␣␣# ---- nonmon hparams component ----
␣␣␣␣␣␣␣␣␣␣␣␣if self.nonmon_hparams:
␣␣␣␣␣␣␣␣␣␣␣␣␣␣␣␣a_4 = make_a("a_4")
␣␣␣␣␣␣␣␣␣␣␣␣␣␣␣␣log_Q3_hparam_force = cat(R_components("_r") + [eps_array, a_4, eps_array])
␣␣␣␣␣␣␣␣␣␣␣␣␣␣␣␣log_Q3_hparam_force = -jax.scipy.special.logsumexp(
␣␣␣␣␣␣␣␣␣␣␣␣␣␣␣␣␣␣␣␣log_Q3_hparam_force, axis=-1, keepdims=True
␣␣␣␣␣␣␣␣␣␣␣␣␣␣␣␣)
␣␣␣␣␣␣␣␣␣␣␣␣else:
␣␣␣␣␣␣␣␣␣␣␣␣␣␣␣␣log_Q3_hparam_force = eps_array
␣
␣␣␣␣␣␣␣␣␣␣␣␣# ---- main body ----
␣␣␣␣␣␣␣␣␣␣␣␣log_R3 = cat(R_components("") + [eps_array])
␣␣␣␣␣␣␣␣␣␣␣␣log_R3 = jax.scipy.special.logsumexp(log_R3, axis=-1, keepdims=True)
␣␣␣␣␣␣␣␣␣␣␣␣log_Q3_first = upper_limit(log_R3, a_3, eps_array)
␣
␣␣␣␣␣␣␣␣␣␣␣␣log_Q3_plus_overfit = cat([log_Q3_first, log_Q3_hparam_force, log_overfit_force, eps_array])
␣␣␣␣␣␣␣␣␣␣␣␣log_Q3_plus_overfit = jax.scipy.special.logsumexp(
␣␣␣␣␣␣␣␣␣␣␣␣␣␣␣␣log_Q3_plus_overfit, axis=-1, keepdims=True
␣␣␣␣␣␣␣␣␣␣␣␣)
␣
␣␣␣␣␣␣␣␣␣␣␣␣if eval_metric_upper_limit:
␣␣␣␣␣␣␣␣␣␣␣␣␣␣␣␣log_eq1_outer = upper_limit(log_Q3_plus_overfit, a_2, eps_array)
␣␣␣␣␣␣␣␣␣␣␣␣else:
␣␣␣␣␣␣␣␣␣␣␣␣␣␣␣␣log_eq1_outer = log_Q3_plus_overfit
␣
␣␣␣␣␣␣␣␣␣␣␣␣log_y = cat([log_eq1_outer, a_0, eps_array])
␣␣␣␣␣␣␣␣␣␣␣␣log_y = jax.scipy.special.logsumexp(log_y, axis=-1, keepdims=False)
␣
␣␣␣␣␣␣␣␣elif self.func_form == "chinchilla":
␣␣␣␣␣␣␣␣␣␣␣␣log_y = cat([mbnsl(x_1, 0, "x_1"), mbnsl(x_2, 0, "x_2"), a_0, eps_array])
␣␣␣␣␣␣␣␣␣␣␣␣log_y = jax.scipy.special.logsumexp(log_y, axis=-1, keepdims=False)
␣
␣␣␣␣␣␣␣␣else:
␣␣␣␣␣␣␣␣␣␣␣␣raise ValueError("functional_form " + str(self.func_form) + " not available")
␣
␣␣␣␣␣␣␣␣# (inverse) normalize output via mean of data
␣␣␣␣␣␣␣␣log_y = log_y + u
␣␣␣␣␣␣␣␣return jnp.exp(log_y)
␣
␣
def functional_form__rmsle(params, _model, batch, batch_train):
␣␣␣␣offset = 1e-36
␣␣␣␣x, y = batch
␣␣␣␣y_pred = _model.apply(params, x, batch_train)
␣␣␣␣pred, target = jnp.log(y_pred + offset), jnp.log(y + offset)
␣␣␣␣sle = jnp.square(pred - target)
␣␣␣␣return jnp.sqrt(jnp.mean(sle))
␣
␣
def l2_regularization(params, exclude_bias=True):
␣␣␣␣l2_loss = 0
␣␣␣␣for _, module_params in params["params"].items():
␣␣␣␣␣␣␣␣for param_name, param_value in module_params.items():
␣␣␣␣␣␣␣␣␣␣␣␣if exclude_bias and param_name == "bias":
␣␣␣␣␣␣␣␣␣␣␣␣␣␣␣␣continue
␣␣␣␣␣␣␣␣␣␣␣␣l2_loss += jnp.sum(jnp.square(param_value))
␣␣␣␣return 0.5 * l2_loss
␣
def count_params(params):
␣␣return sum([jnp.prod(jnp.array(x.shape))
␣␣␣␣␣␣␣␣␣␣␣␣␣␣for x in jax.tree_util.tree_leaves(params)])
\end{ipycell}

\begin{ipycell}
#@title Fits the functional form; NOTE: will stall if any jax arrays are outside of optimize_model function
␣
import gc; gc.collect()
␣
def optimize_model(args):
␣
␣␣␣␣seed, _train, _val, _test = args
␣
␣␣␣␣x_train, y_train = _train
␣␣␣␣x_val, y_val = _val
␣␣␣␣x_test, y_test = _test
␣
␣␣␣␣x_train = jnp.array(x_train)
␣␣␣␣y_train = jnp.array(y_train)
␣
␣␣␣␣x_val = jnp.array(x_val)
␣␣␣␣y_val = jnp.array(y_val)
␣
␣␣␣␣x_test = jnp.array(x_test)
␣␣␣␣y_test = jnp.array(y_test)
␣
␣␣␣␣#n_breaks = 2
␣␣␣␣n_breaks = 1
␣␣␣␣#n_breaks = 0
␣
␣␣␣␣model = FunctionalForm(func_form=func_form,
␣␣␣␣␣␣␣␣␣␣␣␣␣␣␣␣␣␣␣␣␣␣␣␣␣␣n_breaks=n_breaks,
␣␣␣␣␣␣␣␣␣␣␣␣␣␣␣␣␣␣␣␣␣␣␣␣␣␣trivariate=trivariate,
␣␣␣␣␣␣␣␣␣␣␣␣␣␣␣␣␣␣␣␣␣␣␣␣␣␣nonmon_hparams=nonmon_hparams,
␣␣␣␣␣␣␣␣␣␣␣␣␣␣␣␣␣␣␣␣␣␣␣␣␣␣overfit=overfit,
␣␣␣␣␣␣␣␣␣␣␣␣␣␣␣␣␣␣␣␣␣␣␣␣␣␣eval_metric_upper_limit=eval_metric_upper_limit)
␣
␣␣␣␣#L2_reg = 0.0
␣␣␣␣#L2_reg = 1e-7
␣␣␣␣#L2_reg = 1e-6
␣␣␣␣L2_reg = 3e-6
␣␣␣␣#L2_reg = 1e-5
␣␣␣␣#L2_reg = 1e-4
␣␣␣␣#L2_reg = 1e-3
␣
␣␣␣␣rng = random.PRNGKey(seed)
␣␣␣␣rng, init_rng = random.split(rng)
␣␣␣␣params = model.init(init_rng, x_train, (x_train, y_train))
␣
␣␣␣␣def loss(params, batch, offset=1e-20):
␣
␣␣␣␣␣␣␣␣x, y = batch
␣␣␣␣␣␣␣␣y_pred = model.apply(params, x, batch)
␣␣␣␣␣␣␣␣pred, target = jnp.log(offset + y_pred), jnp.log(offset + y)
␣
␣␣␣␣␣␣␣␣y_std = jnp.std(target)
␣
␣␣␣␣␣␣␣␣#This causes the learning rate to be scaled based on std of data
␣␣␣␣␣␣␣␣#"""
␣␣␣␣␣␣␣␣target = target / y_std
␣␣␣␣␣␣␣␣pred = pred / y_std
␣␣␣␣␣␣␣␣#"""
␣
␣␣␣␣␣␣␣␣kfac_jax.register_squared_error_loss(pred, target)
␣
␣␣␣␣␣␣␣␣if L2_reg > 0.0:
␣␣␣␣␣␣␣␣␣␣␣␣l2_loss = l2_regularization(params)
␣␣␣␣␣␣␣␣␣␣␣␣return jnp.mean(jnp.square(pred - target)) + L2_reg * l2_loss
␣␣␣␣␣␣␣␣else:
␣␣␣␣␣␣␣␣␣␣␣␣return jnp.mean(jnp.square(pred - target))
␣
␣␣␣␣# using a second order optimizer makes training way faster
␣␣␣␣optimizer = kfac_jax.Optimizer(
␣␣␣␣␣␣value_and_grad_func=jax.value_and_grad(loss),
␣␣␣␣␣␣l2_reg=L2_reg,
␣␣␣␣␣␣value_func_has_aux=False,
␣␣␣␣␣␣value_func_has_state=False,
␣␣␣␣␣␣value_func_has_rng=False,
␣␣␣␣␣␣use_adaptive_learning_rate=True,
␣␣␣␣␣␣use_adaptive_momentum=True,
␣␣␣␣␣␣use_adaptive_damping=True,
␣␣␣␣␣␣initial_damping=1.0,
␣␣␣␣␣␣num_burnin_steps=0,
␣␣␣␣␣␣#multi_device=False,
␣␣␣␣)
␣
␣␣␣␣rng, init_rng = random.split(rng)
␣␣␣␣opt_state = optimizer.init(params, init_rng, (x_train, y_train))
␣
␣␣␣␣#n_steps = int(1e1) #for debug
␣
␣␣␣␣#n_steps = int(1e4)
␣␣␣␣n_steps = int(2e4)
␣␣␣␣#n_steps = int(5e4)
␣
␣␣␣␣best_loss = 1e6
␣␣␣␣best_params = params
␣␣␣␣nan_bool = False
␣
␣␣␣␣#update_rate = 25
␣␣␣␣update_rate = 1
␣
␣␣␣␣for i in range(n_steps):
␣␣␣␣␣␣␣␣rng, step_rng = jax.random.split(rng)
␣␣␣␣␣␣␣␣params, opt_state, stats = optimizer.step(
␣␣␣␣␣␣␣␣␣␣␣␣params, opt_state, step_rng, batch=(x_train, y_train), global_step_int=i)
␣
␣␣␣␣␣␣␣␣if jnp.isnan(stats['loss']):
␣␣␣␣␣␣␣␣␣␣␣␣nan_bool = True
␣␣␣␣␣␣␣␣␣␣␣␣break
␣
␣␣␣␣␣␣␣␣if i % update_rate == 0:
␣␣␣␣␣␣␣␣␣␣␣␣#rmsle_val = functional_form__rmsle(params, model, (x_val, y_val), (x_train, y_train))
␣␣␣␣␣␣␣␣␣␣␣␣#rmsle__not_test = jnp.sqrt(stats['loss']) * .5 + rmsle_val * .5
␣
␣␣␣␣␣␣␣␣␣␣␣␣rmsle__not_test = stats['loss']
␣␣␣␣␣␣␣␣␣␣␣␣if rmsle__not_test < best_loss:
␣␣␣␣␣␣␣␣␣␣␣␣␣␣␣␣best_params = copy.deepcopy(params)
␣␣␣␣␣␣␣␣␣␣␣␣␣␣␣␣best_loss = rmsle__not_test
␣
␣␣␣␣rmsle_train = functional_form__rmsle(best_params, model, (x_train, y_train), (x_train, y_train))
␣␣␣␣rmsle_train = rmsle_train.item()
␣
␣␣␣␣best_params = jax.tree_util.tree_map(lambda x: np.array(x), best_params)
␣
␣␣␣␣return best_params, best_loss.item(), rmsle_train, model, nan_bool
␣
␣
if __name__ == '__main__':
␣
␣␣␣␣_cpu_ids = list(os.sched_getaffinity(0))
␣
␣␣␣␣n_b = int(1e2)
␣␣␣␣#n_b = int(1e9)
␣
␣␣␣␣n_s = max(int(len(_cpu_ids)/2), 1)
␣␣␣␣n_e = int(os.cpu_count())
␣
␣␣␣␣best_loss = 1e6
␣␣␣␣best_rmsle_train = 1e6
␣␣␣␣best_params = None
␣
␣␣␣␣start_time = time.time()
␣
␣␣␣␣#dummy_debug = optimize_model((1, (x_train, y_train), (x_val, y_val), (x_test, y_test)))
␣
␣␣␣␣print("rmsle_train", "rmsle_val", "rmsle_test", "loss", "     ", "time", "time_per_seed", \
␣␣␣␣␣␣␣␣"num_runs_converged", "", "seeds")
␣
␣␣␣␣for j in range(n_b):
␣␣␣␣␣␣␣␣batch_start_time = time.time()
␣␣␣␣␣␣␣␣executor = get_reusable_executor(max_workers=n_e)
␣␣␣␣␣␣␣␣futures = [executor.submit(optimize_model, (i, (x_train, y_train), (x_val, y_val), ( \
␣␣␣␣␣␣␣␣␣␣␣␣x_test, y_test))) for i in range(int(n_s*j), int(n_s*(j+1)))]
␣␣␣␣␣␣␣␣results = [f.result() for f in futures]
␣
␣␣␣␣␣␣␣␣# Unpack results
␣␣␣␣␣␣␣␣params_array = [result[0] for result in results]
␣␣␣␣␣␣␣␣losses_array = [result[1] for result in results]
␣␣␣␣␣␣␣␣rmsle_train_array = [result[2] for result in results]
␣␣␣␣␣␣␣␣model = results[0][3]
␣
␣␣␣␣␣␣␣␣nan_count = np.mean(np.array([result[4] for result in results]))
␣
␣␣␣␣␣␣␣␣# Determine the run with the best loss
␣␣␣␣␣␣␣␣best_run_index = np.argmin(np.array(losses_array))
␣␣␣␣␣␣␣␣_best_params = params_array[best_run_index]
␣␣␣␣␣␣␣␣_best_loss = losses_array[best_run_index]
␣␣␣␣␣␣␣␣_best_rmsle_train = rmsle_train_array[best_run_index]
␣
␣␣␣␣␣␣␣␣if _best_loss < best_loss:
␣␣␣␣␣␣␣␣␣␣␣␣best_loss = _best_loss
␣␣␣␣␣␣␣␣␣␣␣␣best_params = _best_params
␣␣␣␣␣␣␣␣␣␣␣␣best_rmsle_train = _best_rmsle_train
␣
␣␣␣␣␣␣␣␣best_rmsle_train = functional_form__rmsle(best_params, model, (x_train, y_train), ( \
␣␣␣␣␣␣␣␣␣␣␣␣x_train, y_train))
␣␣␣␣␣␣␣␣best_rmsle_val = functional_form__rmsle(best_params, model, (x_val, y_val), (x_train, \
␣␣␣␣␣␣␣␣␣␣␣␣y_train))
␣␣␣␣␣␣␣␣best_rmsle_test = functional_form__rmsle(best_params, model, (x_test, y_test), (x_train, \
␣␣␣␣␣␣␣␣␣␣␣␣y_train))
␣
␣␣␣␣␣␣␣␣print(f"{best_rmsle_train:.5e}", f"{best_rmsle_val:.5e}", f"{best_rmsle_test:.5e}", f"{ \
␣␣␣␣␣␣␣␣␣␣␣␣best_loss:.5e}", "   ", f"{(time.time() - start_time):.3e}", f"{(time.time() - \
␣␣␣␣␣␣␣␣␣␣␣␣batch_start_time)/n_s:.3e}", f"{nan_count:.1e}", "", int(n_s*(j+1)))
\end{ipycell}

\begin{ipycell}
#@title Convert Everything to Jax
␣
best_params = jax.tree_util.tree_map(lambda x: jnp.array(x), best_params)
␣
␣
x1_train = jnp.array(x1_train)
x2_train = jnp.array(x2_train)
x3_train = jnp.array(x3_train)
y_train = jnp.array(y_train)
␣
x1_test = jnp.array(x1_test)
x2_test = jnp.array(x2_test)
x3_test = jnp.array(x3_test)
y_test = jnp.array(y_test)
␣
␣
x1 = jnp.array(x1)
x2 = jnp.array(x2)
x3 = jnp.array(x3)
y = jnp.array(y)
␣
␣
x1_u = jnp.array(x1_u)
x2_u = jnp.array(x2_u)
x3_u = jnp.array(x3_u)
␣
bc = best_params
\end{ipycell}

\begin{ipycell}
#@title Plots Bivariate
␣
if not trivariate:
␣
␣␣␣␣import matplotlib
␣␣␣␣import matplotlib.pyplot as plt
␣␣␣␣import matplotlib.cm as cm
␣␣␣␣import matplotlib.colors as colors
␣
␣␣␣␣rmsle = functional_form__rmsle(bc, model, (jnp.stack([x1, x2, x3], -1), y), (x_train, y_train))
␣␣␣␣rmsle_train = functional_form__rmsle(bc, model, (x_train, y_train), (x_train, y_train))
␣␣␣␣rmsle_test = functional_form__rmsle(bc, model, (x_test, y_test), (x_train, y_train))
␣␣␣␣print("rmsle all:   ", rmsle)
␣␣␣␣print("rmsle train: ", rmsle_train)
␣␣␣␣print("rmsle test:  ", rmsle_test)
␣
␣␣␣␣print()
␣␣␣␣print(scaling_behavior)
␣␣␣␣#print(count_params(bc), 'params')
␣␣␣␣#print()
␣
␣␣␣␣fig, axs = plt.subplots(1, 2, figsize=(15, 5))
␣
␣␣␣␣base_cmap = matplotlib.colormaps['inferno']
␣␣␣␣color_min = 0.0
␣␣␣␣color_max = 0.91
␣␣␣␣cmap = colors.ListedColormap(
␣␣␣␣␣␣base_cmap(np.linspace(color_min, color_max, 256))
␣␣␣␣)
␣
␣␣␣␣norm_x1 = colors.LogNorm(x1.min(), x1.max())
␣␣␣␣norm_x2 = colors.LogNorm(x2.min(), x2.max())
␣␣␣␣norm_x3 = colors.LogNorm(x3.min(), x3.max())
␣
␣␣␣␣print()
␣␣␣␣print("len(x1_u) =", len(x1_u))
␣␣␣␣print("len(x2_u) =", len(x2_u))
␣␣␣␣print()
␣
␣␣␣␣x3_slice = x3_u[-1]
␣
␣␣␣␣points = 4096
␣␣␣␣x_tile = jnp.logspace(-13, 13, points)
␣
␣␣␣␣linewidth = 0.85
␣
␣
␣␣␣␣for _x2_u in x2_u:
␣␣␣␣␣␣␣␣axs[0].plot(x_tile,
␣␣␣␣␣␣␣␣␣␣␣␣␣␣␣␣␣␣␣␣model.apply(bc, jnp.stack([x_tile, jnp.full(points, _x2_u), jnp.full(points, \
␣␣␣␣␣␣␣␣␣␣␣␣␣␣␣␣␣␣␣␣␣␣␣␣x3_slice)], -1), (x_train, y_train)),
␣␣␣␣␣␣␣␣␣␣␣␣␣␣␣␣␣␣␣␣'-', color=cmap(norm_x2(_x2_u)), linewidth=linewidth)
␣␣␣␣for _x1_u in x1_u:
␣␣␣␣␣␣␣␣axs[1].plot(x_tile,
␣␣␣␣␣␣␣␣␣␣␣␣␣␣␣␣␣␣␣␣model.apply(bc, jnp.stack([jnp.full(points, _x1_u), x_tile, jnp.full(points, \
␣␣␣␣␣␣␣␣␣␣␣␣␣␣␣␣␣␣␣␣␣␣␣␣x3_slice)], -1), (x_train, y_train)),
␣␣␣␣␣␣␣␣␣␣␣␣␣␣␣␣␣␣␣␣'-', color=cmap(norm_x1(_x1_u)), linewidth=linewidth)
␣
␣␣␣␣markersize = 7.9
␣␣␣␣markeredgewidth = 0.9
␣
␣␣␣␣markeredgecolor_black=[0.0, 0.0, 0.0]
␣␣␣␣markeredgecolor_white=[1.0, 1.0, 1.0]
␣
␣␣␣␣edge_width_mul = 2
␣
␣␣␣␣marker_train = '^'
␣␣␣␣marker_test = 'o'
␣
␣␣␣␣for _x1_train, _x2_train, _x3_train, _y_train in zip(x1_train, x2_train, x3_train, y_train):
␣
␣␣␣␣␣␣␣␣axs[0].plot(_x1_train, _y_train, marker_train, markerfacecolor=cmap(norm_x2(_x2_train)), \
␣␣␣␣␣␣␣␣␣␣␣␣markersize=markersize, markeredgecolor=markeredgecolor_white, \
␣␣␣␣␣␣␣␣␣␣␣␣markeredgewidth=markeredgewidth)
␣␣␣␣␣␣␣␣axs[1].plot(_x2_train, _y_train, marker_train, markerfacecolor=cmap(norm_x1(_x1_train)), \
␣␣␣␣␣␣␣␣␣␣␣␣markersize=markersize, markeredgecolor=markeredgecolor_white, \
␣␣␣␣␣␣␣␣␣␣␣␣markeredgewidth=markeredgewidth)
␣
␣␣␣␣␣␣␣␣axs[0].plot(_x1_train, _y_train, marker_train, markerfacecolor='none', \
␣␣␣␣␣␣␣␣␣␣␣␣markersize=markersize+(markeredgewidth*edge_width_mul), \
␣␣␣␣␣␣␣␣␣␣␣␣markeredgecolor=markeredgecolor_black, markeredgewidth=markeredgewidth)
␣␣␣␣␣␣␣␣axs[1].plot(_x2_train, _y_train, marker_train, markerfacecolor='none', \
␣␣␣␣␣␣␣␣␣␣␣␣markersize=markersize+(markeredgewidth*edge_width_mul), \
␣␣␣␣␣␣␣␣␣␣␣␣markeredgecolor=markeredgecolor_black, markeredgewidth=markeredgewidth)
␣
␣␣␣␣for _x1_test, _x2_test, _x3_test, _y_test in zip(x1_test, x2_test, x3_test, y_test):
␣
␣␣␣␣␣␣␣␣axs[0].plot(_x1_test, _y_test, marker_test, markerfacecolor=cmap(norm_x2(_x2_test)), \
␣␣␣␣␣␣␣␣␣␣␣␣markersize=markersize, markeredgecolor=markeredgecolor_white, \
␣␣␣␣␣␣␣␣␣␣␣␣markeredgewidth=markeredgewidth)
␣␣␣␣␣␣␣␣axs[1].plot(_x2_test, _y_test, marker_test, markerfacecolor=cmap(norm_x1(_x1_test)), \
␣␣␣␣␣␣␣␣␣␣␣␣markersize=markersize, markeredgecolor=markeredgecolor_white, \
␣␣␣␣␣␣␣␣␣␣␣␣markeredgewidth=markeredgewidth)
␣
␣␣␣␣␣␣␣␣axs[0].plot(_x1_test, _y_test, marker_test, markerfacecolor='none', \
␣␣␣␣␣␣␣␣␣␣␣␣markersize=markersize+(markeredgewidth*edge_width_mul), \
␣␣␣␣␣␣␣␣␣␣␣␣markeredgecolor=markeredgecolor_black, markeredgewidth=markeredgewidth)
␣␣␣␣␣␣␣␣axs[1].plot(_x2_test, _y_test, marker_test, markerfacecolor='none', \
␣␣␣␣␣␣␣␣␣␣␣␣markersize=markersize+(markeredgewidth*edge_width_mul), \
␣␣␣␣␣␣␣␣␣␣␣␣markeredgecolor=markeredgecolor_black, markeredgewidth=markeredgewidth)
␣
␣␣␣␣#uncomment if you want functional_form to be render on top of (rather than behind) ground truth
␣␣␣␣# data
␣␣␣␣#"""
␣␣␣␣for _x2_u in x2_u:
␣␣␣␣␣␣␣␣axs[0].plot(x_tile,
␣␣␣␣␣␣␣␣␣␣␣␣␣␣␣␣␣␣␣␣model.apply(bc, jnp.stack([x_tile, jnp.full(points, _x2_u), jnp.full(points, \
␣␣␣␣␣␣␣␣␣␣␣␣␣␣␣␣␣␣␣␣␣␣␣␣x3_slice)], -1), (x_train, y_train)),
␣␣␣␣␣␣␣␣␣␣␣␣␣␣␣␣␣␣␣␣'-', color=cmap(norm_x2(_x2_u)), linewidth=linewidth)
␣␣␣␣for _x1_u in x1_u:
␣␣␣␣␣␣␣␣axs[1].plot(x_tile,
␣␣␣␣␣␣␣␣␣␣␣␣␣␣␣␣␣␣␣␣model.apply(bc, jnp.stack([jnp.full(points, _x1_u), x_tile, jnp.full(points, \
␣␣␣␣␣␣␣␣␣␣␣␣␣␣␣␣␣␣␣␣␣␣␣␣x3_slice)], -1), (x_train, y_train)),
␣␣␣␣␣␣␣␣␣␣␣␣␣␣␣␣␣␣␣␣'-', color=cmap(norm_x1(_x1_u)), linewidth=linewidth)
␣␣␣␣#"""
␣
␣␣␣␣__irr_ent = 0
␣␣␣␣axs[0].set_xlim(x1.min()*.8, x1.max()*1.3)
␣␣␣␣axs[1].set_xlim(x2.min()*.8, x2.max()*1.3)
␣
␣␣␣␣axs[0].set_ylim((y.min() - __irr_ent)*.7, (y.max() - __irr_ent)*1.3)
␣␣␣␣axs[1].set_ylim((y.min() - __irr_ent)*.7, (y.max() - __irr_ent)*1.3)
␣
␣
␣␣␣␣axs[0].set_xlabel(x1_name)
␣␣␣␣axs[1].set_xlabel(x2_name)
␣
␣
␣␣␣␣cbar00 = plt.colorbar(cm.ScalarMappable(norm=norm_x2, cmap=cmap), ax=axs[0])
␣␣␣␣cbar00.set_label(x2_name)
␣
␣␣␣␣cbar01 = plt.colorbar(cm.ScalarMappable(norm=norm_x1, cmap=cmap), ax=axs[1])
␣␣␣␣cbar01.set_label(x1_name)
␣
␣
␣␣␣␣axs[0].set_ylabel(y_name)
␣␣␣␣axs[1].set_ylabel(y_name)
␣
␣␣␣␣axs[0].set_xscale('log')
␣␣␣␣axs[0].set_yscale('log')
␣
␣␣␣␣axs[1].set_xscale('log')
␣␣␣␣axs[1].set_yscale('log')
␣
␣␣␣␣plt.tight_layout(pad=0.0)
␣
␣␣␣␣_title = ""
␣
␣␣␣␣os.makedirs("plots", exist_ok=True)
␣␣␣␣plt.savefig('plots/unsl__'+str(_title)+'.png')
␣
␣␣␣␣plt.show()
\end{ipycell}

\begin{ipycell}
#@title Plots Trivariate 1
␣
if trivariate:
␣
␣␣␣␣import matplotlib
␣␣␣␣import matplotlib.pyplot as plt
␣␣␣␣import matplotlib.cm as cm
␣␣␣␣import matplotlib.colors as colors
␣
␣␣␣␣rmsle = functional_form__rmsle(bc, model, (jnp.stack([x1, x2, x3], -1), y), (x_train, y_train))
␣␣␣␣rmsle_train = functional_form__rmsle(bc, model, (x_train, y_train), (x_train, y_train))
␣␣␣␣rmsle_test = functional_form__rmsle(bc, model, (x_test, y_test), (x_train, y_train))
␣␣␣␣print("rmsle all:   ", f"{rmsle:.2e}")
␣␣␣␣print("rmsle train: ", f"{rmsle_train:.2e}")
␣␣␣␣print("rmsle test:  ", f"{rmsle_test:.2e}")
␣
␣␣␣␣print()
␣␣␣␣print(scaling_behavior)
␣␣␣␣#print(count_params(bc), 'params')
␣
␣␣␣␣x_stack = jnp.stack([x1, x2, x3], -1)
␣
␣␣␣␣x_unique_lens = jnp.array([len(x1_u), len(x2_u), len(x3_u)])
␣
␣␣␣␣argmin = jnp.argmin(jnp.array([len(x1_u), len(x2_u), len(x3_u)]))
␣
␣␣␣␣print()
␣␣␣␣print(argmin)
␣␣␣␣print(x_stack.shape)
␣␣␣␣print(x_unique_lens[argmin])
␣␣␣␣print(x_stack[:,argmin].shape)
␣␣␣␣print()
␣
␣␣␣␣print("len(y_train)", len(y_train))
␣␣␣␣print("len(y_test)", len(y_test))
␣
␣␣␣␣x2_dict = {}
␣␣␣␣for _ in x2_u:
␣␣␣␣␣␣␣␣x2_dict[_.item()] = [[], [], []]
␣
␣␣␣␣for _x1, _x2, _x3, _y in zip(x1_train, x2_train, x3_train, y_train):
␣
␣␣␣␣␣␣␣␣x2_dict[_x2.item()][0].append(_x1)
␣␣␣␣␣␣␣␣x2_dict[_x2.item()][1].append(_x3)
␣␣␣␣␣␣␣␣x2_dict[_x2.item()][2].append(_y)
␣
␣
␣␣␣␣x2_dict_test = {}
␣␣␣␣for _ in x2_u:
␣␣␣␣␣␣␣␣x2_dict_test[_.item()] = [[], [], []]
␣
␣␣␣␣for _x1, _x2, _x3, _y in zip(x1_test, x2_test, x3_test, y_test):
␣
␣␣␣␣␣␣␣␣x2_dict_test[_x2.item()][0].append(_x1)
␣␣␣␣␣␣␣␣x2_dict_test[_x2.item()][1].append(_x3)
␣␣␣␣␣␣␣␣x2_dict_test[_x2.item()][2].append(_y)
␣
␣
␣␣␣␣fig, axs = plt.subplots(int(len(x2_u)), 2, figsize=(15, 5*int(len(x2_u))))
␣
␣␣␣␣base_cmap = matplotlib.colormaps['inferno']
␣␣␣␣color_min = 0.0
␣␣␣␣color_max = 0.91
␣␣␣␣cmap = colors.ListedColormap(
␣␣␣␣␣␣base_cmap(np.linspace(color_min, color_max, 256))
␣␣␣␣)
␣
␣␣␣␣norm_x1 = colors.LogNorm(x1.min(), x1.max())
␣␣␣␣norm_x2 = colors.LogNorm(x2.min(), x2.max())
␣␣␣␣norm_x3 = colors.LogNorm(x3.min(), x3.max())
␣
␣␣␣␣print()
␣␣␣␣print("len(x1_u) =", len(x1_u))
␣␣␣␣print("len(x2_u) =", len(x2_u))
␣␣␣␣print("len(x3_u) =", len(x3_u))
␣␣␣␣print()
␣
␣
␣
␣␣␣␣points = 4096
␣␣␣␣x_tile = jnp.logspace(-13, 13, points)
␣
␣␣␣␣linewidth = 0.85
␣
␣␣␣␣markersize = 7.9
␣␣␣␣markeredgewidth = 0.9
␣
␣␣␣␣markeredgecolor_black=[0.0, 0.0, 0.0]
␣␣␣␣markeredgecolor_white=[1.0, 1.0, 1.0]
␣
␣␣␣␣edge_width_mul = 2
␣
␣␣␣␣marker_train = '^'
␣␣␣␣marker_test = 'o'
␣
␣␣␣␣irr_ent = 0
␣
␣␣␣␣#alpha = .85
␣␣␣␣alpha = 1.0
␣
␣␣␣␣count = 0
␣␣␣␣for x2_slice in x2_u:
␣
␣␣␣␣␣␣␣␣_x1_train_ = x2_dict[x2_slice.item()][0]
␣␣␣␣␣␣␣␣_x3_train_ = x2_dict[x2_slice.item()][1]
␣␣␣␣␣␣␣␣_y_train_ = x2_dict[x2_slice.item()][2]
␣
␣␣␣␣␣␣␣␣_x1_test_ = x2_dict_test[x2_slice.item()][0]
␣␣␣␣␣␣␣␣_x3_test_ = x2_dict_test[x2_slice.item()][1]
␣␣␣␣␣␣␣␣_y_test_ = x2_dict_test[x2_slice.item()][2]
␣
␣␣␣␣␣␣␣␣print("len(_y_train_)", len(_y_train_))
␣
␣␣␣␣␣␣␣␣for _x3_u in x3_u:
␣␣␣␣␣␣␣␣␣␣␣␣axs[count, 0].plot(x_tile,
␣␣␣␣␣␣␣␣␣␣␣␣␣␣␣␣␣␣␣␣␣␣␣␣model.apply(bc, jnp.stack([x_tile, jnp.full(points, x2_slice), jnp.full( \
␣␣␣␣␣␣␣␣␣␣␣␣␣␣␣␣␣␣␣␣␣␣␣␣␣␣␣␣points, _x3_u)], -1), (x_train, y_train)),
␣␣␣␣␣␣␣␣␣␣␣␣␣␣␣␣␣␣␣␣␣␣␣␣'-', color=cmap(norm_x3(_x3_u)), linewidth=linewidth, alpha=alpha)
␣␣␣␣␣␣␣␣for _x1_u in x1_u:
␣␣␣␣␣␣␣␣␣␣␣␣axs[count, 1].plot(x_tile,
␣␣␣␣␣␣␣␣␣␣␣␣␣␣␣␣␣␣␣␣␣␣␣␣model.apply(bc, jnp.stack([jnp.full(points, _x1_u), jnp.full(points, \
␣␣␣␣␣␣␣␣␣␣␣␣␣␣␣␣␣␣␣␣␣␣␣␣␣␣␣␣x2_slice), x_tile], -1), (x_train, y_train)),
␣␣␣␣␣␣␣␣␣␣␣␣␣␣␣␣␣␣␣␣␣␣␣␣'-', color=cmap(norm_x1(_x1_u)), linewidth=linewidth, alpha=alpha)
␣
␣
␣␣␣␣␣␣␣␣for _x1_train, _x3_train, _y_train in zip(_x1_train_, _x3_train_, _y_train_):
␣
␣␣␣␣␣␣␣␣␣␣axs[count, 0].plot(_x1_train, _y_train, marker_train, markerfacecolor=cmap(norm_x3( \
␣␣␣␣␣␣␣␣␣␣␣␣␣␣_x3_train)), markersize=markersize, markeredgecolor=markeredgecolor_white, \
␣␣␣␣␣␣␣␣␣␣␣␣␣␣markeredgewidth=markeredgewidth)
␣␣␣␣␣␣␣␣␣␣axs[count, 1].plot(_x3_train, _y_train, marker_train, markerfacecolor=cmap(norm_x1( \
␣␣␣␣␣␣␣␣␣␣␣␣␣␣_x1_train)), markersize=markersize, markeredgecolor=markeredgecolor_white, \
␣␣␣␣␣␣␣␣␣␣␣␣␣␣markeredgewidth=markeredgewidth)
␣
␣␣␣␣␣␣␣␣␣␣axs[count, 0].plot(_x1_train, _y_train, marker_train, markerfacecolor='none', \
␣␣␣␣␣␣␣␣␣␣␣␣␣␣markersize=markersize+(markeredgewidth*edge_width_mul), \
␣␣␣␣␣␣␣␣␣␣␣␣␣␣markeredgecolor=markeredgecolor_black, markeredgewidth=markeredgewidth)
␣␣␣␣␣␣␣␣␣␣axs[count, 1].plot(_x3_train, _y_train, marker_train, markerfacecolor='none', \
␣␣␣␣␣␣␣␣␣␣␣␣␣␣markersize=markersize+(markeredgewidth*edge_width_mul), \
␣␣␣␣␣␣␣␣␣␣␣␣␣␣markeredgecolor=markeredgecolor_black, markeredgewidth=markeredgewidth)
␣
␣␣␣␣␣␣␣␣for _x1_test, _x3_test, _y_test in zip(_x1_test_, _x3_test_, _y_test_):
␣
␣␣␣␣␣␣␣␣␣␣axs[count, 0].plot(_x1_test, _y_test, marker_test, markerfacecolor=cmap(norm_x3( \
␣␣␣␣␣␣␣␣␣␣␣␣␣␣_x3_test)), markersize=markersize, markeredgecolor=markeredgecolor_white, \
␣␣␣␣␣␣␣␣␣␣␣␣␣␣markeredgewidth=markeredgewidth)
␣␣␣␣␣␣␣␣␣␣axs[count, 1].plot(_x3_test, _y_test, marker_test, markerfacecolor=cmap(norm_x1( \
␣␣␣␣␣␣␣␣␣␣␣␣␣␣_x1_test)), markersize=markersize, markeredgecolor=markeredgecolor_white, \
␣␣␣␣␣␣␣␣␣␣␣␣␣␣markeredgewidth=markeredgewidth)
␣
␣␣␣␣␣␣␣␣␣␣axs[count, 0].plot(_x1_test, _y_test, marker_test, markerfacecolor='none', \
␣␣␣␣␣␣␣␣␣␣␣␣␣␣markersize=markersize+(markeredgewidth*edge_width_mul), \
␣␣␣␣␣␣␣␣␣␣␣␣␣␣markeredgecolor=markeredgecolor_black, markeredgewidth=markeredgewidth)
␣␣␣␣␣␣␣␣␣␣axs[count, 1].plot(_x3_test, _y_test, marker_test, markerfacecolor='none', \
␣␣␣␣␣␣␣␣␣␣␣␣␣␣markersize=markersize+(markeredgewidth*edge_width_mul), \
␣␣␣␣␣␣␣␣␣␣␣␣␣␣markeredgecolor=markeredgecolor_black, markeredgewidth=markeredgewidth)
␣
␣␣␣␣␣␣␣␣axs[count, 0].set_xlim(x1.min()*.85, x1.max()*1.3)
␣␣␣␣␣␣␣␣axs[count, 1].set_xlim(x3.min()*.85, x3.max()*1.3)
␣
␣␣␣␣␣␣␣␣axs[count, 0].set_ylim((y.min()  - irr_ent)*.9, (y.max() - irr_ent)*1.1)
␣␣␣␣␣␣␣␣axs[count, 1].set_ylim((y.min()  - irr_ent)*.9, (y.max() - irr_ent)*1.1)
␣
␣
␣␣␣␣␣␣␣␣axs[count, 0].set_xscale('log')
␣␣␣␣␣␣␣␣axs[count, 0].set_yscale('log')
␣
␣␣␣␣␣␣␣␣axs[count, 1].set_xscale('log')
␣␣␣␣␣␣␣␣axs[count, 1].set_yscale('log')
␣
␣␣␣␣␣␣␣␣axs[count, 0].set_xlabel(x1_name)
␣␣␣␣␣␣␣␣axs[count, 1].set_xlabel(x3_name)
␣
␣␣␣␣␣␣␣␣axs[count, 0].set_ylabel(y_name)
␣␣␣␣␣␣␣␣axs[count, 1].set_ylabel(y_name)
␣
␣␣␣␣␣␣␣␣axs[count, 0].set_title(x2_name+" = "+str(f"{x2_slice.item():.5e}"))
␣␣␣␣␣␣␣␣axs[count, 1].set_title(x2_name+" = "+str(f"{x2_slice.item():.5e}"))
␣
␣␣␣␣␣␣␣␣cbar10 = plt.colorbar(cm.ScalarMappable(norm=norm_x3, cmap=cmap), ax=axs[count, 0])
␣␣␣␣␣␣␣␣cbar10.set_label(x3_name)
␣
␣␣␣␣␣␣␣␣cbar11 = plt.colorbar(cm.ScalarMappable(norm=norm_x1, cmap=cmap), ax=axs[count, 1])
␣␣␣␣␣␣␣␣cbar11.set_label(x1_name)
␣
␣␣␣␣␣␣␣␣count += 1
␣
␣␣␣␣plt.tight_layout(pad=0.0)
␣
␣␣␣␣_title = ""
␣
␣␣␣␣os.makedirs("plots", exist_ok=True)
␣␣␣␣plt.savefig('plots/unsl__'+str(_title)+'_1'+'.png')
␣
␣␣␣␣plt.show()
\end{ipycell}
  

\bibliography{iclr2026_conference}
\bibliographystyle{iclr2026_conference}

\end{document}